%% file: mypaper.tex
\documentclass[mnsc,nonblindrev,hidelinks]{informs3_homepage}

\OneAndAHalfSpacedXI

\usepackage{natbib}
 \bibpunct[, ]{(}{)}{,}{a}{}{,}%
 \def\bibfont{\small}%

\TheoremsNumberedBySection  
\ECRepeatTheorems

\EquationsNumberedBySection 

\usepackage{xcolor}
\definecolor{pastelBlue}{rgb}{0.0,0.4,0.7}
\usepackage{hyperref}
\hypersetup{
colorlinks=true,
allcolors = pastelBlue}

\usepackage{float}
\usepackage[caption=false]{subfig}
\usepackage{amsmath}
\usepackage{algorithm}
\usepackage{algpseudocode}
\usepackage{ifthen}
\usepackage{enumitem}
\usepackage{cases}
\usepackage{mathtools}
\usepackage{amssymb}
\usepackage{graphicx}
\usepackage{empheq}
\usepackage{tikz}
\usetikzlibrary{shapes.geometric, fit}
\usetikzlibrary{shapes}
\usetikzlibrary{arrows}
\usetikzlibrary{calc,positioning}
\usepackage{bm}
\usepackage{dsfont}
\usepackage{macros}
\usepackage{mathrsfs}
\usepackage{adjustbox}
\usepackage{booktabs}
\usepackage{multirow}
\usepackage{fix-cm}
\DeclareFontShape{OT1}{cmr}{bx}{sc}{<->cmcsc10}{}

\begin{document}

\RUNTITLE{Can We Validate Counterfactual Estimations in the Presence of Unknown Network Interference?}

\TITLE{Can We Validate Counterfactual Estimations in the Presence of General Network Interference?}
  
\ARTICLEAUTHORS{%
\AUTHOR{Sadegh Shirani$^1$, Yuwei Luo$^1$, William Overman$^1$, Ruoxuan Xiong$^2$, and Mohsen Bayati$^1$}
\AFF{$^1$Stanford University, $^2$Emory University} %
}
\ABSTRACT{%
Randomized experiments have become a cornerstone of evidence-based decision-making in contexts ranging from online platforms and transportation networks to public health and social programs. However, in experimental settings with network interference, a unit's treatment can influence outcomes of other units, challenging both causal effect estimation and its validation. Classic validation approaches fail as outcomes are only observable under a single treatment scenario and exhibit complex correlation patterns due to interference. To address these challenges, we introduce a framework that facilitates the use of machine learning tools for both estimation and validation in causal inference. Central to our approach is the new \textit{distribution-preserving network bootstrap}, a theoretically-grounded technique that generates multiple statistically-valid subpopulations from a single experiment's data. This amplification of experimental samples enables our second contribution: a counterfactual cross-validation procedure. This procedure adapts the principles of model validation to the unique constraints of causal settings, providing a rigorous, data-driven method for selecting and evaluating estimators. We extend recent causal message-passing developments by incorporating heterogeneous unit-level characteristics and varying local interactions, ensuring reliable finite-sample performance through non-asymptotic analysis. Additionally, we develop and publicly release a comprehensive benchmark toolbox featuring diverse experimental environments, from networks of interacting AI agents to opinion formation in real-world communities and ride-sharing applications.\footnote{Code and environments available at \url{https://github.com/CausalMP/CausalMP.git}} These environments provide known ground truth values while maintaining realistic complexities, enabling systematic evaluation of causal inference methods. Extensive testing across these environments demonstrates our method's robustness to diverse forms of network interference. Our work provides researchers with both a practical estimation framework and a standardized platform for testing future methodological developments.}


\maketitle

\vspace{-1.2cm}
\input{Introduction.tex}
\input{Problem_Setup.tex}
\input{Main_results}
\input{Numerical}
\input{Results.tex}

\input{Conclusion.tex}

\bibliography{mypaper}
\bibliographystyle{apalike}

\begin{APPENDICES}
\renewcommand{\theHsection}{A\arabic{section}}
\input{APPENDICES.tex}
\end{APPENDICES}

\end{document}

%% file: Introduction.tex
\section{Introduction}
\label{sec:Intro}
Randomized Controlled Trials (RCTs) are the gold standard for evaluating intervention effectiveness \citep{imbens2015causal}, such as assessing the impact of a public health initiative like deworming program to reduce parasitic infections in school-aged children. However, classic RCT methods often overlook the complex dynamics of disease transmission within communities. In these settings, experimental units, for example, individual children, are influenced by the infection status of their peers, family members, and broader community networks. A study of deworming interventions by \citet{miguel2004} demonstrated that treating some children for parasitic worms also influenced other children in neighboring schools through reduced environmental contamination and transmission pathways.

This phenomenon, known as network interference, violates the Stable Unit Treatment Value Assumption (SUTVA), a key principle of causal inference requiring that one unit's treatment assignment not affect other units' outcomes. When SUTVA is violated, the conceptual boundary between treatment and control groups dissolves, a methodological challenge found across many scientific domains. In public health, one person's vaccination alters infection risk for others \citep{hudgens2008toward};
in economics, housing subsidies reshape neighborhood dynamics \citep{sobel2006randomized}; in healthcare, behavioral changes spread through social ties \citep{valente2012network};
in two-sided marketplaces, interventions targeting one side of the market create spillover effects on the other side \citep{johari2022experimental};
and in dynamic settings, current treatment effects depend on the history of past interventions.
This pervasiveness raises a fundamental question: ``How can we efficiently estimate and validate the causal effect of an intervention within a network of interacting units?" \citep{ugander2013graph,eckles2016design,abaluck2022impact,farias2022markovian,farias2023correcting,ogburn2024causal,imbens2024causal,johari2024does}.

\begin{figure}
    \centering
    \includegraphics[width=0.95\linewidth]{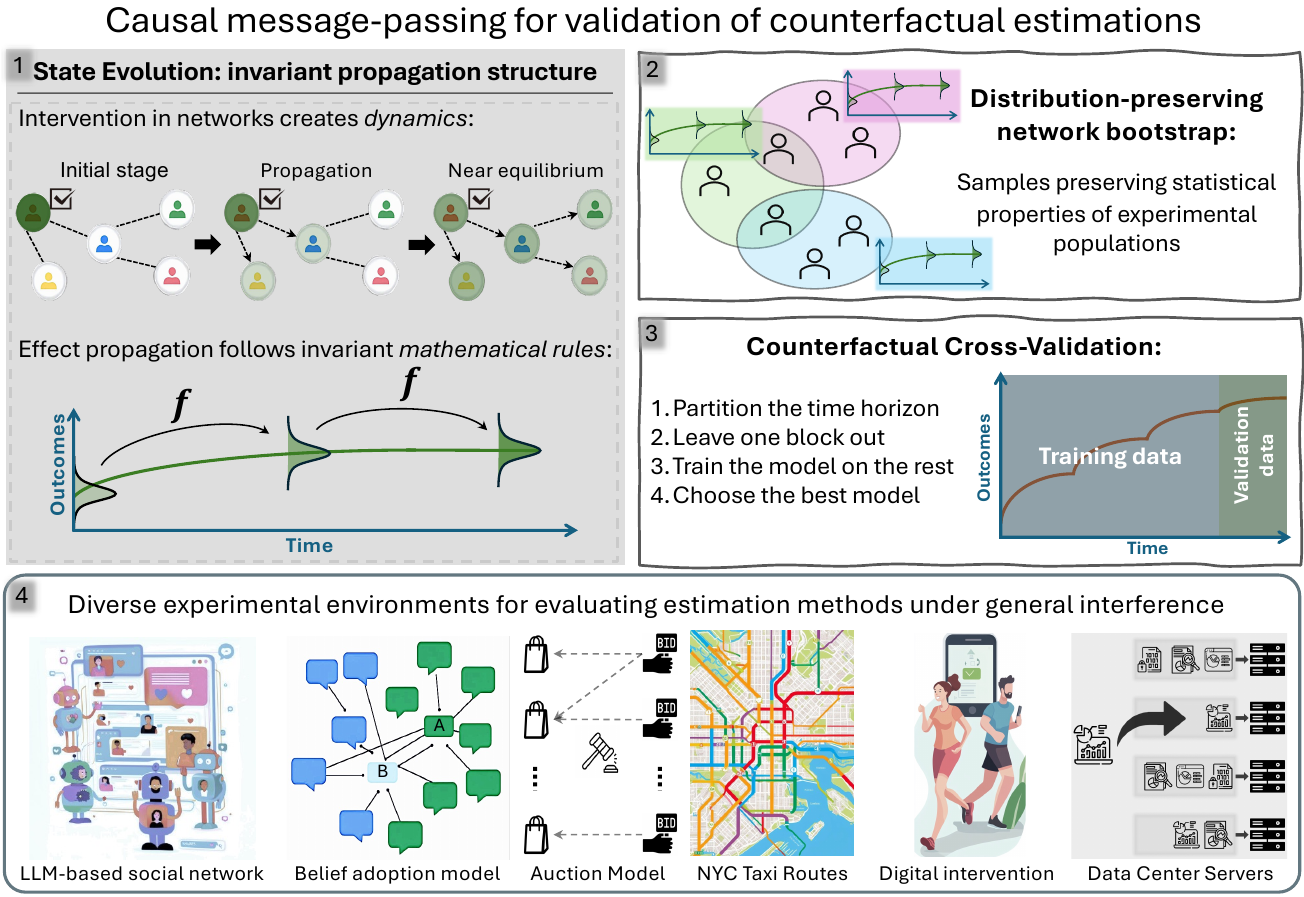}
    \caption{Validating Counterfactual Estimations: (1)~Network interventions evolve through a propagation phase, governed by invariant mathematical rules known as state evolution equations. (2)~Distribution-preserving network bootstrap generates different samples that retain the statistical properties of the original experimental population. (3) Counterfactual cross-validation partitions the time horizon to train and validate estimation models. (4) Six experimental environments spanning diverse domains and network structures to test estimation methods.}
    \label{fig:Figure_One}
\end{figure}

To articulate the challenges in the estimation and validation of causal effects, we first introduce a formal notation and visualization. Consider a randomized experiment conducted over time stamps $t = 0, 1, \ldots, T$, involving a population of $\UN$ units (e.g., individual children in the deworming study), indexed by $i = 1, \ldots, \UN$. At each time $t$, unit~$i$ receives a treatment (e.g., the medication) according to a random variable $\treatment{i}{t}$. In the deworming example, these treatment variables are binary: 1 indicates assignment to the treatment group, and 0 indicates assignment to the control group.

\begin{figure}
    \centering
    \includegraphics[width=0.5\linewidth]{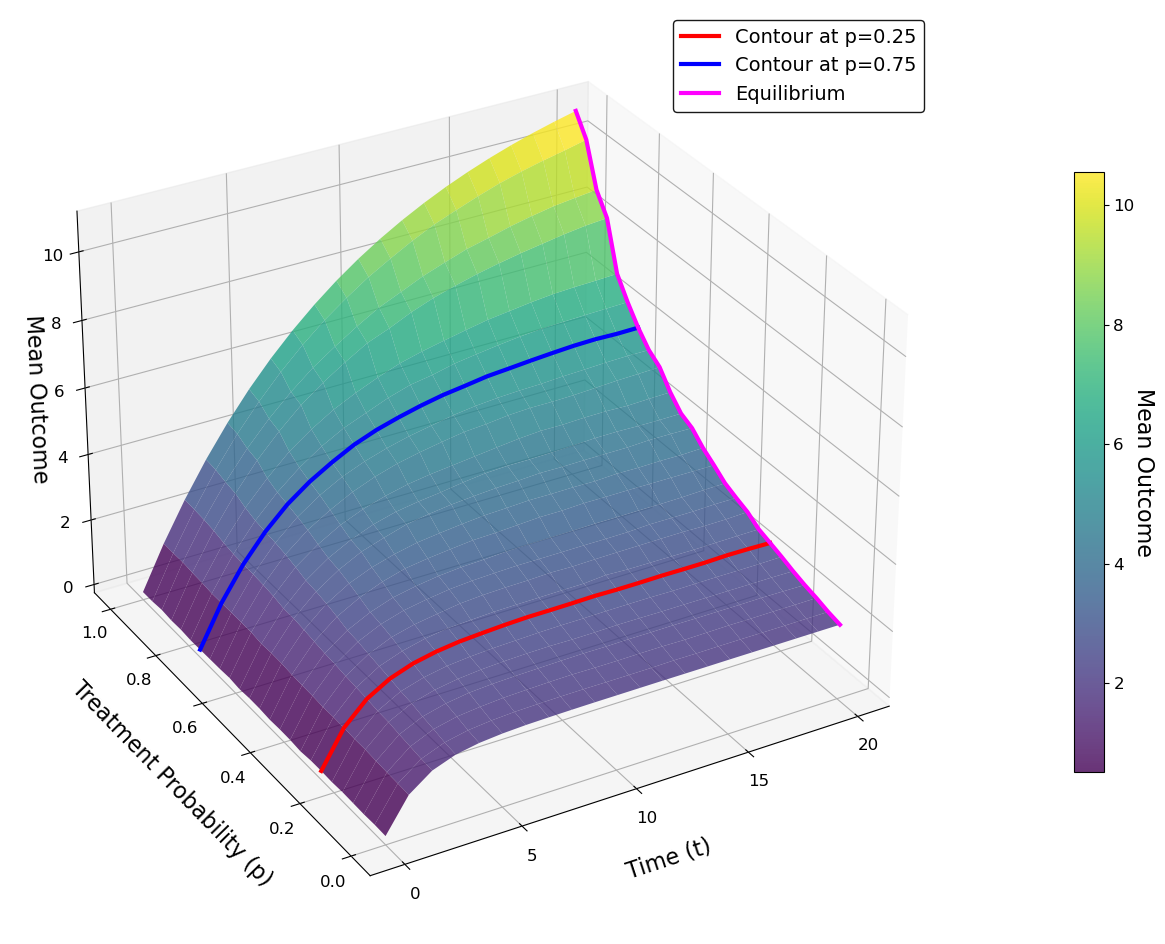}
    \caption{Evolution of outcomes sample mean (z-axis) with respect to time $t$ and treatment probability $p$. Red and blue contours highlight the counterfactual evolutions at treatment probabilities $p=0.25$ and $p=0.75$, respectively. The magenta line represents the equilibrium state, where the treatment effect has stabilized.}
    \label{fig:SE_surface}
\end{figure}

The treatment allocation across the entire population and time frame is represented by an $\UN \times (T+1)$ matrix, denoted by $\Mtreatment{}{}$. Following the Rubin causal framework \citep{imbens2015causal}, we denote by $\outcomeD{}{i}{t}(\Mtreatment{}{})$ the potential outcome of unit $i$ at time $t$. The counterfactual evolution, denoted by $\CFE{\Mtreatment{}{}}{}{t}$, represents the sequence of sample means of potential outcomes under the treatment assignment $\Mtreatment{}{}$ over time\footnote{Typically, outcomes are ``non-anticipating,'' meaning $\CFE{\Mtreatment{}{}}{}{t}$ and $\outcomeDW{\Mtreatment{}{}}{i}{t}$ depend only on treatments up to time~$t$. Our notation and theoretical results allow for more general scenarios where outcomes may be influenced by future treatments. However, for ease of reading, readers may assume the simpler case of non-anticipation where only the first $t+1$ columns of $\Mtreatment{}{}$ impact $\outcomeDW{\Mtreatment{}{}}{i}{t}$. Moreover, the first column of $\Mtreatment{}{}$ is by convention equal to all zeros (or no-treatment state), and the first column of $\Moutcome{}{}{}$ corresponds to outcomes in the all-control state.}:
\begin{align}
    \label{eq:sample_mean_outcomes}
    \CFE{\Mtreatment{}{}}{}{t} :=
    \frac{1}{N} \sum_{i=1}^\UN \outcomeDW{\Mtreatment{}{}}{i}{t},
    \quad\quad\quad
    t = 0,1, \ldots, T.
\end{align}
With $\OMtreatment{}{}$ being a specific realization of $\Mtreatment{}{}$, let $\Moutcome{}{}{}(\Mtreatment{}{} = \OMtreatment{}{})$ denote an $N \times (T+1)$ matrix that collectively represents the observed outcomes under the treatment allocation $\OMtreatment{}{}$.  For example, if per-unit treatment probability at all time periods is equal to $\expr$, the sample mean $\CFE{\Mtreatment{}{}}{}{t}$ is a random variable whose randomness arises from the treatment allocation $\Mtreatment{}{}$ and potentially from inherent stochasticity in the outcomes themselves. However, for a large $N$, this sample mean converges to its expectation. This deterministic evolution is visualized as the smooth surface in Figure~\ref{fig:SE_surface}, where each contour (at a fixed $\expr$) represents the expected dynamic trajectory.\footnote{Note that treatment probabilities can vary over time, making the space of possible counterfactual evolutions $\CFE{\Mtreatment{}{} = \OMtreatment{}{}}{}{t}$ considerably more complex than the simplified visualization shown in Figure~\ref{fig:SE_surface}.}

Our objective is to estimate expected counterfactual evolutions $\CFE{\OMtreatment{}{}'}{}{t}$ under alternative treatment allocations $\OMtreatment{}{}' \neq \OMtreatment{}{}$ and validate the accuracy of these estimations. This enables estimation of treatment effects over time through comparing counterfactual evolutions under different treatment conditions; for example, we can contrast evolution under delivering the medication to everyone against the status quo control. However, this objective faces a fundamental challenge: we can observe the system only under a single scenario, reflecting ``the fundamental problem of causal inference'' \citep{holland1986statistics}. Specifically, with treatment probability fixed at $\expr$, we observe only the corresponding contour in the surface shown in Figure~\ref{fig:SE_surface}. This limitation constrains both our ability to estimate counterfactual evolutions and validate such estimates, as we cannot directly observe outcomes under alternative scenarios.

To address the estimation challenge, we develop our approach by drawing an analogy between a network of units exchanging causal effects and a molecular system exchanging energy. In molecular systems, energy perturbations trigger redistribution according to invariant physical laws governing molecular interactions. Similarly, interventions propagate through network connections following consistent mechanical or behavioral principles. This analogy shows an important insight: \emph{the invariance of treatment propagation mechanisms}. Just as the equations governing molecular energy redistribution maintain their structural form, the mechanisms driving the propagation of the intervention effects preserve their fundamental structure both over time and across treatment scenarios (Part 1 in Figure~\ref{fig:Figure_One}). 
This structural invariance gives us an opportunity: by observing system evolution over time, we can employ machine learning to learn the functions characterizing how intervention effects propagate and aggregate into counterfactual evolutions.

To address the validation challenge, a natural candidate is cross-validation \citep{hastie2009elements} whose theoretical foundations \citep{stone1974cross,allen1974relationship} build upon Fisher's insight about the necessity of randomization for validation \citep{fisher1935design}. However, the application of cross-validation in causal inference has been notably limited by data scarcity. Using observations at an equilibrium reached at large~$t$, depicted by the ``equilibrium'' curve in Figure~\ref{fig:SE_surface} \citep{basse2019randomization,jackson2020adjusting,li2022random}, may not provide sufficient samples for reliable validation. Even using temporal pre-equilibrium data is insufficient, given that extending experiment duration is often prohibitively costly \citep{holtz2020reducing,cooprider2023science,xiong2024optimal}. To overcome this limitation, we introduce a new \emph{\batching{}} method (\batchingAcronym{}) that generates sufficient data to enable systematic validation of our estimation methods (Parts 2 and 3 in Figure~\ref{fig:Figure_One}).

By combining the invariant propagation principle with \batchingAcronym{}, this work advances our ability to address both \emph{estimation} and \emph{validation} challenges. Specifically, this work makes four main contributions. First, we extend the Causal Message-Passing \citep{shirani2024causal} by incorporating unit- and network-level heterogeneities, capturing more general outcome evolution equations. Second, we develop non-asymptotic analyses of these dynamics using techniques from \cite{li2022non}. This provides a principled way to study how two aspects of the interference network, its intensity and its heterogeneity, affect standard estimators such as difference-in-means. Specifically, we show that intensity drives the bias, while heterogeneity drives the variance. This analysis further leads to our \batchingAcronym{} method for generating and analyzing multiple subpopulations from the experimental population. This approach provides sufficient observations at each time period to enable accurate estimation through machine learning techniques. Third, we combine temporal data with our \batchingAcronym{} method to develop a cross-validation framework for counterfactual estimation that implements data-driven model selection. Applications of \batchingAcronym{} can go beyond causal inference, particularly in generating representative subsamples in various network problems without knowledge of the underlying network structure.

Fourth, this work contributes to addressing a fundamental challenge in causal inference: rigorously evaluating estimation methods when ground-truth counterfactuals are typically unobservable. We develop a benchmark toolbox with six different experimental environments that capture different forms of network interference and temporal dynamics. These environments span applications from social network influence to data center load balancing and ride-sharing systems. A notable highlight is our social network environment, which simulates a social media platform using thousands of AI agents interacting through their local network connections via content feeds. Each environment is carefully constructed to maintain known ground truth values while incorporating realistic complexities such as time-varying trends and heterogeneous network effects. We use this toolbox to extensively evaluate our method and make these environments publicly available to facilitate standardized testing of future methodologies.

The rest of the paper is organized as follows. Section \ref{sec:other-lit} reviews related literature on causal inference under network interference, approximate message passing, and network sampling. Section~\ref{sec:Causal_Estimands} presents our generalization of the Causal Message-Passing framework \citep{shirani2024causal}. Section \ref{sec:DPNB} develops our finite sample analysis of network interference dynamics, which leads to our \batching{} method for generating sufficient data to learn invariant aggregate dynamics. Section \ref{sec:C-CV} presents our cross-validation framework for counterfactual estimation. Sections \ref{sec:Benchmark_Toolbox} and \ref{sec:results} describe our benchmark toolbox and empirical validation results. Section \ref{sec:conclusion} concludes with discussion and future directions. Technical proofs and supplementary results appear in the appendices.

\section{Related Literature}
\label{sec:other-lit}

Recent literature on causal inference under network interference has evolved along several dimensions. One common approach addresses settings with known network clusters \citep{sobel2006randomized,rosenbaum2007interference,hudgens2008toward,tchetgen2012causal,auerbach2021local,su2021model,viviano2023causal,jia2024clustered,eichhorn2024low}, where researchers randomly assign entire clusters to different treatment intensities. This cluster-based design helps contain interference within cluster boundaries, though it requires prior knowledge of network clusters. Another significant development employs ``exposure mappings'' \citep{manski2013identification,aronow2017estimating,leung2022causal,harshaw2023design,savje2024causal}, which quantify how a unit's treatment effect varies based on its neighbors' treatment status.

For scenarios with unknown interference patterns, researchers have proposed alternative strategies using multi-period observations or design-based experiments \citep{yu2022estimating,cortez2022staggered,hu2022switchback}. A particularly promising direction has emerged through application-specific approaches that develop customized interference models for various contexts such as marketplace dynamics \citep{holtz2020reducing,bajari2021multiple,wager2021experimenting,munro2021treatment,johari2022experimental,bright2022reducing}. These domain-specific solutions offer tailored frameworks for managing interference in their respective experimental settings (e.g., by leveraging the two-sided structure of the marketplace or assuming interference is mediated by market price). Our work contributes to this literature by developing methods that do not require knowledge of the interference structure or domain-specific context while still capturing complex network effects across different applications.

Our methodology is supported by rigorous theoretical results derived from Approximate Message-Passing (AMP), whose foundations trace back to  \cite{thouless1977solution,mezard1987spin,kabashima2003cdma,mezard2009information,donoho2009message} and were formally established by \cite{bolthausen2014iterative}, \cite{bayati2011dynamics} and a large body of recent literature \citep{javanmard2013state,bayati2015universality,berthier2020state,chen2020universality,xinyi2021approximate,wang2022universality,dudeja2023universality,rush2018finite,li2022non}.
AMP traditionally addresses high-dimensional estimation problems through iterative updates with known matrices. Our approach is different and builds on the work of \cite{shirani2024causal}; specifically, we assume the matrix is unknown and develop techniques to analyze observed outcomes and infer causal~effects.

Our work on \batching{} relates to a rich body of research on network sampling and statistical inference in networked settings. The challenge of generating representative samples from network data has been important in network analysis, due to the inherent dependencies in network structures \citep{wormald1999models,newman2018networks}. Network inference methods have evolved from basic configuration models \citep{bollobas1980probabilistic} to sophisticated approaches handling non-independent observations \citep{snijders1999non,bickel2009nonparametric} and robust bootstrapping techniques \citep{bhattacharyya2015subsampling,green2022bootstrapping}. Another line of research focuses on the efficient algorithmic generation of random graphs that satisfy specific structural constraints, which is helpful for bootstrapping or creating null models \citep{milo2002network}. A central problem in this area is generating graphs with a fixed degree sequence. Classic approaches, such as Markov chain Monte Carlo methods \citep{jerrum1989approximate} or switch-based algorithms \citep{mckay1990uniform}, can produce asymptotically uniform samples but are often computationally intensive, with runtimes that are high-degree polynomials. To address this, faster sequential construction algorithms have been developed, including those based on sequential importance sampling \citep{chen2005sequential,blitzstein2010sequential} and other dynamic generation techniques \citep{steger1999generating,kim2004sandwiching,bayati2010sequential,bayati2018generating}. More recent advances also encompass general null models for complex network structures \citep{karrer2011stochastic,bianconi2018multilayer} and machine learning approaches using graph neural networks \citep{kipf2017semisupervised,hamilton2020graph}. Our approach differs fundamentally from this literature as it operates solely on node-level outcomes observed over time, without requiring access to the underlying network structure.

%% file: Problem_Setup.tex
\section{General Counterfactual Estimation Problem}
\label{sec:Causal_Estimands}
Estimating counterfactual evolution $\left\{ \CFE{\OMtreatment{}{}'}{}{t} \right\}_{t=0}^T$ based on the observed outcomes $\Moutcome{}{}{}(\Mtreatment{}{} = \OMtreatment{}{})$ enables addressing a broad range of causal questions, such as, “\emph{What if we had delivered the treatments according to $\OMtreatment{}{}'$ instead of $\OMtreatment{}{}$}?" For example, delivering the deworming medication to 20\% of children, we may want to explore how the population would have responded if 40\% had received the medication. In addition, by estimating the dynamics of the counterfactuals over time, we gain insights into how the treatment effect may strengthen or weaken as time progresses.

The total treatment effect (TTE)\footnote{Also known as the global treatment effect (GTE).} provides a formal way to compare any two counterfactual scenarios. For two treatment allocations $\OMtreatment{}{}'$ and $\OMtreatment{}{}''$, the TTE measures the difference in population average outcomes:
\begin{align}
    \label{eq:TTE_def}
    \TTE{t}{\OMtreatment{}{}'',\OMtreatment{}{}'} = \CFE{\OMtreatment{}{}''}{}{t} - \CFE{\OMtreatment{}{}'}{}{t},
    \quad\quad\quad
    t = 0,1, \ldots, T\,.
\end{align}
The literature typically focuses on a special case: when all entries of $\OMtreatment{}{}''$ equal one and all entries of $\OMtreatment{}{}'$ equal zero, with all outcomes at equilibrium \citep{candogan2023correlated,ni2023design,ugander2023randomized}. However, this scenario of treating the entire population versus treating no one may be impractical in certain settings. Our goal is to estimate any desired counterfactual evolution $\CFE{\OMtreatment{}{}'}{}{t}$ over the experimental horizon $t=0,1, \ldots, T$, as specified by Eq.~\eqref{eq:sample_mean_outcomes}. Such estimates enable the calculation of TTEs between arbitrary treatment levels and provide decision-makers with more realistic comparisons \citep{muralidharan2017experimentation,egger2022general}.

\subsection{Experimental Design Framework}
\label{sec:Experimental_Design}

We formalize the concept of experimental design by its properties in the large-population limit. Consider the $\UN \times (T+1)$ treatment matrix $\Mtreatment{}{}$ with entries $\treatment{i}{t}$. Each row, $\vec{\treatment{i}{}} := (\treatment{i}{0}, \dots, \treatment{i}{T})$, represents the full history of treatment assignments for unit $i$. We assume that as $N \to \infty$, the empirical distribution of these $N$ row vectors converges weakly to a well-defined probability distribution. We define this limiting distribution, denoted by $\expd_T$, as the \emph{experimental design}.

In many applications, treatments are binary (e.g., $\treatment{i}{t} \in \{0,1\}$), in which case the support of $\expd_T$ is $\{0,1\}^{T+1}$. If the randomization is performed such that the proportion of units treated at time~$t$ approaches $\expr_t$, then the marginal expectation of the $t^{th}$ coordinate of a random vector drawn from $\expd_T$ is precisely $\expr_t$. While we use this Bernoulli case for illustration throughout the main text, our framework accommodates more general scenarios, as detailed in Remark~\ref{rem:general_design}.
\begin{remark}
\label{rem:general_design}
Our framework generalizes beyond the binary case. In Appendix~\ref{sec:Technical_Results}, we allow for integer 
(e.g., different treatment types) or continuous-valued (e.g., varying treatment doses) treatments, a setting more general than much of the current literature (see, e.g., the survey by \cite{arkhangelsky2023causal}).
\end{remark}

Importantly, this definition of the experimental design through its limiting distribution is agnostic to the dependence structure across units or time periods. The assignments $\treatment{i}{t}$ and $\treatment{i'}{t'}$ may be arbitrarily correlated for any two distinct units $i, i'$ and any time periods $t, t'$. This generality encompasses a broad range of experimental designs, including staggered roll-out design \citep{xiong2024optimal}, micro-randomized trials \citep{li2022network}, and switchback experiments \citep{hu2022switchback,bojinov2023design,jia2024clustered}.

For a fixed integer $\dcovar$, we can also consider covariates (might be observed or not) in the form of a $\dcovar$ by $N$ matrix $\covar$, where each column (denoted by $\Vcovar{i} := (\Ecovar{1}{i}, \ldots,  \Ecovar{\dcovar}{i})^\top$) represents characteristics of unit $i$ (e.g., age and gender). The experimental data thus consists of treatment assignments~$\OMtreatment{}{}$, observed outcomes 
$
\Moutcome{}{}{}(\OMtreatment{}{}) := \Moutcome{}{}{}(\Mtreatment{}{}=\OMtreatment{}{})\,,
$
and covariates $\covar$. Therefore, we observe outcomes only under one specific treatment allocation $\OMtreatment{}{}$. This represents a single realization among $2^{N(T+1)}$ distinct potential outcomes, where this number grows exponentially with both population size and time horizon. Consequently, estimating causal effects under general interference becomes impossible due to non-identifiability issues \citep{karwa2018systematic}. To overcome this challenge, we propose a tractable outcome specification that aligns with and extends the causal message-passing framework \citep{shirani2024causal}.

\subsection{Potential Outcome Specification}
\label{sec:Outcome_Specification}
Let $\Mtreatment{t}{} := [\Vtreatment{}{0}{}| \ldots| \Vtreatment{}{t}{}]$ denote the treatment assignments up to time $t$; we represent by $\VoutcomeD{}{}{t}(\Mtreatment{t}{}) = \big(\outcomeD{}{1}{t}(\Mtreatment{t}{}), \ldots, \outcomeD{}{N}{t}(\Mtreatment{t}{})\big)^\top$ the potential outcome vector at time $t$. Consider unknown functions $\outcomeg{t}{}$ and $\outcomeh{t}{}$ which operate component-wise, and the expressions $\outcomeg{t}{}\big(\VoutcomeD{}{}{t}(\Mtreatment{t}{}), \Vtreatment{}{t+1}{}, \covar\big)$ and $\outcomeh{t}{}\big(\VoutcomeD{}{}{t}(\Mtreatment{t}{}), \Vtreatment{}{t+1}{}, \covar \big)$ represent the corresponding column vectors. Given $\VoutcomeD{}{}{0}(\Mtreatment{0}{})$ as the initial outcome vector, we consider the following specification for potential outcomes for $t=0,1,\ldots,T-1$:
\begin{align}
    \label{eq:outcome_function_matrix}
    \VoutcomeD{}{}{t+1}(\Mtreatment{t+1}{}) =
    \big(\IM+\IMatT{t}\big)\outcomeg{t}{}\left(\VoutcomeD{}{}{t}(\Mtreatment{t}{}) ,\Vtreatment{}{t+1}{}, \covar\right)
    +
    \outcomeh{t}{}\left(\VoutcomeD{}{}{t}(\Mtreatment{t}{}) ,\Vtreatment{}{t+1}{}, \covar\right)
    +
    \Vnoise{}{t},
\end{align}
where $\IM$ and $\IMatT{t}$ are $N\times N$ unknown matrices capturing the fixed and time-dependent interference effects, respectively. Additionally, $\Vnoise{}{t} = \big(\noise{1}{t},\ldots,\noise{N}{t}\big)^\top$ is the zero-mean random variable  accounting for observation noise. Denoting by $\IMatl{ij}$ and $\IMatTl{ij}{t}$ the element in the $i^{th}$ row and $j^{th}$ column of $\IM$ and $\IMatT{t}$, respectively, the value $\IMatl{ij}+\IMatTl{ij}{t}$ quantifies the impact of unit $j$ on unit $i$ at time $t$.

The specification in Eq. \eqref{eq:outcome_function_matrix} captures several aspects of experimental data. It accommodates various types of interference, including treatment spillover effects, carryover effects, peer effects, and autocorrelation; Figure~1 of \citet{shirani2024causal} illustrates these effects in a simple example. Notably, this specification accounts for outcome dynamics by acknowledging the temporal interrelation of units' outcomes. This contrasts with existing approaches to panel data analysis, which assume that time labels can be shuffled without affecting causal effects \citep{arkhangelsky2023causal}.

\begin{remark}
In Appendix~\ref{sec:Technical_Results}, we further generalize the outcome specification in Eq.~\eqref{eq:outcome_function_matrix} to incorporate additional lag terms (e.g., $\VoutcomeD{}{}{t-1}(\Mtreatment{t-1}{})$) in the functions $\outcomeg{t}{}$ and $\outcomeh{t}{}$, the complete treatment matrix at any time point (allowing for anticipation effects), and time-dependent covariates.
\end{remark}

Next, we analyze the state evolution of the experimental population, characterizing the asymptotic dynamics of unit outcomes that will provide theoretical foundations for developing robust counterfactual estimators in subsequent sections.

\subsection{Experimental State Evolution}
\label{sec:ESE}
In this section, we analyze the distribution of unit outcomes over time. This analysis requires the following two assumptions about the interference matrices.
\begin{assumption}[Gaussian interference structure]
    \label{asmp:Gaussian Interference Matrice}
    For all $i,j$, the element $\IMatl{ij}$ in the $i^{th}$ row and $j^{th}$ column of $\IM$ is an independent Gaussian random variable with mean $\mu^{ij}/N$ and variance $\sigma^2/N$. Similarly, $\IMatTl{ij}{t}$, the element in the $i^{th}$ row and $j^{th}$ column of $\IMatT{t}$, is an independent Gaussian random variable with mean $\mu^{ij}_t/N$ and variance $\sigma^2_t/N$.
\end{assumption}
\begin{assumption}[Convergent interference pattern]
    \label{asmp:Stable Interference Pattern}
    For all unit $i$ and any time $t$, the elements of vector $(\mu^{i1}, \ldots, \mu^{iN})$ admit a weak limit\footnote{This means that the empirical distribution of $\{\mu^{i1}, \ldots, \mu^{iN}\}$ converges to a probability distribution as $N$ increases.} and, separately, the elements of vector $(\mu^{i1}_t, \ldots, \mu^{iN}_t)$ admit a weak limit, where both limits are invariant in $i$. We later relax this $i$-invariance in Appendix \ref{apndx:batch_state_evolution}-Assumption \ref{asmp:weak_limits}.
\end{assumption}
According to Assumptions~\ref{asmp:Gaussian Interference Matrice} and \ref{asmp:Stable Interference Pattern}, the impact of unit $j$ on unit $i$ is captured by $\mu^{ij}$, $\mu^{ij}_t$, and two centered Gaussian random variables. This generalizes the model of \cite{shirani2024causal}, which assumes i.i.d. interference matrix elements across all units. This extension accommodates more heterogeneous local interactions and varying levels of uncertainty about the interference structure. Specifically, a fully known interference network has exact values for $\mu^{ij}$ and $\mu^{ij}_t$ with $\sigma^2 = 0$ and $\sigma^2_t = 0$, while a completely unknown interference means no knowledge of these quantities. Importantly, our estimation method is designed to handle the cases that we have \emph{no} knowledge of these underlying quantities for implementation.
\begin{example}
    Consider the case where $\mu^{ij}$ and $\mu^{ij}_t$ take values of $0$ and $1$, generating time-dependent adjacency matrices of the graph representing the interference structure. If we have high confidence that interactions occur only through these adjacency matrices, we can imagine negligible values for $\sigma^2$ and $\sigma^2_t$. Conversely, significant uncertainty about potential interactions would be reflected in larger values of $\sigma^2$ and $\sigma^2_t$. 
\end{example}
\begin{remark}
    The formal version of Assumption~\ref{asmp:Stable Interference Pattern} appears in Appendix~\ref{apndx:batch_state_evolution}-Assumption~\ref{asmp:weak_limits}, where we generalize the condition by allowing greater variation across units.
    Furthermore, while this interference model assumes Gaussian distributions conditional on $\mu^{ij}$ and $\mu^{ij}_t$, the appendix of \cite{shirani2024causal} discusses how this conditional Gaussianity assumption can be relaxed to accommodate more general distributional forms.
\end{remark}

We next present the following informal result for the large-sample regime that characterizes outcome dynamics between consecutive periods, with a formal statement presented in Theorem~\ref{thm:BSE_informal}.
\subsubsection*{State evolution - informal statement.}
    Let $N \to \infty$ and suppose Assumptions~\ref{asmp:Gaussian Interference Matrice} and \ref{asmp:Stable Interference Pattern} hold. There exist mappings $\outcomef{t}$, which depend on $\IM, \IMatT{t}, \outcomeg{t}{}, \outcomeh{t}{}, \covar$ and $\Vnoise{}{t}$, $t=0, \ldots, T-1$, such that the 
    distribution of outcomes at time $t+1$ is determined by:
    \begin{equation}
        \label{eq:SE_informal}
        \outcomeD{}{}{t+1}\stackrel{d}{=}
        \outcomef{t}\left(
        \outcomeD{}{}{t}
        ,   
        \treatment{}{t+1}{}
        \right),
    \end{equation}
    where $\outcomeD{}{}{t}$ denotes the weak limit of outcomes $\outcomeD{}{1}{t}(\Mtreatment{t}{}), \ldots, \outcomeD{}{N}{t}(\Mtreatment{t}{})$, $\treatment{}{t+1}{}$ follows a Bernoulli distribution with mean $\expr_{t+1}$, and $\stackrel{d}{=}$ refers to equality in distribution.

The mapping $\outcomef{t}$ in Eq. \eqref{eq:SE_informal} characterizes how aggregate outcomes evolve over time. Specifically, in the large-sample limit, even though each unit's outcome depends on a complex network of interactions, the population-level distribution of outcomes follows a simpler evolution that depends only on the previous distribution and treatment assignment distribution. Building on \cite{shirani2024causal}, while extending their results to a broader setting, we refer to the relationship in Eq.~\eqref{eq:SE_informal} as the experimental state evolution (SE) equation.

\subsection{General Estimation Theory}
\label{sec:gen_estimation}
\begin{figure}
    \centering
    \includegraphics[width=1\linewidth]{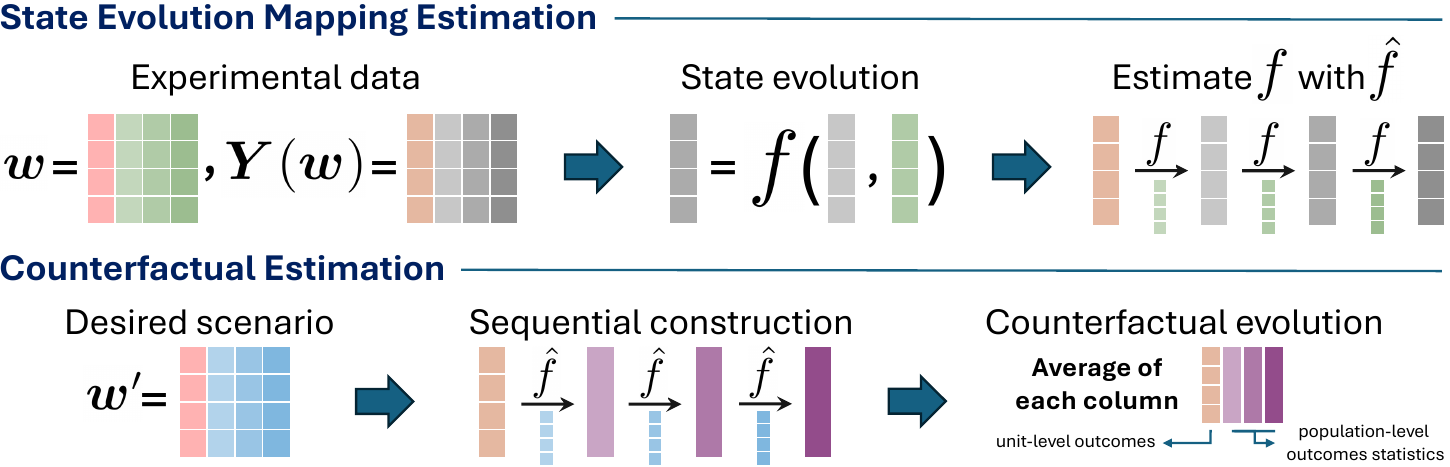}
    \caption{Estimation strategy: Experimental data is used to estimate state evolution mappings through supervised learning, which are then applied recursively to generate desired counterfactual evolutions. Treatment allocations $\bm{w}$ and $\bm{w'}$ share identical initial columns, serving as initialization for our recursive approach. While observations are collected at the unit level, the estimated counterfactuals represent population-level quantities.}
    \label{fig:estimation_flow}
\end{figure}

The state evolution equation~\eqref{eq:SE_informal} motivates a natural estimation strategy. In large experimental populations, the empirical distributions of observed outcomes at each time period approximate the theoretical distributions specified in Eq. \eqref{eq:SE_informal}. 
In other words, and loosely speaking, we can obtain approximations for $\outcomeD{}{}{t+1}$, $\outcomeD{}{}{t}$, and $\treatment{}{t+1}{}$, from experimental data. Leveraging this approximation, we can employ supervised learning methods to estimate the state evolution mappings $\outcomef{t}$. Once estimated, these mappings enable recursive generation of any desired counterfactual evolution $\left\{ \CFE{\OMtreatment{}{}'}{}{t} \right\}_{t=0}^T$ by repeated application of Eq. \eqref{eq:SE_informal}. Our counterfactual estimation problem therefore reduces to consistently estimating $\outcomef{t}$. We present this result below with an informal statement, while providing the rigorous theorem and proof in Theorem~\ref{thm:consistency} of Appendix~\ref{sec:estimation_theory}.
\begin{theorem}[Consistency - informal statement]
\label{thm:consistency_informal}
Suppose that functions $\outcomeg{t}{}$ and $\outcomeh{t}{}$ are affine in their first argument.\footnote{There exist functions $\outcomeg{t}{1}, \outcomeg{t}{2}, \outcomeh{t}{1}$ and $\outcomeh{t}{2}$ such that $\outcomeg{t}{}(y, \cdot) = y \outcomeg{t}{1}(\cdot) + \outcomeg{t}{2}(\cdot)$ as well as $\outcomeh{t}{}(y, \cdot) = y \outcomeh{t}{1}(\cdot) + \outcomeh{t}{2}(\cdot)$.}
Under certain regularity conditions, given consistent estimation of the mappings $\outcomef{t}$ in the state evolution equation~\eqref{eq:SE_informal}, any desired counterfactual evolution can be consistently estimated. Specifically, we can construct an estimator $\ECF{}{t}{\OMtreatment{}{}'}$ such that as $N \rightarrow \infty$, for all $t$:
$$ \ECF{}{t}{\OMtreatment{}{}'} - \CFE{\OMtreatment{}{}'}{}{t} \xrightarrow{P} 0,$$
where the convergence is in probability. When the estimation of the mappings $\outcomef{t}$ achieves strong consistency, this convergence can be strengthened to almost sure convergence.
\end{theorem}
Theorem~\ref{thm:consistency_informal} reframes the counterfactual estimation problem as a supervised learning task, and our immediate goal is to estimate the functions $\outcomef{t}$ from observed data. However, we observe only a single experimental scenario with interference-induced dependence; consequently, successful generalization relies on balancing the flexibility of the hypothesis class for functions $\outcomef{t}$ against the available sample size. Inspired by the invariance of treatment propagation mechanisms over time, we pool statistical strength across time and adopt Assumption~\ref{asmp:SE_decomposition}. Specifically, this assumption requires that the underlying interaction mechanisms remain stable over the experimentation horizon. In other words, we can decompose $\outcomef{t}$ into a substantial time-invariant component that captures the core causal mechanisms and a smaller time-varying component that accounts for temporal variations. Nevertheless, as we discuss in Section~\ref{subsec:practical-limits-pop-level}, Assumption~\ref{asmp:SE_decomposition} alone may not suffice under data scarcity; Sections~\ref{sec:DPNB} and \ref{sec:C-CV} propose remedies.
\begin{assumption}[Weakly additive time trend]
    \label{asmp:SE_decomposition}
    Considering the state evolution equation \eqref{eq:SE_informal}, there exist mappings $\outcomef{}$ and $\Toutcomef{t}$ such that for each $t$:
    \begin{align}
        \label{eq:SE_decomposition}
        \outcomef{t}\left(
        \outcomeD{}{}{t}
        ,
        \treatment{}{t+1}{}
        \right)
        =
        \outcomef{}\left(
        \outcomeD{}{}{t}
        ,
        \treatment{}{t+1}{}
        \right)
        +
        \Toutcomef{t}\left(
        \outcomeD{}{}{t}
        \right)
    \end{align}
\end{assumption}
Under Assumption~\ref{asmp:SE_decomposition}, our primary objective becomes estimating the time-invariant mapping~$\outcomef{}$, which captures the essential treatment effects according to Eq. \eqref{eq:SE_decomposition}. When temporal variations are minimal (e.g., when $\Toutcomef{t}$ is a random function and $\E[\Toutcomef{t}\left(\outcomeD{}{}{t}\right)]=0$), we can safely ignore the time-dependent component $\Toutcomef{t}$ and directly estimate $\outcomef{}$ from the available data (Figure~\ref{fig:estimation_flow}). To build intuition, we present a simplified version of this approach below, adapted from \cite{shirani2024causal}. Conversely, when strong temporal patterns are present, we first estimate and remove these trends by modeling $\Toutcomef{t}$, thereby isolating the time-invariant component $\outcomef{}$. We can then apply supervised learning techniques to estimate this core mapping. We defer the technical details of this more complex case to Appendix~\ref{sec:preprocessing}.

\subsubsection{An Illustrative Example}
\label{sec:example}
Consider a special case of outcome specification \eqref{eq:outcome_function_matrix} where functions $\outcomeh{t}{}$ and $\outcomeg{t}{}$ satisfy,
\begin{equation}
\label{eq:simple_function_structure}
    \begin{aligned}
        \outcomeg{t}{} \big(
        \outcomeD{}{i}{t}(\Mtreatment{t}{}),
        \treatment{i}{t+1}, \Vcovar{i}
        \big)
        =
        \ABE_t
        +
        \ACE_t \outcomeD{}{i}{t}(\Mtreatment{t}{})
        +
        \ADE_t \treatment{i}{t+1},
        \quad%
        \outcomeh{t}{} \big(
        \outcomeD{}{i}{t}(\Mtreatment{t}{}),
        \treatment{i}{t+1}, \Vcovar{i}
        \big)
        =
        \APE_t \outcomeD{}{i}{t}(\Mtreatment{t}{}) \treatment{i}{t+1},
    \end{aligned}
\end{equation}
where $\ABE_t, \ACE_t, \ADE_t,$ and $\APE_t$ are unknown random variables independent of everything else. 

\begin{proposition}
    \label{prp:estimation_example}
    Suppose that Eq. \eqref{eq:simple_function_structure} holds and for each unit~$i$ and time~$t$, and the empirical distributions of $(\mu^{i1}, \ldots, \mu^{iN})$ and 
    $(\mu^{i1}_t, \ldots, \mu^{iN}_t)$
    converge weakly to the distributions of random variables $\MIM{}{}$ and $\MIM{t}{}$, respectively, as $N \rightarrow \infty$. Additionally, assume the following holds for all $t$:
    \begin{align}
    \label{eq:simple_example_quantities}
        \ABE := \E[(\MIM{}{}+\MIM{t}{}) \ABE_t],
        \quad
        \ACE := \E[(\MIM{}{}+\MIM{t}{}) \ACE_t],
        \quad
        \ADE := \E[(\MIM{}{}+\MIM{t}{}) \ADE_t],
        \quad
        \APE := \E[(\MIM{}{}+\MIM{t}{}) \APE_t].
    \end{align}
   Then, the sample mean of outcomes $\sum_{i=1}^N\outcomeD{}{i}{t}(\Mtreatment{t}{})/N$ converges almost surely to a deterministic limit as $N \to \infty$. Denoting this limit by $\AVO{}{}{t}$, it satisfies the following dynamics:
    \begin{align}
        \label{eq:simple_SE_function}
        \AVO{}{}{t+1}
        =\outcomef{}(\AVO{}{}{t},\expr_{t+1})\,,
    \end{align}
    where $\outcomef{}(a,b)=\ABE + \ACE a + \ADE b +
    \APE ab$, that plays the role of the time-invariant function characterizing the evolution of outcomes over time. Moreover, if $\OMtreatment{}{}$ and $\OMtreatment{}{}'$ share identical columns for treatment assignment at time~$t=0$ and the set of marginal expectations of $\expd_T$, $\{\expr_0, \ldots, \expr_T\}$, contains at least two distinct values, Algorithm~\ref{alg:example_algorithm} provides consistent estimator for counterfactual evolutions.
\end{proposition}

Proposition~\ref{prp:estimation_example} exemplifies our estimation strategy in a simple setting and presents the type of assumptions required to demonstrate consistency results. Specifically, according to Eq. \eqref{eq:simple_function_structure} and Proposition~\ref{prp:estimation_example}, although we allow for varying interaction mechanisms across units and time, stability at the population level (see Eq.~\ref{eq:simple_example_quantities}) ensures time-invariant evolution mechanisms as given in Eq.~\eqref{eq:simple_SE_function}.
We present a detailed analysis of a more general setting in Appendix~\ref{sec:application_to_BRD}.

\begin{algorithm}
\caption{Causal message passing counterfactual estimator (simple case)}
\label{alg:example_algorithm}
\begin{algorithmic}
\Require $\Moutcome{}{}{}(\OMtreatment{}{}), \OMtreatment{}{} = [\Otreatment{i}{t}{}]_{i,t}$, and $\OMtreatment{}{}' = [\Otreatment{'i}{t}{}]_{i,t}$

\State \hspace{-1.3em} \textbf{Step 1: Data processing}
\For{$t = 0, \ldots, T$}
\State $\HAVO{}{}{t} \gets \frac{1}{N} \sum_{i=1}^N \outcomeD{}{i}{t} (\OMtreatment{t}{})$, \quad\quad $\Oexpr{}{}{t} \gets \frac{1}{N} \sum_{i=1}^N \Otreatment{i}{t}{}$, \quad\quad $\Dexpr{}{}{t} \gets \frac{1}{N} \sum_{i=1}^N \Otreatment{'i}{t}{}$
\EndFor

\State \hspace{-1.3em} \textbf{Step 2: Parameters estimation}
\State $(\EABE, \EACE, \EADE, \EAPE)$ $\gets$ Estimation of $(\ABE, \ACE, \ADE, \APE)$ using OLS in $\HAVO{}{}{t+1} = \ABE + \ACE \HAVO{}{}{t} + \ADE \Oexpr{}{}{t+1} + \APE \HAVO{}{}{t} \Oexpr{}{}{t+1}$

\State \hspace{-1.3em} \textbf{Step 3: Counterfactual estimation}

\State $\ECF{}{0}{\OMtreatment{}{}'} \gets \HAVO{}{}{0}$

\For{$t = 1, \ldots, T$}
    \State $\ECF{}{t}{\OMtreatment{}{}'} \gets \EABE + \EACE \ECF{}{t-1}{\OMtreatment{}{}'} + \EADE \Dexpr{}{}{t} + \EAPE \ECF{}{t-1}{\OMtreatment{}{}'} \Dexpr{}{}{t}$
\EndFor

\Ensure Estimated counterfactual evolution: $\{\ECF{}{t}{\OMtreatment{}{}'}\}_{t=0}^T$.
\end{algorithmic}
\end{algorithm}

\subsection{Practical Limitations}\label{subsec:practical-limits-pop-level} 
The effectiveness of approaches like Algorithm~\ref{alg:example_algorithm} in identifying treatment effects across various settings has been shown empirically by \cite{shirani2024causal}. However, two significant constraints require consideration. First, the estimation procedure in the second step of Algorithm~\ref{alg:example_algorithm} faces inherent sample size limitations that are bounded by the temporal horizon of the experiment. Second, as demonstrated by \cite{bayati2024higher}, certain settings require accounting for more complex structural relationships, necessitating a richer functional class for the outcome specification. Furthermore, estimation validity can be compromised by potential model misspecification \citep{karwa2018systematic}, highlighting the need for robust validation techniques.

These considerations raise two questions: First, how can we effectively expand the sample size to enhance estimation accuracy? Second, how can we determine the appropriate model specification for the underlying experiment, analogous to model selection procedures in supervised learning? The following two sections aim at addressing these questions.

%% file: Main_results.tex
\section{Distribution-preserving Network Bootstrap}
\label{sec:DPNB}
Our estimation strategy is grounded in the state evolution equation \eqref{eq:SE_informal}, as illustrated in Figure~\ref{fig:estimation_flow}. In this section, we enhance our estimation capabilities by developing a resampling method that generates additional samples from the state evolution equation. The key observation underlying our approach is that the SE equation captures the dynamics of outcome distributions as described by Eq. \eqref{eq:SE_informal}. Therefore, different samples should provide different outcome distributions, but these distributions must evolve according to the same state evolution mappings $\outcomef{t}$.

Our methodology batches experimental units into subpopulations, which we denote by $\batch$. Each subpopulation $\batch$ comprises a subset of units selected from $\{1, \ldots, N\}$, where membership is determined exclusively by the treatment allocation $\Mtreatment{}{}$. Under the randomized treatment assumption, which specifies that $\Mtreatment{}{}$ is random and independent of all other variables, each subpopulation $\batch$ constitutes a random sample from the experimental population. This ensures sufficient similarity between distinct subpopulations to expect them to evolve under the same mappings. However, we must ensure that each particular configuration of $\batch$ yields a unique observation of the SE equation. Consequently, we require that the empirical treatment distributions exhibit sufficient variation across subpopulations. This variation guarantees that different subpopulations provide genuinely distinct information about the underlying evolutionary dynamics.

We develop our \batching{} methodology through a sequence of theoretical results that build toward our main practical proposal. Specifically, we begin by extending the state evolution framework to characterize the dynamics of subpopulation-level quantities. We then introduce a finite-sample decomposition rule. This rule is central to our argument, as it mathematically grounds a fundamental bias-variance trade-off: it reveals how the unobserved mean and heterogeneity of the interference structure drive estimation bias and variance, respectively. This insight highlights the difficulty of determining an optimal estimation strategy `a priori' and motivates the primary contribution of this section: a pragmatic framework that leverages \batchingAcronym{} as a tool to generate sufficient data for robust model selection via cross-validation. Finally, we show how this framework, developed for the general case of an unknown network, can be readily adapted to incorporate partial network knowledge for improved accuracy.

\begin{remark}
The theoretical frameworks presented here extend to arbitrary subpopulations. While our analysis focuses specifically on treatment-allocation-based subpopulations in this section, the appendices provide complete theoretical statements applicable to generic subpopulations.
\end{remark}

\subsection{Subpopulation-specific State Evolution}
\label{sec:BSE}
We begin by extending the result of Eq.~\eqref{eq:SE_informal} to subpopulations of experimental units.
\begin{theorem}
    \label{thm:BSE_informal}
Consider a subpopulation $\batch$ whose size grows to infinity as $N \to \infty$. Under Assumptions~\ref{asmp:Gaussian Interference Matrice} and \ref{asmp:Stable Interference Pattern} and certain regularity conditions, for $t= 0, 1, \ldots, T-1$, we have
\begin{equation}
    \label{eq:state_evolution_mainbody}
    \begin{aligned}
        \Houtcome{\batch}{}{0} &= \outcomeh{0}{}\big(\outcomeD{}{\batch}{0}, \treatment{\batch}{1}, \Vcovar{}\big) + \noise{}{0},
        \\
        \MAVO{}{}{1} &=
        \E\left[ (\MIM{}{}+\MIM{0}{})
        \outcomeg{0}{}\big(\outcomeD{}{}{0}, \treatment{}{1}, \Vcovar{}\big)\right],
        \\
        (\MVVO{}{}{1})^2 &=
        (\sigma^2 + \sigma_0^2)
        \E\left[
        \outcomeg{0}{}\big(\outcomeD{}{}{0}, \treatment{}{1}, \Vcovar{}\big)^2\right],
        \\
        \Houtcome{\batch}{}{t} &= \outcomeh{t}{}\big(\MAVO{}{}{t} + \MVVO{}{}{t} Z_t + \Houtcome{\batch}{}{t-1}, \treatment{\batch}{t+1}, \Vcovar{}\big) + \noise{}{t},
        \\
        \MAVO{}{}{t+1} &=
        \E\left[ (\MIM{}{}+\MIM{t}{})
        \outcomeg{t}{}\big(\MAVO{}{}{t} + \MVVO{}{}{t} Z_t + \Houtcome{}{}{t-1}, \treatment{}{t+1}, \Vcovar{}\big)
        \right],
        \\
        (\MVVO{}{}{t+1})^2 &=
        (\sigma^2+\sigma_t^2) \E\left[
        \outcomeg{t}{} \big(\MAVO{}{}{t} + \MVVO{}{}{t} Z_t + \Houtcome{}{}{t-1}, \treatment{}{t+1}, \Vcovar{}\big)^2
        \right],
    \end{aligned}
\end{equation}
where 
\begin{itemize}
    \item $\outcomeD{}{}{0}$ and $\outcomeD{}{\batch}{0}$ represent the weak limits of the population and subpopulation initial outcomes;

    \item $\treatment{}{t}$ and $\treatment{\batch}{t}$ are the weak limits of the population and subpopulation treatment assignments;

    \item $\Vcovar{}$ represents the weak limit of the population covariates;

    \item $\MIM{}{}$ and $\MIM{t}{}$ represent the weak limits of $(\mu^{i1}, \ldots, \mu^{iN})$ and $(\mu^{i1}_t, \ldots, \mu^{iN}_t)$, respectively;

    \item $\noise{}{t}$ is the weak limit of noise terms $(\noise{1}{t}, \ldots, \noise{N}{t})$;

    \item $Z_t$ follows a standard Gaussian distribution;

    \item $\Houtcome{}{}{t}$ denotes the population-level counterpart of $\Houtcome{\batch}{}{t}$, when $\batch$ is equal to the entire population.
\end{itemize}
Furthermore, if we denote by $\outcomeD{}{\batch}{t}$ the weak limit of outcomes for units belonging to~$\batch$ as $N \rightarrow \infty$, then $\outcomeD{}{\batch}{t}$ follows the same distribution as $\MAVO{}{}{t} + \MVVO{}{}{t} Z_t + \Houtcome{\batch}{}{t-1}$. Similarly, denoting by $\outcomeD{}{}{t}$ the weak limit of outcomes at the population level, $\outcomeD{}{}{t}$ follows the same distribution as $\MAVO{}{}{t} + \MVVO{}{}{t} Z_t + \Houtcome{}{}{t-1}$.
\end{theorem}
We can express the state evolution equation \eqref{eq:state_evolution_mainbody} compactly using mappings $\outcomefb{t},\; t=0, \ldots, T-1$:
\begin{equation}
    \label{eq:BSE_informal}
    \outcomeD{}{\batch}{t+1}
    \stackrel{d}{=}
    \outcomefb{t}
    \left(
    \outcomeD{}{}{t}
    ,
    \treatment{}{t+1}{}
    ,
    \outcomeD{}{\batch}{t}
    ,
    \treatment{\batch}{t+1}{}
    \right),
\end{equation}
where $\stackrel{d}{=}$ refers to equality in distribution and each mapping $\outcomefb{t}$ involves the functions $\outcomeg{t}{}$ and $\outcomeh{t}{}$, the random objects $\MIM{}{}$, $\MIM{t}{}$, $Z_{t+1}$, $\Vcovar{}$ and $\noise{}{t}$, as well as the constants $\sigma$ and $\sigma_t$. Then, by Eq.~\eqref{eq:BSE_informal},
the outcomes of units within $\batch$ evolve through a dynamic interplay between population-level and subpopulation-specific outcomes and treatments. The formal and more general statement of Theorem~\ref{thm:BSE_informal} is presented in Theorem~\ref{thm:Batch_SE} of Appendix~\ref{apndx:batch_state_evolution}.

To demonstrate the efficacy of \batchingAcronym{}, consider two subpopulations $\batch_1$ and $\batch_2$ with distinct treatment allocations ($\treatment{\batch_1}{t+1}{} \neq \treatment{\batch_2}{t+1}{}$ in Eq. \eqref{eq:BSE_informal}). Then, two sequences $\outcomeD{}{\batch_1}{0}, \ldots, \outcomeD{}{\batch_1}{T}$ and $\outcomeD{}{\batch_2}{0}, \ldots, \outcomeD{}{\batch_2}{T}$ yield different samples of the state evolution described by \eqref{eq:BSE_informal}. Notably, for each time step $t$, the mapping $\outcomefb{t}$ shares fundamental structural properties with $\outcomef{t}$, with the experimental state evolution equation in \eqref{eq:SE_informal} emerging as a special case of Eq. \eqref{eq:BSE_informal} when applied to the entire population. This implies that by strategically selecting subpopulations with varying treatment histories, we obtain multiple state evolution trajectories, facilitating the estimation of $\outcomefb{t}$ and, consequently, $\outcomef{t}$.
\begin{example}
    \label{exmpl:BernoulliD_subpopulation}
    Consider two distinct subpopulations based on treatment assignment $\treatment{}{t+1}{}$: treated units ($\batch=\Tc$) and control units ($\batch=\Cc$). From Eq. \eqref{eq:BSE_informal}, we have:
    \begin{equation}
    \label{eq:BD_SP_SEs}
        \begin{aligned}
        \outcomeD{}{\Tc}{t+1}
        \stackrel{d}{=}
        \outcomefb{t}\left(
        \outcomeD{}{}{t}
        ,
        \treatment{}{t+1}{}
        ,
        \outcomeD{}{\Tc}{t}
        ,
        1
        \right),
        \quad\quad
        \outcomeD{}{\Cc}{t+1}
        \stackrel{d}{=}
        \outcomefb{t}\left(
        \outcomeD{}{}{t}
        ,
        \treatment{}{t+1}{}
        ,
        \outcomeD{}{\Cc}{t}
        ,
        0
        \right).
        \end{aligned}
    \end{equation}
    When the direct treatment effect is non-zero, the outcome distributions for the treated group $\outcomeD{}{\Tc}{t}$ and the control group $\outcomeD{}{\Cc}{t}$ are distinct. Nevertheless, by \eqref{eq:BD_SP_SEs}, both distributions evolve according to the same mappings $\outcomefb{t}$, thereby providing two different observations from the state evolution.
\end{example}

Example~\ref{exmpl:BernoulliD_subpopulation} demonstrates how \batchingAcronym{} enables creating multiple observations of the state evolution equation with different treatment intensity levels. Now, we show that \batchingAcronym{} indeed generates representative samples to effectively support robust estimation of state evolution mappings. To illustrate, we must first introduce a theoretical result, a finite-sample analysis that yields a structural decomposition of the outcome specification. This decomposition disentangles the network (or indirect) treatment effect from the direct effect and identifies an endogenous noise term whose magnitude reflects the heterogeneity of the underlying network structure. Then, we discuss how \batchingAcronym{} provides a way to reduce the impact of this endogenous noise through averaging, while simultaneously generating sufficient representative samples for reliable estimation.

\subsection{Outcomes Decomposition and Its Implications}
\label{sec:decomposition_implications}
Building on the finite-sample analysis of AMP algorithms \citep{li2022non}, we obtain the following decomposition rule for the outcome specification in Eq.~\eqref{eq:outcome_function_matrix}.
\begin{theorem}[Unit-level decomposition rule]
    \label{thm:outcome_decomposition}
    Suppose Assumption~\ref{asmp:Gaussian Interference Matrice} holds. For any unit~$i$ and fixed $t \in \{0, \ldots, T-1\}$, we have
    \begin{equation}
        \label{eq:outcome_decomposition_rule}
        \begin{aligned}
        \outcomeD{}{i}{t+1}(\Mtreatment{t+1}{}) =
        \;&\frac{1}{N} \sum_{j=1}^N \left(\mu^{ij}+\mu_t^{ij}\right) \outcomeg{t}{}\left(\outcomeD{}{j}{t}(\Mtreatment{t}{}), \treatment{j}{t+1}{}, \Vcovar{j} \right) +
        \outcomeh{t}{}\left(\outcomeD{}{i}{t}(\Mtreatment{t}{}), \treatment{i}{t+1}{}, \Vcovar{i} \right)
        \\
        &+
        \sqrt{\sigma^2+\sigma_t^2} \norm{\outcomeg{t}{}\left(\VoutcomeD{}{}{t}(\Mtreatment{t}{}) ,\Vtreatment{}{t+1}{}, \covar\right)} \sumvec{i}{t}  + \noise{i}{t},
        \end{aligned}
    \end{equation}
    where $\sumvec{i}{t}$ is a random variable such that $\Sumvec{}{t} := \left(\sumvec{1}{t}, \ldots, \sumvec{N}{t}\right)^\top = \sum_{i=0}^{t} \NPC{i}{t} \Vec{Z}_i$, with 
     $\Vec{Z}_0, \Vec{Z}_1, \ldots, \Vec{Z}_{T-1} \in \R^N$ are i.i.d. random vectors following $\Nc(0,\frac{1}{N}\I_N)$ distribution and
    $\VNPC{}{t} = (\NPC{0}{t}, \ldots, \NPC{t}{t}, 0, \ldots, 0)^\top \in \R^N$ denotes a random vector that is correlated with $\Vec{Z}_0, \Vec{Z}_1, \ldots, \Vec{Z}_{t}$ and satisfies $\normWO{\VNPC{}{t}} = 1$. Furthermore,
    \begin{equation*}
        \Wc_1\left(\law\left(\Sumvec{}{t}\right),\Nc\left(0,\frac{1}{N} \I\right)\right) \leq c \sqrt{\frac{t \log N}{N}},
    \end{equation*}
    where $c$ is a constant independent of $N$ and $t$, $\law(\Sumvec{}{t})$ denotes the probability distribution of~$\Sumvec{}{t}$, and $\Wc_1$ is Wasserstein-1 distance.
\end{theorem}

We now discuss several implications of Theorem~\ref{thm:outcome_decomposition}. For expositional clarity, we assume a time-invariant network structure and set $\mu^{ij}_t=0$ and $\sigma_t=0$ for all $t$, while supposing $\mu^{ij}\equiv \mu$ for all~$i,j$. Then, the only heterogeneity in the network structure arises from i.i.d.\ Gaussian random variables with variance $\sigma^2/N$, while $\mu$ reflects the average intensity of interactions (see Assumptions~\ref{asmp:Gaussian Interference Matrice}). Omitting the explicit outcome--treatment dependence in the notation, Eq.~\eqref{eq:outcome_decomposition_rule} reduces to
\begin{equation}
\label{eq:obs_simplified_decomp}
    \outcomeD{}{i}{t+1}
    =
    \underbrace{\mu \,\bar G_t}_{\text{global interference}}
    \;+\;
    \underbrace{\outcomeh{t}{}\big(\outcomeD{}{i}{t}, \treatment{i}{t+1}{}, \Vcovar{i}\big)}_{\text{unit/direct term}}
    \;+\;
    \underbrace{\sigma \,\norm{\outcomeg{t}{}\!\left(\VoutcomeD{}{}{t},\Vtreatment{}{t+1}{},\covar\right)} \sumvec{i}{t}}_{\text{network heterogeneity noise}}
    \;+\;
    \noise{i}{t},
\end{equation}
where $\displaystyle \bar G_t := \frac{1}{N}\sum_{j=1}^N \outcomeg{t}{}\!\big(\outcomeD{}{j}{t}, \treatment{j}{t+1}{}, \Vcovar{j}\big)$. By Theorem~\ref{thm:outcome_decomposition}, we can approximately view $\sumvec{i}{t}$ as an additive, mean-zero perturbation with variance of order $1/N$.
Consequently, the third term in \eqref{eq:obs_simplified_decomp} behaves like additional noise with variance
\begin{equation}
\label{eq:var_endogenous_noise}
    \frac{\sigma^2}{N}\,
    \norm{\outcomeg{t}{}\!\left(\VoutcomeD{}{}{t},\Vtreatment{}{t+1}{},\covar\right)}^2.
\end{equation}
Note that $\norm{\outcomeg{t}{}\!\left(\VoutcomeD{}{}{t}, \Vtreatment{}{t+1}{}, \covar\right)}^2$ also scales with $N$, implying that the term in Eq.~\eqref{eq:var_endogenous_noise} does not vanish as $N \to \infty$. This persistence of noise has direct implications for standard estimators such as difference-in-means (DM), which is defined at time~$t+1$ as follows:
\[
\widehat{\tau}^{\,\mathrm{DM}}_{t+1}
=
\frac{1}{n_\Tc}\sum_{i\in \Tc}\outcomeD{}{i}{t+1}
-
\frac{1}{n_\Cc}\sum_{i\in \Cc}\outcomeD{}{i}{t+1},
\]
with $\Tc$ and $\Cc$ representing the treatment and control groups, respectively.
In Eq.~\eqref{eq:obs_simplified_decomp}, the global interference term is equal to $\mu\,\bar G_t$ for every $i$ and cancels in the difference. Then, $\widehat{\tau}^{\,\mathrm{DM}}_{t+1}$ behaves ``as if'' there were no interference but with \textit{inflated noise}.
Using \eqref{eq:var_endogenous_noise} and the approximation $\sumvec{i}{t}\stackrel{\mathrm{approx}}{\sim} \Nc(0,1/N)$, the interference-induced variance inflation of $\widehat{\tau}^{\,\mathrm{DM}}_t$ is approximately
\begin{equation}
\label{eq:var_inflation_DM}
    \frac{\sigma^2}{N}\,
    \norm{\outcomeg{t}{}\!\left(\VoutcomeD{}{}{t},\Vtreatment{}{t+1}{},\covar\right)}^2
    \left(\frac{1}{n_\Tc}+\frac{1}{n_\Cc}\right),
\end{equation}
in addition to the variance contributed by observation noise $\noise{i}{t}$.
That means the heterogeneity of the (unknown) interference, captured by $\sigma$, acts as extra noise even when the estimator ignores interference. However, the price of ``ignoring'' $\mu\,\bar G_t$ is that $\widehat{\tau}^{\,\mathrm{DM}}_{t+1}$ incurs a bias in estimating any counterfactual $\CFE{\OMtreatment{}{}'}{}{t+1}$, depending on the magnitude of $\mu\,\bar G_t$ at treatment level $\OMtreatment{}{}'$.

These observations lead to the following takeaways. First, $\mu$ primarily drives \emph{bias}, whereas $\sigma$ primarily drives \emph{variance} through network heterogeneity noise. Second, structural knowledge of the average interaction term captured by $\mu$ enables us to estimate and subtract the first term in \eqref{eq:outcome_decomposition_rule} and reducing bias. However, this does not reduce the variance contribution in \eqref{eq:var_inflation_DM}, as it originates from $\sigma$ and the near-Gaussian perturbations~$\Sumvec{}{t}$. We illustrate these phenomena through a simple simulation that varies $\mu$ and $\sigma$, and then discuss how to address the bias component.

We consider a simple outcome specification where
$\outcomeh{t}{}\big(
\outcomeD{}{i}{t}(\Mtreatment{t}{}),
\treatment{i}{t+1}, \Vcovar{i}
\big)=1-1.2\treatment{i}{t+1}$
and
$\outcomeg{t}{}\big(
\outcomeD{}{i}{t}(\Mtreatment{t}{}),
\treatment{i}{t+1}, \Vcovar{i}
\big)=\treatment{i}{t+1}$.
The experiment runs for eight time periods with a population of $N=500$ units. The treatment assignment for each unit follows a Bernoulli distribution, with probability $0.25$ for the first four periods and $0.75$ for the remaining four periods. Finally, the standard deviation of the observation noise is set to $\sigma_\epsilon=0.1$.

\begin{figure}
\centering
\includegraphics[width=0.8\linewidth]{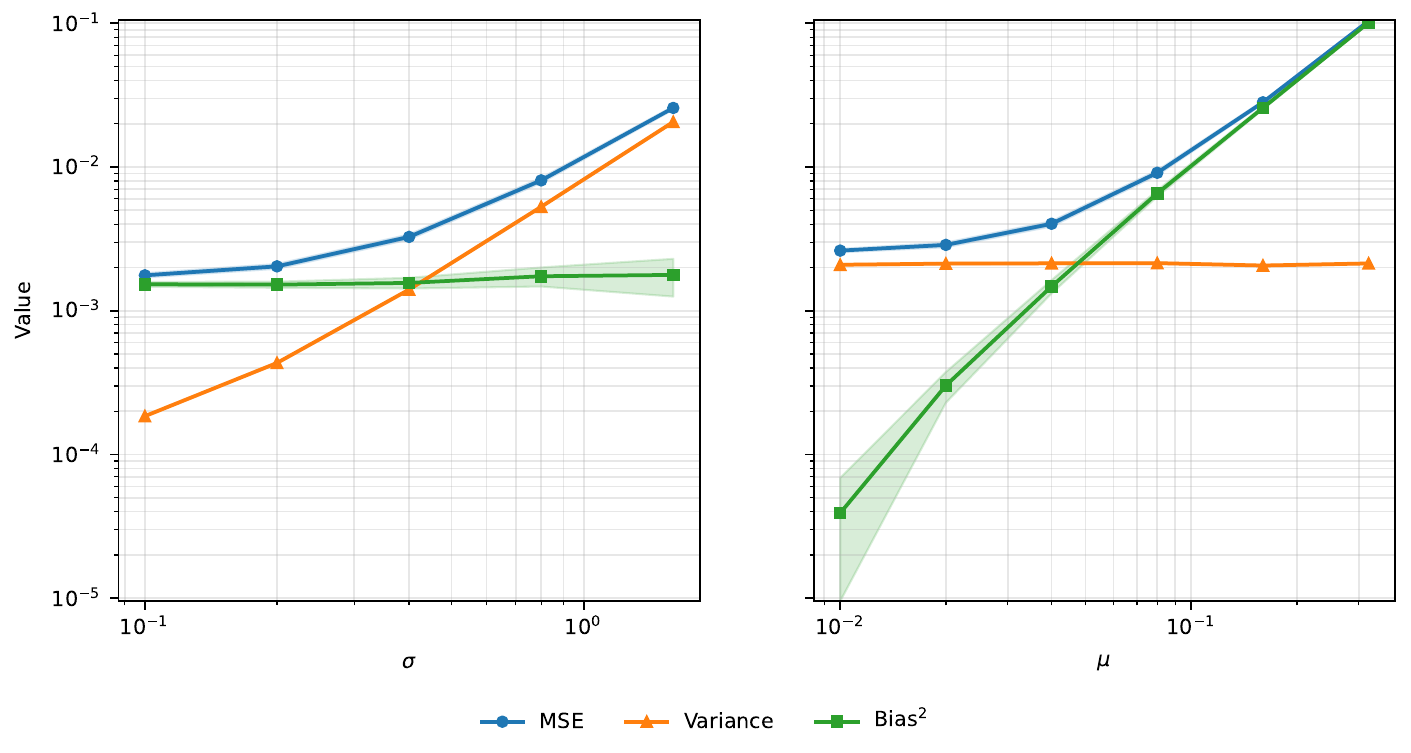}
\caption{Difference‑in‑Means (DM) performance under interference. Left: $\sigma$ sweep at fixed $\mu=0.04$. Right: $\mu$ sweep at fixed $\sigma=0.5$. Curves show MSE, variance, and squared bias with $\pm 1$ SE bands from a nested bootstrap. Axes are log–log.}
\label{fig:dm_simulation}
\end{figure}

We perform two sets of simulations on the effect of interference matrix parameters $\mu$ and~$\sigma$ and display the results in Figure~\ref{fig:dm_simulation}. In the first set (left panel), we fix $\mu=0.04$ and vary $\sigma \in \{0.1,,0.2,,0.4,,0.8,,1.6\}$. In the second set (right panel), we fix $\sigma=0.5$ and vary $\mu \in \{0.01,,0.02,,0.04,,0.08,,0.16,,0.32\}$. For each configuration, we compute the ground‑truth TTE and the DM estimate at the final time period. To quantify uncertainty and decompose error, we use a nested bootstrap: we simulate $100$ independent “worlds” (fresh interference matrix $A$ and initial outcomes $Y_0$) and $200$ resamples of treatment allocations $\Mtreatment{}{}$ per world; we then compute $\pm 1$ SE bands for Mean Squared Error (\MSE), variance, and squared bias of TTE estimation using $400$ nested bootstrap resamples that re‑sample worlds and runs. 

The simulation results in Figure~\ref{fig:dm_simulation} empirically validate our theoretical findings. The left panel shows that when $\mu$ is fixed and $\sigma$ increases, the variance of the DM estimator grows, while its squared bias remains constant. This demonstrates that network heterogeneity, parameterized by~$\sigma$, inflates the variance of simple estimators. In contrast, the right panel shows that when $\sigma$ is fixed and $\mu$ increases, the squared bias grows, dominating the MSE.

\subsection{Navigating Estimation Trade-offs with \batchingAcronym{}}
\label{sec:dpnb_framework}
The preceding analysis highlights the limitations of standard estimators such as DM. While variance inflation from network heterogeneity ($\sigma$) reduces statistical power, the bias from mean interference ($\mu$) poses a more fundamental problem that can lead to incorrect causal conclusions. Strategies like Algorithm~\ref{alg:example_algorithm} address this bias by modeling the unknown interference structure. However, such population-level approaches are limited by the sample size $T$. To utilize the full statistical power across all $N \times T$ observations, we must apply DPNB and develop a more sophisticated estimation framework. Below, our subpopulation decomposition rule provides the precise theoretical tool to understand the statistical trade-offs inherent in this task. It characterizes how averaging over subpopulations of different sizes affects the endogenous noise introduced by network heterogeneity.
\begin{corollary}[Subpopulation-level decomposition rule]
    \label{crl:SampleMean_decomposition}
    Under the assumptions of Theorem~\ref{thm:outcome_decomposition}, for any subpopulation $\batch$, we have
    \begin{equation}
        \label{eq:SampleMean_decomposition}
        \begin{aligned}
        \frac{1}{\cardinality{\batch}} \sum_{i \in \batch} \outcomeD{}{i}{t+1}(\Mtreatment{t+1}{})
        =
        \;&\frac{1}{N \cardinality{\batch}}
        \sum_{i \in \batch}
        \sum_{j=1}^N \left(\mu^{ij}+\mu_t^{ij}\right) \outcomeg{t}{}\left(\outcomeD{}{j}{t}(\Mtreatment{t}{}), \treatment{j}{t+1}{}, \Vcovar{j} \right)
        \\
        \;&+
        \frac{1}{\cardinality{\batch}} \sum_{i \in \batch}
        \outcomeh{t}{}\left(\outcomeD{}{i}{t}(\Mtreatment{t}{}), \treatment{i}{t+1}{}, \Vcovar{i} \right)
        \\
        \;&+
        \sqrt{ \frac{\sigma^2+\sigma_t^2}{{\cardinality{\batch}}}} \norm{\outcomeg{t}{}(\VoutcomeD{}{}{t}(\Mtreatment{t}{}) ,\Vtreatment{}{t+1}{}, \covar)} \avesumvec{}{t}
        +
        \frac{1}{\cardinality{\batch}}
        \sum_{i \in \batch} \noise{i}{t},
        \end{aligned}
    \end{equation}
    where $\cardinality{\batch}$ denotes the size of the subpopulation $\batch$ and $\avesumvec{}{t}$ is a random variable satisfying
    \begin{equation*}
        \Wc_1\left(\law\left(\avesumvec{}{t}\right),\Nc\left(0,\frac{1}{N} \right)\right) \leq c \sqrt{\frac{t \log N}{N}}.
    \end{equation*}
\end{corollary}
This corollary mathematically grounds a trade-off between two estimation extremes. The third term in the right-hand side of Eq.~\eqref{eq:SampleMean_decomposition} shows that the standard deviation of the endogenous network heterogeneity noise scales with $1/\sqrt{\cardinality{\batch}}$. 
At one extreme, a unit-level estimation (ULE) uses each unit as a subpopulation of size one ($\cardinality{\batch}=1$). This maximizes the number of observations ($N\times T$) but makes the model susceptible to the large endogenous noise term, which can bias estimation. At the other extreme, a population-level estimation (PLE) uses the entire population as a single batch ($\cardinality{\batch}=N$). This maximally suppresses the noise but leaves only a single time series ($T$ data points), limiting statistical power and increasing the risk of misspecification.

Our proposed \batching{} method provides a flexible framework that interpolates between these two extremes. By generating multiple samples from subpopulation averages, it aims to increase the effective sample size beyond $T$ while still leveraging the noise-reducing benefits of outcome aggregations.

However, the choice of batch configuration itself introduces a nuanced bias-variance trade-off. Larger batches behave more like PLE. They more effectively average out the endogenous noise, reducing estimation bias. But this comes at the cost of generating fewer, less variable batches, which can limit statistical power and make it difficult to identify complex functional forms. Smaller batches provide more samples and greater cross-sectional variation, which is helpful for learning the underlying dynamics and enhancing statistical power. But, these smaller averages are more susceptible to the endogenous noise, potentially reintroducing the bias that impacts unit-level~models.

A formal characterization of the optimal batching strategy is a significant challenge, as it depends on features of the experimental environment that are typically unknown, such as the relative magnitudes of mean interference ($\mu$) and network heterogeneity ($\sigma$). While rigorous theoretical analysis of these trade-offs represents a promising direction for future research, we take a data-driven approach in the next section. We embrace the complexity of this trade-off and use \batchingAcronym{} to generate multiple observational trajectories from a single experiment, enabling our cross-validation framework with more samples of the state evolution equation. This framework forms the core of our proposed methodology for settings with unknown networks. Before detailing this framework, we briefly discuss how our approach can be adapted when partial network information is available.

\subsection{Incorporating Partial Network Knowledge}
Settings where the network structure is entirely unknown represent a challenging scenario common in many experiments. However, when partial knowledge of the network is available, this information should be leveraged to improve estimation accuracy. Eq.~\eqref{eq:outcome_decomposition_rule} provides a decomposition of unit outcomes where the first term captures the interference effect. We can adjust this term to reflect any available knowledge about the interference network, as illustrated in the following example.
\begin{example}[Known clusters]
    Consider a scenario where the network structure is partially known with clusters $C^1, \ldots, C^K$ (e.g., the experimental population consists of individuals from different villages or online communities). In this context, the interference term in Eq.~\eqref{eq:outcome_decomposition_rule} can be rewritten to distinguish between within-cluster and between-cluster effects:
    \begin{equation*}
        \begin{aligned}
        \outcomeD{}{i}{t+1}(\Mtreatment{t+1}{}) =
        \;&\frac{1}{N} \sum_{k=1}^K 
        \Bigg[
        \underbrace{\sum_{j \in C^k, i \notin C^k} \left(\mu^{ij}+\mu_t^{ij}\right) \outcomeg{t}{}\left(\outcomeD{}{j}{t}(\Mtreatment{t}{}), \treatment{j}{t+1}{}, \Vcovar{j} \right)}_{\text{Between-cluster interference}}
        \\
        \;&+
        \underbrace{\sum_{i,j \in C^k} \left(\mu^{ij}+\mu_t^{ij}\right) \outcomeg{t}{}\left(\outcomeD{}{j}{t}(\Mtreatment{t}{}), \treatment{j}{t+1}{}, \Vcovar{j} \right)}_{\text{Within-cluster interference}}
        \Bigg]
        \\
        \;&+
        \outcomeh{t}{}\left(\outcomeD{}{i}{t}(\Mtreatment{t}{}), \treatment{i}{t+1}{}, \Vcovar{i} \right)
        +
        \sqrt{\sigma^2+\sigma_t^2} \norm{\outcomeg{t}{}\left(\VoutcomeD{}{}{t}(\Mtreatment{t}{}) ,\Vtreatment{}{t+1}{}, \covar\right)} \sumvec{i}{t}  + \noise{i}{t},
        \end{aligned}
    \end{equation*}
    where we would typically expect the magnitude of $\left(\mu^{ij} + \mu_t^{ij}\right)$ to be larger when units $i$ and $j$ belong to the same cluster, reflecting stronger intra-cluster ties.
\end{example}

The reformulated decomposition in the example above has practical implications for estimation. Instead of relying on a single population-level average (as in PLE) or averages over arbitrary subpopulations (as in \batchingAcronym{}), one can now leverage the known cluster structure. By computing outcome and treatment averages for each cluster, we can generate a richer set of summary statistics to learn the state evolution dynamics. Ultimately, whether operating with no network knowledge or incorporating partial structural information, our proposed methodology provides the necessary data for robust estimation. The following sections detail the cross-validation framework that leverages this data and then demonstrate the performance of our framework empirically.

\section{Counterfactual Cross Validation}
\label{sec:C-CV}
The precision of estimated counterfactuals, as established in Theorem~\ref{thm:consistency_informal}, depends on accurately estimating the SE mappings $\outcomef{t}$. This estimation process faces two main challenges that require systematic validation:

\subsubsection*{Function approximation and model selection} 
The true specification of potential outcomes is often unknown, requiring us to approximate the SE mappings $\outcomef{t}$. This approximation involves two related aspects. First, we must choose appropriate summary statistics of the joint distribution of outcomes and treatments to serve as inputs to our model, an approach analogous to feature engineering in supervised learning. For example, \cite{shirani2024causal} used sample means of outcomes, treatments, and their products. While domain knowledge can guide this selection, we need a systematic way to validate these choices. Second, depending on the complexity of experimental settings, we may need to employ a range of estimation techniques, from simple linear regression to more sophisticated methods such as neural networks. The choice of technique and its specific implementation (e.g., architecture, hyperparameters) must be validated to prevent issues like model instability and estimation bias. These two aspects are interconnected, as both contribute to how well we can approximate the true SE mappings $\outcomef{t}$. A data-driven validation methodology helps us jointly optimize these choices while reducing misspecification risks.

\subsubsection*{Optimal batch configuration}
As established in \S\ref{sec:dpnb_framework}, \batchingAcronym{} offers an effective strategy to control endogenous noise by batching experimental units. However, the choice of batch size presents a trade-off. Larger batches better average out noise, but decrease the number of distinct batches available for learning. Conversely, smaller batches provide more samples for learning, but may retain more noise that potentially introduces bias. The number of batches creates a similar trade-off that interacts with batch size selection. Therefore, data-driven validation is essential to find the optimal combination of batch~size and number of batches that balances these competing effects within a given experimental context.

\subsection{Counterfactual Cross Validation Algorithm}
\label{sec:CCV_Algorithm}
To address these challenges, we propose a counterfactual cross-validation algorithm that divides the time horizon into a list of time blocks $\tblockList$, which serve as natural cross-validation folds. For each time block, we use all other blocks as training data and the held-out block as validation data. The training process involves generating batches as discussed above, with their size and number serving as tuning parameters. For validation, we use a fixed set of pre-specified validation batches $\batch_1^v, \ldots, \batch_{b_v}^v$ constructed as follows. We begin by computing each unit's average treatment exposure across the experimental horizon (e.g., a unit that is treated in 60\% of the time periods receives a treatment exposure of 0.6). We then rank the units in descending order based on their treatment exposure values and partition them into $b_v$ equal-sized groups, ensuring validation batches cover the full spectrum of treatment exposure. For each fold, the validation metric is averaged across these validation batches to provide a robust performance measure.

The parameters being selected through the cross-validation include both the choice of estimator from a candidate set $\estimatorList$ and the batching configuration from a candidate set $\batchList$. Each estimator $\estimator \in \estimatorList$ represents a specific combination of feature generation (what summary statistics to use) and estimation method (e.g., linear regression or neural network). The batching parameters $(\batchSize,\batchCount)\in\batchList$ specify the size $\batchSize$ and number $\batchCount$ of training batches (e.g., 100 batches each containing 20\% of units or 500 batches each containing 10\% of units). For each held-out time block, we train models using all possible combinations of these parameters on the remaining blocks and evaluate their performance on the validation batches obtained from the held-out block.

\begin{figure}
    \centering
    \includegraphics[width=\linewidth]{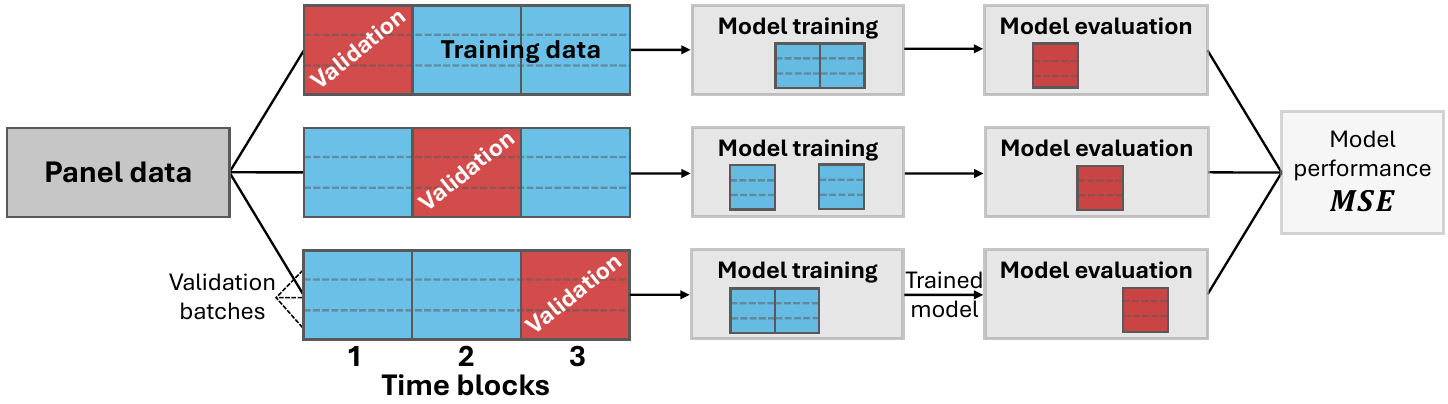}
    \caption{Counterfactual Cross-Validation. Time horizon is partitioned into blocks for leave-one-out validation. Models are trained on remaining blocks and evaluated via MSE to select optimal configurations.}
    \label{fig:CCV}
\end{figure}

\subsubsection*{Step 1: Reference ground truth construction}
After the experiment is completed, we obtain the treatment matrix $\OMtreatment{}{}$ and observed outcomes panel $\Moutcome{}{}{}(\OMtreatment{}{})$. 
We first compute the sample mean of outcomes over time as ground truth for evaluating estimations in subsequent steps. More precisely, for each time period $t\in\{0,1,\ldots,T\}$, and each validation batch $\batch_{j}^v$ with $j\in \{1,\ldots,b_v\}$, we calculate average of outcomes at time $t$ across units in $\batch_{j}^v$ and denote that by $\CFETest{t}{\batch_{j}^v}$.

\subsubsection*{Step 2: Leave-one-out and counterfactual estimation}
For each time block $\tblock\in\tblockList$, we construct training datasets $\outcomeTrain^{-\tblock}$ and $\treatmentTrain^{-\tblock}$ as submatrices of $\Moutcome{}{}{}(\OMtreatment{}{})$ and $\OMtreatment{}{}$, respectively, excluding columns within $\tblock$. We define $\treatmentTest^{\tblock}$ as the submatrix of $\OMtreatment{}{}$ containing only columns in $\tblock$. Each estimator $\estimator$ is then trained using $\outcomeTrain^{-\tblock}$ and $\treatmentTrain^{-\tblock}$, and used to estimate counterfactuals for treatment allocation $\treatmentTest^{\tblock}$ during the held-out time block $\tblock$.

\subsubsection*{Step 3: Optimal Estimator Selection}
Following the estimation across all time blocks, we identify the optimal estimator and batch parameters by comparing the results with observed counterfactuals using a predetermined loss function. For instance, using mean square error:
\begin{align*}
    \MSE_{\estimator,\batchSize,\batchCount}
    =
    \frac{1}{b_v (T+1)}
    \sum_{j=1}^{b_v}
    \sum_{t=0}^T
    \Big[\CFETest{t}{\batch_{j}^v} - 
    \ECFTrain{t}{\batch_{j}^v}{\estimator,\batchSize,\batchCount}
    \Big]^2.
\end{align*}

Algorithm~\ref{alg:C-CV} presents our counterfactual cross-validation procedure. To ensure consistency of the final selected estimator, we construct the initial estimator list $\estimatorList$ such that each candidate estimator is consistent under certain assumptions. Our cross-validation algorithm then optimizes estimator selection for real-world scenarios with finite experimental units.
For example, we can use variations of Algorithm~\ref{alg:example_algorithm} based on Ridge regression with different penalty terms as candidate estimators, with our cross-validation optimizing the Ridge penalty selection. Similarly, Algorithm~\ref{alg:FO-recursive} in Appendix~\ref{sec:estimation_theory} provides additional candidate estimators based on linear regression that extend Algorithm~\ref{alg:example_algorithm} by incorporating subpopulation-specific terms. In this case, the optimization process also involves selecting the optimal batch configuration.

When additional data are available (e.g., multiple clusters or pre-experiment observations), Algorithm~\ref{alg:C-CV} can be enhanced to leverage these datasets for more sophisticated training and validation.

\begin{algorithm}
\caption{Counterfactual cross-validation}\label{alg:C-CV}
\footnotesize
\begin{algorithmic}
\Require Treatment allocation $\OMtreatment{}{}$, observed outcomes $\Moutcome{}{}{}(\OMtreatment{}{})$, validation batches $\{\batch_j^v\}_{j=1}^{b_v}$, time blocks $\tblockList$, loss function $\criteria:\R^{(T+1)\times b_v}\times\R^{(T+1)\times b_v}\to\R_+$, and candidate estimators $\estimatorList$ and batch parameters $\batchList$
\State \hspace{-1.3em} \textbf{Step 1: Reference Ground Truth Construction}
\State $\big\{ \CFETest{t}{\batch_{j}^v} \big\}_{t = 0}^T \gets$ average of observed outcomes over time for units belonging to $\batch_{j}^v,\; j=1, \ldots, b_v$
\State \hspace{-1.3em} \textbf{Step 2: Leave-one-out and Counterfactual Estimation}
\For{$\estimator,\batchSize,\batchCount \in \estimatorList\times\batchList$}
    \For{$\tblock \in \tblockList$}
        \State $\outcomeTrain^{-\tblock} \gets$ columns of $\Moutcome{}{}{}(\OMtreatment{}{})$ outside of $\tblock$
        \State $\treatmentTrain^{-\tblock} \gets$ columns of $\OMtreatment{}{}$ outside of $\tblock$
        \State $\treatmentTest^{\tblock} \gets$ columns of $\OMtreatment{}{}$ in $\tblock$
        \State Train $\estimator$ with batching parameters  $(\batchSize,\batchCount)$ on data $\outcomeTrain^{-\tblock}$ and $\treatmentTrain^{-\tblock}$ 
        \For{$j \in \{1,\ldots,b_v\}$}
            \State $\big\{ \ECFTrain{t}{\batch_{j}^v}{\estimator,\batchSize,\batchCount} \big\}_{t \in \tblock} \gets$ estimate CFE for $\treatmentTest^{\tblock}$ via the trained model, for units in $\batch_j^v$
        \EndFor 
    \EndFor
\EndFor
\State \hspace{-1.3em} \textbf{Step 3: Optimal Estimator Selection}
\State $\estimator^{*},\batchSize^{*},\batchCount^{*} \gets \arg\min_{(\estimator,\batchSize,\batchCount)\in\estimatorList\times\batchList} \criteria \left(\Big\{\big\{ \CFETest{t}{\batch_{j}^v} \big\}_{t = 0}^T\Big\}_{j=1}^{b_v}, \Big\{\big\{ \ECFTrain{t}{\batch_{j}^v}{\estimator,\batchSize,\batchCount} \big\}_{t = 0}^T\Big\}_{j=1}^{b_v}\right)$
\end{algorithmic}
\end{algorithm}

\begin{remark}
    \label{rem:proprocessing}
    In experimental settings with strong temporal patterns such as seasonality, we first \emph{detrend} the observed outcomes, following Assumption~\ref{asmp:SE_decomposition}. We can incorporate detrending into Algorithm~\ref{alg:C-CV} by augmenting Step~1 as follows:
    \begin{itemize}
        \item Fit an auxiliary model on the full observed data to capture temporal patterns and compute the baseline counterfactual as the sample mean of outcomes under all-control.
        \item Create filtered data by subtracting the baseline counterfactual from observed outcomes.
        \item Feed the filtered data into Step~2 that targets treatment-effect identification.
        \item Combine the baseline counterfactual with estimated treatment effects for final estimates.
    \end{itemize}
    Algorithm~\ref{alg:FO_with_preprocessing} in Appendix~\ref{sec:preprocessing} presents an example of counterfactual estimation with detrending, where the consistency result relies on a weak additive separability assumption between baseline dynamics and treatment effects (see Appendix~\ref{sec:preprocessing}).
\end{remark}

%% file: Numerical.tex
\section{Interference Gym}
\label{sec:Benchmark_Toolbox}
In this section, we introduce six semi-synthetic experiments that combine simulated environments with real-world data (Table~\ref{tab:detailed_environments}). This approach offers two key advantages: it maintains \emph{realistic data characteristics} while allowing us to \emph{compute ground truth values} for our estimands. Unlike real experimental settings where outcomes are observed under a single treatment allocation, these settings provide ground truth values for any desired scenario. This enables rigorous evaluation of estimation methods under realistic conditions. We specify the treatment allocation, outcomes, and network structure for each experimental setting below. Each setting is accompanied by a corresponding Jupyter notebook accessible via these icons:
{\begingroup
\setbox0=\hbox{\includegraphics[height=3ex]{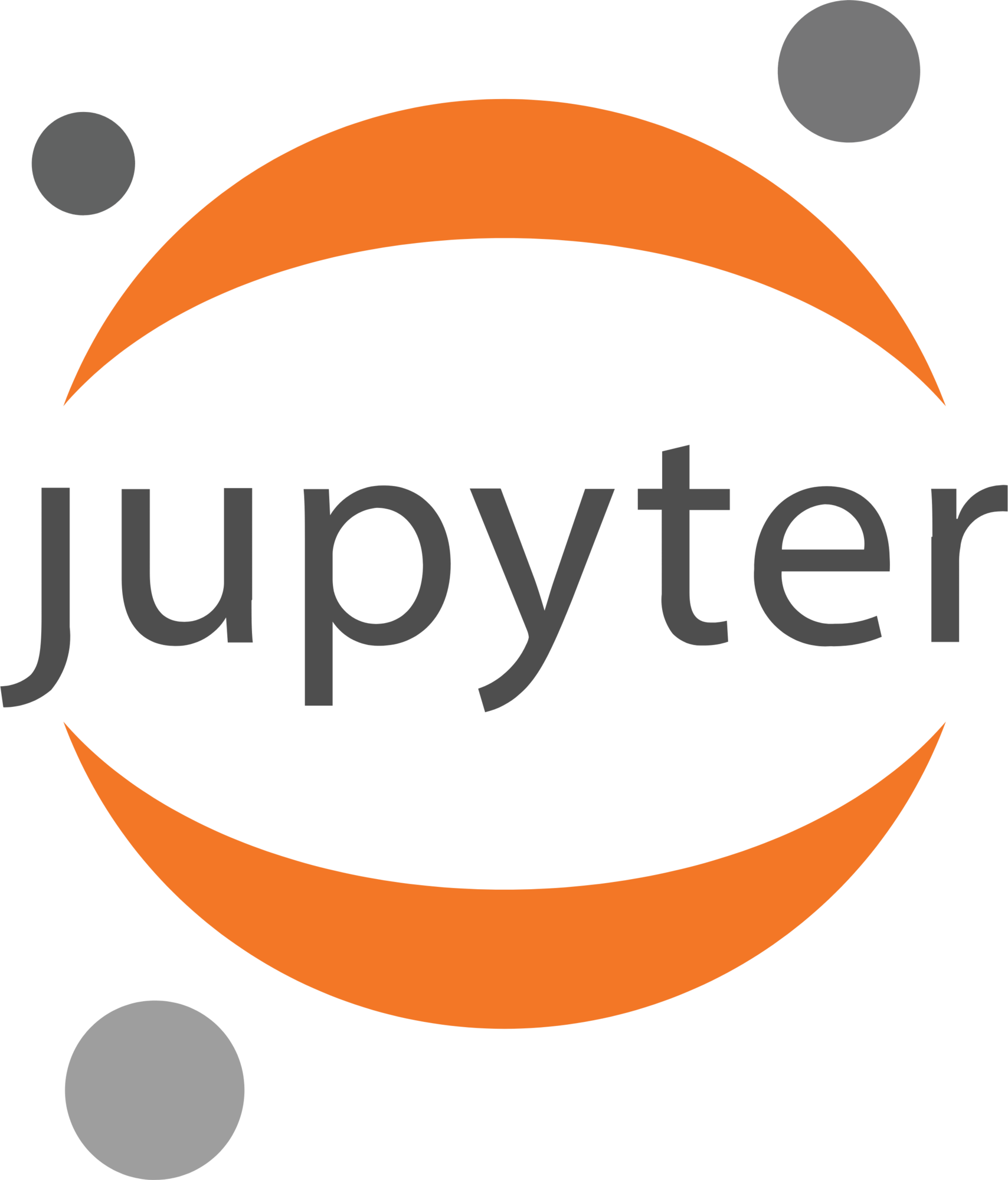}}%
\parbox{\wd0}{\box0}\endgroup}.
These notebooks are part of our \href{https://github.com/CausalMP/CausalMP.git}{Python package} and allow running our estimation framework or extracting data for other applications.

\begin{table}[htbp]
    \centering
    \footnotesize
    \begin{tabular*}{\textwidth}{@{\extracolsep{\fill}}p{0.22\textwidth}p{0.75\textwidth}@{}}
        \toprule
        
        \multicolumn{2}{c}{\small\textbf{LLM-based Social Network} -- Social media platform (like Facebook) with AI agents as users} \\
        \midrule
        \textbf{Treatment} & Feed ranking: random ordering (control) vs. friend engagement weighting (treatment) \\
        \cmidrule(l){1-2}
        \textbf{Outcome} & User engagement metrics (sum of likes and replies per user) \\
        \cmidrule(l){1-2}
        \textbf{Interference Structure} & Preferential attachment model with directed follower-following relationships \\
        \cmidrule(l){1-2}
        \textbf{Key Characteristic} & AI agents with demographically diverse personas based on US Census data \\
        
        \midrule
        \multicolumn{2}{c}{\small\textbf{Belief Adoption Model} -- Opinion diffusion through social coordination games} \\
        \midrule
        \textbf{Treatment} & Promotional campaign to promote Opinion A (new belief) \\
        \cmidrule(l){1-2}
        \textbf{Outcome} & Binary adoption decision (Opinion A vs. Opinion B) \\
        \cmidrule(l){1-2}
        \textbf{Interference Structure} & Coordination game with payoff-based decisions influenced by friends' opinions \\
        \cmidrule(l){1-2}
        \textbf{Key Characteristic} & Pokec social network (3 regional networks: 3,366-42,971 users) with demographic profiles \\
        
        \midrule
        \multicolumn{2}{c}{\small\textbf{NYC Taxi Routes} -- Urban transportation network with route dependencies} \\
        \midrule
        \textbf{Treatment} & Pricing experimentation on randomly selected routes \\
        \cmidrule(l){1-2}
        \textbf{Outcome} & Trip count per route during 6-hour periods \\
        \cmidrule(l){1-2}
        \textbf{Interference Structure} & Route adjacency based on geographic proximity and functional relationships \\
        \cmidrule(l){1-2}
        \textbf{Key Characteristic} & Real temporal pattern using NYC taxi records (58M trips, Jan-Mar 2024) \\
        
        \midrule
        \multicolumn{2}{c}{\small\textbf{Exercise Encouragement} -- Digital health intervention with social influence} \\
        \midrule
        \textbf{Treatment} & Digital motivational messages for physical activity \\
        \cmidrule(l){1-2}
        \textbf{Outcome} & Binary exercise decision (Bernoulli distribution) \\
        \cmidrule(l){1-2}
        \textbf{Interference Structure} & Twitter social circles with peer influence (avg. 21.74 connections) \\
        \cmidrule(l){1-2}
        \textbf{Key Characteristic} & Demographic-based response
        patterns with weekly cycles based on US Census data \\
        
        \midrule
        \multicolumn{2}{c}{\small\textbf{Data Center} -- Server farm with workload redistribution effects} \\
        \midrule
        \textbf{Treatment} & Processing power enhancements for selected servers \\
        \cmidrule(l){1-2}
        \textbf{Outcome} & Server utilization rate (proportion of time server remains busy during each interval) \\
        \cmidrule(l){1-2}
        \textbf{Interference Structure} & Join-the-shortest-queue routing creates implicit interference patterns \\
        \cmidrule(l){1-2}
        \textbf{Key Characteristic} & Time-dependent Poisson arrivals with realistic daily pattern and job classification \\

        \midrule
        \multicolumn{2}{c}{\small\textbf{Auction Model} -- Competitive auction market with bidders competing for different objects} \\
        \midrule
        \textbf{Treatment} & Promotional interventions increasing bidder valuations by $\tau$\% for selected objects \\
        \cmidrule(l){1-2}
        \textbf{Outcome} & Value of each object in each auction round \\
        \cmidrule(l){1-2}
        \textbf{Interference Structure} & Indirect effects through strategic bidding and price-driven market dynamics \\
        \cmidrule(l){1-2}
        \textbf{Key Characteristic} & Market-based pricing with heterogeneous bidder types and realistic valuation strategies \\
        
        \bottomrule
    \end{tabular*}
    \caption{Description of Experimental Environments}
    \label{tab:detailed_environments}
\end{table}

\subsection{LLM-based Social Network 
\texorpdfstring{\jupyter{https://github.com/CausalMP/CausalMP/blob/main/nb_llm_based_social_network.ipynb}}{jupyter}}
\label{sec:LLM}
This environment simulates a social media platform like Facebook or Quora where user interactions occur through content feeds, designed to study the effects of feed ranking algorithms on user engagement. The environment employs Large Language Model (LLM) agents to represent users, with demographically realistic personas derived from US Census data \citep{uscensus2023} following \cite{chang2024llms}'s methodology. Each agent possesses two interests selected from a set of \emph{keywords}: Technology, Sports, Politics, Entertainment, Science, Health, and Fashion.

The treatment variable is the \emph{feed ranking algorithm}, which determines the ordering of content presented to each user. In the control condition, users receive content in random order, while in the treatment condition, content is ranked by weighting based on friend engagement levels. The outcome of interest is \emph{user engagement metrics}, measured as the total number of likes and replies by each user within a time period. The network structure is then captured by a directed follower-following graph constructed using a preferential attachment model \citep{barabasi1999emergence}.

The dynamics of the environment center on content generation and user interactions. Content originates from a bank focused on a specified topic (e.g., climate change) with cross-keyword variations (e.g., technology or politics). The system then generates user-specific feeds by combining interest-based content (matching user interests), trending content (high engagement), and random content. In each round, users interact with their feeds through an LLM-driven decision process that incorporates content relevance, friend engagement, feed position, and demographic characteristics.

The simulation maintains comprehensive state information including engagement metrics, user interaction histories, network relationships, content visibility, and conversation threads. At each time step, agents have a 1\% probability of generating new content, contributing to organic content evolution. This interaction framework ensures that user behaviors and social influence patterns evolve naturally through the network while maintaining computational traceability.
\begin{figure}
    \centering
    \includegraphics[width=0.7\linewidth]{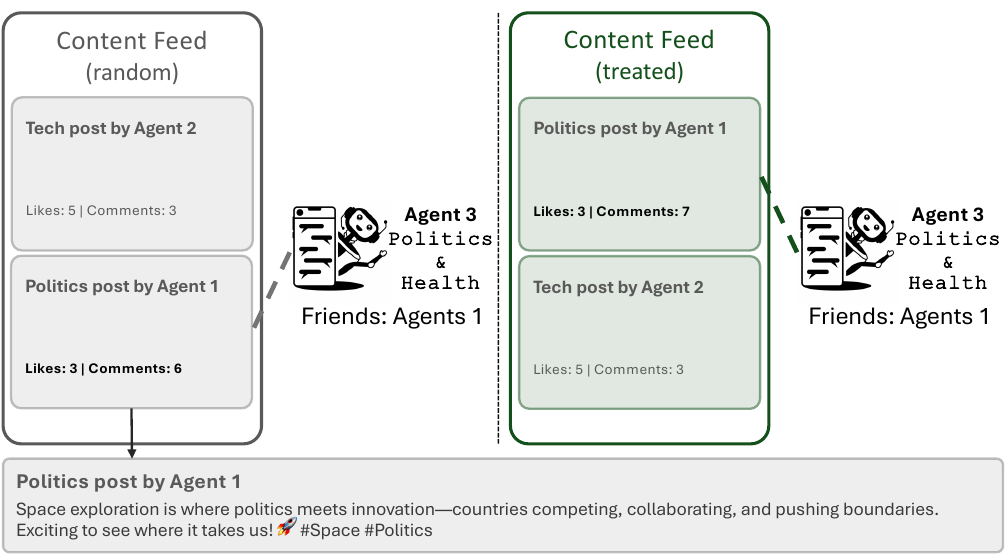}
    \caption{LLM-based Social Network. Demographically diverse AI agents interact through content feeds. Treatment involves optimized vs. random feed ranking, with outcomes measured as user engagement metrics.}
    \label{fig:LLM_model}
\end{figure}

\subsection{Belief Adoption Model
\texorpdfstring{\jupyter{https://github.com/CausalMP/CausalMP/blob/main/nb_belief_adoption__Krupina.ipynb}}{jupyter}
\texorpdfstring{\jupyter{https://github.com/CausalMP/CausalMP/blob/main/nb_belief_adoption__Topolcany.ipynb}}{jupyter}
\texorpdfstring{\jupyter{https://github.com/CausalMP/CausalMP/blob/main/nb_belief_adoption__Zilina.ipynb}}{jupyter}
}
\label{sec:Belief_Adoption}
This environment models the diffusion of competing opinions within interconnected communities, implementing the cascade model of \cite{montanari2010spread}. The setting examines how opinions spread through social networks when individuals make decisions through coordination games with their neighbors. Through this framework, we evaluate the effectiveness of promotional campaigns in influencing opinion adoption patterns.

The environment considers two competing stances: Opinion A (e.g., voting in the upcoming election) and Opinion B (e.g., not voting in the election). The treatment represents a \emph{campaign aimed at increasing Opinion~$\OPA$ adoption}. The outcome for each unit in each period is binary: \emph{1 if they adopt Opinion~$\OPA$, and 0 otherwise}.

The opinion evolution follows a network-based coordination game where each individual $i$ assigns base payoff values $\payoff^i_\OPA$ and $\payoff^i_\OPB$ to both opinions. The probability of Opinion~$\OPA$ adoption in the next period depends on the neighbor configuration and relative payoffs:
\begin{align*}
    \P\big(\text{adopting Opinion~$\OPA$} \big| \neighbor^i_\OPA,\neighbor^i_\OPB \big) =
    \frac{1}{1 + e^{-2\beta  \left(\neighbor^i h^i + \neighbor^i_\OPA - \neighbor^i_\OPB  \right)}},
    \quad\quad
    h^i = \frac{\payoff^i_\OPA - \payoff^i_\OPB}{\payoff^i_\OPA + \payoff^i_\OPB},
\end{align*}
where $\neighbor^i_\OPA$ represents the number of neighbors holding Opinion~$\OPA$ out of $\neighbor^i$ total neighbors in the current period, and $\beta$ is a predetermined constant. The underlying network in our simulator is derived from the \emph{Pokec social network} dataset \citep{takac2012data,snapnets}, focusing on three regional networks: Krupina (3,366 users), Topolcany (18,246 users), and Zilina (42,971 users). For treated individuals, the final payoff equals the base payoff plus the treatment effect, where both components are characterized for each user through three key demographic variables: age, profile activity, and gender from the Pokec social network dataset.

\begin{figure}[htbp]
\centering
\resizebox{0.8\linewidth}{!}{
\begin{tikzpicture}[
    box/.style={draw, rounded corners=5pt, thick, text width=7.1cm, align=center, minimum height=2cm},
    title/.style={font=\bfseries},
    arrow/.style={->, thick, color=orange!80!red, line width=2pt}
]

\node[box, fill=blue!15] (left) at (0.2,0) {
    \textcolor{blue!70!black}{\textbf{Individual Level}} \\[0.1cm]
    
    \textbf{\footnotesize Demographics:}
    {\scriptsize Age and profile activity} \\
    
    {\scriptsize $\downarrow$} \\
    
    \textbf{\footnotesize Base payoffs} 
    {\footnotesize and}
    \textbf{\footnotesize treatment effect} \\

    {\scriptsize $\downarrow$} \\
    
    \textbf{\footnotesize Heterogeneous responses}
};

\node[box, fill=orange!15] (right) at (8.6,0) {
    \textcolor{orange!70!black}{\textbf{Network Level}} \\[0.1cm]
    
    \textbf{\footnotesize Network effects amplify individual changes} 
    \\
    {\scriptsize through coordination games with the neighbors}\\
    
    {\scriptsize $\downarrow$} \\
    
    \textbf{\footnotesize Opinion cascades}
    \\
    {\scriptsize through social connections} \\

};

\draw[arrow] (left.east) -- (right.west);

\end{tikzpicture}
}

\caption{Belief Adoption Model. Individual demographics determine heterogeneous base payoffs and treatment responses, which then propagate through network to create opinion cascades across the network.}
\label{fig:belief-adoption-mechanism}
\end{figure}

Base payoffs for Opinion $\OPA$ are strategically calibrated to reflect realistic demographic patterns. Users aged 25-55 years receive elevated base payoffs for Opinion $\OPA$, capturing the tendency for individuals in this age range to exhibit greater civic engagement. Furthermore, users with higher profile activity levels, quantified by the completeness of their profile information, are assigned increased base payoffs for voting. This reflects the fact that socially engaged users typically demonstrate heightened civic participation and political awareness. These demographic adjustments create heterogeneous baseline voting propensities across the network, with distinct population segments exhibiting varying predispositions toward electoral participation.

\begin{figure}[ht]
    \centering
    \includegraphics[width=0.85\linewidth]{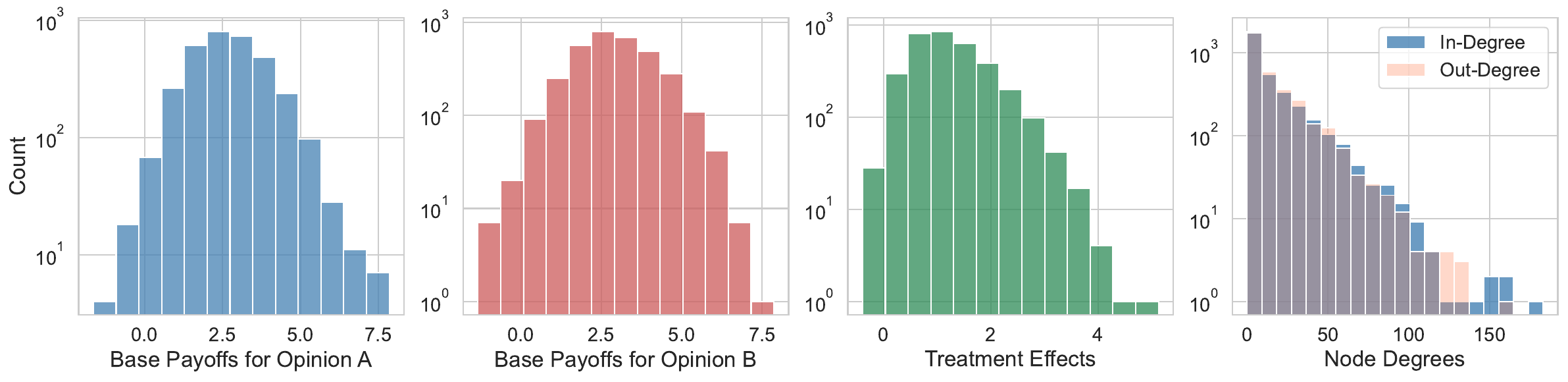}
    \includegraphics[width=0.85\linewidth]{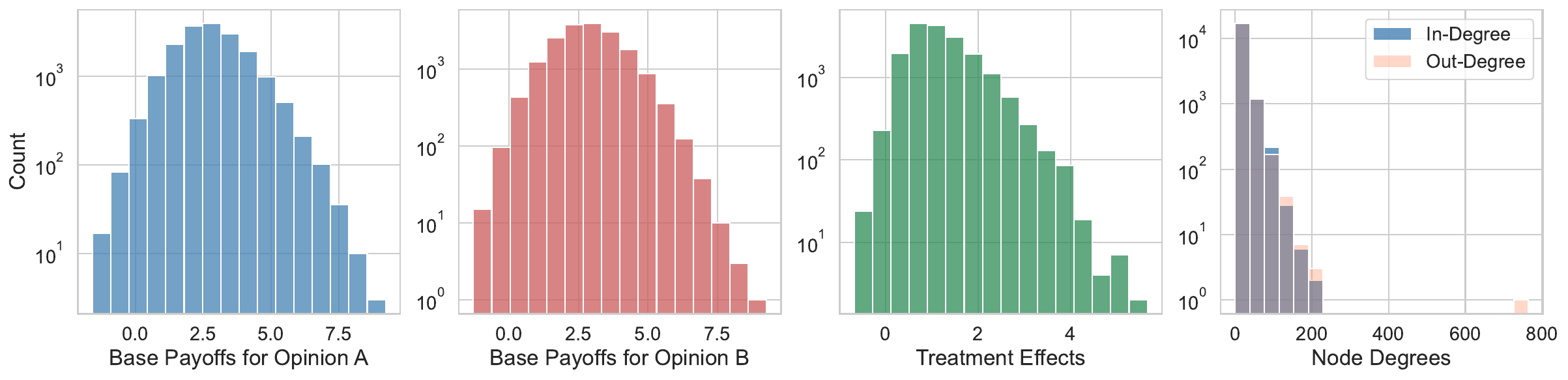}
    \includegraphics[width=0.85\linewidth]{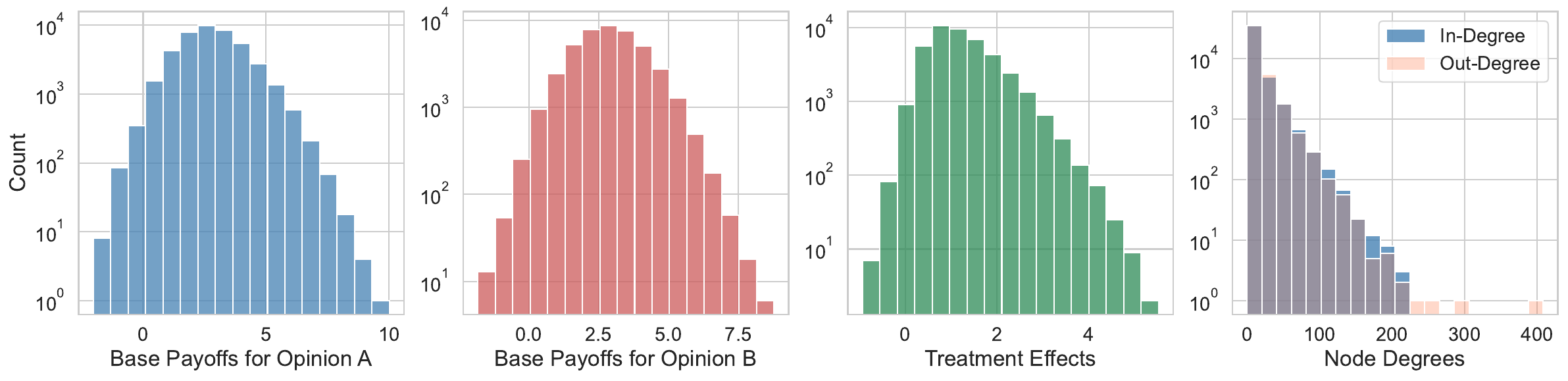}
    \caption{Distribution of base payoffs, treatment effects, and node degrees for individuals across three regions: Krupina with 3,366 users (top), Topolcany with 18,246 users  (middle), and Zilina with 42,971 users  (bottom).}
    \label{fig:BAM-all}
\end{figure}

Treatment effectiveness employs a complementary modeling approach. The treatment effect follows a Gaussian distribution centered at age 35, representing peak responsiveness to new content (e.g., get-out-the-vote messaging) within the adult population. Complementing these age-based effects, treatment impact also scales proportionally with each user's profile activity level. Users maintaining more comprehensive profiles (indicative of greater platform engagement and social connectivity) experience amplified treatment effects. This captures the phenomenon that active social media participants exhibit increased susceptibility to civic campaigns and peer-mediated influence. This approach generates realistic heterogeneity in treatment responsiveness across the network, as demonstrated in Figure~\ref{fig:BAM-all}.

\subsection{New York City Taxi Routes
\texorpdfstring{\jupyter{https://github.com/CausalMP/CausalMP/blob/main/nb_nyc_taxi_routes.ipynb}}{jupyter}
}
\label{sec:LiM}
This environment is based on ride-sharing dynamics across New York City taxi zones using TLC Trip Record Data \citep{nyc_tlc_trip_data}. The framework adopts the established linear-in-means outcome specification \citep{eckles2016design,leung2022causal} to model passenger response patterns to a pricing experimentation, while incorporating realistic noise levels, seasonality patterns, and temporal correlations observed in the actual data. By leveraging actual trip records, passenger density metrics, and inter-zone connectivity, the simulation captures the complex network of interactions between taxi routes throughout the city across different periods.

\begin{figure}[htbp]
\centering
\begin{tikzpicture}[
    box/.style={draw, rounded corners=8pt, thick, text width=4.5cm, align=center, minimum height=2cm},
    title/.style={font=\bfseries\footnotesize}
]

\node[box, fill=blue!20] (temporal) at (0,0) {
    \textcolor{blue!70!black}{\textbf{\small Temporal Variations}} \\[0.1cm]
    
    \textcolor{black!80}{\textbf{\footnotesize Real seasonality:}} \\
    {\scriptsize • Daily rush patterns} \\
    {\scriptsize • Weekly demand cycles} \\[0.05cm]
    
    \textcolor{black!80}{\textbf{\footnotesize Data scale:}}\\
    {\scriptsize 58M taxi trips in NYC}
};

\node[box, fill=green!20] (network) at (5.0,0) {
    \textcolor{green!70!black}{\textbf{\small Complex Interference}} \\[0.1cm]
    
    \textcolor{black!80}{\textbf{\footnotesize Route connections:}} \\
    {\scriptsize • Geographic proximity} \\
    {\scriptsize • Shared infrastructure} \\[0.05cm]
    
    \textcolor{black!80}{\textbf{\footnotesize Network structure:}}\\
    {\scriptsize 18,768 routes, avg degree: 8.32}
};

\node[box, fill=purple!20] (heterogeneity) at (10,0) {
    \textcolor{purple!70!black}{\textbf{\small Response Heterogeneity}} \\[0.1cm]
    
    \textcolor{black!80}{\textbf{\footnotesize Route differences:}} \\
    {\scriptsize • Passenger density} \\
    {\scriptsize • Location characteristics} \\[0.05cm]
    
    \textcolor{black!80}{\textbf{\footnotesize Treatment effects:}}\\
    {\scriptsize AI-generated route-specific scores}
};

\end{tikzpicture}
\caption{NYC Taxi Model. The environment incorporates authentic temporal patterns from 58M trips, realistic network interference through route adjacency, and heterogeneous treatment responses.}
\label{fig:nyc-taxi-challenges}
\end{figure}

In this setting, we conceptualize the treatment as a \emph{new pricing algorithm} that incorporates route-specific characteristics (e.g., distance, average demand and supply at origin and destination). Each route corresponds to an origin-destination pair between the city's 263 taxi zones. The new pricing algorithm is then implemented on randomly selected routes to evaluate travelers' response patterns. We segment time into 6-hour periods, with the outcome variable measuring the \emph{number of trips} along each route during each period.

Precisely, given baseline outcomes $[\OoutcomeD{}{i}{t}]_{i,t}$, the system's evolution follows Equation \eqref{eq:linear-in-mean}:
\begin{equation}
\label{eq:linear-in-mean}
\begin{aligned}
    \outcomeD{}{i}{t+1}
    =
    \OoutcomeD{}{i}{t+1}
    +
    \ACE
    \sum_{j=1}^N \adjMe^{ij}
    (\outcomeD{}{j}{t} - \OoutcomeD{}{i}{t})
    +
    \ADE_{\pl} \sum_{j=1}^N \adjMe^{ij}\treatment{j}{t+1}
    +
    \ADE^i_{\ul} \treatment{i}{t+1},
    \quad
    t \geq 1,
\end{aligned}
\end{equation}
where we initiate the recursion by setting $\outcomeD{}{i}{0} = \OoutcomeD{}{i}{0}$ and $\treatment{i}{0} = 0$ for all $i$.
Here, $\adjM = [\adjMe^{ij}]_{i,j}$ represents the normalized route adjacency matrix, $\ADE_{\ul}^i$ represents route-specific direct treatment effects, and parameters $(\ACE,\ADE_{\pl}) = (0.4,0.2)$ control autocorrelation and spillover effects.
\begin{figure}
    \centering
    \includegraphics[width=0.6\linewidth]{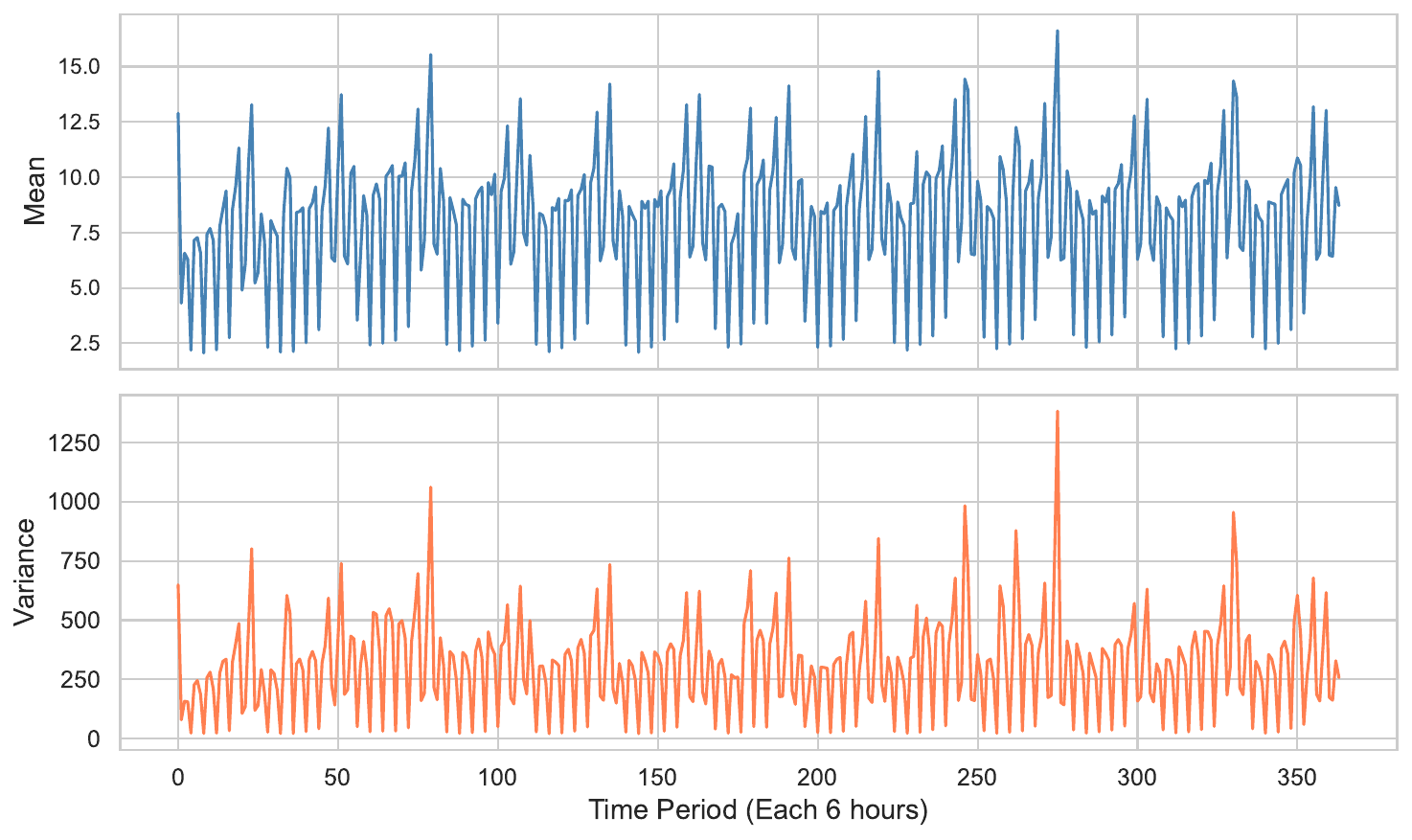}
    \caption{Mean and variance of baseline outcomes $y_t^i$ (the number of trips on route $i$ during period $t$), revealing strong daily and weekly seasonality patterns.}
    \label{fig:NYC_taxi_panel}
\end{figure}
\begin{figure}[ht]
    \centering
    \includegraphics[width=0.3\linewidth]{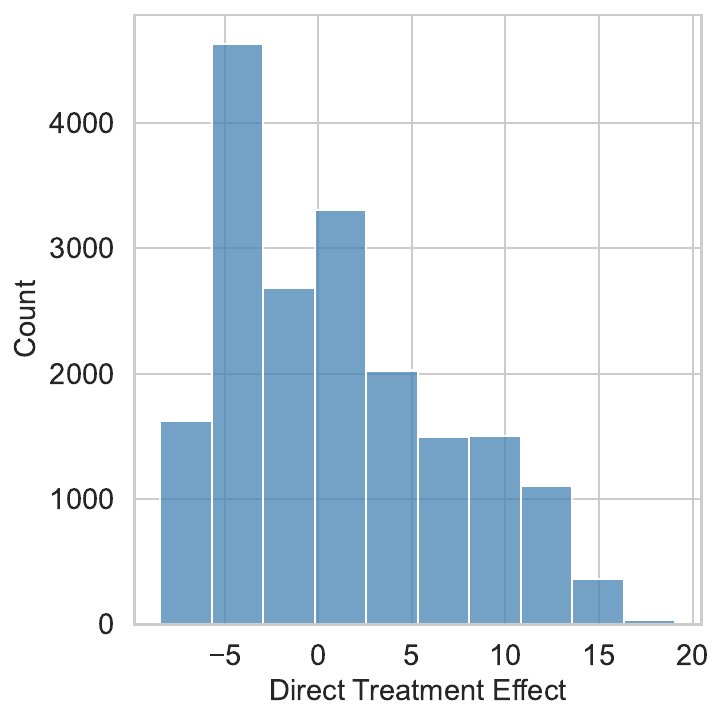}
    \includegraphics[width=0.3\linewidth]{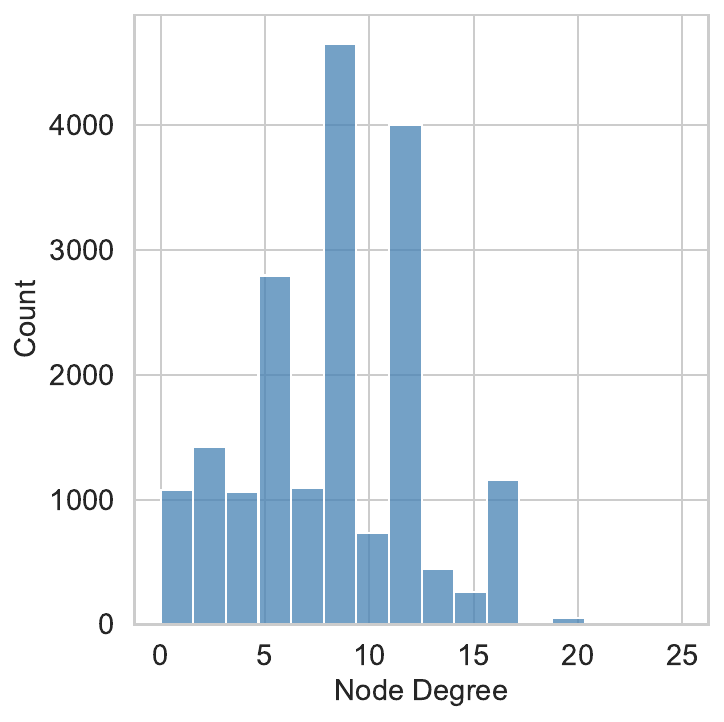}
    \caption{Distribution of route-specific direct treatment effects (left) and node degree distribution in the route interference network (right), both illustrating heterogeneity in treatment responses and local interactions.}
    \label{fig:NYC_analysis}
\end{figure}

In \eqref{eq:linear-in-mean}, the outcome $\outcomeD{}{i}{t}$ represents the number of trips over route $i$ during period $t$ under a specific treatment allocation. To naturally reflect potential real-world outcome variations both across units and over time, we extract baseline outcomes $\OoutcomeD{}{i}{t}$ from ``High Volume For-Hire Vehicle Trip Records" for January-March 2024 (57,974,677 trips), resulting in 18,768 active routes across 366 periods. As shown in Figure~\ref{fig:NYC_taxi_panel}, the baseline outcomes demonstrate strong seasonality patterns, which are common in ride-hailing applications \citep{xiong2024data}.

The adjacency network captured by $\adjM$ in \eqref{eq:linear-in-mean} is constructed using AI models (Claude 3.5 Sonnet, Anthropic, 2024) based on the ``Taxi Zone Lookup Table." Specifically, the adjacency of two routes is determined according to geographic proximity, transit connections, shared roads, and functional relationships, yielding a network with an average node degree of 8.32 (Figure~\ref{fig:NYC_analysis}, right panel). Finally, route-specific direct treatment effects $\ADE_{\ul}^i$ are generated using passenger density scores in both origin and destination, also determined by AI models (Figure~\ref{fig:NYC_analysis}, left panel). This approach is to ensure the simulation reproduces key real-world temporal variations and noise patterns.

\subsection{Exercise Encouragement Program
\texorpdfstring{\jupyter{https://github.com/CausalMP/CausalMP/blob/main/nb_exercise_encouragement_program.ipynb}}{jupyter}
}
\label{sec:BOM}
This environment models a digital health intervention program designed to assess how sending digital encouragement messages influences users’ exercise decisions \citep{liao2016sample,klasnja2015microrandomized,klasnja2019efficacy}. To enhance realism, we integrate individual-level features from the Census Bureau database \citep{kohavi1994data} with Twitter social circles data \citep{leskovec2012learning}. This combination allows us to capture user diversity and their heterogeneous social interactions.
\begin{figure}[htbp]
    \centering
    \includegraphics[width=1\linewidth]{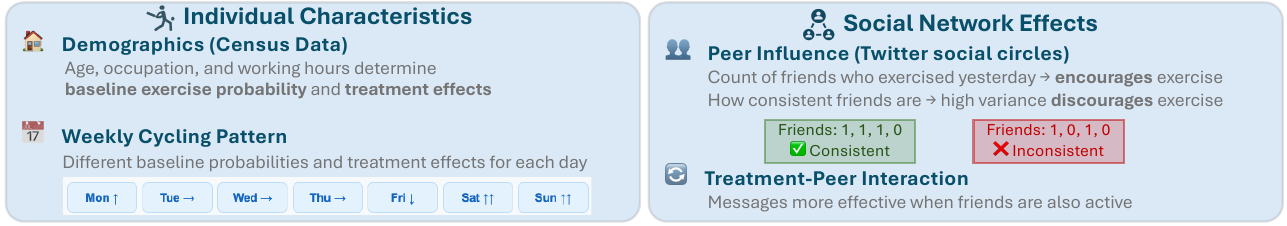}
    \caption{Exercise Encouragement Program. Individual decisions depend on personal characteristics, treatment status, and friends' exercise patterns, creating dynamic loops where exercise behavior spreads through social ties.}
    \label{fig:exercise_simulator}
\end{figure}

\begin{figure}[htbp]
    \centering
    \includegraphics[width=0.4\linewidth]{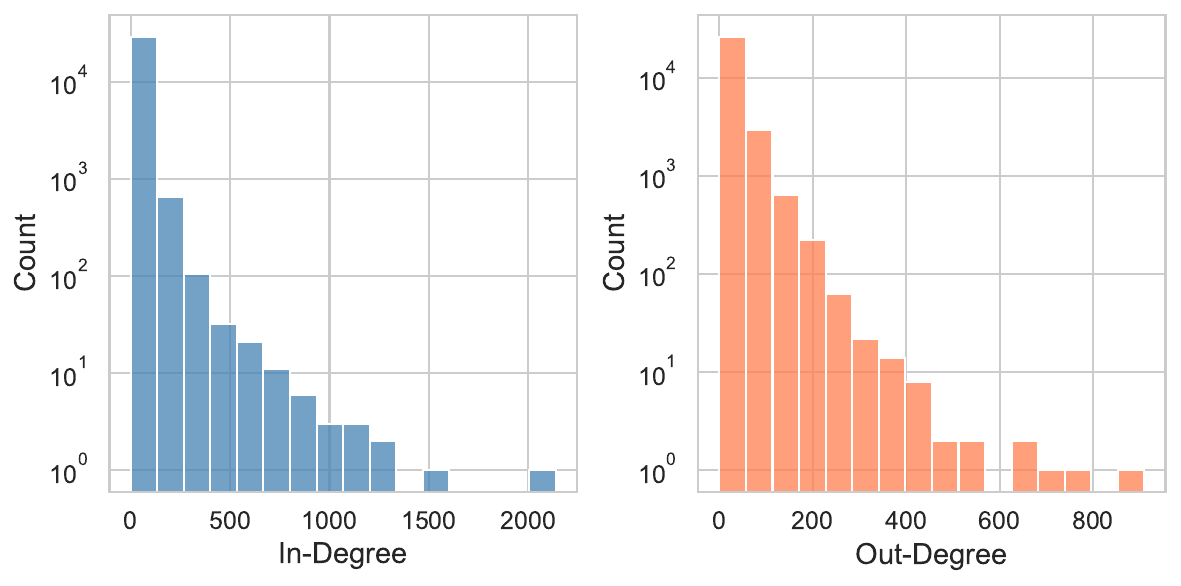}
    \caption{Distribution of node degrees in the Twitter social network.}
    \label{fig:twitter_network_hist}
\end{figure}

In this setting, the experimental units are individuals, with binary outcomes representing their \emph{exercise decisions in each period} (1 for exercise, 0 for no exercise). The treatment consists of sending \emph{digital intervention messages} designed to encourage physical activity. Inspired by \cite{li2022network}, we let the outcomes follow a Bernoulli distribution defined as follows:
\begin{align}
    \label{eq:MRT}
    \outcomeD{}{i}{t+1} \sim \text{Bernoulli}\left(
    \frac{1}
    {
    1
    +
    \exp{[-(
    \ABE_t^i
    +
    \ADE_t^i
    \treatment{i}{t+1}
    +
    \ACE
    \outcomeD{}{i}{t} Z^i_t
    +
    \APE \treatment{i}{t+1} \outcomeD{}{i}{t} Z^i_t
    -
    \eta V_i^t
    )}]}\right),
\end{align}
where $Z^i_t = \sum_{j=1}^N \adjMe^{ij}\outcomeD{}{j}{t}$ represents the count of neighboring individuals who exercised in the previous period, and $V_t^i$ represents the variance of exercise behavior among individual $i$'s neighbors in the previous period. Here, $\adjMe^{ij}$'s are the elements of the adjacency matrix from \emph{Twitter social circles} data, with an average of 21.74 connections per individual (Figure~\ref{fig:twitter_network_hist}). 

\begin{figure}[htbp]
    \centering
    \includegraphics[width=1\linewidth]{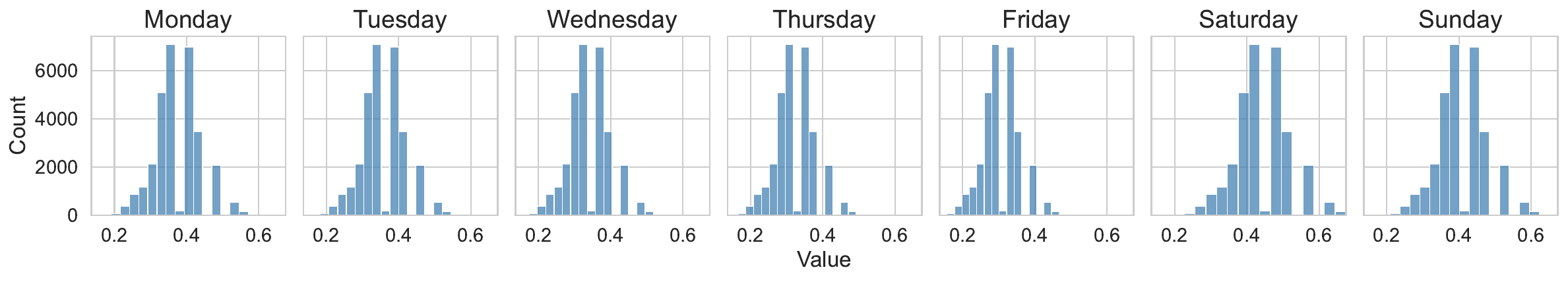}
    \caption{Distribution of baseline exercise score $\ABE_t^i$ across the population over weekdays.}
    \label{fig:exercise_probability}
\end{figure}

In Equation~\eqref{eq:MRT}, each component captures distinct aspects of exercise behavior. The baseline score~($\ABE_t^i$) represents an individual's inherent propensity to exercise, uniquely determined for each participant based on Census Bureau demographic data. This score reflects factors such as age (with younger individuals assigned higher values), working hours (which show an inverse relationship with exercise likelihood), and occupation type (favoring active or professional occupations). These baseline scores exhibit pronounced weekly patterns: they peak on weekends due to increased free time, rise on Mondays from renewed weekly motivation, stabilize mid-week, and slightly decline on Fridays, reflecting end-of-week fatigue (Figure~\ref{fig:exercise_probability}).

\begin{figure}[ht]
    \centering
    \includegraphics[width=1\linewidth]{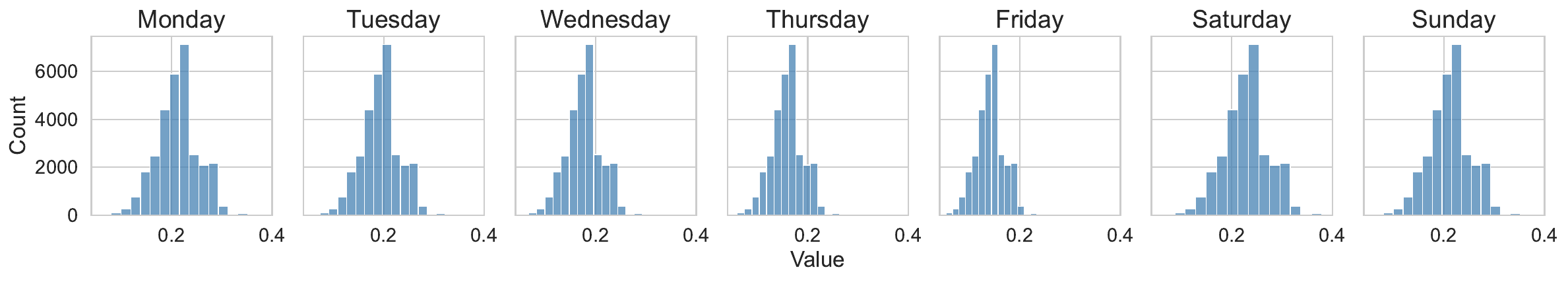}
    \caption{Distribution of intervention message effects across the population over weekdays.}
    \label{fig:message_impact}
\end{figure}

The intervention effectiveness ($\ADE_t^i$) quantifies how individuals respond to digital messages. This response varies based on multiple demographic factors. Younger individuals show higher responsiveness to digital interventions and education level correlates positively with intervention effectiveness. Job-related factors, such as occupation type and working hours, further influence response rates by reflecting flexibility and availability to act on the interventions. Moreover, the impact of messages follows a weekly cycle: effectiveness peaks on weekends and Mondays, then gradually declines through mid-week (Figure~\ref{fig:message_impact}). The model also incorporates parameters $(\ACE, \APE, \eta) = (0.04, 0.01, 0.02)$ to capture various types of peer influence effects.

\subsection{Data Center Server Utilization
\texorpdfstring{\jupyter{https://github.com/CausalMP/CausalMP/blob/main/nb_data_center_model.ipynb}}{jupyter}
}
\label{sec:servers}
This environment simulates a server farm to evaluate interventions designed to improve server utilization. With the increasing demand for cloud computing resources, optimizing data center utilization has become critical for addressing global sustainability concerns \citep{zhang2023global,saxena2023sustainable}. Within this system, servers influence each other's performance through the system's physical characteristics, particularly via the task routing policy. This operational dynamic creates an implicit interference pattern in the absence of a pre-specified explicit network structure.

\begin{figure}[htbp]
    \centering
    \includegraphics[width=0.6\linewidth]{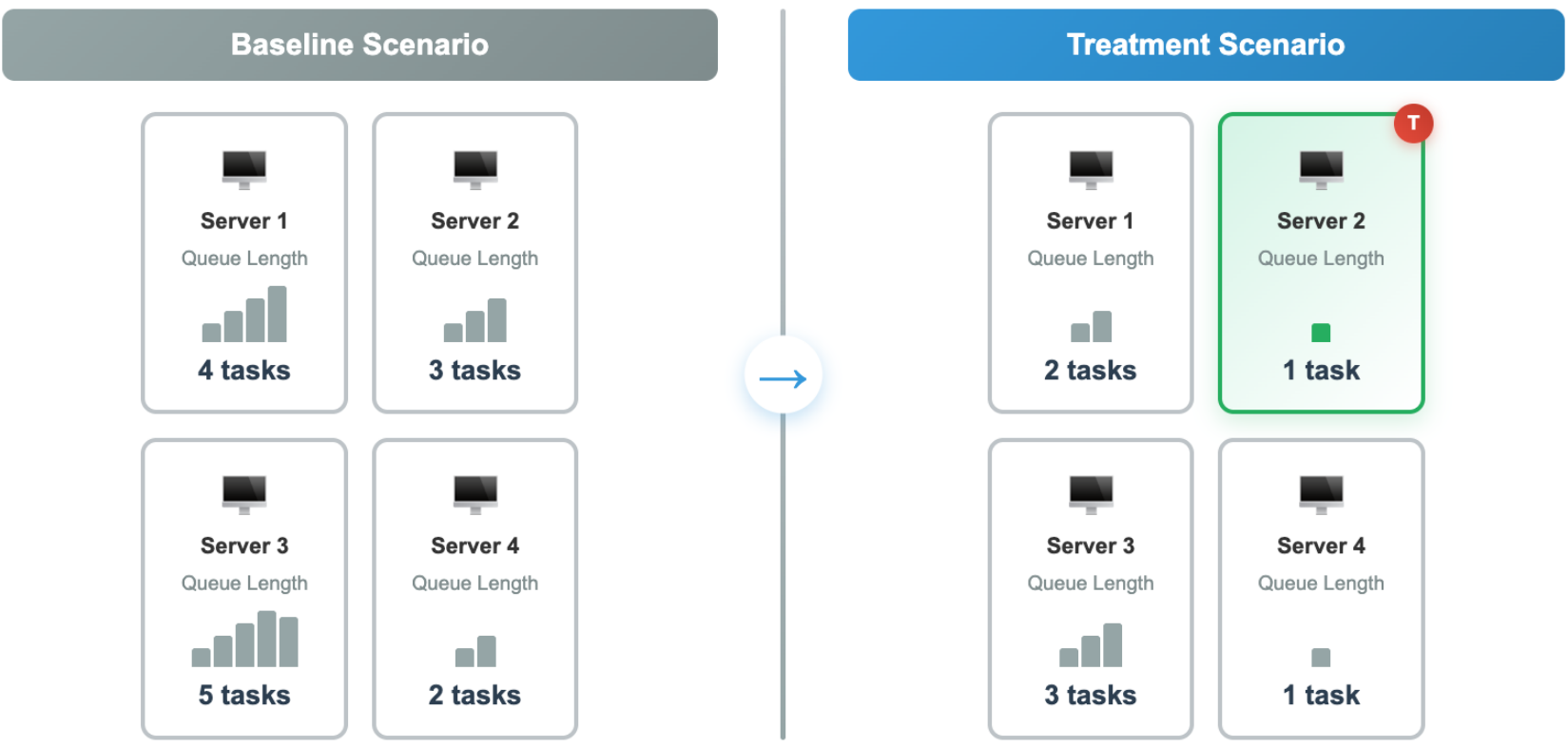}
    \caption{A simple four-server data center. Join-the-shortest-queue routing creates implicit interference: Server 2 with enhanced processing power (treatment) maintains consistently shorter queues, causing the routing system to direct more tasks to this server, reducing available demand for untreated servers.}
    \label{fig:four-server}
\end{figure}

The experimental units in this setting are individual servers in a parallel processing system of $N$ servers. The outcome variable $\outcomeD{}{i}{t}$ represents \emph{server $i$'s utilization} during the interval $[t, t+1)$, measured as the proportion of time the server is busy. The treatment consists of interventions that \emph{enhance the processing power} of selected servers. Then, the system evolves according to a structured routing mechanism. When tasks arrive, the system identifies capable servers for each task type and randomly samples among them. Following the join-the-shortest-queue policy \citep{gupta2007analysis}, tasks are assigned to the servers with the shortest queues within this sample, with random assignment used to break ties. This routing policy naturally generates interference effects, as performance improvements in treated servers shift task distribution across the system (Figure~\ref{fig:four-server}).

\begin{figure}
    \centering
    \includegraphics[width=0.5\linewidth]{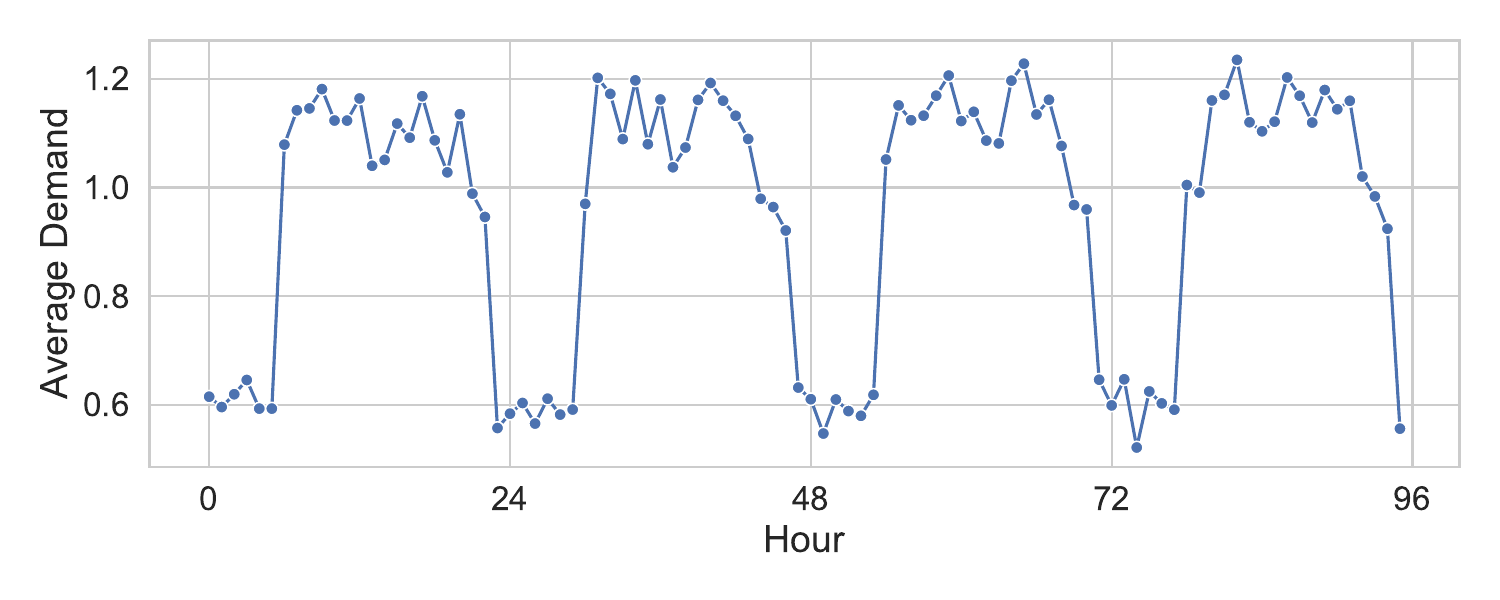}
    \caption{Average demand for the data center over time shows a strong seasonality.}
    \label{fig:Data_center_time_trend}
\end{figure}

The environment captures realistic workload patterns using a time-dependent Poisson arrival process. This demand model features daily cycles (nighttime lows, morning ramp-ups, midday peaks, and evening declines) overlaid with stochastic fluctuations to mimic real-world variability (Figure~\ref{fig:Data_center_time_trend}). Beyond these dynamic patterns, tasks are categorized into multiple types, each with its own compatibility constraints. Servers are also assigned a subset of job types they can process, creating a diverse landscape of server capabilities. Each server processes tasks with exponentially distributed service times, where the processing rates can be enhanced by treatment interventions. Collectively, these mechanisms replicate critical aspects of real-world data centers, such as dynamic routing and resource contention, while enabling controlled experimentation and measurement of treatment effects.

\subsection{Auction Model
\texorpdfstring{\jupyter{https://github.com/CausalMP/CausalMP/blob/main/nb_auction_model.ipynb}}{jupyter}
}
\label{sec:Auction}
This environment simulates a stylized auction model with explicit interference through a bipartite structure. In this setting, multiple bidders participate in an auction for objects, following the model of \cite{bertsekas1990auction}. The auction mechanism creates a simplified pricing system where bidder interactions generate basic patterns of market influence without direct object-to-object relationships.

The setting operates with $N$ objects and $N$ bidders. Each object represents an experimental unit, with its \emph{final value} in each period serving as the outcome variable. The treatment consists of \emph{promotional interventions} that increase bidder valuations by $\tau$\% for randomly selected objects. This treatment affects all bidders interested in the selected objects.

\begin{figure}
    \centering
    \includegraphics[width=0.9\linewidth]{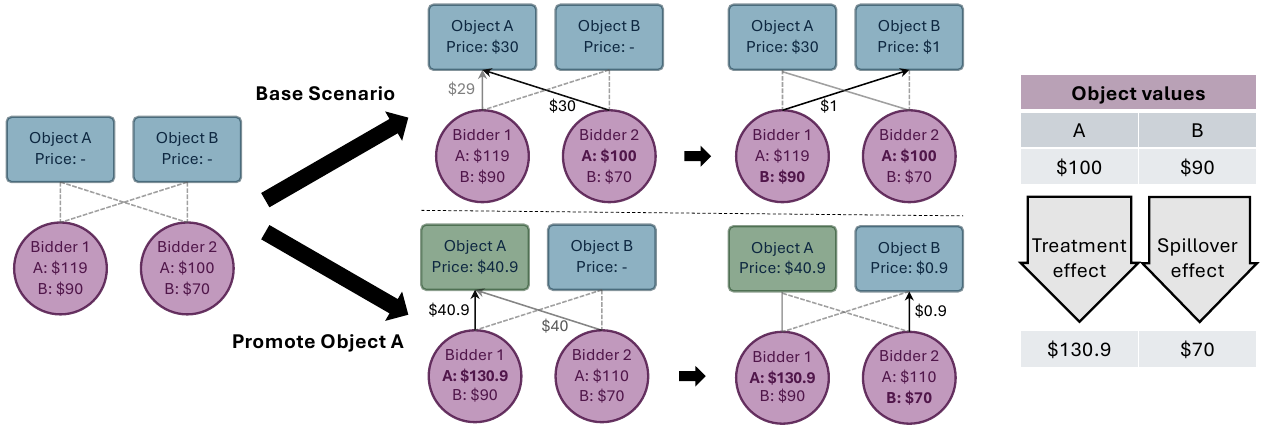}
    \caption{Auction model for two objects in two bidding rounds. Upper panel: base scenario with no treatment. Lower panel: treating Object A with a 10\% valuation increase raises its final value from \$100 to \$130.9, while Object B's value decreases from \$90 to \$70, demonstrating indirect market interference effects.}
    \label{fig:auction_example}
\end{figure}

The market evolution follows a simple bidding process. In each round, bidders evaluate objects based on their private valuations and current market prices. They submit bids for their preferred objects, with objects being assigned to the highest bidders. These assignments establish new price levels, which influence subsequent bidding behavior. As prices increase through competitive bidding, objects become progressively less attractive to competing bidders. This leads to a unique form of interference: while objects do not directly influence each other, \emph{treatment effects propagate through the market via bidders' strategic responses to price changes.} This creates a network of indirect treatment effects, as promotional interventions for certain objects can influence market outcomes for others through shifts in bidder behavior and price dynamics (Figure~\ref{fig:auction_example}).

To enhance realism, the environment implements different bidder types with distinct valuation patterns: standard bidders provide baseline valuations, collectors value items higher with more variability, dealers maintain slightly lower but more consistent valuations, and investors exhibit higher but more variable valuations. This approach captures how different market participants value the same items differently, creating heterogeneous pricing strategies.

%% file: Results.tex
\section{Results and Discussions}
\label{sec:results}
\begin{figure}
    \centering
    \includegraphics[width=\linewidth]{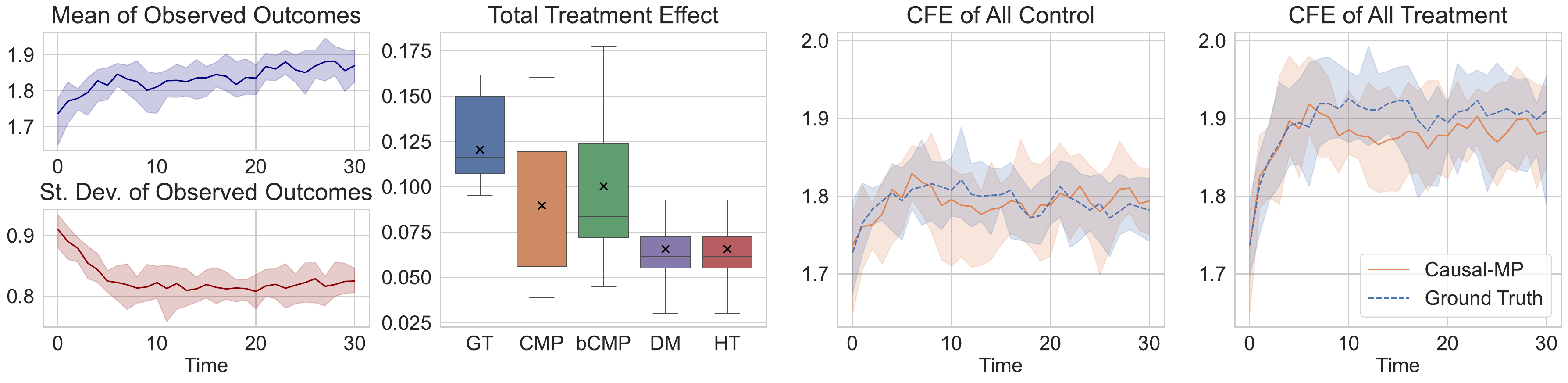}
    \caption{LLM-based social network with $N=1,000$ agents.}
    \label{fig:LLM}
\end{figure}

\begin{figure}
    \centering
    \includegraphics[width=\linewidth]{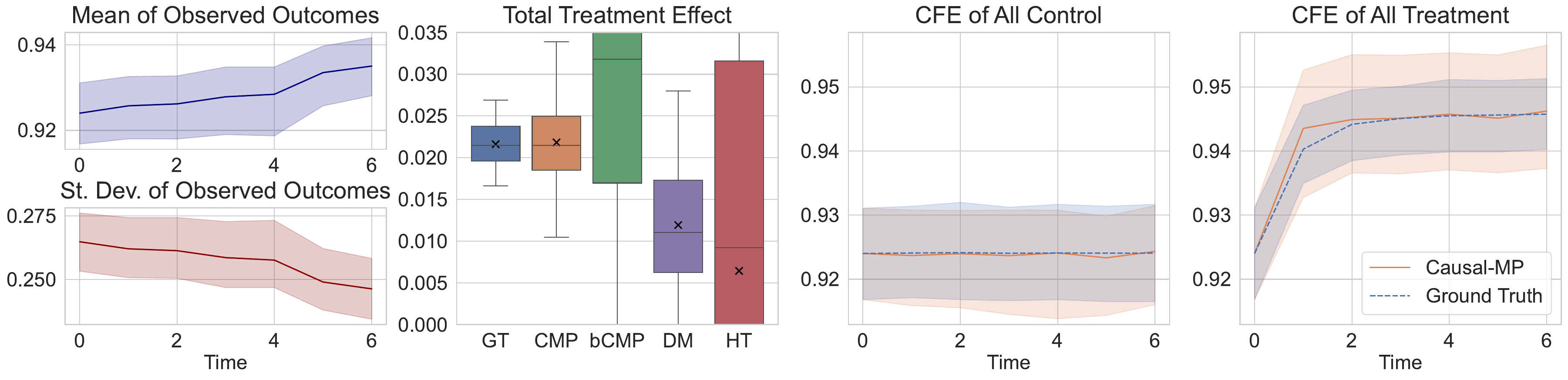}
    \caption{Belief adoption model with Krupina network with $N=3,366$ users.}
    \label{fig:BAM1}
\end{figure}

\begin{figure}
    \centering
    \includegraphics[width=\linewidth]{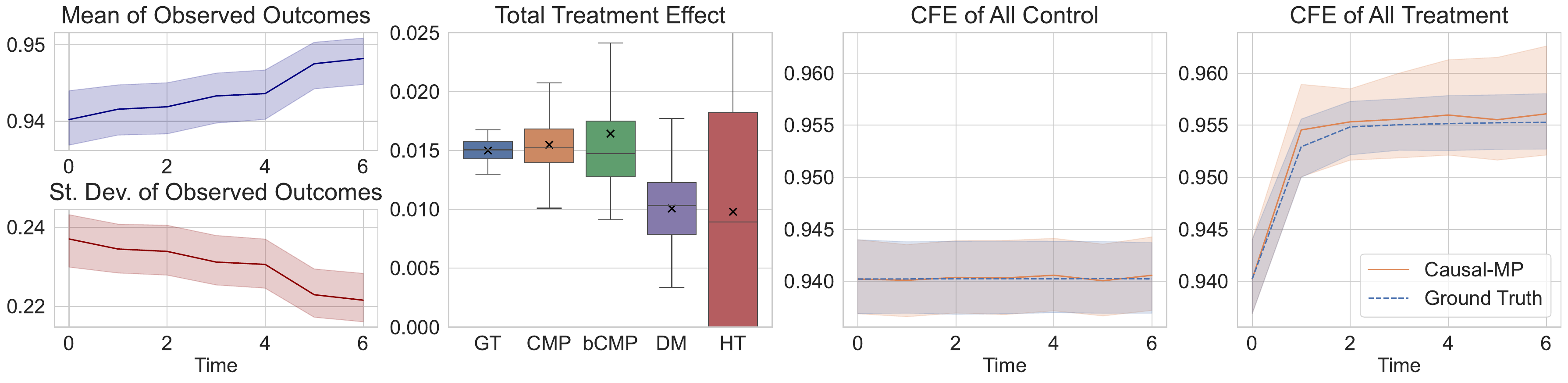}
    \caption{Belief adoption model with Topolcany network with $N=18,246$ users.}
    \label{fig:BAM2}
\end{figure}

\begin{figure}
    \centering
    \includegraphics[width=\linewidth]{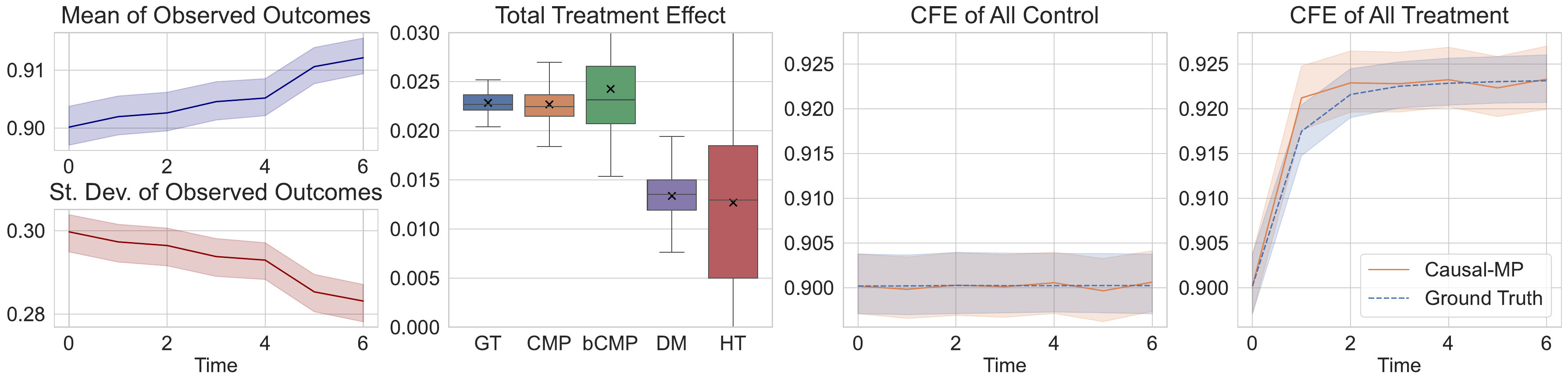}
    \caption{Belief adoption model with Zilina network with $N=42,971$ users.}
    \label{fig:BAM3}
\end{figure}

\begin{figure}
    \centering
    \includegraphics[width=0.99\linewidth]{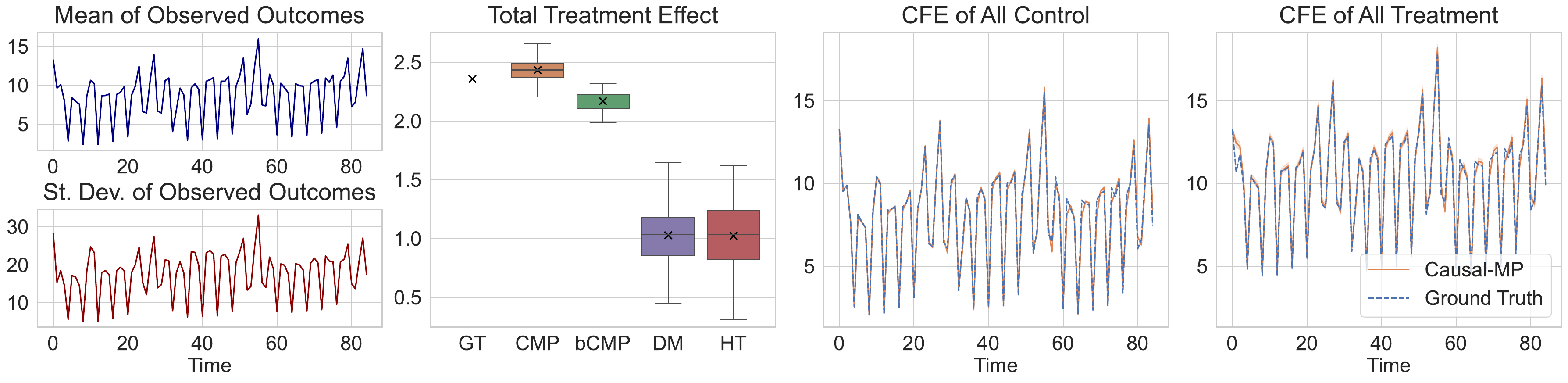}
    \caption{New York City Taxi model with $N=18,768$ Routes.}
    \label{fig:NYC_taxi}
\end{figure}

\begin{figure}
    \centering
    \includegraphics[width=\linewidth]{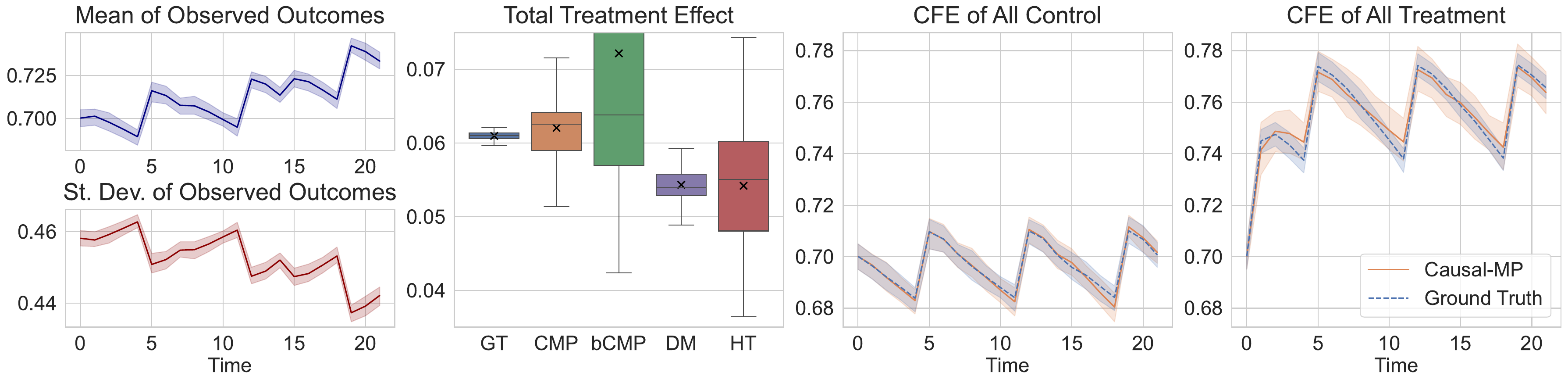}
    \caption{Exercise encouragement program with $N=30,162$ users.}
    \label{fig:EP}
\end{figure}

\begin{figure}
    \centering
    \includegraphics[width=\linewidth]{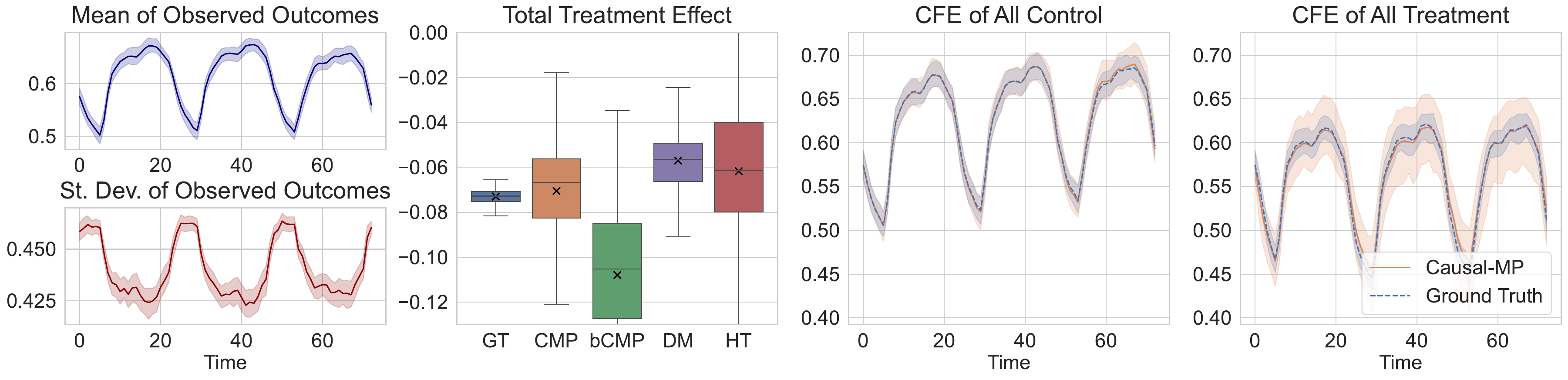}
    \caption{Data Center model with $N=2,000$ servers.}
    \label{fig:DC}
\end{figure}

Table~\ref{tab:implementation_details} provides implementation details for our experimental setup across six diverse scenarios described in \S\ref{sec:Benchmark_Toolbox}, with results presented in Figures~\ref{fig:LLM}--\ref{fig:DC}. Each experimental run executes one iteration of the respective setting, followed by applying our counterfactual cross-validation framework (Algorithm~\ref{alg:C-CV}) for estimation and validation. We outline our implementation strategy below and summarize the key empirical findings.

\begin{table}[htbp]
\centering
\footnotesize
\begin{tabular}{p{3.5cm}p{7.5cm}}
\toprule
\multicolumn{2}{c}{\textbf{Common Implementation Parameters}} \\
\midrule
Initial Condition & All scenarios start from $t=0$ with no treatment \\
\midrule
Treatment Design & Staggered rollout over three stages \\
\midrule
Baseline Methods & basic Causal Message-Passing (bCMP)\\ & Difference-in-Means (DM), Horvitz-Thompson (HT) \\
\bottomrule
\end{tabular}%

\vspace{0.02cm}

\begin{tabular}{p{3.5cm}p{7.5cm}}
\toprule
\multicolumn{2}{c}{\textbf{Validation Setup for Counterfactual Cross-Validation (Algorithm~\ref{alg:C-CV})}} \\
\midrule
Validation Batches & $b_v = 2$ batches \\
\midrule
Time Blocks & Three equally-length blocks \\
\midrule
Number of Batches ($\batchCount$) & 100, 500, 1000 \\
\midrule
Batch Sizes ($\batchSize$) &  5\%, 10\%, 20\%, 30\%, 50\% of population $N$ \\
\midrule
Model & Ridge regression with $\alpha \in \{10^{-4}, 10^{-2}, 1, 100\}$ \\
\bottomrule
\end{tabular}%

\vspace{0.02cm}

\begin{tabular}{p{4cm}>{\centering\arraybackslash}p{2.5cm}>{\centering\arraybackslash}p{3.5cm}>{\centering\arraybackslash}p{1cm}>{\centering\arraybackslash}p{1.8cm}}
\toprule
\multicolumn{5}{c}{\textbf{Scenario-Specific Implementation Details}} \\
\midrule
\textbf{Scenario} & \textbf{Population ($N$)} & \textbf{Each Stage Length} & \textbf{Runs} & \textbf{Design $\vec{p}$} \\
\midrule\midrule
LLM Social Network & 1,000 & 10 (total: $T=31$) & 10  & $(0.2, 0.5, 0.8)$ \\
\midrule
Belief Adoption - Krupina & 3,366 & \multirow{4}{*}{2 (total: $T=7$)} & \multirow{10}{*}{100} & \multirow{10}{*}{$(0.1, 0.2, 0.5)$} \\
\cmidrule{1-2}
Belief Adoption - Topolcany & 18,246 & & & \\
\cmidrule{1-2}
Belief Adoption - Zilina & 42,971 & & & \\
\cmidrule{1-3}
NYC Taxi Routes & 18,768 & 28 (total: $T=85$) & & \\
\cmidrule{1-3}
Exercise Encouragement & 30,162 & 7 (total: $T=22$) & & \\
\cmidrule{1-3}
Data Center & 2,000 & 24 (total: $T=73$) & & \\
\cmidrule{1-3}
Auction Model & 200 & 5 (total: $T=16$) & & \\

\bottomrule
\end{tabular}

\caption{Implementation Details}
\label{tab:implementation_details}
\end{table}

For the LLM-based social network, we conducted 10 distinct runs, constrained by OpenAI API limitations. Each run with $N=1,000$ and $T=31$ required approximately 100,000 GPT-3.5 API~calls to generate experimental and ground truth results. Each run employs a unique treatment allocation following a staggered rollout design across three stages with $\vec{p} = (0.2, 0.5, 0.8)$, each spanning 10 periods. This design ensures that on average 20\% of units receive intervention in the initial 10 periods, with an additional 30\% in the subsequent 10 periods, and so forth. All runs begin with a single historical observation with no treatment to initialize the recursive estimation algorithm.

For the remaining five experiments, we conducted 100 independent runs for each setting with treatment probabilities $\vec{p} = (0.1, 0.2, 0.5)$ across three stages (see Table~\ref{tab:implementation_details} for complete details). In the leftmost panel in Figures~\ref{fig:LLM}--\ref{fig:DC}, we display the temporal trajectories of observed outcomes through their mean and standard deviation, along with the 95th percentile across runs.

The second panels of Figures~\ref{fig:LLM}--\ref{fig:DC} display box plots of the average total treatment effect (TTE):
\begin{align}
    \label{eq:TTE}
    \text{TTE} :=
    \frac{1}{LN} \sum_{t=T-L+1}^T \sum_{i=1}^\UN 
    \left[
    \outcomeDW{\mathbf{1}}{i}{t}
    -
    \outcomeDW{\mathbf{0}}{i}{t}
    \right].
\end{align}
In each setting, we select $L$ so that the TTE in \eqref{eq:TTE} covers all periods in the last experimental stage. The results compare ground truth (GT) values against estimates from four methods: our proposed causal message-passing approach (CMP), which employs DPNB and cross-validation; basic causal message-passing (bCMP), which extends the estimator of \citet{shirani2024causal} without DPNB or cross-validation; the difference-in-means estimator (DM); and the Horvitz-Thompson estimator (HT)%
\footnote{Difference-in-means (DM) and Horvitz-Thompson (HT) are expressed as:
\begin{align*}
\DIME :=  \frac{1}{L} \sum_{t=T-L+1}^T 
\Big(
\frac{\sum_{i=1}^N\outcomeD{}{i}{t}\treatment{i}{t}}{\sum_{i=1}^N\treatment{i}{t}} - \frac{\sum_{i=1}^N\outcomeD{}{i}{t}(1-\treatment{i}{t})}{\sum_{i=1}^N(1-\treatment{i}{t})} \Big),
\quad\quad
\HTE :=  \frac{1}{LN} \sum_{t=T-L+1}^T \sum_{i=1}^N \left( \frac{\outcomeD{}{i}{t} \treatment{i}{t}}{\E[\treatment{i}{t}]} - \frac{\outcomeD{}{i}{t} (1 - \treatment{i}{t})}{\E[1 - \treatment{i}{t}]} \right).
\end{align*}}
\citep{savje2021average}. Finally, the rightmost panels display the counterfactual evolution (CFE) under ground truth and CMP estimates for all-control and all-treatment conditions, along with their respective 95th percentiles across different runs.


In implementing Algorithm~\ref{alg:C-CV}, we select candidate estimators, starting with a base model where each outcome is expressed as a linear function of two components: the sample mean of previous round outcomes and the current treatment allocation means. We then systematically modify this model by incorporating higher-order and interaction terms. Then, the parameters in Table~\ref{tab:implementation_details} are comprehensively combined to generate a diverse set of configurations, with time blocks aligned to experimental stages. For example, when $T=7$,
$\tblockList$ has three elements: $([0,3], [3, 5], [5,7])$.

Our framework demonstrates robust performance across all scenarios, successfully estimating counterfactual evolutions in the presence of strong seasonality patterns and without requiring information about the underlying interference network. As illustrated in Figures~\ref{fig:LLM}--\ref{fig:DC}, CMP yields estimates with smaller bias, closely tracking the ground truth in most scenarios. The effectiveness of our method is particularly evident in the challenging scenarios presented in Figures~\ref{fig:BAM1}--\ref{fig:BAM3}, where conventional estimators struggle to reliably determine the direction of treatment effects. Meanwhile, our method successfully estimates the treatment effect despite its small signal magnitude.

These results demonstrate our framework's ability to accurately estimate counterfactual evolutions and treatment effects across different experimental contexts. The robust performance, spanning social networks to distributed service systems like data centers, can empower decision makers with reliable causal insights when network interference undermines conventional approaches.

\begin{figure}
    \centering
    \includegraphics[width=\linewidth]{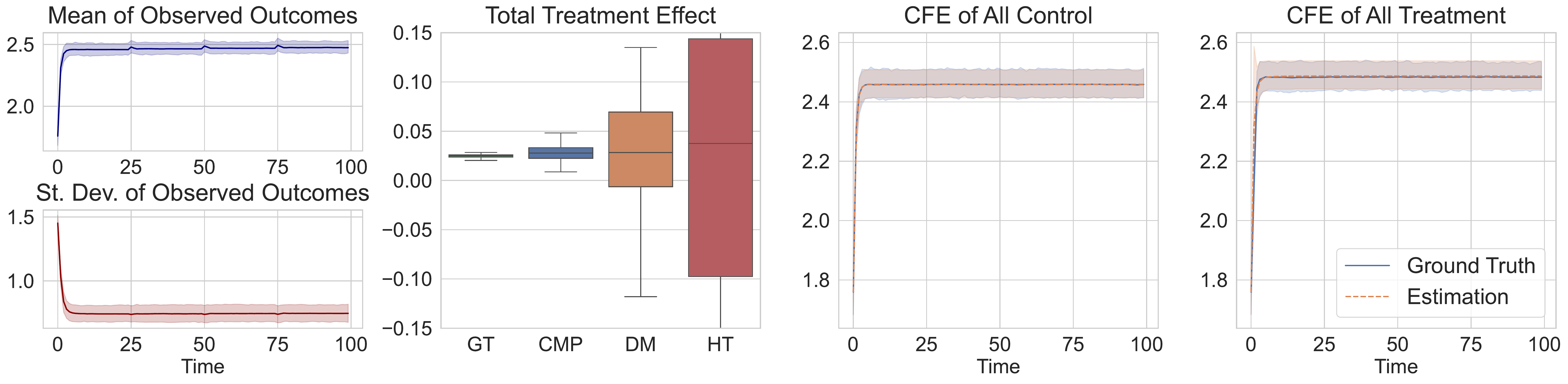}
    \caption{Auction model with $N=200$ objects.}
    \label{fig:auction}
\end{figure}

\begin{remark}
    \label{rem:batch_generation}
    Selecting a predetermined number of batches for a given batch size $\batchCount$ presents a significant computational challenge, particularly in large-scale problems with time-varying treatment allocations across units. For staggered rollout designs, we implement a heuristic approach while deferring comprehensive analysis to future research. Our heuristic consists of three steps. First, we order units by their treatment duration, defined as the number of time periods under treatment. Second, we select two blocks of size $\batchCount$: one that slides through the ordered list to cover all treatment durations, and another chosen randomly to ensure sufficient between-batch variation. Third, we select individual units from these merged blocks with equal probability to generate batches with average size $\batchCount$. This procedure maintains computational efficiency while ensuring batches with diverse treatment allocations with high probability. For more details, see Appendix~\ref{apndx:batching}.
\end{remark}

%% file: Conclusion.tex
\section{Concluding Remarks}
\label{sec:conclusion}
This work presents several contributions to the study of causal inference under network interference: a distribution-preserving network bootstrap approach that enables resampling from networked populations with unobserved interaction patterns, a counterfactual cross-validation framework to validate estimation methods, and an interference gym containing six experimental environments. Our empirical results suggest the framework can be effective across diverse settings, showing promising performance when handling temporal patterns and complex unobserved network structures. 

Future work could explore advanced machine learning architectures and incorporate domain-specific knowledge to further enhance the framework's practical utility while maintaining its theoretical properties. An important direction involves handling scenarios with more complex time trends, particularly where temporal patterns are significantly altered by treatment allocation, which would extend the framework's applicability to more challenging experimental contexts.

%% file: APPENDICES.tex
\newpage

\begin{center}
    {\Large \textbf{Appendices}}
\end{center}

The following appendices provide supporting material for the main paper, including detailed theoretical statements, rigorous proofs, and technical derivations that establish the foundation for our results.

\section*{Organization}
\label{appendices}
The first appendix outlines our heuristic batching technique for the DPNB implementation. The second appendix presents the technical results, beginning with essential notation and a reformulation of the outcome specification. We then establish the outcome decomposition rules through a sequence of proofs. This begins with Lemmas~\ref{lm:outcome_decomposition}-\ref{lm:apndx_outcome_decomposition}, culminating in the proof of Theorem~\ref{thm:outcome_decomposition} and Corollary~\ref{crl:SampleMean_decomposition}. The discussion then progresses to a rigorous statement of batch-level state evolution in \S\ref{apndx:batch_state_evolution}, that a simpler version of which was informally introduced in Theorem~\ref{thm:BSE_informal}. This is followed by a brief overview of the conditioning technique in \S\ref{apndx:Conditioning_Technique}, which is essential for proving the state evolution equation, as detailed in \S\ref{apndx:big_lemma}.

Appendix~\ref{sec:estimation_theory} demonstrates how consistent estimation of state evolution parameters enables consistent estimation of desired counterfactuals. We present the necessary assumptions, prove the main theoretical results, and detail the corresponding algorithms. We then analyze the application of these results to Bernoulli randomized designs, concluding in presenting three families of estimators in \S\ref{apndx:semi-recursive_estimators}, \S\ref{apndx:recursive_estimators}, and \S\ref{apndx:HO_recursive_estimators}.

Building on these results, \S\ref{sec:preprocessing} presents an extension to the causal message-passing methodology that addresses strong time-trends, enabling counterfactual estimation in the presence of seasonality or temporal patterns. The appendices conclude with \S\ref{apndx:auxiliary_results}, which presents auxiliary theorems necessary for our main proofs.

\section{Heuristic Batching Technique for DPNB}
\label{apndx:batching}
To implement DPNB effectively, we must construct batches that exhibit sufficient variation in treatment exposure levels. The following pseudocode demonstrates our approach for generating such batches under staggered rollout designs, where treatment allocation probabilities evolve systematically over time.

\begin{algorithm}
\caption{Create Random Batches for DPNB Implementation}
\label{alg:batching}
\begin{algorithmic}
    \Require Batch size $\batchSize$ and number of batches $\batchCount$

    \State Sort units by their treatment duration (number of time periods under treatment)

    \State Create $\batchCount$ evenly spaced positions along the sorted units
    
    \For{each batch $i$}
        \State Select a \emph{systematic} block of size $\batchSize$ starting at position $i$ in evenly spaced positions
        
        \State Select a \emph{random} block of size $\batchSize$ from anywhere in the dataset
        
        \State Merge the two blocks and remove duplicates to form candidate pool
        
        \State Calculate selection probability to achieve desired batch size on average
        
        \State Randomly select units from the candidate pool   
    \EndFor
    
    \Ensure $\batchCount$ distinct batches with average size of $\batchSize$
\end{algorithmic}
\end{algorithm}
Algorithm~\ref{alg:batching} works in three key steps. First, it orders all participants based on how long they've been receiving treatment. Second, it selects two groups of participants: one group that systematically slides through the ordered list to ensure coverage of all treatment durations, and another group chosen completely randomly to add variation. Finally, it combines these two groups and randomly selects individual participants from this pool with equal probability, creating batches that naturally contain diverse treatment allocations while maintaining computational efficiency, see Figure~\ref{fig:batch-algorithm}.
\usetikzlibrary{positioning, fit, shapes.geometric, calc, backgrounds}

\definecolor{sortedcolor}{RGB}{233, 236, 239}
\definecolor{systematiccolor}{RGB}{77, 171, 247}
\definecolor{randomcolor}{RGB}{255, 169, 77}
\definecolor{combinedcolor}{RGB}{130, 201, 30}
\definecolor{selectedcolor}{RGB}{190, 75, 219}

\begin{figure}[htbp]
\centering
\begin{tikzpicture}[
    unit/.style={draw, thick, minimum width=0.8cm, minimum height=0.8cm},
    unitbg/.style={unit, fill=sortedcolor},
    systematic/.style={unit, fill=systematiccolor},
    random/.style={unit, fill=randomcolor},
    combined/.style={unit, fill=combinedcolor},
    selected/.style={unit, fill=selectedcolor},
    box/.style={draw, rounded corners, inner sep=6pt},
    steptitle/.style={font=\bfseries},
    scale=0.9,
    transform shape
]

\node[steptitle, anchor=north west] (step1) at (-8,-1.5) {Step 1: Sort units by treatment duration};

\foreach \i [count=\xi] in {1,1,2,3,4,4,5,5,6,6,7,7,8,8,9,9,10,10} {
    \node[unitbg] (unit\xi) at (\xi*1-9,-2.5) {};
    \node[below=0.2cm of unit\xi] {\i};
}

\node[steptitle, anchor=north west] (step2a) at (-8,-4) {Step 2a: Select systematic block (slides through ordered list)};

\foreach \i [count=\xi] in {1,1,2,3,4,4,5,5,6,6,7,7,8,8,9,9,10,10} {
    \ifnum\xi>3
        \ifnum\xi<9
            \node[systematic] (sys\xi) at (\xi*1-9,-5) {};
        \else
            \node[unitbg] (sys\xi) at (\xi*1-9,-5) {};
        \fi
    \else
        \node[unitbg] (sys\xi) at (\xi*1-9,-5) {};
    \fi
    \node[below=0.2cm of sys\xi] {\i};
}

\node[steptitle, anchor=north west] (step2b) at (-8,-6.5) {Step 2b: Select random block};

\foreach \i [count=\xi] in {1,1,2,3,4,4,5,5,6,6,7,7,8,8,9,9,10,10} {
    \ifnum\xi>11
        \ifnum\xi<17
            \node[random] (rand\xi) at (\xi*1-9,-7.5) {};
        \else
            \node[unitbg] (rand\xi) at (\xi*1-9,-7.5) {};
        \fi
    \else
        \node[unitbg] (rand\xi) at (\xi*1-9,-7.5) {};
    \fi
    \node[below=0.2cm of rand\xi] {\i};
}

\node[steptitle, anchor=north west] (step3a) at (-8,-9) {Step 3a: Merge blocks to form candidate pool};

\foreach \i [count=\xi] in {1,1,2,3,4,4,5,5,6,6,7,7,8,8,9,9,10,10} {
    \ifnum\xi>3
        \ifnum\xi<9
            \node[combined] (comb\xi) at (\xi*1-9,-10) {};
        \else
            \ifnum\xi>11
                \ifnum\xi<17
                    \node[combined] (comb\xi) at (\xi*1-9,-10) {};
                \else
                    \node[unitbg] (comb\xi) at (\xi*1-9,-10) {};
                \fi
            \else
                \node[unitbg] (comb\xi) at (\xi*1-9,-10) {};
            \fi
        \fi
    \else
        \node[unitbg] (comb\xi) at (\xi*1-9,-10) {};
    \fi
    \node[below=0.2cm of comb\xi] {\i};
}

\node[steptitle, anchor=north west] (step3b) at (-8,-11.5) {Step 3b: Randomly select units from candidate pool};

\foreach \i [count=\xi] in {1,1,2,3,4,4,5,5,6,6,7,7,8,8,9,9,10,10} {
    \ifnum\xi=5
        \node[selected] (sel\xi) at (\xi*1-9,-12.5) {};
    \else
        \ifnum\xi=7
            \node[selected] (sel\xi) at (\xi*1-9,-12.5) {};
        \else
            \ifnum\xi=13
                \node[selected] (sel\xi) at (\xi*1-9,-12.5) {};
            \else
                \ifnum\xi=15
                    \node[selected] (sel\xi) at (\xi*1-9,-12.5) {};
                \else
                    \node[unitbg] (sel\xi) at (\xi*1-9,-12.5) {};
                \fi
            \fi
        \fi
    \fi
    \node[below=0.2cm of sel\xi] {\i};
}

\end{tikzpicture}
\caption{Visualization of the Random Batch Creation Algorithm for Staggered Rollout Designs.}
\label{fig:batch-algorithm}
\end{figure}

\section{Technical Results}
\label{sec:Technical_Results}
In this section, we delve into the analysis of the outcome specification in Eq.~\eqref{eq:outcome_function_matrix}. Our analysis draws upon various results from the literature on approximate message-passing algorithms \citep{donoho2009message,bayati2011dynamics, rush2018finite,li2022non}.

In the following, we first introduce several notations necessary for the rigorous presentation of our theoretical results. We then rewrite the potential outcome specification, facilitating the ensuing discussions. Next, we provide a rigorous proof for the outcome decomposition rule. This is rooted in the non-asymptotic results for the analysis of AMP algorithms \citep{li2022non} and can be seen as the finite sample analysis of the causal message-passing framework \citep{shirani2024causal}, within a broader class of outcome specifications.

\subsection{Notations}
\label{apndx:notations}
For any vector $\Vec{v} \in \R^n$, we denote its Euclidean norm as $\norm{\Vec{v}}$. For a fixed $k \geq 1$, we define $\poly{k}$ as the class of functions $f: \R^n \rightarrow \R$ that are continuous and exhibit polynomial growth of order $k$. In other words, there exists a constant $c$ such that $|f(\Vec{v})| \leq c(1+\norm{\Vec{v}}^k)$. Moreover, we consider a probability space $(\Omega, \F, \P)$, where $\Omega$ represents the sample space, $\F$ is the sigma-algebra of events, and $\P$ is the probability measure. We denote the expectation with respect to $\P$ as $\E$. Additionally, for any other probability measure~$p$, we use $\E_p$ to denote the expectation with respect to $p$. 

For any set $S$, the indicator function $\1_S(\omega)$ evaluates to $1$ if $\omega$ belongs to $S$, and $0$ otherwise. We define $\R^{n\times m}$ as the set of matrices with $n$ rows and $m$ columns. Given a matrix $\bm{M}$, we denote its transpose as~$\bm{M}^\top$. Additionally, we represent a matrix of ones with dimensions $n\times m$ as $\ones{n\times m} \in \R^{n\times m}$. The symbol $\eqd$ is used to denote equality in distribution, while $\eqas$ is used for equalities that hold almost surely with respect to the reference probability measure $\P$.

\subsection{Preliminaries}
\label{apndx:Preliminaries}
We initiate by rewriting the outcome model in Eq.~\eqref{eq:outcome_function_matrix}. Specifically, we consider the following more general specification:
\begin{equation}
    \label{eq:apndx_outcome_function}
    \begin{aligned}
        \VoutcomeD{}{}{t+1}(\Mtreatment{}{}) =
        \Voutcome{}{}{t+1} +
        \IMatMean{t} \outcomeg{t}{}\left(\VoutcomeD{}{}{0}(\Mtreatment{}{}), \ldots, \VoutcomeD{}{}{t}(\Mtreatment{}{}), \Mtreatment{}{}, \covar\right) +
        \outcomeh{t}{}\left(\VoutcomeD{}{}{0}(\Mtreatment{}{}), \ldots, \VoutcomeD{}{}{t}(\Mtreatment{}{}), \Mtreatment{}{}, \covar \right)
        + \Vnoise{}{t},
    \end{aligned}
\end{equation}
where
\begin{align}
    \label{eq:apndx_outcome_function_WOD}
    \Voutcome{}{}{t+1} = \big( \IM + \IMatT{t} \big) \outcomeg{t}{}\left(\VoutcomeD{}{}{0}(\Mtreatment{}{}), \ldots, \VoutcomeD{}{}{t}(\Mtreatment{}{}), \Mtreatment{}{}, \covar\right).
\end{align}
Above, $\IMatMean{t}$ is the matrix with the element $(\mu^{ij} + \mu_t^{ij})/N$ in the $i^{th}$ row and $j^{th}$ column. With a slight abuse of notation, we consider the entries of matrices $\IM$ and $\IMatT{t}$ to have zero mean. Formally, we state the following assumption regarding the interference matrices.
\begin{assumption}
    \label{asmp:apndx_Gaussian Interference Matrice}
    Entries of $\IM$ are i.i.d. Gaussian variables with zero mean and variance $\sigma^2/N$. Similarly, for any $t$, entries of $\IMatT{t}$ are i.i.d. Gaussian variables with zero mean and variance $\sigma^2_t/N$, independent of other components of the model.
\end{assumption}

Compared to Eq.~\eqref{eq:outcome_function_matrix}, the specification in Eqs.~\eqref{eq:apndx_outcome_function} and \eqref{eq:apndx_outcome_function_WOD} is more comprehensive, incorporating both the complete outcome history and the full treatment allocation matrix. This generalization enables us to capture more complex temporal dynamics, including additional lag terms and potential anticipation effects of treatments. The functions $\outcomeg{t}{}$ and $\outcomeh{t}{}$ can also be specified to include only a finite number $l \in \N$ of historical lag terms. While all subsequent results hold for the general model in \eqref{eq:apndx_outcome_function} and \eqref{eq:apndx_outcome_function_WOD}, for notational simplicity, we present the proofs using only one lag term $\VoutcomeD{}{}{t}(\Mtreatment{}{})$. Furthermore, when the context is clear, we omit the treatment matrix notation $\Mtreatment{}{}$ and write simply $\VoutcomeD{}{}{t}$ as the vector of outcomes at time $t$.

\subsection{Outcome Decomposition Rule}
\label{apndx:OD_Rule}
Fixing $N$, we proceed by setting more notations. Given $\VoutcomeD{}{}{0}$ as the vector of initial outcomes,  for $1\leq t < N$, define:
\begin{equation}
    \label{eq:apndx_orthonormal_outcomes_vectors}
    \VOCO{}{}{0} :=
    \frac{\outcomeg{0}{}\left(\VoutcomeD{}{}{0} ,\Mtreatment{}{}{}, \covar\right)}{\norm{\outcomeg{0}{}\left(\VoutcomeD{}{}{0} ,\Mtreatment{}{}{}, \covar\right)}},
    \quad\quad\quad\quad
    \VOCO{}{}{t} :=
    \frac{\left(\I_{N} - \MOCO{}{}{t-1}\MOCO{}{\top}{t-1}\right) \outcomeg{t}{}\left(\VoutcomeD{}{}{t} ,\Mtreatment{}{}{}, \covar\right)}{\norm{\left(\I_{N} - \MOCO{}{}{t-1}\MOCO{}{\top}{t-1}\right) \outcomeg{t}{}\left(\VoutcomeD{}{}{t} ,\Mtreatment{}{}{}, \covar\right)}},
\end{equation}
where, $\I_{N}$ is $N \x N$ identity matrix, and
\begin{equation}
    \label{eq:apndx_orthonormal_outcomes_matrix}
    \MOCO{}{}{t-1} = \left[\VOCO{}{}{0}\Big|\ldots\Big|\VOCO{}{}{t-1}\right].
\end{equation}
Note that $\I_{N} - \MOCO{}{}{t-1}\MOCO{}{\top}{t-1}$ functions as a projection onto the subspace that is orthogonal to the column space of $\MOCO{}{}{t-1}$. As a result, the vectors $\{\VOCO{}{}{0}, \ldots, \VOCO{}{}{N-1}\}$ constitute an orthonormal basis by definition. Therefore, we can represent the vector $\outcomeg{t}{}\left(\VoutcomeD{}{}{t} ,\Mtreatment{}{}{}, \covar\right)$ with respect to this basis as $\VPC{}{t} := (\PC{0}{t},\ldots,\PC{t}{t}, 0,\ldots,0)^\top \in \R^{N}$; that is:
\begin{equation}
    \label{eq:apndx_representation_in_the_new_basis}
    \outcomeg{t}{}\left(\VoutcomeD{}{}{t} ,\Mtreatment{}{}{}, \covar\right) =
    \sum_{j=0}^t \PC{j}{t} \VOCO{}{}{j},
    \quad\quad\quad\quad
    \PC{j}{t} = \pdot{\outcomeg{t}{}\left(\VoutcomeD{}{}{t} ,\Mtreatment{}{}{}, \covar\right)}{\VOCO{}{}{j}}.
\end{equation}
Then, it is immediate to get $\norm{\VPC{}{t}} = \normWO{\outcomeg{t}{}(\VoutcomeD{}{}{t} ,\Mtreatment{}{}{}, \covar)}$. In addition, we let $\MOCO{}{\perp}{t-1} \in \R^{N\times (N-t)}$ denote the orthogonal complement of $\MOCO{}{}{t-1}$ such that $(\MOCO{}{\perp}{t-1})^\top \MOCO{}{\perp}{t-1} = \I_{N-t}$.

We also define the following sequence of matrices based on the fixed interference matrix $\IM$:
\begin{equation}
    \label{eq:apndx_IM_recursion}
    \IMat{0} := \IM,
    \quad\quad\quad\quad
    \IMat{t} := \IMat{t-1} \left(\I_{N} - \VOCO{}{}{t-1}\VOCO{}{\top}{t-1}\right).
\end{equation}
Then,
Eq.~\eqref{eq:apndx_IM_recursion} enables us to write:
\begin{equation}
    \label{eq:apndx_representation_of_IM}
    \IMat{0} = \IMat{t} + \sum_{j=0}^{t-1} \left(\IMat{j}-\IMat{j+1}\right) = \IMat{t} + \sum_{j=0}^{t-1} \IMat{j} \VOCO{}{}{j}\VOCO{}{\top}{j}.
\end{equation}
Further, for $t=1, \ldots,N$, we define the following sequence of matrices:
\begin{equation}
    \label{eq:apndx_new_IM_matrix} 
    \IMatnew{t} := \IMat{t} \MOCO{}{\perp}{t-1}
    = \IMat{t-1} \left(\I_{N} - \VOCO{}{}{t-1}\VOCO{}{\top}{t-1}\right) \MOCO{}{\perp}{t-1}
    =
    \IMat{t-1} \MOCO{}{\perp}{t-1}
    = \ldots = \IM \MOCO{}{\perp}{t-1}
    \in \R^{N \x (N-t)},
\end{equation}
where we used the fact that $\VOCO{}{\top}{s-1} \MOCO{}{\perp}{t-1} = \Vec{0} \in \R^{N-t}$, for any $s \leq t$. Also, by \eqref{eq:apndx_IM_recursion}, we can write
\begin{equation}
    \label{eq:apndx_new_IM_matrix_2}
    \IMat{t} = \IMat{t-1} \left(\I_{N} - \VOCO{}{}{t-1}\VOCO{}{\top}{t-1}\right) = \ldots = \IM \left(\I_{N} - \MOCO{}{}{t-1}\MOCO{}{\top}{t-1}\right) = \IM \MOCO{}{\perp}{t-1} (\MOCO{}{\perp}{t-1})^\top = \IMatnew{t} (\MOCO{}{\perp}{t-1})^\top.
\end{equation}

We prove the outcome decomposition rule in multiple steps, beginning with the following lemma.
\begin{lemma}
    \label{lm:outcome_decomposition}
    For any $t = 0, \ldots, N-1$, we have
    \begin{equation}
        \label{eq:apndx_outcome_decomposition}
        \VoutcomeD{}{}{t+1} =
        \IMatMean{t} \outcomeg{t}{}\left(\VoutcomeD{}{}{t}, \Mtreatment{}{}{}, \covar{} \right)
        +
        \sum_{j=0}^{t} \PC{j}{t} \left(\IMat{j} + \IMatT{t}\right) \VOCO{}{}{j} 
        + \outcomeh{t}{}\left(\VoutcomeD{}{}{t}, \Mtreatment{}{}{}, \covar{} \right)
        + \Vnoise{}{t}.
    \end{equation}
\end{lemma}
\textbf{Proof.} By outcome model given in \eqref{eq:apndx_outcome_function} and \eqref{eq:apndx_representation_of_IM}, we can write:
\begin{equation*}
    \begin{aligned}
    \VoutcomeD{}{}{t+1} 
    &=
    \IMat{t} \outcomeg{t}{}\left(\VoutcomeD{}{}{t}, \Mtreatment{}{}{}, \covar\right)
    + \sum_{j=0}^{t-1} \IMat{j} \VOCO{}{}{j}\VOCO{}{\top}{j} \outcomeg{t}{}\left(\VoutcomeD{}{}{t}, \Mtreatment{}{}{}, \covar\right)
    + \big(\IMatMean{t} + \IMatT{t}\big) \outcomeg{t}{}\left(\VoutcomeD{}{}{t}, \Mtreatment{}{}{}, \covar\right)
    + \outcomeh{t}{}\left(\VoutcomeD{}{}{t}, \Mtreatment{}{}{}, \covar{} \right)
    + \Vnoise{}{t}
    \\
    &=
    \IMat{t} \outcomeg{t}{}\left(\VoutcomeD{}{}{t}, \Mtreatment{}{}{}, \covar\right)
    + \sum_{j=0}^{t-1} \PC{j}{t} \IMat{j} \VOCO{}{}{j}
    + \sum_{j=0}^{t} \PC{j}{t} \IMatT{t} \VOCO{}{}{j}
    + \IMatMean{t} \outcomeg{t}{}\left(\VoutcomeD{}{}{t}, \Mtreatment{}{}{}, \covar{} \right)
    + \outcomeh{t}{}\left(\VoutcomeD{}{}{t}, \Mtreatment{}{}{}, \covar{} \right)
    + \Vnoise{}{t}
    \\
    &=
    \IMatMean{t} \outcomeg{t}{}\left(\VoutcomeD{}{}{t}, \Mtreatment{}{}{}, \covar{} \right)
    +
    \sum_{j=0}^{t} \PC{j}{t} (\IMat{j}+\IMatT{t}) \VOCO{}{}{j}
    + \outcomeh{t}{}\left(\VoutcomeD{}{}{t}, \Mtreatment{}{}{}, \covar{} \right)
    + \Vnoise{}{t},
    \end{aligned}
\end{equation*}
where in the third line, we used \eqref{eq:apndx_representation_in_the_new_basis} and the fact that the vectors $\{\VOCO{}{}{0}, \ldots, \VOCO{}{}{t}\}$ constitute an orthonormal set. Also, in the last line, we utilized the following fact
\begin{equation*}
    \IMat{t} \outcomeg{t}{}\left(\VoutcomeD{}{}{t}, \Mtreatment{}{}{}, \covar\right)
    = \IMat{t} \left(\I_{N} - \MOCO{}{}{t-1}\MOCO{}{\top}{t-1}\right) \outcomeg{t}{}\left(\VoutcomeD{}{}{t}, \Mtreatment{}{}{}, \covar\right)
    = \PC{t}{t} \IMat{t} \VOCO{}{}{t},
\end{equation*}
that holds true because $\IMat{t} \MOCO{}{}{t-1}\MOCO{}{\top}{t-1} = 0$ by \eqref{eq:apndx_IM_recursion}; additionally, the term $\left(\I_{N} - \MOCO{}{}{t-1}\MOCO{}{\top}{t-1}\right) \outcomeg{t}{}\left(\VoutcomeD{}{}{t}, \Mtreatment{}{}{}, \covar\right)$ is equal to the projection of $\outcomeg{t}{}\left(\VoutcomeD{}{}{t}, \Mtreatment{}{}{}, \covar\right)$ on the subspace perpendicular to the column space of $\MOCO{}{}{t-1}$, which is $\PC{t}{t} \VOCO{}{}{t}$. It completes the proof. \ep

The next lemma characterizes the distribution of $\IMatnew{t}$, defined in \eqref{eq:apndx_new_IM_matrix}, based on the rotational invariance property of Gaussian matrices. 
\begin{lemma}
    \label{lm:new_IM_distribution}
    Fix $1 \leq t < N$. Conditional on $\VoutcomeD{}{}{0}$, $\VoutcomeD{}{}{1}, \ldots, \VoutcomeD{}{}{t-1}$, $\Mtreatment{}{}{}$, and $\covar$, entries of the matrix $\IMatnew{t} \in \R^{N \times (N-t)}$ are i.i.d. with distribution $\Nc(0,{\sigma^2}/ {N})$, and the matrix $\IMatnew{t}$ (and so $\IMat{t}$) is independent of $\VoutcomeD{}{}{t},\VOCO{}{}{t},$ as well as $\IMat{0}\VOCO{}{}{0}, \ldots, \IMat{t-1}\VOCO{}{}{t-1}$. 
\end{lemma}
\textbf{Proof.} First, note that given $\Mtreatment{}{}{}$ and $\covar$, by \eqref{eq:apndx_orthonormal_outcomes_vectors}, conditioning on $\VoutcomeD{}{}{0}$, $\VoutcomeD{}{}{1}, \ldots, \VoutcomeD{}{}{t-1}$, is equivalent to conditioning on $\VOCO{}{}{0}$, $\VOCO{}{}{1}, \ldots, \VOCO{}{}{t-1}$. Now, we use an induction on $t$ to prove the result.

\textbf{Step 1.} Let $t=1$. By \eqref{eq:apndx_new_IM_matrix} and the rotational invariance property of Gaussian matrices, we have
\begin{equation}
    \label{eq:apndx_proof_new_IM_distribution_1}
    \IMatnew{1} = \IM \MOCO{}{\perp}{0} \eqd \IM \sbasis{\perp}{1},
\end{equation}
where $\sbasis{\perp}{1}$ denotes the orthogonal complement of $\sbasis{}{1}$, which is the first standard basis vector. Letting $\sbasis{\perp}{1} = [\sbasis{}{2}|\ldots|\sbasis{}{N}]$, by Assumption~\ref{asmp:apndx_Gaussian Interference Matrice}, we get that the entries of $\IMatnew{1} \in \R^{N \times (N-1)}$ are i.i.d. with distribution $\Nc(0,{\sigma^2}/ {N})$. Furthermore, considering that $\MOCO{}{}{0} = \VOCO{}{}{0}$, the matrix $\IMatnew{1}$ is independent from $\IMat{0}\VOCO{}{}{0}$. Then, conditional on $\VoutcomeD{}{}{0}$, the outcome model in \eqref{eq:apndx_outcome_function} implies that $\IMatnew{1}$ is independent of $\VoutcomeD{}{}{1}$ and $\VOCO{}{}{1}$ (note that the randomness of $\VoutcomeD{}{}{1}$ comes from $\IMat{0}\VOCO{}{}{0}$, $\IMatT{0}$, and $\Vnoise{}{0}$). Finally, considering \eqref{eq:apndx_new_IM_matrix_2}, the same results about the independency hold true for the matrix $\IMat{1}$.

\textbf{Step 2.} Suppose that the result is true for $t=1, \ldots, s-1$. Given $\VoutcomeD{}{}{0}, \ldots, \VoutcomeD{}{}{s-1}$, we show the result also holds for $t=s$. Note that by the induction hypothesis, conditional on $\VoutcomeD{}{}{0}, \ldots, \VoutcomeD{}{}{s-2}$, the matrix $\IMat{s-1}$ is independent of $\VoutcomeD{}{}{s-1}$ and $\IMat{0}\VOCO{}{}{0}, \ldots, \IMat{s-2}\VOCO{}{}{s-2}$. Thus, $\IMat{s-1}$ and $\VoutcomeD{}{}{s-1}$ are conditionally independent. This implies that conditional on $\VoutcomeD{}{}{0}, \ldots, \VoutcomeD{}{}{s-1}$ (we added $\VoutcomeD{}{}{s-1}$), the matrix $\IMat{s-1}$ (and so $\IMat{s} := \IMat{s-1} (\I_N - \VOCO{}{}{s-1}\VOCO{}{\top}{s-1})$, see \eqref{eq:apndx_IM_recursion}, as well as $\IMatnew{s}$) is still independent of $\IMat{0}\VOCO{}{}{0}, \ldots, \IMat{s-2}\VOCO{}{}{s-2}$.

Next, we show that $\IMat{s}$ is also independent from $\IMat{s-1}\VOCO{}{}{s-1}$. By \eqref{eq:apndx_new_IM_matrix} and the rotational invariance property of Gaussian matrices, we can write
\begin{equation*}
    \IMatnew{s} = \IM \MOCO{}{\perp}{s-1} \eqd \IM [\sbasis{}{1}| \ldots| \sbasis{}{s}]^\perp.
\end{equation*}
Here, $[\sbasis{}{1}| \ldots| \sbasis{}{s}]^\perp$ represents the orthogonal complement of the first $s$ standard basis vectors. Then, a similar argument to the one in Step 1 implies that the matrix $\IMatnew{s} \in \R^{N\times (N-s)}$ has i.i.d. entries with a distribution of $\Nc(0,{\sigma^2}/ {N})$. Furthermore, it yields that $\IMatnew{s}$ (and consequently $\IMat{s}$, see Eq.~\eqref{eq:apndx_new_IM_matrix_2}) is independent of the vector $\IMat{s-1}\VOCO{}{}{s-1} = \IM (\I_N - \MOCO{}{}{s-2} \MOCO{}{\top}{s-2}) \VOCO{}{}{s-1} = \IM \VOCO{}{}{s-1}$; this holds true because of Eq.~\eqref{eq:apndx_new_IM_matrix_2} and the fact that $\VOCO{}{\top}{j} \VOCO{}{}{s-1} =0$, for $j = 0, \ldots, s-2$. 

Now, by Lemma~\ref{lm:outcome_decomposition}, we have
\begin{equation*}
    \VoutcomeD{}{}{s} =
    \IMatMean{s-1} \outcomeg{s-1}{}\left(\VoutcomeD{}{}{s-1}, \Mtreatment{}{}{}, \covar{} \right)
    \sum_{j=0}^{s-1} \PC{j}{s-1} \left(\IMat{j} + \IMatT{s-1}\right) \VOCO{}{}{j}
    + \outcomeh{s-1}{}\left(\VoutcomeD{}{}{s-1}, \Mtreatment{}{}{}, \covar{} \right)
    + \Vnoise{}{s-1}.
\end{equation*}
As a result, $\IMat{s}$ and $\IMatnew{s}$ are also independent of $\VoutcomeD{}{}{s}$ and $\VOCO{}{}{s}$. This concludes the proof. \ep

By combining the results of Lemmas~\ref{lm:outcome_decomposition} and~\ref{lm:new_IM_distribution}, we arrive at the conclusion of Lemma~\ref{lm:apndx_outcome_decomposition}.
\begin{lemma}
    \label{lm:apndx_outcome_decomposition}
    For any $t = 0, \ldots, N-1$, we have
    \begin{equation}
        \label{eq:apndx_outcome_decomposition_distribtuion}
        \VoutcomeD{}{}{t+1} =
        \IMatMean{t} \outcomeg{t}{}\left(\VoutcomeD{}{}{t}, \Mtreatment{}{}{}, \covar{} \right) +
        \outcomeh{t}{}\left(\VoutcomeD{}{}{t}, \Mtreatment{}{}{}, \covar{} \right) +
        \sqrt{\sigma^2+\sigma_t^2} \norm{\outcomeg{t}{}(\VoutcomeD{}{}{t} ,\Mtreatment{}{}{}, \covar)} \sum_{j=0}^{t} \NPC{j}{t} \Vec{Z}_j  + \Vnoise{}{t},
    \end{equation}
    where $\Vec{Z}_0, \Vec{Z}_1, \ldots, \Vec{Z}_t$ are i.i.d. random vectors in $\R^N$ following $\Nc(0,\frac{1}{N}\I_N)$ distribution. Additionally, $\NPC{j}{t} := \PC{j}{t}/\normWO{\outcomeg{t}{}(\VoutcomeD{}{}{t}, \Mtreatment{}{}, \covar)}$, making $\VNPC{}{t} = (\NPC{0}{t}, \ldots, \NPC{t}{t}, 0, \ldots, 0)^\top \in \R^N$ a unit random vector (i.e., $\normWO{\VNPC{}{t}} = 1$). Note that $\NPC{j}{t}$ and $\Vec{Z}_j$ are not independent.
\end{lemma}
\textbf{Proof.}
Fixing $t$, we prove the result in two steps. First, we demonstrate that $\IMat{j} \VOCO{}{}{j}$, for $j=0, \ldots, t$, follows a Gaussian distribution with specified mean and variance. Then, we show that $\IMat{0} \VOCO{}{}{0}, \ldots, \IMat{t} \VOCO{}{}{t}$ are independent. 

Conditional on $\VoutcomeD{}{}{0}$, $\VoutcomeD{}{}{1}, \ldots, \VoutcomeD{}{}{j-1}$, $\Mtreatment{}{}{}$, and $\covar$, Lemma~\ref{lm:new_IM_distribution} implies that the matrix $\IMat{j}$ and the vector $\VOCO{}{}{j}$ are independent. Also, by  \eqref{eq:apndx_new_IM_matrix_2}, we can write
\begin{equation*}
    \IMat{j} \VOCO{}{}{j} =
    \IM \MOCO{}{\perp}{j-1} (\MOCO{}{\perp}{j-1})^\top  \VOCO{}{}{j} = \IM \VOCO{}{}{j},
\end{equation*}
where we used the fact that the vector $\VOCO{}{}{j}$ is perpendicular to the column space of the matrix $\MOCO{}{}{j-1}$. Thus, conditional on the value of $\VOCO{}{}{j}$, as well as $\VoutcomeD{}{}{0}$, $\VoutcomeD{}{}{1}, \ldots, \VoutcomeD{}{}{j-1}$, $\Mtreatment{}{}{}$, and $\covar$, by the rotational invariance property of Gaussian matrices, the elements of $\IMat{j} \VOCO{}{}{j}$ are i.i.d. random variables with distribution $\Nc(0,\frac{\sigma^2}{N})$. Furthermore, note that this conditional distribution of $\IMat{j} \VOCO{}{}{j}$ remains the same regardless of the value of $\VOCO{}{}{j}$. As a result, we can conclude that the elements of $\IMat{j} \VOCO{}{}{j}$ are i.i.d. Gaussian, even without conditioning on $\VOCO{}{}{j}$ as well as $\VoutcomeD{}{}{0}$, $\VoutcomeD{}{}{1}, \ldots, \VoutcomeD{}{}{j-1}$, $\Mtreatment{}{}{}$, and $\covar$. Precisely, for a deterministic vector $\Vec{v}$, we can write:
\begin{align*}
    \P
    \left(
    \IMat{j} \VOCO{}{}{j} \leq \Vec{v}
    \right)
    =
    \E
    \left[
    \E
    \left[
    \P
    \left(
    \IMat{j} \VOCO{}{}{j} \leq \Vec{v}
    \right)
    \Big|
    \VOCO{}{}{j}
    \right]
    \bigg|
    \VoutcomeD{}{}{0}, \ldots, \VoutcomeD{}{}{j-1}, \Mtreatment{}{}{}, \covar
    \right]
    =
    \E
    \left[
    \E
    \left[
    \Phi(\Vec{v})
    \Big|
    \VOCO{}{}{j}
    \right]
    \bigg|
    \VoutcomeD{}{}{0}, \ldots, \VoutcomeD{}{}{j-1}, \Mtreatment{}{}{}, \covar
    \right]
    =
    \Phi(\Vec{v}),
\end{align*}
where $\Phi$ denotes the CDF of a vector whose entries follow a normal distribution $\Nc(0, \frac{\sigma^2}{N})$.

We proceed by establishing the independence property. Note that by Lemma~\ref{lm:new_IM_distribution}, conditional on the values of $\VoutcomeD{}{}{0}, \ldots, \VoutcomeD{}{}{t}, \Mtreatment{}{}{}, \covar$ (and so on the values of $\VOCO{}{}{0}, \ldots, \VOCO{}{}{t-1}, \VOCO{}{}{t}$), it follows that $\IMat{t}$ (and so $\IMat{t} \VOCO{}{}{t}$) is independent of $\IMat{0}\VOCO{}{}{0}, \ldots, \IMat{t-1}\VOCO{}{}{t-1}$. Importantly, our previous demonstration confirmed that the distribution of $\IMat{t} \VOCO{}{}{t}$ remains unchanged across different values of $\VoutcomeD{}{}{0}, \ldots, \VoutcomeD{}{}{t}, \Mtreatment{}{}{}, \covar$. Consequently, we can assert that the random vector $\IMat{t}\VOCO{}{}{t}$ is independent of $\IMat{0}\VOCO{}{}{0}, \ldots, \IMat{t-1}\VOCO{}{}{t-1}$. More precisely, we can repeat this argument multiple times and, for deterministic vectors $\Vec{v}_0, \ldots, \Vec{v_t}$, show that
\begin{align*}
    \P
    \left(
    \IMat{0} \VOCO{}{}{0} \leq \Vec{v}_0,
    \ldots,
    \IMat{t} \VOCO{}{}{t} \leq \Vec{v}_t
    \right)
    &=
    \E
    \left[
    \P
    \left(
    \IMat{0} \VOCO{}{}{0} \leq \Vec{v}_0,
    \ldots,
    \IMat{t} \VOCO{}{}{t} \leq \Vec{v}_t
    \right)
    \Big|
    \VoutcomeD{}{}{0}, \ldots, \VoutcomeD{}{}{t}, \Mtreatment{}{}{}, \covar
    \right]
    \\
    &=
    \E
    \left[
    \P
    \left(
    \IMat{0} \VOCO{}{}{0} \leq \Vec{v}_0,
    \ldots,
    \IMat{t-1} \VOCO{}{}{t-1} \leq \Vec{v}_{t-1}
    \right)
    \Phi(\Vec{v}_t)
    \Big|
    \VoutcomeD{}{}{0}, \ldots, \VoutcomeD{}{}{t}, \Mtreatment{}{}{}, \covar
    \right]
    \\
    &=
    \Phi(\Vec{v}_t)
    \E
    \left[
    \P
    \left(
    \IMat{0} \VOCO{}{}{0} \leq \Vec{v}_0,
    \ldots,
    \IMat{t-1} \VOCO{}{}{t-1} \leq \Vec{v}_{t-1}
    \right)
    \Big|
    \VoutcomeD{}{}{0}, \ldots, \VoutcomeD{}{}{t-1}, \Mtreatment{}{}{}, \covar
    \right]
    \\
    &=
    \ldots
    \\
    &=
    \Phi(\Vec{v}_0)
    \Phi(\Vec{v}_1)
    \ldots
    \Phi(\Vec{v}_t).
\end{align*}

Based on the fact that the time-dependent interference matrix $\IMatT{t}$ and the noise vector $\Vnoise{}{t}$ are independent of everything else in the model, the proof is complete and we obtain the desired result in \eqref{eq:apndx_outcome_decomposition_distribtuion}. \ep

The dependence between $\NPC{j}{t}$ and $\Vec{Z}_j$ in Lemma~\ref{lm:apndx_outcome_decomposition} implies that the Gaussianity of the elements in $\VoutcomeD{}{}{t+1}$ cannot be inferred directly from the result of this lemma alone.

\noindent
\textbf{Proof of Theorem~\ref{thm:outcome_decomposition}.}
To obtain the desired result, we apply Lemma~\ref{lm:apndx_outcome_decomposition} and Lemma~\ref{lm:Gap_with_Gaussian_vector} together. To be more specific, in view of Lemma~\ref{lm:apndx_outcome_decomposition}, we know that
\begin{equation*}
    \VoutcomeD{}{}{t+1} =
    \IMatMean{t} \outcomeg{t}{}\left(\VoutcomeD{}{}{t}, \Mtreatment{}{}{}, \covar{} \right) +
    \outcomeh{t}{}\left(\VoutcomeD{}{}{t}, \Mtreatment{}{}{}, \covar{} \right) +
    \sqrt{\sigma^2+\sigma_t^2} \norm{\outcomeg{t}{}(\VoutcomeD{}{}{t} ,\Mtreatment{}{}{}, \covar)}  \Sumvec{}{t} + \Vnoise{}{t},
\end{equation*}
where $\Sumvec{}{t} = \sum_{i=0}^{t} \NPC{i}{t} \Vec{Z}_i$. But, by Lemma~\ref{lm:Gap_with_Gaussian_vector}, we have
\begin{equation*}
    W_1\left(\law\left(\Sumvec{}{t}\right),\Nc\left(0,\frac{1}{N}\I_N\right)\right) \leq c \sqrt{\frac{t \log N}{N}},
\end{equation*}
which concludes the proof. \ep

\noindent
\textbf{Proof of Corollary~\ref{crl:SampleMean_decomposition}.}
By the result of Theorem~\ref{thm:outcome_decomposition}, we can write
\begin{align*}
    \frac{1}{\cardinality{\batch}} \sum_{i \in \batch} \outcomeD{}{i}{t+1}
    =
    \;&\frac{1}{N \cardinality{\batch}} \sum_{i \in \batch} \sum_{j=1}^N \left(\mu^{ij}+\mu_t^{ij}\right) \outcomeg{t}{}\left(\outcomeD{}{j}{t}, \Vtreatment{j}{}{}, \Vcovar{j} \right)
    +
    \frac{1}{\cardinality{\batch}} \sum_{i \in \batch} \outcomeh{t}{}\left(\outcomeD{}{i}{t}, \Vtreatment{i}{}{}, \Vcovar{i} \right)
    \\ \;&+
    \sqrt{\frac{\sigma^2+\sigma_t^2}{\cardinality{\batch}}} \norm{\outcomeg{t}{}(\VoutcomeD{}{}{t} ,\Mtreatment{}{}{}, \covar)} \frac{1}{\sqrt{\cardinality{\batch}}} \sum_{i \in \batch} \sumvec{i}{t}
    +
    \frac{1}{\cardinality{\batch}} \sum_{i \in \batch} \noise{i}{t}.
\end{align*}
Letting $\avesumvec{}{t} := \frac{1}{\sqrt{\cardinality{\batch}}} \sum_{i \in \batch} \sumvec{i}{t}$ and applying Lemma~\ref{lm:Gap_with_Gaussian} for $\Vec{\Phi} = \Sumvec{}{t}$, we get the result. \ep

\subsection{Batch-level State Evolution}
\label{apndx:batch_state_evolution}
Next, we analyze the large-sample behavior of the outcomes for a subpopulation of units. Specifically, let $\batch \subset [N]$ represent an arbitrary subpopulation\footnote{This generalizes the result in \S\ref{sec:DPNB} which focuses on treatment-allocation-based subpopulations.}, with its size $\cardinality{\batch}$ increasing indefinitely as the population size $N$ grows large to infinity. We investigate the asymptotic behavior of the elements in $\Moutcome{}{}{}$ as $N$ approaches infinity. This provides valuable insights into the evolution of outcomes within the subpopulation.
\begin{assumption}
    \label{asmp:BL}
    Fixing $T \in \N$ and $k \geq 2$, we assume that
    \begin{enumerate}[label=(\roman*)]
        \item \label{asmp:BL-pl functions} For all $t\in[T]$, the function $\outcomeg{t}{}:\R \times \R^{T+1} \times \R^{\dcovar}\rightarrow \R$ is a $\poly{\frac{k}{2}}$ function.

        \item \label{asmp:BL-pl h-functions} For all $t\in[T]$, the function $\outcomeh{t}{}:\R \times \R^{T+1} \times \R^{\dcovar}\rightarrow \R$ is a $\poly{1}$ function.

        \item \label{asmp:BL-bound on initials} The sequence of initial outcome vectors $\VoutcomeD{}{}{0}$, the treatment allocation matrices~$\Mtreatment{}{}$, the covariates $\covar$, and the function $\outcomeg{0}{}$ are such that for a deterministic value $\MVVO{}{}{1}>0$, we have
        \begin{align*}
            (\MVVO{}{}{1})^2
            &=
            \lim_{N\rightarrow \infty}
            \frac{\sigma^2+\sigma_0^2}{N} \sum_{i=1}^N
            \outcomeg{0}{}\big(
            \outcomeD{}{i}{0},\Vtreatment{i}{},\Vcovar{i}
            \big)^2 < \infty.
        \end{align*}
    \end{enumerate}
\end{assumption}
Assumption~\ref{asmp:BL} comprises a collection of regularity conditions on the model attributes. The first two parts ensure that the functions $\outcomeg{t}{}$ and $\outcomeh{t}{}$ do not demonstrate fast explosive behavior, thereby ensuring the well-posedness of the large system asymptotic. The final part pertains to the initial observation of the network and corroborates that the function $\outcomeg{0}{}$ is non-degenerate. This ensures that initial observations provide meaningful information and contribute to the evolution of outcomes.
\begin{assumption}
    \label{asmp:weak_limits}
    Fix $T \in \N$, $k \geq 2$, and a subpopulation $\batch \subset [N]$, where the size $\cardinality{\batch}$ grows to infinity as $N \rightarrow \infty$. We assume the following statements hold:
    \begin{enumerate}[label=(\roman*)]
        \item Let $p_{y}^{\batch}(N)$ denote the empirical distribution of the \textbf{initial outcomes} $\outcomeD{}{i}{0}$ with $i \in \batch$, in a system with $N$ units. Then, $p_{y}^{\batch}(N)$ converges weakly to a probability measure $p_{y}^{\batch}$.

        \item Let $p_{x}^{\batch}(N)$ denote the empirical distribution of the \textbf{covariate} vectors $\Vcovar{i}$ with $i \in \batch$, in a system with $N$ units. Then, $p_{x}^{\batch}(N)$ converges weakly to a probability measure $p_{x}^{\batch}$.

        \item Let $p_{w}^{\batch}(N)$ denote the empirical distribution of the \textbf{treatment} vectors $\Vtreatment{i}{}$ with $i \in \batch$, in a system with $N$ units. Then, $p_{w}^{\batch}(N)$ converges weakly to a probability measure $p_{w}^{\batch}$.
        
        \item \label{asmp:interference_element_convergence} For all $i$ and any $t\in [T]_0$, let $p_{\mu^i}(N)$ and $p_{\mu_t^i}(N)$ be, respectively, the empirical distribution of the elements of the vectors $\Vec{\mu}^{\;i\cdot} := (\mu^{i1}, \ldots, \mu^{iN})^\top$ and $\Vec{\mu}^{\;i\cdot}_t := (\mu^{i1}_t, \ldots, \mu^{iN}_t)^\top$. Then, $p_{\mu^i}(N)$ converges weakly to $p_{\mu^i}$ and $p_{\mu_t^i}(N)$ converges weakly to $p_{\mu_t^i}$.
        
        \item \label{asmp:mean_interference_element_convergence}
        Let $\MIM{}{i} \sim p_{\mu^i}$ and $\MIM{t}{i} \sim p_{\mu^i_t}$; then, $\MIM{}{i}$ and $\MIM{t}{i}$ are independent of all other randomnesses in the model. Additionally, denoting $\bar{\mu}^i = \E[\MIM{}{i}]$ and $\bar{\mu}^i_t = \E[\MIM{t}{i}]$, let $p_{\mu}^{\batch}(N)$ and $p_{\mu_t}^{\batch}(N)$ be respectively the empirical distributions of $\bar{\mu}^i$ and $\bar{\mu}^i_t$ with $i \in \batch$ and $i \leq N$. Then, $p_{\mu}^{\batch}(N)$ and $p_{\mu_t}^{\batch}(N)$ converge weakly to probability measures $p_{\mu}^{\batch}$ and $p_{\mu_t}^{\batch}$, respectively.

        \item \label{asmp:k_moment_convergence} For all $i$ and $t\in [T]_0$, suppose that
        \begin{align*}
            \left(\outcomeD{}{\batch}{0}(N), \Vcovar{\batch}(N), \Vtreatment{\batch}{}(N), \MIM{}{i}(N), \MIM{t}{i}(N), \MIM{}{\batch}(N), \MIM{t}{\batch}(N)\right)^\top
            \\
            \sim p_{y}^{\batch}(N) \times p_{x}^{\batch}(N) \times p_{w}^{\batch}(N) \times p_{\mu^i}(N) \times p_{\mu_t^i}(N) \times p_{\mu}^{\batch}(N) \times p_{\mu_t}^{\batch}(N),
        \end{align*}
        as well as 
        $$(\outcomeD{}{\batch}{0},\Vcovar{\batch}, \Vtreatment{\batch}{}, \MIM{}{i}, \MIM{t}{i}, \MIM{}{\batch}, \MIM{t}{\batch})^\top \sim p_{y}^{\batch} \times p_{x}^{\batch} \times p_{w}^{\batch} \times p_{\mu^i} \times p_{\mu_t^i} \times p_{\mu}^{\batch} \times p_{\mu_t}^{\batch}$$

        Then, as $N \rightarrow \infty$, we have
        \begin{align*}
            &\E \left[\norm{\left(\outcomeD{}{\batch}{0}(N), \Vcovar{\batch}(N), \Vtreatment{\batch}{}(N), \MIM{}{i}(N), \MIM{t}{i}(N), \MIM{}{\batch}(N), \MIM{t}{\batch}(N)\right)^\top }^k\right]
            \\
            \rightarrow~
            &\E \left[\norm{(\outcomeD{}{\batch}{0},\Vcovar{\batch}, \Vtreatment{\batch}{}, \MIM{}{i}, \MIM{t}{i}, \MIM{}{\batch}, \MIM{t}{\batch})^\top }^k\right] < \infty.
        \end{align*}
    \end{enumerate}
\end{assumption}
Assumption~\ref{asmp:weak_limits} is a standard assumption in statistical theory, ensuring that the empirical distributions of system attributes remain stable and do not diverge as the sample size increases. This assumption holds, for example, when units' attributes $\left\{(\outcomeD{}{i}{0}, \Vcovar{i}, \Vtreatment{i}{}, \Vec{\mu}^{\;i\cdot}, \Vec{\mu}^{\;i\cdot}_t)\right\}_{i}$ follow an i.i.d. distribution with finite moments of order $k$. Moreover, a wide range of treatment assignments satisfies the conditions of Assumption~\ref{asmp:weak_limits}, including cases where the support of $\expd$ (as specified by Remark~\ref{rem:general_design}) is bounded, such as the Bernoulli design. By imposing such conditions, we ensure the reliability of estimation by keeping the experimental design moments finite and manageable.

To be more specific, Parts~\ref{asmp:interference_element_convergence} and \ref{asmp:mean_interference_element_convergence} of Assumption~\ref{asmp:weak_limits} formalize and extend Assumption~\ref{asmp:Stable Interference Pattern} from the main text. Indeed, we can simplify the framework by dropping Part~\ref{asmp:mean_interference_element_convergence} under the assumption that $p_{\mu^i}$ and $p_{\mu_t^i}$ are identical across all units $i$. Figure~\ref{fig:asmp:interference_means} illustrates these assumptions by showing how the underlying $N \times N$ interference matrices must satisfy certain convergent weak limits across both their columns and rows as the population size $N$ increases.

\begin{figure}[ht]
\centering
\begin{adjustbox}{max width=\linewidth}
\begin{tikzpicture}[every node/.style={anchor=center, font=\footnotesize}, x=0.9cm, y=0.9cm]
    \pgfmathsetseed{42}

    \definecolor{warm0}{RGB}{255,245,235}
    \definecolor{warm1}{RGB}{255,235,210}
    \definecolor{warm2}{RGB}{255,215,180}
    \definecolor{warm3}{RGB}{255,190,150}
    \definecolor{warm4}{RGB}{255,160,120}
    \definecolor{warm5}{RGB}{255,130,90}
    \definecolor{warm6}{RGB}{255,100,60}
    \definecolor{warm7}{RGB}{230,80,40}
    \def\colormap{{"warm0","warm1","warm2","warm3","warm4","warm5","warm6","warm7"}}

    \def\N{4} 

    \foreach \i in {1,...,\N} {
        \foreach \j in {1,...,\N} {
            \newcommand{\entry}{}
            \def\drawcolor{1}

            \ifnum\i=3
                \ifnum\j=3
                    \def\entry{$\ddots$}
                    \def\drawcolor{0}
                \else
                    \def\entry{$\vdots$}
                    \def\drawcolor{0}
                \fi
            \else
                \ifnum\j=3
                    \def\entry{$\ldots$}
                    \def\drawcolor{0}
                \else
                    \ifnum\j=\N
                        \ifnum\i=1
                            \def\entry{$\mu^{1N}$}
                        \else
                            \ifnum\i=2
                                \def\entry{$\mu^{2N}$}
                            \else
                                \def\entry{$\mu^{NN}$}
                            \fi
                        \fi
                    \else
                        \ifnum\i=1
                            \def\entry{$\mu^{1\j}$}
                        \else
                            \ifnum\i=2
                                \def\entry{$\mu^{2\j}$}
                            \else
                                \def\entry{$\mu^{N\j}$}
                            \fi
                        \fi
                    \fi
                \fi
            \fi

            \ifnum\drawcolor=1
                \pgfmathsetmacro{\ridx}{int(random*7)}
                \pgfmathparse{\colormap[\ridx]}
                \edef\cellcolor{\pgfmathresult}
                \filldraw[fill=\cellcolor, draw=black] (\j, -\i) rectangle ++(1, -1);
            \else
                \draw[black] (\j, -\i) rectangle ++(1, -1);
            \fi
            \node at (\j+0.5, -\i-0.5) {\entry};
        }

        \node[right=0.0cm] at (\N+1.4, -\i-0.5) {
            \ifnum\i=3
                $\vdots$
            \else
                \ifnum\i=1
                    $\MIM{}{1}$
                \else
                    \ifnum\i=2
                        $\MIM{}{2}$
                    \else
                        $\MIM{}{N}$
                    \fi
                \fi
            \fi
        };
        
        \ifnum\i=3
        \else
            \draw[->] (\N+1.1, -\i-0.5) -- (\N+1.4, -\i-0.5);
        \fi

        \node[right=0cm] at (\N+2.6, -\i-0.5) {
            \ifnum\i=3
                $\vdots$
            \else
                \ifnum\i=1
                    $\bar{\mu}^{1}$
                \else
                    \ifnum\i=2
                        $\bar{\mu}^{2}$
                    \else
                        $\bar{\mu}^{N}$
                    \fi
                \fi
            \fi
        };

        \ifnum\i=3
        \else
            \draw[->, thick, dashed] (\N+2.2, -\i-0.5) -- (\N+2.6, -\i-0.5);
        \fi
    }

    \node (mim) at (\N+2.85, -\N-1.5) {$\MIM{}{}$};
    \draw[->] (\N+2.85, -\N-0.80) -- (\N+2.85, -\N-1.1);
    \draw[draw=black,thick] (mim) circle [radius=0.275cm];

    \def\xshift{7}
    \foreach \i in {1,...,\N} {
        \foreach \j in {1,...,\N} {
            \newcommand{\entry}{}
            \def\drawcolor{1}

            \ifnum\i=3
                \ifnum\j=3
                    \def\entry{$\ddots$}
                    \def\drawcolor{0}
                \else
                    \def\entry{$\vdots$}
                    \def\drawcolor{0}
                \fi
            \else
                \ifnum\j=3
                    \def\entry{$\ldots$}
                    \def\drawcolor{0}
                \else
                    \ifnum\j=\N
                        \ifnum\i=1
                            \def\entry{$\mu_t^{1N}$}
                        \else
                            \ifnum\i=2
                                \def\entry{$\mu_t^{2N}$}
                            \else
                                \def\entry{$\mu_t^{NN}$}
                            \fi
                        \fi
                    \else
                        \ifnum\i=1
                            \def\entry{$\mu_t^{1\j}$}
                        \else
                            \ifnum\i=2
                                \def\entry{$\mu_t^{2\j}$}
                            \else
                                \def\entry{$\mu_t^{N\j}$}
                            \fi
                        \fi
                    \fi
                \fi
            \fi

            \ifnum\drawcolor=1
                \pgfmathsetmacro{\ridx}{int(random*7)}
                \pgfmathparse{\colormap[\ridx]}
                \edef\cellcolor{\pgfmathresult}
                \filldraw[fill=\cellcolor, draw=black] (\j+\xshift, -\i) rectangle ++(1, -1);
            \else
                \draw[black] (\j+\xshift, -\i) rectangle ++(1, -1);
            \fi
            \node at (\j+\xshift+0.5, -\i-0.5) {\entry};
        }

        \node[right=0.0cm] at (\N+8.4, -\i-0.5) {
            \ifnum\i=3
                $\vdots$
            \else
                \ifnum\i=1
                    $\MIM{t}{1}$
                \else
                    \ifnum\i=2
                        $\MIM{t}{2}$
                    \else
                        $\MIM{t}{N}$
                    \fi
                \fi
            \fi
        };

        \ifnum\i=3
        \else
            \draw[->] (\N+8.1, -\i-0.5) -- (\N+8.4, -\i-0.5);
        \fi

        \node[right=0cm] at (\N+9.6, -\i-0.5) {
            \ifnum\i=3
                $\vdots$
            \else
                \ifnum\i=1
                    $\bar{\mu}_t^{1}$
                \else
                    \ifnum\i=2
                        $\bar{\mu}_t^{2}$
                    \else
                        $\bar{\mu}_t^{N}$
                    \fi
                \fi
            \fi
        };

        \ifnum\i=3
        \else
            \draw[->, thick, dashed] (\N+9.2, -\i-0.5) -- (\N+9.6, -\i-0.5);
        \fi
    }

    \node (mimt) at (\N+9.85, -\N-1.5) {$\MIM{t}{}$};
    \draw[->] (\N+9.85, -\N-0.80) -- (\N+9.85, -\N-1.1);
    \draw[draw=black,thick] (mimt) circle [radius=0.275cm];

    \draw[black] (\N+11.25, -2.55) rectangle ++(2.9, -0.85);
    \node at (\N+12.6, -2.8) {\scriptsize Weak limit};
    \draw[->] (\N+11.35, -2.8) -- (\N+11.75, -2.8);
    \node at (\N+12.95, -3.2) {\scriptsize Expectation $\mathbb{E}[\cdot]$};
    \draw[->, thick, dashed] (\N+11.35, -3.2) -- (\N+11.75, -3.2);

\end{tikzpicture}
\end{adjustbox}
\caption{Conceptual illustration of interference matrix convergence assumptions. The diagram shows how elements of the $N \times N$ interference matrices must exhibit consistent distributional behavior across rows (units) and columns (interactions) as $N \to \infty$, ensuring that both individual unit interference patterns and cross-unit interaction structures converge to well-defined limiting distributions.}
\label{fig:asmp:interference_means}
\end{figure}

\begin{example}
    Consider the fixed interference matrix (the left matrix in Figure~\ref{fig:asmp:interference_means}) in a simple setting with two interference levels: $\mu^{ij} \in \{0.3, 0.8\}$. For the sequence of settings illustrated in Figure~\ref{fig:matrix_repetition}, the random variable $\MIM{}{}$ follows the distribution $\P(\MIM{}{}=0.3) = \frac{2}{3}$ and $\P(\MIM{}{}=0.8) = \frac{1}{3}$, with $\MIM{}{i} = \MIM{}{}$ for all units~$i$.
\end{example}

\begin{figure}[ht]
\centering
\begin{adjustbox}{max width=\linewidth}
\begin{tikzpicture}[every node/.style={anchor=center, font=\tiny}, x=0.4cm, y=0.4cm]
    
    \definecolor{warm0}{RGB}{255,245,235}
    \definecolor{warm1}{RGB}{255,235,210}
    \definecolor{warm2}{RGB}{255,215,180}
    \definecolor{warm3}{RGB}{255,190,150}
    \definecolor{warm4}{RGB}{255,160,120}
    \definecolor{warm5}{RGB}{255,130,90}
    \definecolor{warm6}{RGB}{255,100,60}
    \definecolor{warm7}{RGB}{230,80,40}
    
    \node[above] at (1.5, 0) {\scriptsize $N=3$};
    
    \filldraw[fill=warm2, draw=black, line width=0.3pt] (0, 0) rectangle (1, -1);
    \node at (0.5, -0.5) {$0.3$};
    \filldraw[fill=warm2, draw=black, line width=0.3pt] (1, 0) rectangle (2, -1);
    \node at (1.5, -0.5) {$0.3$};
    \filldraw[fill=warm6, draw=black, line width=0.3pt] (2, 0) rectangle (3, -1);
    \node at (2.5, -0.5) {$0.8$};
    
    \filldraw[fill=warm6, draw=black, line width=0.3pt] (0, -1) rectangle (1, -2);
    \node at (0.5, -1.5) {$0.8$};
    \filldraw[fill=warm2, draw=black, line width=0.3pt] (1, -1) rectangle (2, -2);
    \node at (1.5, -1.5) {$0.3$};
    \filldraw[fill=warm2, draw=black, line width=0.3pt] (2, -1) rectangle (3, -2);
    \node at (2.5, -1.5) {$0.3$};
    
    \filldraw[fill=warm2, draw=black, line width=0.3pt] (0, -2) rectangle (1, -3);
    \node at (0.5, -2.5) {$0.3$};
    \filldraw[fill=warm6, draw=black, line width=0.3pt] (1, -2) rectangle (2, -3);
    \node at (1.5, -2.5) {$0.8$};
    \filldraw[fill=warm2, draw=black, line width=0.3pt] (2, -2) rectangle (3, -3);
    \node at (2.5, -2.5) {$0.3$};
    
    \draw[->, thick] (3.5, -1.5) -- (4.5, -1.5);
    
    \node[above] at (8, 0) {\scriptsize $N=6$};
    
    \def\drawpattern#1#2{
        \filldraw[fill=warm2, draw=black, line width=0.2pt] (#1, #2) rectangle (#1+1, #2-1);
        \node at (#1+0.5, #2-0.5) {\tiny $0.3$};
        \filldraw[fill=warm2, draw=black, line width=0.2pt] (#1+1, #2) rectangle (#1+2, #2-1);
        \node at (#1+1.5, #2-0.5) {\tiny $0.3$};
        \filldraw[fill=warm6, draw=black, line width=0.2pt] (#1+2, #2) rectangle (#1+3, #2-1);
        \node at (#1+2.5, #2-0.5) {\tiny $0.8$};
        
        \filldraw[fill=warm6, draw=black, line width=0.2pt] (#1, #2-1) rectangle (#1+1, #2-2);
        \node at (#1+0.5, #2-1.5) {\tiny $0.8$};
        \filldraw[fill=warm2, draw=black, line width=0.2pt] (#1+1, #2-1) rectangle (#1+2, #2-2);
        \node at (#1+1.5, #2-1.5) {\tiny $0.3$};
        \filldraw[fill=warm2, draw=black, line width=0.2pt] (#1+2, #2-1) rectangle (#1+3, #2-2);
        \node at (#1+2.5, #2-1.5) {\tiny $0.3$};
        
        \filldraw[fill=warm2, draw=black, line width=0.2pt] (#1, #2-2) rectangle (#1+1, #2-3);
        \node at (#1+0.5, #2-2.5) {\tiny $0.3$};
        \filldraw[fill=warm6, draw=black, line width=0.2pt] (#1+1, #2-2) rectangle (#1+2, #2-3);
        \node at (#1+1.5, #2-2.5) {\tiny $0.8$};
        \filldraw[fill=warm2, draw=black, line width=0.2pt] (#1+2, #2-2) rectangle (#1+3, #2-3);
        \node at (#1+2.5, #2-2.5) {\tiny $0.3$};
    }
    
    \drawpattern{5}{0}    
    \drawpattern{8}{0}    
    \drawpattern{5}{-3}   
    \drawpattern{8}{-3}   
    
    \draw[thick] (5, 0) rectangle (8, -3);     
    \draw[thick] (8, 0) rectangle (11, -3);    
    \draw[thick] (5, -3) rectangle (8, -6);    
    \draw[thick] (8, -3) rectangle (11, -6);   
    
    \draw[->, thick] (12.0, -1.5) -- (13.0, -1.5);
    
    \node[above] at (18.5, 0) {\scriptsize $N=9$};
    
    \def\drawsmallpattern#1#2{
        \filldraw[fill=warm2, draw=black, line width=0.15pt] (#1, #2) rectangle (#1+1, #2-1);
        \node at (#1+0.5, #2-0.5) {\fontsize{4pt}{5pt}\selectfont $0.3$};
        \filldraw[fill=warm2, draw=black, line width=0.15pt] (#1+1, #2) rectangle (#1+2, #2-1);
        \node at (#1+1.5, #2-0.5) {\fontsize{4pt}{5pt}\selectfont $0.3$};
        \filldraw[fill=warm6, draw=black, line width=0.15pt] (#1+2, #2) rectangle (#1+3, #2-1);
        \node at (#1+2.5, #2-0.5) {\fontsize{4pt}{5pt}\selectfont $0.8$};
        
        \filldraw[fill=warm6, draw=black, line width=0.15pt] (#1, #2-1) rectangle (#1+1, #2-2);
        \node at (#1+0.5, #2-1.5) {\fontsize{4pt}{5pt}\selectfont $0.8$};
        \filldraw[fill=warm2, draw=black, line width=0.15pt] (#1+1, #2-1) rectangle (#1+2, #2-2);
        \node at (#1+1.5, #2-1.5) {\fontsize{4pt}{5pt}\selectfont $0.3$};
        \filldraw[fill=warm2, draw=black, line width=0.15pt] (#1+2, #2-1) rectangle (#1+3, #2-2);
        \node at (#1+2.5, #2-1.5) {\fontsize{4pt}{5pt}\selectfont $0.3$};
        
        \filldraw[fill=warm2, draw=black, line width=0.15pt] (#1, #2-2) rectangle (#1+1, #2-3);
        \node at (#1+0.5, #2-2.5) {\fontsize{4pt}{5pt}\selectfont $0.3$};
        \filldraw[fill=warm6, draw=black, line width=0.15pt] (#1+1, #2-2) rectangle (#1+2, #2-3);
        \node at (#1+1.5, #2-2.5) {\fontsize{4pt}{5pt}\selectfont $0.8$};
        \filldraw[fill=warm2, draw=black, line width=0.15pt] (#1+2, #2-2) rectangle (#1+3, #2-3);
        \node at (#1+2.5, #2-2.5) {\fontsize{4pt}{5pt}\selectfont $0.3$};
    }

    \draw[->, thick] (24.0, -1.5) -- (25.0, -1.5);
    \node at (27, -1.5) {\Large $\ldots$};
    
    \drawsmallpattern{14}{0}    
    \drawsmallpattern{17}{0}
    \drawsmallpattern{20}{0}
    \drawsmallpattern{14}{-3}   
    \drawsmallpattern{17}{-3}
    \drawsmallpattern{20}{-3}
    \drawsmallpattern{14}{-6}   
    \drawsmallpattern{17}{-6}
    \drawsmallpattern{20}{-6}
    
    \draw[thick] (14, 0) rectangle (17, -3);   
    \draw[thick] (17, 0) rectangle (20, -3);
    \draw[thick] (20, 0) rectangle (23, -3);
    \draw[thick] (14, -3) rectangle (17, -6);  
    \draw[thick] (17, -3) rectangle (20, -6);
    \draw[thick] (20, -3) rectangle (23, -6);
    \draw[thick] (14, -6) rectangle (17, -9);  
    \draw[thick] (17, -6) rectangle (20, -9);
    \draw[thick] (20, -6) rectangle (23, -9);

\end{tikzpicture}
\end{adjustbox}
\caption{A sequence of interference matrices exhibiting the same interference distribution pattern.}

\label{fig:matrix_repetition}
\end{figure}

\begin{remark}
    We can drop Assumptions \ref{asmp:BL} and \ref{asmp:weak_limits}-\ref{asmp:k_moment_convergence} by confining the functions $\outcomeg{t}{}$ and $\outcomeh{t}{}$ to be bounded and continuous.
\end{remark}
\begin{remark}[Time-dependent covariates]
    For each time period $t = 0, 1, \ldots, T$, let $\Xc_t \in \R^{\dcovar_t \times N}$ denote the time-dependent covariate matrix, where the $i^{th}$ column corresponds to the covariates for unit~$i$ at time $t$. We extend the original covariate matrix $\covar$ by incorporating these time-dependent covariates, resulting in an augmented covariate matrix:
    $$
    [\covar^\top | \Xc_0^\top | \ldots | \Xc_{T-1}^\top]^\top \in \R^{(\dcovar + \dcovar_0 + \ldots + \dcovar_T) \times N}.
    $$
    The functions $\outcomeg{t}{}$ and $\outcomeh{t}{}$ are then modified to reference the appropriate portion of this extended covariate matrix at each time period. As an example and to streamline subsequent analysis, we exclude the noise vectors $\Vnoise{}{t}$ from discussions of the potential outcome specification in Eq.~\eqref{eq:apndx_outcome_function} by constructing a new covariate matrix that absorbs the noise terms:
    $$\covar^{\text{new}} := [\covar^\top | \Vnoise{}{0} | \ldots | \Vnoise{}{T-1}]^\top \in \R^{(\dcovar_t+T) \times N}.$$
    We then redefine the outcome functions as $\outcomeg{t}{\text{new}}(\cdot, \cdot, \covar^{\text{new}}) := \outcomeg{t}{}(\cdot, \cdot, \covar)$ and $\outcomeh{t}{\text{new}}(\cdot, \cdot, \covar^{\text{new}}) := \outcomeh{t}{}(\cdot, \cdot, \covar) + \Vnoise{}{t}$.
\end{remark}
\begin{remark}[Notation convention]
    \label{rem:notation}
    Whenever $\batch$ represents the entire experimental population, we omit the superscript $\batch$ from all notations.
\end{remark}

To state the main theoretical results, we need to define the \textbf{Batch State Evolution (BSE)} equations as follows:
\begin{equation}
    \label{eq:state evolution}
    \begin{aligned}
        \MAVO{}{\batch}{1} &=
        \E\left[ (\MIM{}{\batch}+\MIM{0}{\batch})
        \outcomeg{0}{}\big(\outcomeD{}{}{0}, \Vtreatment{}{},\Vcovar{}\big)\right],
        \quad
        (\MVVO{}{}{1})^2 =
        (\sigma^2 + \sigma_0^2)
        \E\left[
        \outcomeg{0}{}\big(\outcomeD{}{}{0}, \Vtreatment{}{},\Vcovar{}\big)^2\right],
        \quad
        \Houtcome{\batch}{}{0} = \outcomeh{0}{}\big(\outcomeD{}{\batch}{0}, \Vtreatment{\batch}{}, \Vcovar{\batch}\big),
        \\
        \Houtcome{\batch}{}{t} &= \outcomeh{t}{}\big(\MAVO{}{\batch}{t} + \MVVO{}{}{t} Z_t + \Houtcome{\batch}{}{t-1}, \Vtreatment{\batch}{}, \Vcovar{\batch}\big),
        \\
        \MAVO{}{\batch}{t+1} &=
        \E\left[ (\MIM{}{\batch}+\MIM{t}{\batch})
        \outcomeg{t}{}\big(\MAVO{}{}{t} + \MVVO{}{}{t} Z_t + \Houtcome{}{}{t-1}, \Vtreatment{}{}, \Vcovar{}\big)
        \right],
        \\
        (\MVVO{}{}{t+1})^2 &=
        (\sigma^2+\sigma_t^2) \E\left[
        \outcomeg{t}{} \big(\MAVO{}{}{t} + \MVVO{}{}{t} Z_t + \Houtcome{}{}{t-1}, \Vtreatment{}{},\Vcovar{}\big)^2
        \right],
        \\
        \AVO{}{\batch}{t+1} &= \MAVO{}{\batch}{t+1} + \E\left[\Houtcome{\batch}{}{t}\right],
    \end{aligned}
\end{equation}
where 
\begin{itemize}
    \item $\outcomeD{}{}{0} \sim p_y$ and $\outcomeD{}{\batch}{0} \sim p_y^\batch$ represent the weak limits of the population and subpopulation initial outcomes;

    \item $\Vtreatment{}{} \sim p_w$ and $\Vtreatment{\batch}{} \sim p_w^\batch$ are the weak limits of the population and subpopulation treatment assignments;

    \item $\Vcovar{} \sim p_x$ and $\Vcovar{\batch} \sim p_x^\batch$ represent the weak limits of the population and subpopulation covariates;

    \item $\MIM{}{\batch} \sim p_{\mu}^\batch$ and $\MIM{t}{\batch} \sim p_{\mu_t}^\batch$ represent the weak limits of interference elements at the subpopulation level, as specified in Assumption~\ref{asmp:weak_limits};

    \item $Z_t$ follows a standard Gaussian distribution;

    \item and $\MAVO{}{}{t}$ and $\Houtcome{}{}{t}$ represent the same quantities as $\MAVO{}{\batch}{t}$ and $\Houtcome{\batch}{}{t}$, respectively, but for $\batch$ to include the entire population (see Remark~\ref{rem:notation}).
\end{itemize}

\begin{remark}
    While we focus on the special case of the general model in \eqref{eq:apndx_outcome_function} and \eqref{eq:apndx_outcome_function_WOD} with only one lag term $\VoutcomeD{}{}{t}$, we can readily extend \eqref{eq:state evolution} to incorporate more lag terms. For instance, to include an additional lag term in function $\outcomeg{t}{}$, we would modify \eqref{eq:state evolution} by including $\MAVO{}{}{t-1} + \MVVO{}{}{t-1} Z_{t-1} + \Houtcome{}{}{t-2}$ as an argument in $\outcomeg{t}{}$. We also need to revise the initial conditions (first line in \eqref{eq:state evolution}) according to the number of lag terms.
\end{remark}

Now, the following theorem characterizes the distribution of units' outcomes $\outcomeD{}{i}{1}, \ldots, \outcomeD{}{i}{t+1}$ within the large sample asymptotic, based on the BSE equations outlined in Eq.~\eqref{eq:state evolution}.
\begin{theorem}
    \label{thm:Batch_SE}
    Fixing $k\geq 2$, consider the sequence of units' attributes $\left\{(\outcomeD{}{i}{0}, \Vcovar{i}, \Vtreatment{i}{}, \Vec{\mu}^{\;i\cdot}, \Vec{\mu}^{\;i\cdot}_t)\right\}_{i,t}$ and suppose that Assumptions~\ref{asmp:apndx_Gaussian Interference Matrice} and \ref{asmp:BL} hold, and Assumption~\ref{asmp:weak_limits} is satisfied once for $\batch$ and once for the enitre population. Then, in view of the BSE equations given in \eqref{eq:state evolution}, for any function $\psi \in \poly{k}$, we have,
    \begin{equation}
        \begin{aligned}
            \label{eq:BSE-limit}
            &\;\lim_{N \rightarrow \infty}
            \frac{1}{\cardinality{\batch}} \sum_{i \in \batch}
            \psi\big(
            \outcomeD{}{i}{0},
            \outcomeD{}{i}{1},
            \ldots,
            \outcomeD{}{i}{T},
            \Vtreatment{i}{}, \Vcovar{i}
            \big)
            \\
            \eqas
            \;&\E
            \Big[
            \psi\big(
            \outcomeD{}{\batch}{0},
            \MAVO{}{\batch}{1} + \MVVO{}{}{1} Z_1 + \Houtcome{\batch}{}{0},
            \ldots,
            \MAVO{}{\batch}{T} + \MVVO{}{}{T} Z_{T} + \Houtcome{\batch}{}{T-1},
            \Vtreatment{\batch}{}, \Vcovar{\batch}
            \big)
            \Big],
        \end{aligned}
    \end{equation}
    where $Z_t \sim \Nc(0,1),\; t= 1,\ldots,T,$ is independent of $\Vtreatment{\batch}{} \sim p_w^\batch$ and $\Vcovar{\batch} \sim p_x^\batch$.
\end{theorem}
We prove the result of Theorem~\ref{thm:Batch_SE} by extending the theoretical results of \cite{shirani2024causal}, which in turn build on the AMP framework developed by \cite{bayati2011dynamics}. The proof mainly relies on a \emph{conditioning technique} introduced by \cite{bolthausen2014iterative}. Below, we first present a version of the conditioning technique adapted to our specific setting, followed by a detailed proof of the theorem.

\begin{remark}
    To derive the result of Theorem~\ref{thm:BSE_informal} from Theorem~\ref{thm:Batch_SE}, we note that when $\batch$ depends solely on the treatment allocation, which is independently distributed from all other variables, $\batch$ effectively functions as a random sample from the experimental population. As a result, the distributions of $\outcomeD{}{\batch}{0}$, $\Vcovar{\batch}$, $\MIM{}{\batch}$, and $\MIM{t}{\batch}$ are equivalent to those of $\outcomeD{}{}{0}$, $\Vcovar{}$, $\MIM{}{}$, and $\MIM{t}{}$, respectively. This equivalence follows from the fact that the empirical distribution of quantities in a random sample converges to the empirical distribution of the entire population.
\end{remark}

\subsection{Conditioning Technique}
\label{apndx:Conditioning_Technique}
Recalling \eqref{eq:apndx_outcome_function_WOD} and
letting $\VUoutcome{}{}{t} = \outcomeg{t}{}\big(\VoutcomeD{}{}{t},\Mtreatment{}{},\covar\big)$ (and $\Uoutcome{i}{}{t} = \outcomeg{t}{}\big(\outcomeD{}{i}{t},\Vtreatment{i}{},\Vcovar{i}\big)$), we denote
\begin{equation}
\label{eq:Q and R}
\begin{aligned}
    \bm{Q}_t
    :=
    \left[
    \VUoutcome{}{}{0}
    \Big|
    \VUoutcome{}{}{1}
    \Big|
    \ldots
    \Big|
    \VUoutcome{}{}{t-1}
    \right],
    \quad\quad
    \bm{R}_t
    :=
    \left[
    \Voutcome{}{}{1} -
    \IMatT{0} \VUoutcome{}{}{0}
    \Big|
    \ldots
    \Big|
    \Voutcome{}{}{t} -
    \IMatT{t-1} \VUoutcome{}{}{t-1}
    \right].
\end{aligned}
\end{equation}
According to Eq.~\eqref{eq:Q and R}, $\bm{Q}_t$ and $\bm{R}_t$ are matrices with columns of $\VUoutcome{}{}{s-1}$ and $\Voutcome{}{}{s} - \IMatT{s-1} \VUoutcome{}{}{s-1}$, when $s=1,\ldots,t$, respectively. Then, we denote by $\VUoutcome{\parallel}{}{t}$ the projection of $\VUoutcome{}{}{t}$ onto the space generated by the columns of $\bm{Q}_t$ and define $\VUoutcome{\perp}{}{t} = \VUoutcome{}{}{t} - \VUoutcome{\parallel}{}{t}$. Assuming $\bm{Q}_t$ is a full-row rank matrix (we prove this is indeed the case in Lemma~\ref{lm:Big lemma}-\ref{part:BL-c}), we also define $\VAPC{}{t} = (\APC{0}{t}, \APC{1}{t}, \ldots, \APC{t-1}{t})^\top$ such that
\begin{align}
    \label{eq:projection sum}
    \VUoutcome{\parallel}{}{t}
    =
    \sum_{s=0}^{t-1} \APC{s}{t} \VUoutcome{}{}{s}
    =
    \sum_{s=0}^{t-1} \APC{s}{t}
    \outcomeg{s}{}
    \big(
    \VoutcomeD{}{}{s}, \Mtreatment{}{}, \covar
    \big),
\end{align}
where
\begin{align}
    \label{eq:projection coefficients}
    \VAPC{}{t}
    =
    \left(
    \bm{Q}_t^\top \bm{Q}_t
    \right)^{-1}
    \bm{Q}_t^\top \VUoutcome{}{}{t}.
\end{align}
Now, note that the available observation at any time $t$ implicitly reveals information about the fixed interference matrix $\IM$. To manage this intricate randomness, we define $\Gc_t$ as the $\sigma$-algebra generated by $\VoutcomeD{}{}{0}$, $\Voutcome{}{}{1}, \ldots, \Voutcome{}{}{t}$, $\IMatMean{t}$, $\IMatT{0}, \ldots, \IMatT{t-1}$, $\Mtreatment{}{}$, and $\covar$. We then compute the conditional distribution of $\IM$ given $\Gc_t$. Conditioning on $\Gc_t$, by definition and based on Eq.~\eqref{eq:apndx_outcome_function_WOD}, is equivalent to conditioning on the following event:
$$
\{
\IM \VUoutcome{}{}{0} = \Voutcome{}{}{1} - \IMatT{0} \VUoutcome{}{}{0},\;
\IM \VUoutcome{}{}{1} = \Voutcome{}{}{2} - \IMatT{1} \VUoutcome{}{}{1},\;
\ldots,\;
\IM \VUoutcome{}{}{t-1} = \Voutcome{}{}{t} - \IMatT{t-1} \VUoutcome{}{}{t-1}
\}.
$$
which encapsulates all information we have about the fixed interference matrix $\IM$ up to time $t$. By Eq.~\eqref{eq:Q and R}, this is also equivalent to the event $\IM \bm{Q}_t = \bm{R}_t$. When conditioned on $\Gc_t$, the entries of both $\bm{Q}_t$ and $\bm{R}_t$ become deterministic values, leading to the following lemma.

\begin{lemma}
    \label{lm:conditional dist of IM}
    Fix $t$ and assume that $\bm{Q}_t$ is a full-row rank matrix. Then, for the conditional distribution of the fixed interference matrix $\IM$ given $\IM \bm{Q}_t=\bm{R}_t$, we have
        \begin{align}
        \label{eq:conditional dist of IM}
        \IM|_{\IM \bm{Q}_t=\bm{R}_t}
        \eqd
        \bm{R}_t
        \left(
        \bm{Q}_t^\top \bm{Q}_t
        \right)^{-1}
        \bm{Q}_t^\top
        +
        \widetilde{\IM} P^\perp.
        \end{align}
    where $\widetilde{\IM} \eqd \IM$ independent of $\IM$ and $P^\perp = (\I-P)$ that P denotes the orthogonal projector onto the column space of $\bm{Q}_t$.
\end{lemma}

The proof of Lemma~\ref{lm:conditional dist of IM} relies on the rotational invariance of the Gaussian distribution and utilizes Lemma~11 from \cite{bayati2011dynamics}. The proofs of this lemma and the subsequent lemma, which describes the distribution of $\Voutcome{}{}{t+1}$ given the event $\IM \bm{Q}_t = \bm{R}_t$, follow a similar approach to the proofs of Lemmas~2 and 3 in \cite{shirani2024causal}, and we refer readers to that work for detailed derivations.

\begin{lemma}
    \label{lm:conditional dist of outcome}
    Fix $t$ and assume that $\bm{Q}_t$ is a full-row rank matrix. The following holds for the conditional distribution of the vector $\Voutcome{}{}{t+1}$:
            \begin{align}
                \label{eq:conditional dist of outcome_nonsym}
                \Voutcome{}{}{t+1}\big|_{\Gc_t}
                \eqd
                &\;
                \widetilde{\IM} 
                \VUoutcome{\perp}{}{t}
                + \bm{R}_t \VAPC{}{t} + \IMatT{t} \VUoutcome{}{}{t},
            \end{align}
    where the matrix $\widetilde{\IM}$ is independent of $\IM$ and has the same distribution.
\end{lemma}

\subsection{Detailed Proof of Theorem~\ref{thm:Batch_SE}}
\label{apndx:big_lemma}
We begin by stating Lemma~\ref{lm:Big lemma}, which provides an expanded version of Theorem~\ref{thm:Batch_SE}. This expanded formulation is necessary due to our inductive proof technique, which requires establishing that $\bm{Q}_t$ is a full-row rank matrix alongside several auxiliary results needed to advance each step of the induction.
To this end, we need some new notations. Specifically, for vectors $\Vec{u},\Vec{v} \in \R^m$, we define the scalar product $\pdot{\Vec{u}}{\Vec{v}}:= \frac{1}{m} \sum_{i=1}^m u_i v_i$. Also, considering \eqref{eq:state evolution}, for $t\geq 1$, we define
\begin{equation}
    \label{eq:state evolution_fixed part}
    \begin{aligned}
        (\BVVO{}{}{1})^2
            &=
            \lim_{N\rightarrow \infty}
            \frac{\sigma^2}{N} \sum_{i=1}^N
            \outcomeg{0}{}\big(
            \outcomeD{}{i}{0},\Vtreatment{i}{},\Vcovar{i}
            \big)^2 < \infty
        \\
        (\BVVO{}{}{t+1})^2 &=
        \sigma^2 \E\left[
        \outcomeg{t}{} \big(\MAVO{}{}{t} + \MVVO{}{}{t} Z_t + \Houtcome{}{}{t-1}, \Vtreatment{}{},\Vcovar{}\big)^2
        \right].
    \end{aligned}
\end{equation}

\begin{lemma}
    \label{lm:Big lemma}
    For a fixed $k\geq 2$ and a specified subpopulation of experimental units $\batch$, suppose that the conditions of Theorem~\ref{thm:Batch_SE} hold. Then, the following statements are valid for all time steps $t$ under the BSE equations defined in \eqref{eq:state evolution}:
    \begin{enumerate}[label=(\alph*)]
        \item \label{part:BL-a} For any function $\psi \in \poly{k}$, we have
        \begin{equation}
            \label{eq:BL-a}
            \begin{aligned}
                \lim_{N \rightarrow \infty}
                \frac{1}{\cardinality{\batch}} \sum_{i \in \batch}
                \psi\big(
                \outcomeD{}{i}{0},
                \outcomeD{}{i}{1},
                \ldots,
                \outcomeD{}{i}{t+1},
                \Vtreatment{i}{},\Vcovar{i}
                \big)
                \\
                \eqas
                \E
                \Big[
                \psi
                \big(
                \outcomeD{}{\batch}{0},
                \MAVO{}{\batch}{1}
                + \MVVO{}{}{1} Z_1
                + \Houtcome{\batch}{}{0},
                \ldots,
                \MAVO{}{\batch}{t+1}
                + \MVVO{}{}{t+1} Z_t
                + \Houtcome{\batch}{}{t},
                \Vtreatment{\batch}{}, \Vcovar{\batch}
                \big)
                \Big],
            \end{aligned}
        \end{equation}
        where $Z_1, \ldots, Z_t$ are standard Gaussian random variables.

        \item \label{part:BL-b} For all $0 \leq r\neq s \leq t$, the following equations hold and all limits exist, are bounded, and have degenerate distribution (i.e. they are constant random variables)
        \begin{subequations}
            \label{eq:BL-b}
            \begin{align}
                \label{eq:BL-b-1}
                \lim_{N \rightarrow \infty} 
                \frac{1}{N}
                \sum_{i=1}^N
                (\outcome{}{i}{r+1})^2
                &\eqas
                (\MVVO{}{}{r+1})^2
                \eqas
                \lim_{N \rightarrow \infty}
                \frac{\sigma^2+\sigma^2_{r}}{N} \sum_{i=1}^N (\Uoutcome{i}{}{r})^2
                \\
                \label{eq:BL-b-2}
                \lim_{N \rightarrow \infty} 
                \frac{1}{N}
                \sum_{i=1}^N
                \outcome{}{i}{r+1}\outcome{}{i}{s+1}
                &\eqas
                \lim_{N \rightarrow \infty}
                \frac{\sigma^2}{N} \sum_{i=1}^N \Uoutcome{i}{}{r}\Uoutcome{i}{}{s},
                \\
                \label{eq:BL-b-3}
                \lim_{N \rightarrow \infty}
                \frac{1}{N}
                \sum_{i=1}^N
                \big(\outcome{}{i}{r+1}
                - \IMatTv{i \cdot}{r} \VUoutcome{}{}{r}
                \big)^2
                &\eqas
                (\BVVO{}{}{r+1})^2
                \eqas
                \lim_{N \rightarrow \infty}
                \frac{\sigma^2}{N}
                \sum_{i=1}^N
                \left(\Uoutcome{i}{}{r}\right)^2,
                \\
                \label{eq:BL-b-4}
                \lim_{N \rightarrow \infty}
                \frac{1}{N}
                \sum_{i=1}^N
                \big(\outcome{}{i}{r+1}
                - \IMatTv{i \cdot}{r} \VUoutcome{}{}{r}
                \big)
                \big(\outcome{}{i}{s+1}
                - \IMatTv{i \cdot}{s} \VUoutcome{}{}{s}
                \big)
                &\eqas
                \lim_{N \rightarrow \infty}
                \frac{\sigma^2}{N}
                \sum_{i=1}^N
                \Uoutcome{i}{}{r} \Uoutcome{i}{}{s},
                \\
                \label{eq:BL-b-5}
                \lim_{N \rightarrow \infty}
                \frac{1}{N}
                \sum_{i=1}^N
                \outcome{}{i}{r+1}
                \big(\outcome{}{i}{s+1}
                - \IMatTv{i \cdot}{s} \VUoutcome{}{}{s}
                \big)
                &\eqas
                \lim_{N \rightarrow \infty}
                \frac{\sigma^2}{N}
                \sum_{i=1}^N
                \Uoutcome{i}{}{r} \Uoutcome{i}{}{s}.
            \end{align}
        \end{subequations}

        \item \label{part:BL-c} Letting $\MOG{t} = [\Voutcome{}{}{1}|\ldots|\Voutcome{}{}{t}]$, the following matrices are positive definite almost surely:
        \begin{align}
            \label{eq:BL-lower bound for perps}
            \lim_{N \rightarrow \infty} \frac{\bm{Q}_{t}^\top \bm{Q}_{t}}{N} \succ 0,
            \quad\quad\quad
            \lim_{N \rightarrow \infty} \frac{\MOG{t}^\top \MOG{t}}{N}
            \succ 0.
        \end{align}
    \end{enumerate}
\end{lemma}

\noindent
\textbf{Proof.}
For all $t$, we assume, without loss of generality, that the mapping $y \mapsto \outcomeg{t}{}(y, \Vtreatment{}{}, \Vcovar{})$ is a non-constant function with positive probability with respect to the randomness of $\Vtreatment{}{}$ and $\Vcovar{}$; otherwise, the result is trivial and does not require further analysis. We prove the result by induction on $t$.

\textbf{Step 1.} Let $t = 0$. By definition, the matrices $\bm{Q}_0$ and $\bm{R}_0$ are empty, and the $\sigma$-algebra $\Gc_0$ is generated by $\VoutcomeD{}{}{0}$, $\Mtreatment{}{}$, and $\covar$. As the induction base case, we establish Parts \ref{part:BL-a} and \ref{part:BL-b} for $t=0$ and Part~\ref{part:BL-c} for $t = 1$.
\begin{enumerate}[label=(\alph*)]
    \item \label{item:BL-average limit} Conditioning on the values of $\VoutcomeD{}{}{0}$, $\Mtreatment{}{}$, $\covar$, and so on the value of $\VUoutcome{}{}{0} = \outcomeg{0}{}\big(\VoutcomeD{}{}{0},\Mtreatment{}{},\covar\big)$, the elements of $\Voutcome{}{}{1}$ are i.i.d. Gaussian random variables with zero mean and variance~$(\MVVO{}{}{1N})^2$:
    \begin{equation}
        \label{eq:BL-a0-Y1 stat}
        \begin{aligned}
            (\MVVO{}{}{1N})^2
            &:=
            \Var
            \left[
            \outcome{}{i}{1}
            \Big|
            \VUoutcome{}{}{0}
            \right]
            =
            \frac{\sigma^2 + \sigma^2_0}{N}
            \sum_{i=1}^N
            \outcomeg{0}{} \left(\outcomeD{}{i}{0},\Vtreatment{i}{},\Vcovar{i}
            \right)^2.
        \end{aligned}
    \end{equation}
    Then, Assumption~\ref{asmp:BL}-\ref{asmp:BL-bound on initials} implies that the value of $(\MVVO{}{}{1N})^2$ is bounded independent from $N$, and
    \begin{equation}
        \label{eq:BL-a0-Y1 stats limits}
        \begin{aligned}
            (\MVVO{}{}{1})^2 :=
            \lim_{N\rightarrow \infty} (\MVVO{}{}{1N})^2.
        \end{aligned}
    \end{equation}

    Now, let $Z$ denote a standard Gaussian random variable. Fixing $l \geq 1$, it is straightforward to show that
    \begin{equation}
    \begin{aligned}
    \label{eq:BL-proof-a0-1}
        \E
        \left[
        \big|\outcome{}{i}{1}\big|^l
        \big|
        \VUoutcome{}{}{0}
        \right]
        &=
        \E
        \left[
        \big|\MVVO{}{}{1N} Z\big|^l
        \big|
        \VUoutcome{}{}{0}
        \right] \leq c,
    \end{aligned}
    \end{equation}
    where $c$ is a constant independent of $N$ and might alter in different lines.

    We next focus on the second term on the right-hand side of Eq.~\eqref{eq:apndx_outcome_function} and define
    \begin{equation}
        \label{eq:BL-a0-Y1 stat-2}
        \begin{aligned}
            \MAVO{}{i}{1N}
            &:=
            \frac{1}{N}
            \sum_{j=1}^N
            (\mu^{ij} + \mu_0^{ij})
            \outcomeg{0}{} \big(\outcomeD{}{j}{0},\Vtreatment{j}{},\Vcovar{j}\big),
            \quad\quad\quad
            i \in [N].
        \end{aligned}
    \end{equation}
    In view of Assumption~\ref{asmp:weak_limits}, we can apply Theorem~\ref{thm:SLLN-2}. Consequently, for all $i$, we can write
    \begin{equation}
        \label{eq:BL-a0-Y1 stat-2_limit}
        \begin{aligned}
            \lim_{N \rightarrow \infty} \MAVO{}{i}{1N}
            \eqas
            \E \left[ (\MIM{}{i}+\MIM{0}{i}) \outcomeg{0}{} \big(\outcomeD{}{}{0},\Vtreatment{}{},\Vcovar{}\big) \right] = (\bar{\mu}^i + \bar{\mu}^i_0) \E \left[ \outcomeg{0}{} \big(\outcomeD{}{}{0},\Vtreatment{}{},\Vcovar{}\big) \right] = \MAVO{}{i}{1} < \infty.
        \end{aligned}
    \end{equation}
    Above, we let $\bar{\mu}^i := \E[\MIM{}{i}]$ and $\bar{\mu}^i_0 := \E[\MIM{0}{i}]$, where $\MIM{}{i} \sim p_{\mu^i}$ and $\MIM{0}{i} \sim p_{\mu^i_0}$; indeed, we used Assumption~\ref{asmp:weak_limits}-\ref{asmp:mean_interference_element_convergence} stating that $\MIM{}{i}$ and $\MIM{0}{i}$ are independent of everything else. Additionally, $\outcomeD{}{}{0}$, $\Vtreatment{}{}$, and $\Vcovar{}$ represent the weak limits of the initial outcomes, treatment allocations, and covariates for the entire population as outlined in Assumption~\ref{asmp:weak_limits}, respectively. In this context, $\bar{\mu}^i+\bar{\mu}^i_0$ determines the average interaction level of unit $i$ at time $0$.

    Also, note that \eqref{eq:BL-a0-Y1 stat-2_limit} yields the boundedness of $\MAVO{}{i}{1N}$ for all $i$ independent of $N$.
    Now, using Assmption~\ref{asmp:weak_limits} and the fact that $\psi \in \poly{k}$, for $\kappa > 0$, we have the following, 
    \begin{align*}
        \frac{1}{\cardinality{\batch}} \sum_{i \in \batch}
        \E
        \left[
        \Big|
        \psi\big(
        \outcomeD{}{i}{0},
        \outcome{}{i}{1},
        \Vtreatment{i}{}, \Vcovar{i}, \MAVO{}{i}{1N}
        \big)
        -
        \E_{\IM,\IMatT{0}}
        \left[
        \psi\big(
        \outcomeD{}{i}{0},
        \outcome{}{i}{1},
        \Vtreatment{i}{}, \Vcovar{i}, \MAVO{}{i}{1N}
        \big)
        \right]
        \Big|^{2+\kappa}
        \right]
        \leq
        c \cardinality{\batch}^{\kappa/2},
    \end{align*}
    where $\E_{\IM,\IMatT{0}}$ is the expectation with respect to the randomness of the interference matrices $\IM,\IMatT{0}$ and $c$ is a constant independent of $N$. Then, applying the Strong Law of Large Numbers (SLLN) for triangular arrays in Theorem~\ref{thm:SLLN}, we obtain the following result:
    \begin{align}
        \label{eq:BL-a0-SLLN}
        \lim_{N \rightarrow \infty}
        \frac{1}{\cardinality{\batch}} \sum_{i \in \batch}
        \Big(
        \psi\big(
        \outcomeD{}{i}{0},
        \outcome{}{i}{1},
        \Vtreatment{i}{}, \Vcovar{i}, \MAVO{}{i}{1N}
        \big)
        -
        \E_{\IM,\IMatT{0}}
        \left[
        \psi\big(
        \outcomeD{}{i}{0},
        \outcome{}{i}{1},
        \Vtreatment{i}{}, \Vcovar{i}, \MAVO{}{i}{1N}
        \big)
        \right]
        \Big)
        \eqas
        0.
    \end{align}
    On the other hand, employing the dominated convergence theorem, e.g., Theorem 16.4 in \cite{billingsley2008probability}, allows us to interchange the limit and the expectation in view of the fact that $\psi \in \poly{k}$. We also utilize the continuous mapping theorem, e.g., Theorem 2.3 in \cite{van2000asymptotic}, to pass the limit through the function. As a result, considering $\outcome{}{i}{1} \eqd \MVVO{}{}{1N} Z$, we get
    \begin{align}
        \label{eq:BL-a0-limit to the function}
        \lim_{N \rightarrow \infty}
        \frac{1}{\cardinality{\batch}} \sum_{i \in \batch}
        \E_{\IM,\IMatT{0}}
        \left[
        \psi\big(
        \outcomeD{}{i}{0},
        \outcome{}{i}{1},
        \Vtreatment{i}{}, \Vcovar{i}, \MAVO{}{i}{1N}
        \big)
        \right]
        \eqas
        \lim_{N \rightarrow \infty}
        \frac{1}{\cardinality{\batch}} \sum_{i \in \batch}
        \E_{Z}
        \left[
        \psi\big(
        \outcomeD{}{i}{0},
        \MVVO{}{}{1} Z,
        \Vtreatment{i}{}, \Vcovar{i}, \MAVO{}{i}{1}
        \big)
        \right].
    \end{align}
    Then, applying Theorem~\ref{thm:SLLN-2} for the function $\text{f}(\outcomeD{}{i}{0}, \Vtreatment{i}{}, \Vcovar{i}, \MAVO{}{i}{1}) = \E_{Z}\left[\psi\big(\outcomeD{}{i}{0}, \MVVO{}{}{1} Z, \Vtreatment{i}{}, \Vcovar{i}, \MAVO{}{i}{1} \big)\right]$, we can write
    \begin{equation}
        \label{eq::BL-proof-a0-dynamics}
        \begin{aligned}
            \lim_{N \rightarrow \infty}
            \frac{1}{\cardinality{\batch}} \sum_{i \in \batch}
            \psi\big(
            \outcomeD{}{i}{0},
            \outcome{}{i}{1},
            \Vtreatment{i}{}, \Vcovar{i}, \MAVO{}{i}{1N}
            \big)
            \eqas
            \E
            \Big[
            \psi
            \big(
            \outcomeD{}{\batch}{0},
            \MVVO{}{}{1} Z,
            \Vtreatment{\batch}{}, \Vcovar{\batch}, \MAVO{}{\batch}{1}
            \big)
            \Big],
        \end{aligned}
    \end{equation}
    where $\outcomeD{}{\batch}{0},\;\Vtreatment{\batch}{},\;\Vcovar{\batch},\; \MAVO{}{\batch}{1}$ represent the weak limits of $\outcomeD{}{i}{0},\;\Vtreatment{i}{},\;\Vcovar{i},\; \MAVO{}{i}{1}$ over the subpopulation units, as specified in Assumption~\ref{asmp:weak_limits} and
    \begin{equation}
        \label{eq:BL-a0-Y1 stat-2_limit_without_n}
        \begin{aligned}
            \MAVO{}{\batch}{1}
            \eqas
            \E \left[ (\MIM{}{\batch}+\MIM{0}{\batch}) \outcomeg{0}{} \big(\outcomeD{}{}{0},\Vtreatment{}{},\Vcovar{}\big) \right],
        \end{aligned}
    \end{equation}
    where $\MIM{}{\batch} + \MIM{0}{\batch}$ captures the weak limit of the average interference level for units in the subpopulation.
    
    In Eq.~\eqref{eq::BL-proof-a0-dynamics}, the function $\text{f}$ is within $\poly{k}$, since $\psi \in \poly{k}$ and expectation is a linear operator. It is important to note that above, $Z$ is independent of $\outcomeD{}{\batch}{0},\; \Vtreatment{\batch}{},\; \Vcovar{\batch}, \MIM{}{\batch}$, and $\MIM{0}{\batch}$.
    This is true because the randomness of $Z$ arises from the interference matrices which are assumed to be independent of everything in the model, see Assumption~\ref{asmp:apndx_Gaussian Interference Matrice}.

    Now, we use Eq.~\eqref{eq::BL-proof-a0-dynamics} to derive the main result. Fix an arbitrary function $\psi \in \poly{k}$ and based on Eqs.~\eqref{eq:apndx_outcome_function} and \eqref{eq:apndx_outcome_function_WOD}, define the function $\widetilde\psi$ such that
    \begin{align}
        \label{eq:BL-a0-function_re_def}
        \psi\big(\outcomeD{}{i}{0},
        \outcomeD{}{i}{1},
        \Vtreatment{i}{}, \Vcovar{i}
        \big)
        =
        \psi\left(
        \outcomeD{}{i}{0},
        \outcome{}{i}{1} +
        \MAVO{}{i}{1N} +
        \outcomeh{0}{} \big(\outcomeD{}{i}{0},\Vtreatment{i}{},\Vcovar{i}\big),
        \Vtreatment{i}{}, \Vcovar{i}
        \right)
        =
        \widetilde{\psi}\left(
        \outcomeD{}{i}{0},
        \outcome{}{i}{1}, 
        \Vtreatment{i}{}, \Vcovar{i}, \MAVO{}{i}{1N}
        \right).
    \end{align}
    The function $\widetilde\psi$ is within $\poly{k}$ by Assumption~\ref{asmp:BL}. Then, applying \eqref{eq::BL-proof-a0-dynamics} for the function $\widetilde{\psi}$, we obtain,
    \begin{equation}
        \label{eq::BL-proof-a0-dynamics2}
        \begin{aligned}
            \lim_{N \rightarrow \infty}
            \frac{1}{\cardinality{\batch}} \sum_{i \in \batch}
            \psi\big(\outcomeD{}{i}{0},
            \outcomeD{}{i}{1},
            \Vtreatment{i}{}, \Vcovar{i}
            \big)
            \eqas
            \E \left[
            \psi\left(
            \outcomeD{}{\batch}{0},
            \MAVO{}{\batch}{1}
            +
            \MVVO{}{}{1} Z + 
            \Houtcome{\batch}{}{0}
            ,
            \Vtreatment{\batch}{}, \Vcovar{\batch}
            \right)
            \right]
        \end{aligned}
    \end{equation}
    where $\Houtcome{\batch}{}{0} = \outcomeh{0}{} \big(\outcomeD{}{\batch}{0},\Vtreatment{\batch}{},\Vcovar{\batch}\big)$ is a random variable independent of $Z$.

    In the second step of the induction, we also require the following results. Note that the result in \eqref{eq::BL-proof-a0-dynamics2} represents a specific instance of the more comprehensive result \eqref{eq:BL-proof-a0-dynamics-with-eps}. These results can be derived by following the same procedure outlined above.
    \begin{equation}
        \label{eq:BL-proof-a0-dynamics-with-eps0}
        \begin{aligned}
            \lim_{N \rightarrow \infty}
            \frac{1}{N} \sum_{i = 1}^N
            &\psi\big(
            \outcomeD{}{i}{0},
            \outcomeD{}{i}{1},
            \outcome{}{i}{1},
            \outcome{}{i}{1}
            - \IMatTv{i \cdot}{0} \VUoutcome{}{}{0},
            \Vtreatment{i}{},\Vcovar{i},
            \mu^{ji}, \mu^{ji}_r
            \big)
            \\
            \eqas
            \E
            \Big[
            &\psi
            \big(
            \outcomeD{}{}{0},
            \MAVO{}{}{1}
            + \MVVO{}{}{1} Z
            + \Houtcome{}{}{0},
            \MVVO{}{}{1} Z,
            \BVVO{}{}{1} Z',
            \Vtreatment{}{},\Vcovar{},
            \MIM{}{j}, \MIM{r}{j}
            \big)
            \Big],
        \end{aligned}
    \end{equation}
    as well as
    \begin{equation}
        \label{eq:BL-proof-a0-dynamics-with-eps}
        \begin{aligned}
            \lim_{N \rightarrow \infty}
            \frac{1}{\cardinality{\batch}} \sum_{i \in \batch}
            &\psi\big(
            \outcomeD{}{i}{0},
            \outcomeD{}{i}{1},
            \outcome{}{i}{1},
            \outcome{}{i}{1}
            - \IMatTv{i \cdot}{0} \VUoutcome{}{}{0},
            \Vtreatment{i}{},\Vcovar{i},
            \bar{\mu}^i, \bar{\mu}^i_r
            \big)
            \\
            \eqas
            \E
            \Big[
            &\psi
            \big(
            \outcomeD{}{\batch}{0},
            \MAVO{}{\batch}{1}
            + \MVVO{}{}{1} Z
            + \Houtcome{\batch}{}{0},
            \MVVO{}{}{1} Z,
            \BVVO{}{}{1} Z',
            \Vtreatment{\batch}{},\Vcovar{\batch},
            \MIM{}{\batch}, \MIM{r}{\batch}
            \big)
            \Big],
        \end{aligned}
    \end{equation}
    for any $j$ and all $r \in [T]_0$; here, $Z'$ is a standard Gaussian random variable and $\IMatTv{i \cdot}{0}$ detnoes the $i^{th}$ row of the time-dependent interference matrix $\IMatT{0}$.

    \item By \eqref{eq:BL-a0-Y1 stats limits} and \eqref{eq:BL-proof-a0-dynamics-with-eps0}, we get
    \begin{equation}
        \label{eq:BL-proof-b0-1}
        \begin{aligned}
            \lim_{N \rightarrow \infty}
            \frac{1}{N}
            \sum_{i=1}^N
            \big(\outcome{}{i}{1}
            \big)^2
            &\eqas
            (\MVVO{}{}{1})^2
            =
            \lim_{N \rightarrow \infty}
            \frac{\sigma^2+\sigma_0^2}{N}
            \sum_{i=1}^N
            \left(\Uoutcome{i}{}{0}\right)^2.
        \end{aligned}
    \end{equation}
    Similarly, by \eqref{eq:state evolution_fixed part} and \eqref{eq:BL-proof-a0-dynamics-with-eps0}, we can write %
    \begin{equation}
        \label{eq:BL-proof-b0-2}
        \begin{aligned}
            \lim_{N \rightarrow \infty}
            \frac{1}{N}
            \sum_{i=1}^N
            \big(\outcome{}{i}{1}
            - \IMatTv{i \cdot}{0} \VUoutcome{}{}{0}
            \big)^2
            &\eqas
            (\BVVO{}{}{1})^2
            =
            \lim_{N \rightarrow \infty}
            \frac{\sigma^2}{N}
            \sum_{i=1}^N
            \left(\Uoutcome{i}{}{0}\right)^2.
        \end{aligned}
    \end{equation}
    
    Finally, given $\VUoutcome{}{}{0}$, we know that $\IMatv{i \cdot} \VUoutcome{}{}{0}$ and $\IMatTv{i \cdot}{0} \VUoutcome{}{}{0}$ are statistically independent. Then, applying \eqref{eq:BL-proof-a0-dynamics-with-eps0} together with \eqref{eq:BL-proof-b0-2}, we obtain the following result:
    \begin{equation*}
        \begin{aligned}
            \lim_{N \rightarrow \infty}
            \frac{1}{N}
            \sum_{i=1}^N
            \outcome{}{i}{1}
            \big(\outcome{}{i}{1}
            - \IMatTv{i \cdot}{0} \VUoutcome{}{}{0}
            \big)
            =
            \lim_{N \rightarrow \infty}
            \frac{1}{N}
            \sum_{i=1}^N
            \big(\IMatv{i \cdot} \VUoutcome{}{}{0}
            +
            \IMatTv{i \cdot}{0} \VUoutcome{}{}{0}
            \big)
            \big(\IMatv{i \cdot} \VUoutcome{}{}{0}
            \big)
            &\eqas
            (\BVVO{}{}{1})^2
            =
            \lim_{N \rightarrow \infty}
            \frac{\sigma^2}{N}
            \sum_{n=1}^N
            \left(\Uoutcome{n}{}{0}\right)^2.
        \end{aligned}
    \end{equation*}
    where $\IMatv{i \cdot}$ and $\IMatTv{i \cdot}{0}$ denote the $i^{th}$ row of the fixed interference matrix $\IMat{}$ and time-dependent interference matrix $\IMatT{}$.

    \item For $t=1$, the matrix $\bm{Q}_1$ is equal to the vector $\VUoutcome{}{}{0}$ and $\bm{V}_1$ is equal to the vector $\VoutcomeD{}{}{1}$. By Assumption~\ref{asmp:BL}-\ref{asmp:BL-bound on initials} and \eqref{eq:BL-proof-b0-1}, we have
    \begin{align*}
        \lim_{N \rightarrow \infty} \frac{\bm{Q}_1^\top \bm{Q}_1}{N}
        =
        \lim_{N \rightarrow \infty}
        \frac{1}{N}
        \sum_{i=1}^N \left(\Uoutcome{i}{}{0}\right)^2 > 0,
        \quad
        \lim_{N \rightarrow \infty} \frac{\MOG{1}^\top \MOG{1}}{N} 
        =
        \lim_{N \rightarrow \infty} \pdot{\Voutcome{}{}{1}}{\Voutcome{}{}{1}}
        > 0.
    \end{align*}
    \end{enumerate}

    \textbf{Induction Hypothesis (IH).} Now, we assume that the following results hold true:
    \begin{equation}
        \label{eq:BL-proof-IH-a}
        \tag{IH-1}
        \begin{aligned}
            \lim_{N \rightarrow \infty}
            \frac{1}{N} \sum_{i = 1}^N
            &\psi\big(
            \outcomeD{}{i}{0},
            \outcomeD{}{i}{1},
            \ldots,
            \outcomeD{}{i}{t},
            \outcome{}{i}{1},
            \ldots,
            \outcome{}{i}{t},
            \outcome{}{i}{1}
            - \IMatTv{i \cdot}{0} \VUoutcome{}{}{0},
            \ldots,
            \outcome{}{i}{t}
            - \IMatTv{i \cdot}{t-1} \VUoutcome{}{}{t},
            \Vtreatment{i}{},\Vcovar{i},
            \mu^{ji}, \mu^{ji}_r
            \big)
            \\
            \eqas
            \E
            \Big[
            &\psi
            \big(
            \outcomeD{}{}{0},
            \MAVO{}{}{1}
            + \MVVO{}{}{1} Z_1
            + \Houtcome{}{}{0},
            \ldots,
            \MAVO{}{}{t}
            + \MVVO{}{}{t} Z_t
            + \Houtcome{}{}{t-1},
            \MVVO{}{}{1} Z_1,
            \ldots,
            \MVVO{}{}{t} Z_t,
            \BVVO{}{}{1} Z'_1,
            \ldots,
            \BVVO{}{}{t} Z'_t,
            \Vtreatment{}{},\Vcovar{},
            \MIM{}{j}, \MIM{r}{j}
            \big)
            \Big],
        \end{aligned}
    \end{equation}
    and
    \begin{equation}
        \label{eq:BL-proof-IH-a1}
        \tag{IH-2}
        \begin{aligned}
            \lim_{N \rightarrow \infty}
            \frac{1}{\cardinality{\batch}} \sum_{i \in \batch}
            &\psi\big(
            \outcomeD{}{i}{0},
            \outcomeD{}{i}{1},
            \ldots,
            \outcomeD{}{i}{t},
            \outcome{}{i}{1},
            \ldots,
            \outcome{}{i}{t},
            \outcome{}{i}{1}
            - \IMatTv{i \cdot}{0} \VUoutcome{}{}{0},
            \ldots,
            \outcome{}{i}{t}
            - \IMatTv{i \cdot}{t-1} \VUoutcome{}{}{t},
            \Vtreatment{i}{},\Vcovar{i},
            \bar{\mu}^i, \bar{\mu}^i_r
            \big)
            \\
            \eqas
            \E
            \Big[
            &\psi
            \big(
            \outcomeD{}{\batch}{0},
            \MAVO{}{\batch}{1}
            + \MVVO{}{}{1} Z_1
            + \Houtcome{\batch}{}{0},
            \ldots,
            \MAVO{}{\batch}{t}
            + \MVVO{}{}{t} Z_t
            + \Houtcome{\batch}{}{t-1},
            \MVVO{}{}{1} Z_1,
            \ldots,
            \MVVO{}{}{t} Z_t,
            \BVVO{}{}{1} Z'_1,
            \ldots,
            \BVVO{}{}{t} Z'_t,
            \Vtreatment{\batch}{},\Vcovar{\batch},
            \MIM{}{\batch}, \MIM{r}{\batch}
            \big)
            \Big].
        \end{aligned}
    \end{equation}
    Also, for $0 \leq s \neq r \leq t-1$, we have
    \begin{align}
            \label{eq:BL-proof-IH-b-1}
            \tag{IH-3}
            \lim_{N \rightarrow \infty} 
            \frac{1}{N}
            \sum_{i=1}^N
            (\outcome{}{i}{r+1})^2
            &\eqas
            (\MVVO{}{}{r+1})^2
            \eqas
            \lim_{N \rightarrow \infty}
            \frac{\sigma^2+\sigma^2_{r}}{N} \sum_{i=1}^N (\Uoutcome{i}{}{r})^2
            \\
            \label{eq:BL-proof-IH-b-2}
            \tag{IH-4}
            \lim_{N \rightarrow \infty} 
            \frac{1}{N}
            \sum_{i=1}^N
            \outcome{}{i}{r+1}\outcome{}{i}{s+1}
            &\eqas
            \lim_{N \rightarrow \infty}
            \frac{\sigma^2}{N} \sum_{i=1}^N \Uoutcome{i}{}{r}\Uoutcome{i}{}{s},
            \\
            \label{eq:BL-proof-IH-b-3}
            \tag{IH-5}
            \lim_{N \rightarrow \infty}
            \frac{1}{N}
            \sum_{i=1}^N
            \big(\outcome{}{i}{r+1}
            - \IMatTv{i \cdot}{r} \VUoutcome{}{}{r}
            \big)^2
            &\eqas
            (\BVVO{}{}{r+1})^2
            \eqas
            \lim_{N \rightarrow \infty}
            \frac{\sigma^2}{N}
            \sum_{i=1}^N
            \left(\Uoutcome{i}{}{r}\right)^2,
            \\
            \label{eq:BL-proof-IH-b-4}
            \tag{IH-6}
            \lim_{N \rightarrow \infty}
            \frac{1}{N}
            \sum_{i=1}^N
            \big(\outcome{}{i}{r+1}
            - \IMatTv{i \cdot}{r} \VUoutcome{}{}{r}
            \big)
            \big(\outcome{}{i}{s+1}
            - \IMatTv{i \cdot}{s} \VUoutcome{}{}{s}
            \big)
            &\eqas
            \lim_{N \rightarrow \infty}
            \frac{\sigma^2}{N}
            \sum_{i=1}^N
            \Uoutcome{i}{}{r} \Uoutcome{i}{}{s},
            \\
            \label{eq:BL-proof-IH-b-5}
            \tag{IH-7}
            \lim_{N \rightarrow \infty}
            \frac{1}{N}
            \sum_{i=1}^N
            \outcome{}{i}{r+1}
            \big(\outcome{}{i}{s+1}
            - \IMatTv{i \cdot}{s} \VUoutcome{}{}{s}
            \big)
            &\eqas
            \lim_{N \rightarrow \infty}
            \frac{\sigma^2}{N}
            \sum_{i=1}^N
            \Uoutcome{i}{}{r} \Uoutcome{i}{}{s}.
    \end{align}
    Finally, the following condition holds almost surely:
    \begin{align}
            \label{eq:BL-proof-IH-c}
            \tag{IH-8}
            \lim_{N \rightarrow \infty} \frac{\MOG{t-1}^\top \MOG{t-1}}{N} 
            \succ 0.
    \end{align}

    \textbf{Step 2.} To establish the second step of the induction, we prove the assertions in reverse order, starting with Part (c), followed by Part (b), and concluding with Part (a).

    \begin{enumerate}[label=(\alph*)]
        \item[(c)] We begin the second step by applying \eqref{eq:BL-proof-IH-a} to the function $g_s\big(\outcomeD{}{i}{s},\Vtreatment{i}{},\Vcovar{i}\big) g_r\big(\outcomeD{}{i}{r},\Vtreatment{i}{},\Vcovar{i}\big)$, for $1 \leq r,s \leq t$. Precisely, by Assumption~\ref{asmp:BL}-\ref{asmp:BL-bound on initials} as well as \eqref{eq:BL-proof-IH-a}, we get
        \begin{equation}
            \label{eq:BL-proof-ct-1}
            \begin{aligned}
            \lim_{N \rightarrow \infty}
            \frac{1}{N}
            \sum_{i=1}^N
            (\Uoutcome{i}{}{0})^2
            &\eqas
            \E\left[
            \outcomeg{0}{}(\outcomeD{}{}{0},\Vtreatment{}{},\Vcovar{})^2\big)
            \right]
            =
            \frac{(\MVVO{}{}{1})^2}{\sigma^2 + \sigma_0^2} > 0,
            \\
            \lim_{N \rightarrow \infty}
            \frac{1}{N}
            \sum_{i=1}^N
            \Uoutcome{i}{}{0} \Uoutcome{i}{}{s}
            &\eqas
            \E\left[
            \outcomeg{0}{}(\outcomeD{}{}{0},\Vtreatment{}{},\Vcovar{}) \outcomeg{s}{}\big(\MAVO{}{}{s}
            +
            \MVVO{}{}{s} Z_{s} + 
            \Houtcome{}{}{s-1},
            \Vtreatment{}{},\Vcovar{}\big)
            \right],
            \\
            \lim_{N \rightarrow \infty}
            \frac{1}{N}
            \sum_{i=1}^N
            \Uoutcome{i}{}{s} \Uoutcome{i}{}{r}
            &\eqas
            \E\Big[
            \outcomeg{s}{}\big(\MAVO{}{}{s}
            +
            \MVVO{}{}{s} Z_{s} + 
            \Houtcome{}{}{s-1},
            \Vtreatment{}{},\Vcovar{}\big) \outcomeg{r}{}\big(\MAVO{}{}{r}+
            \MVVO{}{}{r} Z_{r} + 
            \Houtcome{}{}{r-1},
            \Vtreatment{}{},\Vcovar{}\big)
            \Big].
            \end{aligned}
        \end{equation}
        Now, let $\Vec{u} = (u_1,\ldots,u_t)^\top \in \R^t$ be a non-zero vector. By Assumption~\ref{asmp:BL}-\ref{asmp:BL-bound on initials} and \eqref{eq:BL-proof-ct-1}, we have
        \begin{equation}
            \label{eq:BL-proof-ct-2}
            \begin{aligned}
                \Vec{u}^\top \left(\lim_{N \rightarrow \infty} \frac{\bm{Q}_t^\top \bm{Q}_t}{N}\right) \Vec{u}
                &=
                \lim_{N \rightarrow \infty} \Vec{u}^\top \frac{\bm{Q}_t^\top \bm{Q}_t}{N} \Vec{u}
                \\
                &\eqas
                \E\left[
                \left(
                u_1
                \outcomeg{0}{}(\outcomeD{}{}{0},\Vtreatment{}{},\Vcovar{})
                +
                \sum_{s=1}^{t-1}
                u_{s+1} 
                \outcomeg{s}{}\big(\MAVO{}{}{s}
                +
                \MVVO{}{}{s} Z_{s} + 
                \Houtcome{}{}{s-1},
                \Vtreatment{}{},\Vcovar{}\big)
                \right)^2
                \right] \geq 0.
            \end{aligned}
        \end{equation}
        We show that the inequality in \eqref{eq:BL-proof-ct-2} is strict. To this end, note that $\Vec{u}$ is a non-zero vector, and there exists some $1\leq i \leq t$ such that $u_i \neq 0$. Whenever $u_1 \neq 0 = u_2 = \ldots = u_t$, the result is immediate. Otherwise, recall that $y \mapsto \outcomeg{s}{}(y,\Vtreatment{}{},\Vcovar{})$ is a non-constant function with a positive probability with respect to $(\Vtreatment{}{},\Vcovar{})$; consequently, the mapping $(y_0,\ldots,y_{t-1}) \mapsto \sum_{s=0}^{t-1} u_s \outcomeg{s}{}\big(y_s,\Vtreatment{}{},\Vcovar{}\big)$ is a non-constant function as well. Considering $\Houtcome{}{}{s} = \outcomeh{s}{} \big(\MAVO{}{}{s} + \MVVO{}{}{s} Z_{s} + \Houtcome{}{}{s-1},\Vtreatment{}{},\Vcovar{}\big)$ and $\Houtcome{}{}{0} = \outcomeh{0}{} \big(\outcomeD{}{}{0},\Vtreatment{}{},\Vcovar{}\big)$, the randomness of $u_1\outcomeg{0}{}(\outcomeD{}{}{0},\Vtreatment{}{},\Vcovar{}) + \sum_{s=1}^{t-1} u_{s+1} \outcomeg{s}{}\big(\MAVO{}{}{s} + \MVVO{}{}{s} Z_{s} + \Houtcome{}{}{s-1}, \Vtreatment{}{},\Vcovar{}\big)$ comes solely from $\outcomeD{}{}{0}, Z_1, \ldots, Z_{t-1}, \Vtreatment{}{}, \Vcovar{}$; as a result, there exists a measurable continuous function $\phi$ (which depends on $\Vec{u}, \outcomeg{0}{}, \ldots, \outcomeg{t-1}{}, \outcomeh{0}{}, \ldots, \outcomeh{t-1}{}$) such that we can rewrite the right-hand side of \eqref{eq:BL-proof-ct-2} as follows, 
        \begin{equation*}
            \begin{aligned}
                &\;\E\left[
                \left(
                u_1
                \outcomeg{0}{}(\outcomeD{}{}{0},\Vtreatment{}{},\Vcovar{})
                +
                \sum_{s=1}^{t-1}
                u_{s+1} 
                \outcomeg{s}{}\big(\MAVO{}{}{s}
                +
                \MVVO{}{}{s} Z_{s} + 
                \Houtcome{}{}{s-1},
                \Vtreatment{}{},\Vcovar{}\big)
                \right)^2
                \right]
                \\
                = 
                &\;\E\left[
                \phi\left(
                \outcomeD{}{}{0}, \MAVO{}{}{1}, \ldots, \MAVO{}{}{t-1}, \MVVO{}{}{1} Z_1, \ldots, \MVVO{}{}{t-1} Z_{t-1}, \Vtreatment{}{}, \Vcovar{}
                \right)^2
                \right]
                \\
                =
                &\;\E\left[
                \E\left[
                \phi\left(
                y_0, \MAVO{}{}{1}, \ldots, \MAVO{}{}{t-1}, \MVVO{}{}{1} Z_1, \ldots, \MVVO{}{}{t-1} Z_{t-1}, \Vtreatment{}{}, \Vcovar{}
                \right)^2
                \Bigg|
                \outcomeD{}{}{0} = y_0
                \right]
                \right],
            \end{aligned}
        \end{equation*}
        where in the last equality we used the tower property of conditional expectations. Then, it suffices to show that the random variable $\phi\left( y_0, \MAVO{}{}{1}, \ldots, \MAVO{}{}{t-1}, \MVVO{}{}{1} Z_1, \ldots, \MVVO{}{}{t-1} Z_{t-1}, \Vtreatment{}{}, \Vcovar{} \right)$ has a non-degenerate distribution. To obtain that, by \eqref{eq:BL-proof-IH-a}, it is straightforward to obtain the following, 
        \begin{equation*}
            \begin{aligned}
                \Cov\left[\left(
                \MVVO{}{}{1} Z_{1},
                \ldots, \MVVO{}{}{t-1} Z_{t-1} \right)\right]
                \eqas
                \lim_{N \rightarrow \infty} \frac{\MOG{t-1}^\top \MOG{t-1}}{N},
            \end{aligned}
        \end{equation*}
        which is positive definite by \eqref{eq:BL-proof-IH-c}, and the proof of the first claim is complete.

        To proceed to the proof of the second claim, for $1\leq r,s \leq t$, let us denote
        \begin{align*}
            v_{r,s}
            :=
            \left[
            \frac{\MOG{t}^\top \MOG{t}}{N} 
            \right]^{r,s}
            =
            \frac{\Voutcome{}{\top}{r} \Voutcome{}{}{s}}{N}. 
        \end{align*}
        By \eqref{eq:BL-proof-IH-b-2}, whenever $r\neq s$, we can write
        \begin{align*}
            \lim_{N \rightarrow \infty}
            v_{r,s}
            \eqas
            \lim_{N \rightarrow \infty}
            \frac{\sigma^2}{N} \sum_{i=1}^N \Uoutcome{i}{}{r-1}\Uoutcome{i}{}{s-1},
        \end{align*}
        and if $r=s$, by \eqref{eq:BL-proof-IH-b-1}, we have
        \begin{align*}
            \lim_{N \rightarrow \infty}
            v_{r,r}
            \eqas
            \lim_{N \rightarrow \infty}
            \frac{\sigma^2+\sigma_r^2}{N} \sum_{i=1}^N (\Uoutcome{i}{}{r-1})^2.
        \end{align*}
        Then, the result follows directly, as we have just established the almost sure positive definiteness of $\bm{Q}_t$.
        \begin{corollary}
            \label{cl:alpha is bounded}
            The vector $\VAPC{}{t}$ defined in \eqref{eq:projection coefficients} has a finite limit as $N \rightarrow \infty$.
        \end{corollary}
        Proof. By \eqref{eq:projection coefficients}, we can write
        \begin{align}
            \label{eq:cl-alpha is finite}
            \lim_{N \rightarrow \infty} \VAPC{}{t}
            =
            \lim_{N \rightarrow \infty} \left(
            \bm{Q}_t^\top \bm{Q}_t
            \right)^{-1}
            \bm{Q}_t^\top \VUoutcome{}{}{t}
            =
            \lim_{N \rightarrow \infty} \left(
            \frac{\bm{Q}_t^\top \bm{Q}_t}{N}
            \right)^{-1}
            \lim_{N \rightarrow \infty}
            \frac{\bm{Q}_t^\top \VUoutcome{}{}{t}}{N}.
        \end{align}    
        Using the result of part~\ref{part:BL-c}, for large values of $N$, the matrix $\frac{\bm{Q}_t^\top \bm{Q}_t}{N}$ is positive definite (this is true because the eigenvalues of a matrix vary continuously with respect to its entries). Then, note that the mapping $\bm{G} \mapsto \bm{G}^{-1}$ is continuous for any invertible matrix $\bm{G}$. As a result, we get
        \begin{align*}
            \lim_{N \rightarrow \infty} \left(\frac{\bm{Q}_t^\top \bm{Q}_t}{N}\right)^{-1}
            =
            \left(\lim_{N \rightarrow \infty}\frac{\bm{Q}_t^\top \bm{Q}_t}{N}\right)^{-1}.
        \end{align*}
        Since the matrix $\lim_{N \rightarrow \infty} \frac{\bm{Q}_t^\top \bm{Q}_t}{N}$ is positive definite, the left term in \eqref{eq:cl-alpha is finite} is well-defined and finite. The finiteness of the right term is the consequence of \eqref{eq:BL-proof-ct-1}. \ep

        \item[(b)] We first derive several intermediate results and then utilize them to demonstrate that \eqref{eq:BL-b} holds true for $0 \leq r, s \leq t$. In this process, we apply the Strong Law of Large Numbers (SLLN) from Theorem~\ref{thm:SLLN} multiple times, without explicitly verifying the conditions, as they are straightforward.

        By Lemma~\ref{lm:conditional dist of outcome}, conditioning on $\Gc_t$, the terms $\IMatvnew{i\cdot} \VUoutcome{\perp}{}{t}$ for $i \in [N]$ are i.i.d. Gaussian random variables. Similarly, the terms $\IMatTv{i \cdot}{t} \VUoutcome{}{}{t}$ for $i \in [N]$ are also i.i.d. Gaussian random variables:
        \begin{equation}
            \label{eq:BL-proof-bt-simple-1}
            \begin{aligned}
                \IMatvnew{i\cdot} \VUoutcome{\perp}{}{t} \sim \Nc\left(0, \frac{\sigma^2}{N} \sum_{j=1}^N (\Uoutcome{\perp,j}{}{t})^2\right),
            \quad\quad
                \IMatTv{i \cdot}{t} \VUoutcome{}{}{t} \sim \Nc\left(0, \frac{\sigma_t^2}{N} \sum_{j=1}^N (\Uoutcome{j}{}{t})^2\right).
            \end{aligned}
        \end{equation}
        Now, applying Theorem~\ref{thm:SLLN}, we get the following results:
        \begin{equation}
            \label{eq:BL-proof-bt-simple-4}
            \begin{aligned}
                \lim_{N \rightarrow \infty}
                \frac{1}{N}
                \sum_{i=1}^N
                \left(
                \IMatvnew{i\cdot} 
                \VUoutcome{\perp}{}{t}
                \right)^2
                &\eqas
                \lim_{N \rightarrow \infty} \frac{\sigma^2}{N}
                \sum_{i=1}^N
                \left(\Uoutcome{\perp,i}{}{t}\right)^2,
            \end{aligned}
        \end{equation}
        as well as
        \begin{equation}
            \label{eq:BL-proof-bt-simple-5}
            \begin{aligned}
                \lim_{N \rightarrow \infty}
                \frac{1}{N}
                \sum_{i=1}^N
                \left(
                \IMatTv{i \cdot}{t}
                \VUoutcome{}{}{t}
                \right)^2
                &\eqas
                \lim_{N \rightarrow \infty} \frac{\sigma_t^2}{N}
                \sum_{i=1}^N
                \left(\Uoutcome{i}{}{r}\right)^2.
            \end{aligned}
        \end{equation}
        Also, considering \eqref{eq:Q and R} and applying \eqref{eq:BL-proof-IH-b-4}, we obtain
        \begin{equation}
        \label{eq:BL-proof-bt-simple-7}
            \begin{aligned}
                &\lim_{N \rightarrow \infty}
                \frac{1}{N}
                \sum_{i=1}^N
                \left(
                \left[
                \bm{R}_t
                \VAPC{}{t}
                \right]^i
                \right)^2
                \\
                =
                &\lim_{N \rightarrow \infty}
                \frac{1}{N}
                \sum_{i=1}^N
                \left(
                \sum_{s=0}^{t-1}
                \APC{t}{s}
                \big(
                \outcome{}{i}{s+1}
                - \IMatTv{i \cdot}{s} \VUoutcome{}{}{s}
                \big)
                \right)^2
                \\
                =
                &\lim_{N \rightarrow \infty}
                \frac{1}{N}
                \sum_{i=1}^N
                \sum_{0\leq s,r < t}
                \APC{t}{s}
                \APC{t}{r}
                \big(
                \outcome{}{i}{s+1} - \IMatTv{i \cdot}{s} \VUoutcome{}{}{s}
                \big)
                \big(
                \outcome{}{i}{r+1} - \IMatTv{i \cdot}{r} \VUoutcome{}{}{r}
                \big)
                \\
                =
                &\sum_{0\leq s,r < t}
                \APC{t}{s}
                \APC{t}{r}
                \left(
                \lim_{N \rightarrow \infty}
                \frac{1}{N}
                \sum_{i=1}^N
                \big(
                \outcome{}{i}{s+1} - \IMatTv{i \cdot}{s} \VUoutcome{}{}{s}
                \big)
                \big(
                \outcome{}{i}{r+1} - \IMatTv{i \cdot}{r} \VUoutcome{}{}{r}
                \big)\right)
                \\
                \eqas
                & \lim_{N \rightarrow \infty}
                \frac{\sigma^2}{N}
                \sum_{i=1}^N
                \sum_{0\leq s,r < t}
                \APC{t}{s}
                \APC{t}{r}
                \Uoutcome{i}{}{s} \Uoutcome{i}{}{r}
                \\
                =
                &\lim_{N \rightarrow \infty}
                \frac{\sigma^2}{N}
                \sum_{i=1}^N
                \left(\Uoutcome{\parallel,i}{}{t}\right)^2,
            \end{aligned}
        \end{equation}
        where in the last line we used \eqref{eq:projection sum}.

        Now, we first obtain \eqref{eq:BL-b-1} for $r=t$. By \eqref{eq:conditional dist of outcome_nonsym}, \eqref{eq:BL-proof-bt-simple-4}, \eqref{eq:BL-proof-bt-simple-5}, and \eqref{eq:BL-proof-bt-simple-7}, we can write
        \begin{equation}
            \label{eq:BL-proof-bt-2-1}
            \begin{aligned}
                \lim_{N \rightarrow \infty}
                \frac{1}{N}
                \sum_{i=1}^N
                \big(
                \outcome{}{i}{t+1}
                \big)^2
                &\eqas
                \lim_{N \rightarrow \infty}
                \frac{1}{N}
                \sum_{i=1}^N
                \left(
                \IMatvnew{i\cdot} 
                \VUoutcome{\perp}{}{t}
                +
                \left[
                \bm{R}_t
                \VAPC{}{t}
                \right]^i
                +
                \IMatTv{i \cdot}{t} \VUoutcome{}{}{t}
                \right)^2
                \\
                &\eqas
                \lim_{N \rightarrow \infty} \frac{\sigma^2}{N}
                \sum_{i=1}^N
                \left(\Uoutcome{\perp,i}{}{t}\right)^2
                +
                \lim_{N \rightarrow \infty}
                \frac{\sigma^2}{N}
                \sum_{i=1}^N
                \left(\Uoutcome{\parallel,i}{}{t}\right)^2
                +
                \lim_{N \rightarrow \infty} \frac{\sigma_t^2}{N}
                \sum_{i=1}^N
                \left(\Uoutcome{i}{}{t}\right)^2
                \\
                &\quad
                +
                \lim_{N \rightarrow \infty}
                \frac{2}{N}
                \sum_{i=1}^N
                \left(\IMatvnew{i\cdot} 
                \VUoutcome{\perp}{}{t} \left[
                \bm{R}_t
                \VAPC{}{t}
                \right]^i \right)
                +
                \lim_{N \rightarrow \infty}
                \frac{2}{N}
                \sum_{i=1}^N
                \left(\IMatvnew{i\cdot} 
                \VUoutcome{\perp}{}{t} \IMatTv{i \cdot}{t} \VUoutcome{}{}{t} \right)
                \\
                &\quad
                +
                \lim_{N \rightarrow \infty}
                \frac{2}{N}
                \sum_{i=1}^N
                \left( \left[
                \bm{R}_t
                \VAPC{}{t}
                \right]^i \IMatTv{i \cdot}{t}\VUoutcome{}{}{t} \right).
            \end{aligned}
        \end{equation}
        Note that the only random elements in the right-hand side of \eqref{eq:BL-proof-bt-2-1} are $\IMatvnew{i \cdot}$ and $\IMatTv{i \cdot}{t}$. Thus, by \eqref{eq:BL-proof-bt-simple-1} and applying Theorem~\ref{thm:SLLN}, we can demonstrate that the last three terms vanish, resulting in the following:
        \begin{align}
            \label{eq:BL-proof-bt-2-result}
            \lim_{N \rightarrow \infty}
            \frac{1}{N}
            \sum_{i=1}^N
            \big(
            \outcome{}{i}{t+1}
            \big)^2
            &\eqas
            \lim_{N \rightarrow \infty} \frac{\sigma^2+\sigma_t^2}{N}
            \sum_{i=1}^N
            \left(\Uoutcome{i}{}{t}\right)^2
            \eqas
            (\MVVO{}{}{t+1})^2,
        \end{align}
        where the last equality is immediate by \eqref{eq:BL-proof-IH-a}.
        
        Next, we derive \eqref{eq:BL-b-2} for $r=t$ and $0\leq s \leq t-1$. Considering \eqref{eq:conditional dist of outcome_nonsym}, we can write
        \begin{equation}
        \label{eq:BL-proof-bt-3-1}
            \begin{aligned}
                &\lim_{N \rightarrow \infty}
                \frac{1}{N}
                \sum_{i=1}^N
                \outcome{}{i}{s+1}
                \outcome{}{i}{t+1}
                \\
                \eqas
                &\lim_{N \rightarrow \infty}
                \frac{1}{N}
                \sum_{i=1}^N
                \outcome{}{i}{s+1}
                \left(
                \IMatvnew{i\cdot} 
                \VUoutcome{\perp}{}{t}
                +
                \left[
                \bm{R}_t
                \VAPC{}{t}
                \right]^i
                + \IMatTv{i \cdot}{t} \VUoutcome{}{}{t}
                \right)
                \\
                \eqas
                &\lim_{N \rightarrow \infty}
                \frac{1}{N}
                \sum_{i=1}^N
                \left(
                \IMatvnew{i\cdot} 
                \VUoutcome{\perp}{}{t}
                \outcome{}{i}{s+1}
                +
                \left[
                \bm{R}_t
                \VAPC{}{t}
                \right]^i
                \outcome{}{i}{s+1}
                +
                \IMatTv{i \cdot}{t} 
                \VUoutcome{}{}{t}
                \outcome{}{i}{s+1}
                \right).
            \end{aligned}
        \end{equation}
        Note that by conditioning on $\Gc_t$, the value of $\outcome{}{n}{s+1}$ is deterministic. Then, applying Theorem~\ref{thm:SLLN} and considering \eqref{eq:BL-proof-bt-simple-1}, \eqref{eq:Q and R}, and by \eqref{eq:BL-proof-IH-b-5}, we get the desired result:
        \begin{equation}
            \label{eq:BL-proof-bt-3-4}
            \begin{aligned}
                \lim_{N \rightarrow \infty}
                \frac{1}{N}
                \sum_{i=1}^N
                \outcome{}{i}{s+1}
                \outcome{}{i}{t+1}
                &\eqas
                \lim_{N \rightarrow \infty}
                \frac{1}{N}
                \sum_{i=1}^N
                \left(
                \left[
                \bm{R}_t
                \VAPC{}{t}
                \right]^i
                \right)\outcome{}{i}{s+1}
                \\
                &=
                \lim_{N \rightarrow \infty}
                \frac{1}{N}
                \sum_{i=1}^N
                \left(
                \sum_{r=0}^{t-1}
                \APC{r}{t}
                \big(
                \outcome{}{i}{r+1}
                - \IMatTv{i \cdot}{r} \VUoutcome{}{}{r}
                \big)
                \outcome{}{i}{s+1}
                \right)
                \\
                &\eqas
                \sum_{r=0}^{t-1}
                \APC{r}{t} 
                \left(
                \lim_{N \rightarrow \infty}
                \frac{\sigma^2}{N} \sum_{i=1}^N \Uoutcome{i}{}{r}\Uoutcome{i}{}{s}
                \right)
                \\
                &\eqas
                \lim_{N \rightarrow \infty}\hspace{-1mm}
                \left(
                \frac{\sigma^2}{N} \sum_{i=1}^N \sum_{r=0}^{t-1}
                \APC{r}{t}
                \Uoutcome{i}{}{r}\Uoutcome{i}{}{s}\right)
                \\
                &=
                \lim_{N \rightarrow \infty}
                \frac{\sigma^2}{N}
                \sum_{i=1}^N
                \left(
                \Uoutcome{\parallel,i}{}{t}
                \Uoutcome{i}{}{s}
                \right)
                \\
                &=
                \lim_{N \rightarrow \infty}
                \frac{\sigma^2}{N}
                \sum_{i=1}^N
                \left(
                \Uoutcome{i}{}{t}
                \Uoutcome{i}{}{s}
                \right),
            \end{aligned}
        \end{equation}
        where in the last line, we used the fact that $\pdot{\VUoutcome{}{}{t}}{\VUoutcome{}{}{s}} = \pdot{\VUoutcome{\parallel}{}{t}}{\VUoutcome{}{}{s}}$ as $\VUoutcome{\perp}{}{t} \perp \VUoutcome{}{}{s}$.

        The derivations for \eqref{eq:BL-b-3} and \eqref{eq:BL-b-4} follow a similar procedure, which we omit here for brevity. We then apply a similar approach to obtain \eqref{eq:BL-b-5}. Specifically, fixing $0 \leq r \leq t-1$ and setting $s=t$, we can write the following using \eqref{eq:conditional dist of outcome_nonsym} and \eqref{eq:BL-proof-bt-3-4}:
        \begin{equation*}
            \begin{aligned}
                \lim_{N \rightarrow \infty}
                \frac{1}{N}
                \sum_{i=1}^N
                \big(
                \outcome{}{i}{t+1}
                - \IMatTv{i \cdot}{t} \VUoutcome{}{}{t}
                \big)
                \outcome{}{i}{r+1}
                \eqas
                \lim_{N \rightarrow \infty}
                \frac{1}{N}
                \sum_{i=1}^N
                \left(
                \IMatvnew{i\cdot} 
                \VUoutcome{\perp}{}{t}
                +
                \left[
                \bm{R}_t
                \VAPC{}{t}
                \right]^i
                \right)\outcome{}{i}{r+1}
                \eqas
                \lim_{N \rightarrow \infty}
                \frac{\sigma^2}{N}
                \sum_{i=1}^N
                \left(
                \Uoutcome{i}{}{t}
                \Uoutcome{i}{}{r}
                \right).
            \end{aligned}
        \end{equation*}
        Likewise, we can show the result for the case that $r=t$ and $0\leq s \leq t-1$:
        \begin{equation}
            \label{eq:BL-proof-bt-4-1}
            \begin{aligned}
                &\lim_{N \rightarrow \infty}
                \frac{1}{N}
                \sum_{i=1}^N
                \big(
                \outcome{}{i}{s+1}
                - \IMatTv{i \cdot}{s} \VUoutcome{}{}{s}
                \big)
                \outcome{}{i}{t+1}
                \\
                \eqas
                &\lim_{N \rightarrow \infty}
                \frac{1}{N}
                \sum_{i=1}^N
                \big(
                \outcome{}{i}{s+1}
                - \IMatTv{i \cdot}{s} \VUoutcome{}{}{s}
                \big)
                \left(
                \IMatvnew{i\cdot} 
                \VUoutcome{\perp}{}{t}
                +
                \left[
                \bm{R}_t
                \VAPC{}{t}
                \right]^i
                + \IMatTv{i \cdot}{t} \VUoutcome{}{}{t}
                \right)
                \\
                =
                &\lim_{N \rightarrow \infty}
                \frac{1}{N}
                \sum_{i=1}^N
                \left(
                \IMatvnew{i\cdot} 
                \VUoutcome{\perp}{}{t}
                \right)
                \big(
                \outcome{}{i}{s+1}
                - \IMatTv{i \cdot}{s} \VUoutcome{}{}{s}
                \big)
                \\
                &+
                \lim_{N \rightarrow \infty}
                \frac{1}{N}
                \sum_{i=1}^N
                \left(
                \IMatTv{i \cdot}{t} \VUoutcome{}{}{t}
                \right)
                \big(
                \outcome{}{i}{s+1}
                - \IMatTv{i \cdot}{s} \VUoutcome{}{}{s}
                \big),
                \\
                &+
                \lim_{N \rightarrow \infty}
                \frac{1}{N}
                \sum_{i=1}^N
                \left(
                \left[
                \bm{R}_t
                \VAPC{}{t}
                \right]^i
                \right)
                \big(
                \outcome{}{i}{s+1}
                - \IMatTv{i \cdot}{s} \VUoutcome{}{}{s}
                \big).
            \end{aligned}
        \end{equation}
        Then, by Theorem~\ref{thm:SLLN}, the first and second terms on the right-hand side are zero. Additionally,
        \eqref{eq:Q and R}, \eqref{eq:projection sum}, and \eqref{eq:BL-proof-IH-b-4} imply that
        \begin{equation}
        \label{eq:BL-proof-bt-4-4}
            \begin{aligned}
                \lim_{N \rightarrow \infty}
                \frac{1}{N}
                \sum_{i=1}^N
                \left(
                \left[
                \bm{R}_t
                \VAPC{}{t}
                \right]^i
                \right)
                \big(
                \outcome{}{i}{s+1}
                - \IMatTv{i \cdot}{s} \VUoutcome{}{}{s}
                \big)
                &=
                \lim_{N \rightarrow \infty}
                \frac{1}{N}
                \sum_{i=1}^N
                \sum_{r=0}^{t-1}
                \APC{r}{t}
                \big(
                \outcome{}{i}{r+1}
                - \IMatTv{i \cdot}{r} \VUoutcome{}{}{r}
                \big)
                \big(
                \outcome{}{i}{s+1}
                - \IMatTv{i \cdot}{s} \VUoutcome{}{}{s}
                \big)
                \\
                &\eqas
                \lim_{N \rightarrow \infty}
                \frac{\sigma^2}{N}
                \sum_{i=1}^N
                \Uoutcome{i}{}{s} \Uoutcome{i}{}{r},
            \end{aligned}
        \end{equation}
        where in the last line, we used $\pdot{\VUoutcome{}{}{t}}{\VUoutcome{}{}{s}} = \pdot{\VUoutcome{\parallel}{}{t}}{\VUoutcome{}{}{s}}$. The desired result follows by aggregating \eqref{eq:BL-proof-bt-4-1}-\eqref{eq:BL-proof-bt-4-4}.

        \item We use induction hypotheses to establish the following result:
        \begin{equation}
            \label{eq:BL-a_(t)}
            \begin{aligned}
                \lim_{N \rightarrow \infty}
                \frac{1}{\cardinality{\batch}} \sum_{i \in \batch}
                \psi\big(
                \outcomeD{}{i}{0},
                \outcomeD{}{i}{1},
                \ldots,
                \outcomeD{}{i}{t+1},
                \Vtreatment{i}{},\Vcovar{i}
                \big)
                \\
                \eqas
                \E
                \Big[
                \psi
                \big(
                \outcomeD{}{\batch}{0},
                \MAVO{}{\batch}{1}
                + \MVVO{}{}{1} Z_1
                + \Houtcome{\batch}{}{0},
                \ldots,
                \MAVO{}{\batch}{t+1}
                + \MVVO{}{}{t+1} Z_t
                + \Houtcome{\batch}{}{t},
                \Vtreatment{\batch}{}, \Vcovar{\batch}
                \big)
                \Big].
            \end{aligned}
        \end{equation}
        More general results related to the extension of \eqref{eq:BL-proof-a0-dynamics-with-eps0} and \eqref{eq:BL-proof-a0-dynamics-with-eps} follow a similar procedure and are omitted for brevity.
        
        We proceed by introducing a new notation. Fixing $i$ as an arbitrary unit, we define
        \begin{equation*}
            \begin{aligned}
                \Psi^{i}(N)
                &:=
                \psi\big(
                \outcomeD{}{i}{0},
                \outcomeD{}{i}{1}, \ldots,
                \outcomeD{}{i}{t},
                \outcome{}{i}{t+1},
                \Vtreatment{i}{},
                \Vcovar{i},
                \MAVO{}{i}{(t+1)N}
                \big),
            \end{aligned}
        \end{equation*}
        where
        \begin{equation*}
        \begin{aligned}
            \MAVO{}{i}{(t+1)N} &=
            \frac{1}{N} \sum_{k=1}^N (\mu^{ik} + \mu_t^{ik}) \outcomeg{t}{} \big(\outcomeD{}{k}{t}, \Vtreatment{k}{}, \Vcovar{k}\big).
        \end{aligned}
        \end{equation*}
        Using \eqref{eq:BL-proof-IH-a}, this implies that
        \begin{equation}
        \label{eq:BL-at-stat-2_limit}
        \begin{aligned}
            \lim_{N \rightarrow \infty} \MAVO{}{i}{(t+1)N}
            &\eqas
            \E \left[ (\MIM{}{i}+\MIM{t}{i}) \outcomeg{t}{} \big(\MAVO{}{}{t} + \MVVO{}{}{t} Z + \Houtcome{}{}{t-1},\Vtreatment{}{},\Vcovar{}\big) \right]
            \\ &=
            (\bar{\mu}^i+\bar{\mu}^i_t) \E \left[\outcomeg{t}{} \big(\MAVO{}{}{t} + \MVVO{}{}{t} Z + \Houtcome{}{}{t-1},\Vtreatment{}{},\Vcovar{}\big) \right]
            =
            \MAVO{}{i}{t+1} < \infty.
        \end{aligned}
        \end{equation}
        Now, by \eqref{eq:conditional dist of outcome_nonsym}, we can write
        \begin{equation*}
            \begin{aligned}
                \Psi^{i}(N)
                \Big|_{\Gc_t}
                \eqd
                \psi\left(
                \outcomeD{}{i}{0},
                \outcomeD{}{i}{1}, \ldots,
                \outcomeD{}{i}{t},
                \left[
                \widetilde{\IM}
                \VUoutcome{\perp}{}{t}
                + \bm{R}_t 
                \VAPC{}{t}
                +
                \IMatT{t} \VUoutcome{}{}{t}
                \right]^i,
                \Vtreatment{i}{},
                \Vcovar{i},
                \MAVO{}{i}{(t+1)N}
                \right),
            \end{aligned}
        \end{equation*}
        where $\left[ \widetilde{\IM} \VUoutcome{\perp}{}{t} + \bm{R}_t  \VAPC{}{t} + \IMatT{t} \VUoutcome{}{}{t} \right]^i$ represent the $i^{th}$ element in the vector $ \widetilde{\IM} \VUoutcome{\perp}{}{t} + \bm{R}_t  \VAPC{}{t} + \IMatT{t} \VUoutcome{}{}{t}$. We also let
        \begin{align*}
            \widetilde{\Psi}^{i}(N) = \Psi^{i}(N) - \E_{\IM,\IMatT{t}}[\Psi^{i}(N)].
        \end{align*}
        where $\E_{\IM,\IMatT{t}}$ denotes the expectation with respect to the randomness of the interference matrices $\IM$ and $\IMatT{t}$. We follow the same approach as Step~1-\ref{item:BL-average limit}. Note that given $\Gc_t$, the elements of $\widetilde{\IM} \VUoutcome{\perp}{}{t} + \IMatT{t} \VUoutcome{}{}{t}$ are i.i.d. Gaussian random variables with a zero mean and variance~$(\hat{\rho}_{tN})^2$:
        \begin{equation}
            \label{eq:BL-bt-Yt stat}
            \begin{aligned}
                (\hat{\rho}_{tN})^2
                &:=
                \Var
                \left[
                [
                \widetilde{\IM} \VUoutcome{\perp}{}{t} + \IMatT{t} \VUoutcome{}{}{t}
                ]^i
                \Big|
                \VUoutcome{}{}{t}
                \right]
                =
                \frac{\sigma^2}{N}
                \sum_{j=1}^N
                \left(\Uoutcome{\perp,j}{}{t}\right)^2
                +
                \frac{\sigma_t^2}{N}
                \sum_{j=1}^N
                \left(\Uoutcome{j}{}{t}\right)^2,
            \end{aligned}
        \end{equation}
        where $\Uoutcome{j}{}{t} = \outcomeg{t}{}\big(\outcomeD{}{j}{t}, \Vtreatment{j}{}, \Vcovar{j}\big)$ is the $j^{th}$ element of the column vector $\VUoutcome{}{}{t}$, and similarly, $\Uoutcome{\perp,j}{}{t}$ is the $j^{th}$ element of the column vector $\VUoutcome{\perp}{}{t}$. Letting
        \begin{align}
            \label{eq:BL-bt-(-1)}
            (\hat{\rho}_t)^2 = \lim_{N\rightarrow \infty } (\hat{\rho}_{tN})^2,
        \end{align}
        we have the following lemma regarding the finiteness of $\hat{\rho}_t$.
        \begin{lemma}
            \label{lm:finite variance}
            $\hat{\rho}_t$ is almost surely finite.
        \end{lemma}
        \textbf{Proof of Lemma~\ref{lm:finite variance}.} We show that the following relations hold with a probability of 1:
        \begin{align}
            \label{eq:BL-bt-1}
            \lim_{N \rightarrow \infty}
            \frac{1}{N}
            \sum_{j=1}^N
            \left(\Uoutcome{\perp,j}{}{t}\right)^2 < \infty,
            \;\;\;
            \lim_{N \rightarrow \infty}
            \frac{1}{N}
            \sum_{j=1}^N
            (\Uoutcome{j}{}{t})^2 < \infty.
        \end{align}
        By definition, we can write
        \begin{equation}
        \label{eq:BL-bt-2}
        \begin{aligned}
            \frac{1}{N}
            \sum_{j=1}^N
            \left(\Uoutcome{\perp,j}{}{t}\right)^2
            =
            \pdot{\VUoutcome{\perp}{}{t}}{\VUoutcome{\perp}{}{t}}
            &=
            \pdot{\VUoutcome{}{}{t}}{\VUoutcome{}{}{t}}
            -
            \pdot{\VUoutcome{\parallel}{}{t}}{\VUoutcome{\parallel}{}{t}}
            =
            \frac{1}{N}
            \sum_{j=1}^N
            \left(\Uoutcome{j}{}{t}\right)^2
            -
            \frac{1}{N}
            \sum_{j=1}^N
            \left(\Uoutcome{\parallel,j}{}{t}\right)^2.
        \end{aligned}
        \end{equation}
        Then, by \eqref{eq:BL-proof-IH-a} for the function $\outcomeg{t}{}\big(\outcomeD{}{j}{t}, \Vtreatment{j}{}, \Vcovar{j}\big)^2$, we get
        \begin{equation}
        \label{eq:BL-bt-3}
        \begin{aligned}
            \lim_{N \rightarrow \infty}
            \frac{1}{N}
            \sum_{j=1}^N
            \left(\Uoutcome{j}{}{t}\right)^2
            &=
            \lim_{N \rightarrow \infty}
            \frac{1}{N}
            \sum_{j=1}^N
            \outcomeg{t}{} \big(\outcomeD{}{j}{t}, \Vtreatment{j}{}, \Vcovar{j}\big)^2
            \eqas
            \E\left[
            \outcomeg{t}{} \big(\MAVO{}{}{t} + \MVVO{}{}{t} Z + \Houtcome{}{}{t-1}, \Vtreatment{}{}, \Vcovar{}\big)^2
            \right]  < \infty,
        \end{aligned}
        \end{equation}
        where $Z \sim \Nc(0,1)$. Further, by \eqref{eq:projection sum}, we have
        \begin{align*}
            \frac{1}{N}
            \sum_{j=1}^N
            \Uoutcome{\parallel,j}{}{t}
            &=
            \frac{1}{N}
            \sum_{j=1}^N
            \sum_{s=0}^{t-1} \APC{s}{t} \Uoutcome{j}{}{s}
            =
            \sum_{s=0}^{t-1}
            \frac{\APC{s}{t}}{N}
            \sum_{j=1}^N
            \Uoutcome{j}{}{s}
            \\
            \frac{1}{N}
            \sum_{j=1}^N
            \left(\Uoutcome{\parallel,j}{}{t}\right)^2,
            &=
            \frac{1}{N}
            \sum_{j=1}^N
            \left(
            \sum_{s=0}^{t-1} \APC{s}{t} \Uoutcome{j}{}{s}
            \right)^2
            =
            \sum_{r,s=0}^{t-1} \APC{r}{t} \APC{s}{t} \pdot{\VUoutcome{}{}{r}}{\VUoutcome{}{}{s}}.
        \end{align*}
        Considering Corollary~\ref{cl:alpha is bounded}, the vector $\VAPC{}{t}$ has a finite limit as $N\rightarrow \infty$. Similar to \eqref{eq:BL-bt-3}, the induction hypothesis for the function $\psi = \outcomeg{r}{}\big(\outcomeD{}{j}{r},\Vtreatment{j}{},\Vcovar{j}\big) \outcomeg{s}{}\big(\outcomeD{}{j}{s},\Vtreatment{j}{},\Vcovar{j}\big)$ implies that almost surely
        \begin{align}
            \label{eq:BL-bt-4}
            \quad\quad
            \lim_{N\rightarrow \infty}
            \frac{1}{N}
            \sum_{j=1}^N
            \left(\Uoutcome{\parallel,j}{}{t}\right)^2
            =
            \lim_{N\rightarrow \infty}
            \sum_{r,s=0}^{t-1} \APC{r}{t} \APC{s}{t} \pdot{\VUoutcome{}{}{r}}{\VUoutcome{}{}{s}}
            < \infty.
        \end{align}
        Consequently, by \eqref{eq:BL-bt-2}-\eqref{eq:BL-bt-4}, we get the result in \eqref{eq:BL-bt-1} and the proof is complete. \ep

        An immediate corollary of the result of Lemma~\ref{lm:finite variance} is that $\hat{\rho}_{tN}$, in~\eqref{eq:BL-bt-Yt stat}, is almost surely bounded independent of $N$. Then, for $l\geq 1$, it is straightforward to show that,
        \begin{align}
            \label{eq:BL-proof-t-2}
            \E
            \left[
            \left|
                \big[
                \widetilde{\IM} \VUoutcome{\perp}{}{t} + \IMatT{t} \VUoutcome{}{}{t}
                \big]^i
                + 
                \left[\bm{R}_t 
                \VAPC{}{t}
                \right]^i
            \right|^{l}
            \right]
            \leq
            2^{l-1}
            \E
            \left[
            \left|
                \big[
                \widetilde{\IM} \VUoutcome{\perp}{}{t} + \IMatT{t} \VUoutcome{}{}{t}
                \big]^i
                \right|^{l}
                +
                \left|
                \left[\bm{R}_t 
                \VAPC{}{t}
                \right]^i
            \right|^{l}
            \right]
            \leq c,
        \end{align}
        where $c$ is a constant independent of $N$ and we used the inequality $(v_1+v_2)^l \leq 2^{l-1} (v_1^l+v_2^l),\; v_1,v_2 \geq 0$. Note that in \eqref{eq:BL-proof-t-2}, given $\Gc_t$, the term $\bm{R}_t  \VAPC{}{t}$ is deterministic and bounded by Corollary~\ref{cl:alpha is bounded}. Then, fixing $0 < \kappa < 1$ and using the fact that $\psi \in \poly{k}$ and so $|\psi(\Vec{\omega})|\leq c (1 + \norm{\Vec{\omega}}^k)$, by Assumption~\ref{asmp:weak_limits}, we get,
        \begin{equation}
            \label{eq:BL-proof-t-1}
            \begin{aligned}
                \frac{1}{\cardinality{\batch}}
                \sum_{i \in \batch}
                \E\left[\left| \widetilde{\Psi}^{i}(N)\right|^{2+\kappa} \right]
                \leq c \cardinality{\batch}^{\kappa/2}.
            \end{aligned}
        \end{equation}
        Therefore, we can apply the SLLN for triangular arrays in Theorem~\ref{thm:SLLN} to obtain the following result:
        \begin{equation}
        \label{eq:BL-proof-t-3}
        \begin{aligned}
            \lim_{N \rightarrow \infty}
            \frac{1}{\cardinality{\batch}}
            \sum_{i\in \batch}
            &\psi\big(
                \outcomeD{}{i}{0},
                \outcomeD{}{i}{1}, \ldots,
                \outcomeD{}{i}{t},
                \outcome{}{i}{t+1},
                \Vtreatment{i}{},
                \Vcovar{i},
                \MAVO{}{i}{(t+1)N}
            \big)
            \\
            \eqas
            \;\lim_{N \rightarrow \infty}
            \frac{1}{\cardinality{\batch}}
            \sum_{i\in \batch}
            \E_{\IM,\IMatT{t}}
            \Big[
            \psi&\Big(
                \outcomeD{}{i}{0},
                \outcomeD{}{i}{1}, \ldots,
                \outcomeD{}{i}{t},
                \left[
                \widetilde{\IM}
                \VUoutcome{\perp}{}{t}
                + \bm{R}_t 
                \VAPC{}{t}
                +
                \IMatT{t} \VUoutcome{}{}{t}
                \right]^i,
                \Vtreatment{i}{}, \Vcovar{i},
                \HAVO{}{i}{(t+1)N}
                \Big)
            \Big].
        \end{aligned}
        \end{equation}
        Now, considering \eqref{eq:BL-at-stat-2_limit} and \eqref{eq:BL-bt-(-1)}, we can write
        \begin{equation}
        \label{eq:BL-proof-t-t}
        \begin{aligned}
            &\lim_{N \rightarrow \infty}
            \frac{1}{\cardinality{\batch}}
            \sum_{i\in \batch}
            \E_{\IM,\IMatT{t}}
            \Big[
            \psi\Big(
                \outcomeD{}{i}{0},
                \outcomeD{}{i}{1}, \ldots,
                \outcomeD{}{i}{t},
                \left[
                \widetilde{\IM}
                \VUoutcome{\perp}{}{t}
                + \bm{R}_t 
                \VAPC{}{t}
                +
                \IMatT{t} \VUoutcome{}{}{t}
                \right]^i,
                \Vtreatment{i}{}, \Vcovar{i},
                \MAVO{}{i}{(t+1)N}
                \Big)
            \Big]
            \\
            \eqas
            &\lim_{N \rightarrow \infty}
            \frac{1}{\cardinality{\batch}}
            \sum_{i \in \batch}
            \E_{Z}
            \Big[
            \psi\Big(
                \outcomeD{}{i}{0},
                \outcomeD{}{i}{1}, \ldots,
                \outcomeD{}{i}{t},
                \hat{\rho}_t Z
                +
                \left[
                \bm{R}_t 
                \VAPC{}{t}
                \right]^i,
                \Vtreatment{i}{}, \Vcovar{i},
                \MAVO{}{i}{t+1}
                \Big)
            \Big],
        \end{aligned}
        \end{equation}
        where similar to \eqref{eq:BL-a0-limit to the function} in the first step, we utilized the dominated convergence theorem and the continuous mapping theorem to pass the limit through the expectation and the function, respectively.
        
        From Eq.~\eqref{eq:BL-at-stat-2_limit}, recall that $\MAVO{}{i}{t+1} = (\bar{\mu}^i+\bar{\mu}^i_t) \E \left[\outcomeg{t}{} \big(\MAVO{}{}{t} + \MVVO{}{}{t} Z + \Houtcome{}{}{t-1},\Vtreatment{}{},\Vcovar{}\big) \right]$; accordingly, we define
        \begin{equation*}
        \begin{aligned}
            \widehat{\psi}&
            \big(
            \outcomeD{}{i}{0},
            \outcomeD{}{i}{1},
            \ldots,
            \outcomeD{}{i}{t},
            \outcome{}{i}{1} - \IMatTv{i \cdot}{0}\VUoutcome{}{}{0}
            , \ldots,
            \outcome{}{i}{t} - \IMatTv{i\cdot}{t-1}\VUoutcome{}{}{t-1},
            \Vtreatment{i}{},
            \Vcovar{i},
            \bar{\mu}^i, \bar{\mu}^i_t
            \big)
            \\
            :=
            \E_{Z}
            \Big[
            \psi&\Big(
                \outcomeD{}{i}{0},
                \outcomeD{}{i}{1},
                \ldots,
                \outcomeD{}{i}{t},
                \hat{\rho}_{t} Z
                +
                \sum_{s=0}^{t-1}
                \APC{t}{s}
                \big(
                \outcome{}{i}{s+1}
                - \IMatTv{i \cdot}{s}\VUoutcome{}{}{s}
                \big),
                \Vtreatment{i}{},
                \Vcovar{i},
                \MAVO{}{i}{t+1}
                \Big)
            \Big].
        \end{aligned}
        \end{equation*}
        Considering \eqref{eq:BL-proof-IH-a1} as well as \eqref{eq:BL-proof-t-3}-\eqref{eq:BL-proof-t-t}, for the function $\widehat{\psi}$, we have
        \begin{equation}
        \label{eq:BL-proof-t-new function}
        \begin{aligned}
            &\lim_{N \rightarrow \infty}
            \frac{1}{\cardinality{\batch}}
            \sum_{i\in \batch}
            \psi\big(
                \outcomeD{}{i}{0},
                \outcomeD{}{i}{1}, \ldots,
                \outcomeD{}{i}{t},
                \outcome{}{i}{t+1},
                \Vtreatment{i}{},
                \Vcovar{i},
                \MAVO{}{i}{(t+1)N}
            \big)
            \\
            \eqas
            &\;
            \E
            \bigg[
            \widehat{\psi}\bigg(
                \outcomeD{}{\batch}{0},
                \MAVO{}{\batch}{1}
                +
                \MVVO{}{}{1} Z_1
                +
                \Houtcome{\batch}{}{0},
                \ldots,
                \MAVO{}{\batch}{t}
                +
                \MVVO{}{}{t} Z_t
                +
                \Houtcome{\batch}{}{t-1},
                \BVVO{}{}{1}
                Z'_{1},
                \ldots,
                \BVVO{}{}{t}
                Z'_{t},
                \Vtreatment{\batch}{}, \Vcovar{\batch},
                \MIM{}{\batch}, \MIM{t}{\batch}
                \bigg)
            \bigg]
            \\
            \eqas
            &\;
            \E
            \bigg[
            \psi\bigg(
                \outcomeD{}{\batch}{0},
                \MAVO{}{\batch}{1}
                +
                \MVVO{}{}{1} Z_1
                +
                \Houtcome{\batch}{}{0},
                \ldots,
                \MAVO{}{\batch}{t}
                +
                \MVVO{}{}{t} Z_t
                +
                \Houtcome{\batch}{}{t-1},
                \hat{\rho}_{t} Z
                +
                \sum_{s=0}^{t-1}
                \APC{t}{s}
                \BVVO{}{}{s+1}
                Z'_{s+1},
                \Vtreatment{\batch}{}, \Vcovar{\batch},
                \MAVO{}{\batch}{t+1}
                \bigg)
            \bigg],
        \end{aligned}
        \end{equation}
        where $Z$ is a standard Gaussian random variable, independent of all other variables, as the inherent randomness arises from $\widetilde{\IM}$ and $\IMatT{t}$. Additionally, $\widehat{\psi}$ belongs to $\poly{k}$, since $\MAVO{}{i}{t+1}$ in the calculations of \eqref{eq:BL-proof-t-new function} can be viewed as a linear function depending solely on $\bar{\mu}^i$ and $\bar{\mu}^i_t$.
        
        Now, we need to show that
        \begin{equation}
            \label{eq:BL-proof-bt-dynamics-1}
            \begin{aligned}
                \Var
                \left[
                \hat{\rho}_{t} Z
                +
                \sum_{s=0}^{t-1}
                \APC{t}{s}
                \big(
                \BVVO{}{}{s+1}
                Z'_{s+1}
                \big)
                \right]
                &=
                (\MVVO{}{}{t+1})^2.
            \end{aligned}
        \end{equation}
        Considering that $(Z'_1,\ldots,Z'_t)$ follows a joint Normal distribution independent of $Z$, the random variable $\hat{\rho}_{t} Z + \sum_{s=0}^{t-1} \APC{t}{s} \BVVO{}{}{s+1} Z'_{s+1}$ is Gaussian. To obtain \eqref{eq:BL-proof-bt-dynamics-1}, we let $\psi = (\outcome{}{i}{t+1})^2$ in \eqref{eq:BL-proof-t-new function}. This yields
        \begin{equation}
            \label{eq:BL-proof-bt-dynamics-2}
            \begin{aligned}
                \lim_{N \rightarrow \infty}
                \frac{1}{N}
                \sum_{i=1}^N
                \big(
                \outcome{}{i}{t+1}\big)^2
                &=
                \E
                \left[
                \left(
                \hat{\rho}_{t} Z
                +
                \sum_{s=0}^{t-1}
                \APC{t}{s}
                \BVVO{}{}{s+1}
                Z'_{s+1}
                \right)^2
                \right].
            \end{aligned}
        \end{equation}
        Meanwhile, by \eqref{eq:BL-proof-bt-2-result}, we have
        \begin{align}
        \label{eq:BL-proof-bt-dynamics-3}
            \lim_{N \rightarrow \infty}
            \frac{1}{N}
            \sum_{i=1}^N
            (\outcome{}{i}{t+1})^2
            \eqas
            (\MVVO{}{}{t+1})^2.
        \end{align}
        Combining \eqref{eq:BL-proof-bt-dynamics-2} and \eqref{eq:BL-proof-bt-dynamics-3}, we derive the desired result as stated in \eqref{eq:BL-proof-bt-dynamics-1}.

        Finally, similar to \eqref{eq:BL-a0-function_re_def} and based on outcome representation in \eqref{eq:apndx_outcome_function}, we define the function $\widetilde{\psi}$ such that
        \begin{align*}
            \psi\big(
                \outcomeD{}{i}{0},
                \outcomeD{}{i}{1}, \ldots,
                \outcomeD{}{i}{t},
                \outcomeD{}{i}{t+1},
                \Vtreatment{i}{},
                \Vcovar{i}
            \big)
            &=
            \psi\big(
                \outcomeD{}{i}{0},
                \outcomeD{}{i}{1}, \ldots,
                \outcomeD{}{i}{t},
                \outcome{}{i}{t+1} + \MAVO{}{i}{(t+1)N} + \outcomeh{t}{} \big(\outcomeD{}{i}{t}, \Vtreatment{i}{}, \Vcovar{i}\big),
                \Vtreatment{i}{},
                \Vcovar{i}
            \big)
            \\
            &=
            \widetilde\psi \big(
                \outcomeD{}{i}{0},
                \outcomeD{}{i}{1}, \ldots,
                \outcomeD{}{i}{t},
                \outcome{}{i}{t+1},
                \Vtreatment{i}{},
                \Vcovar{i}, \MAVO{}{i}{(t+1)N}
            \big).
        \end{align*}
        Whenever $\psi \in \poly{k}$, by Assumption~\ref{asmp:BL}, we get that $\widetilde\psi \in \poly{k}$, and applying the result in Eq.~\eqref{eq:BL-proof-t-new function} for the function $\widetilde\psi$ yields the desired result. \ep
        
    \end{enumerate}

\section{Estimation of Counterfactual Evolutions}
\label{sec:estimation_theory}
In this section, we present a general framework for counterfactual estimation based on the outcome specification in Eq.~\eqref{eq:outcome_function_matrix} and the result of Theorem~\ref{thm:Batch_SE}. Specifically, given a desired treatment allocation matrix $\OMtreatment{u}{}$ (corresponding to an \emph{unobserved} counterfactual), and observing $\Moutcome{}{}{}(\Mtreatment{}{}=\OMtreatment{o}{})$, $\OMtreatment{o}{}$, and $\covar$, we aim to estimate the counterfactual evolution denoted by $\CFE{\OMtreatment{u}{}}{}{t}$ as defined in \eqref{eq:sample_mean_outcomes}.

To state the main theoretical result of this section, we parameterize the unknown functions in the state evolution equations. Specifically, let $\outcomeg{t}{}\big(\cdot; \param{\outcomeg{t}{}}\big)$ and $\outcomeh{t}{}\big(\cdot; \param{\outcomeh{t}{}}\big)$ represent the parameterized forms of $\outcomeg{t}{}(\cdot)$ and $\outcomeh{t}{}(\cdot)$ for $t = 0, \ldots, T-1$, respectively. Here, $\Eparam{\outcomeg{t}{}}$ and $\Eparam{\outcomeh{t}{}}$ are vectors of appropriate dimensions, denoting the unknown parameters of the respective functions. 
\begin{assumption}
    \label{asmp:independent_interference_mean}
    Considering \eqref{eq:state evolution}, for all $t$, we assume there exists a modification of the function $\outcomeg{t}{}$, which, with a slight abuse of notation, is also denoted by $\outcomeg{t}{}$ such that:
    \begin{align}
    \label{eq:independent_interference_mean}
        \MAVO{}{}{t+1} =
        \E\left[
        \outcomeg{t}{}\big(\MAVO{}{}{t} + \MVVO{}{}{t} Z_t + \Houtcome{}{}{t-1}, \Vtreatment{}{}, \Vcovar{}\big)
        \right].
    \end{align}
\end{assumption}
An example where Assumption~\ref{asmp:independent_interference_mean} holds is when the random variable $\MIM{}{} + \MIM{t}{}$ is independent of all other sources of randomness. In this case, we can normalize the mean to 1 by adjusting the function $\outcomeg{t}{}$, scaling it by a constant factor equal to $\E[\MIM{}{} + \MIM{t}{}]$.

For further simplicity in notation, we also define $\sigma_t := \sqrt{\sigma^2 + \sigma_t^2}$. Then, we can collect the unknown parameters in the state evolution equations in \eqref{eq:state evolution} as follows:
\begin{align}
    \label{eq:apndx_unknowns}
    \Uc := \left(\{\sigma_t\}, \{\param{\outcomeg{t}{}} \}, \{\param{\outcomeh{t}{}}\}\right).
\end{align}
We denote an estimation of $\Uc$ as $\widehat{\Uc} := \left(\left\{\hat{\sigma}_t\right\}, \{\Eparam{\outcomeg{t}{}}\}, \{\Eparam{\outcomeh{t}{}}\}\right)$. We can then employ Algorithm~\ref{alg:CF estimation} to compute the desired counterfactual.
\begin{algorithm}
\caption{General counterfactual estimation}
\label{alg:CF estimation}
\begin{algorithmic}
\Require $\Moutcome{}{}{}(\Mtreatment{}{}=\OMtreatment{o}{}), \OMtreatment{o}{}, \covar$, and $\OMtreatment{u}{}$

\State \hspace{-1.3em} \textbf{Step 1: Parameters Estimation}

\State Estimate the set of unknown parameters $\Uc$ by $\widehat{\Uc}$.

\State \hspace{-1.3em} \textbf{Step 2: Counterfactual Estimation}

\State $\EAVO{}{}{1} \gets \frac{1}{N} \sum_{i=1}^N \outcomeg{0}{}\big(\outcomeD{}{i}{0}, \OVtreatment{i}{u}, \Vcovar{i}; \Eparam{\outcomeg{0}{}} \big)$ 

\State $\EHoutcome{i}{}{0} \gets \outcomeh{0}{}\big(\outcomeD{}{i}{0}, \OVtreatment{i}{u}, \Vcovar{i}; \Eparam{\outcomeh{0}{}}\big),\; i = 1, \ldots, N$ 

\State $\ECF{}{1}{\OMtreatment{u}{}} \gets \EAVO{}{}{1} + \frac{1}{N} \sum_{i=1}^N \EHoutcome{i}{}{0}$

\For{$t = 1, \ldots, T-1$}

\State $\EHoutcome{i}{}{t} \gets \outcomeh{t}{}\big(\EAVO{}{}{t} +
\EHoutcome{i}{}{t-1}, \OVtreatment{i}{u}, \Vcovar{i}; \Eparam{\outcomeh{t}{}}\big),\; i = 1, \ldots, N$

\State $\EAVO{}{}{t+1} \gets \frac{1}{N} \sum_{i=1}^N \outcomeg{t}{}\big(\EAVO{}{}{t} + 
\EHoutcome{i}{}{t-1}, \OVtreatment{i}{u}, \Vcovar{i}; \Eparam{\outcomeg{t}{}}\big)$

\State $\ECF{}{t+1}{\OMtreatment{u}{}} \gets \EAVO{}{}{t+1} +  \frac{1}{N} \sum_{i=1}^N \EHoutcome{i}{}{t}$

\EndFor

\Ensure $\ECF{}{t}{\OMtreatment{u}{}}$, $t = 1, \ldots, T$.
\end{algorithmic}
\end{algorithm}
\begin{assumption}[Continuous parameterization]
    \label{asmp:continuous_parameterization}
    For all $t$, the mappings $\param{\outcomeg{t}{}} \mapsto \outcomeg{t}{} \big(\cdot; \param{\outcomeg{t}{}} \big)$ and $\param{\outcomeh{t}{}} \mapsto \outcomeh{t}{} \big(\cdot; \param{\outcomeh{t}{}} \big)$ are continuous functions.
\end{assumption}
\begin{assumption}[Affine outcome functions]
    \label{asmp:affine_functions}
    For all $t$, the mappings $y \mapsto \outcomeg{t}{}(y, \cdot)$ and $y \mapsto \outcomeh{t}{}(y, \cdot)$ are affine; that means there exist functions $\outcomeg{t}{1}, \outcomeg{t}{2}, \outcomeh{t}{1}$ and $\outcomeh{t}{2}$ such that $\outcomeg{t}{}(y, \cdot) = y \outcomeg{t}{1}(\cdot) + \outcomeg{t}{2}(\cdot)$ as well as $\outcomeh{t}{}(y, \cdot) = y \outcomeh{t}{1}(\cdot) + \outcomeh{t}{2}(\cdot)$.
\end{assumption}
\begin{assumption}[Consistent parameters estimation]
    \label{asmp:consistent_parameter_estimation}
    $\widehat{\Uc}$ is a consistent estimator of $\Uc$; that is, $\widehat{\Uc} \xrightarrow{P} \Uc$ as $N \rightarrow \infty$.
\end{assumption}
\begin{assumption}[All control initialization]
    \label{asmp:all_control_initialization}
    There is no treatment at time $t=0$; that means $\treatment{i}{0}=0$, for all $i\in[N]$, and no treatment is anticipated.
\end{assumption}

In the following, we demonstrate the consistency of the results from Algorithm~\ref{alg:CF estimation} under the above assumptions.
\begin{theorem}[Consistency]
    \label{thm:consistency}
    Let the conditions of Theorem~\ref{thm:Batch_SE} and Assumptions~\ref{asmp:independent_interference_mean}--\ref{asmp:all_control_initialization} hold. In particular, suppose Assumption~\ref{asmp:weak_limits} holds for both the observed treatment allocation $\OMtreatment{o}{}$ and the target (unobserved) treatment allocation $\OMtreatment{u}{}$. Then, as $N \rightarrow \infty$, the estimator $\ECF{}{t}{\OMtreatment{u}{}}$ specified in Algorithm~\ref{alg:CF estimation} satisfies:
    \begin{align}
        \label{eq:consistency}
        \ECF{}{t}{\OMtreatment{u}{}} - \CFE{\OMtreatment{u}{}}{}{t} \xrightarrow{P} 0,
    \end{align}
    for any time $t$.
\end{theorem}

\begin{remark}
    If the consistency in Assumption~\ref{asmp:consistent_parameter_estimation} holds in the strong sense, i.e., $\widehat{\Uc} \xrightarrow{a.s.} \Uc$ as $N \rightarrow \infty$, then the consistency in Theorem~\ref{thm:consistency} is also strong. That is, the convergence in~\eqref{eq:consistency} holds almost surely.
\end{remark}
\begin{remark}
    We can generalize Algorithm~\ref{alg:CF estimation} in several ways. First, when considering outcome specifications with $l \geq 1$ lag terms, we can modify the algorithm accordingly, and extend Assumption~\ref{asmp:all_control_initialization} to require $l$ historical observations: $\treatment{i}{t}=0$ for all $t \leq l$ and $i\in[N]$. We can also relax Assumption~\ref{asmp:all_control_initialization} by beginning from any arbitrary state, provided we enforce the initial conditions to the desired counterfactual scenario--- specifically, requiring that $\OMtreatment{o}{}$ and $\OMtreatment{u}{}$ match for the first $l$ time periods.
\end{remark}
\noindent
\textbf{Proof.}
We use an induction argument on $t \geq 1$ to prove the following statement. As $N \rightarrow \infty$, we show that
\begin{equation}
    \label{eq:detailed_consistency}
    \begin{aligned}
        \EAVO{}{}{t} \xrightarrow{P} \MAVO{}{}{t},
        \quad\quad\quad
        \frac{1}{N} \sum_{i=1}^N \EHoutcome{i}{}{t-1}\xrightarrow{P} \E\left[ \Houtcome{}{}{t-1} \right],
    \end{aligned}
\end{equation}
This implies that $\ECF{}{t}{\OMtreatment{u}{}}$ also converges to $\AVO{}{}{t} = \MAVO{}{}{t} + \E\left[\Houtcome{}{}{t-1}\right]$ in probability, whenever $N \rightarrow \infty$. Finally, applying Eq.~\eqref{eq:BSE-limit}, we can write:
\begin{align*}
    \lim_{N \rightarrow \infty} \ECF{}{t}{\OMtreatment{u}{}}
    \eqp
    \AVO{}{}{t}
    \eqas
    \lim_{N \rightarrow \infty} \frac{1}{N} \sum_{i=1}^\UN \outcomeDW{\OMtreatment{u}{}}{i}{t}
    =
    \lim_{N \rightarrow \infty} \CFE{\OMtreatment{u}{}}{}{t},
\end{align*}
which concludes the proof. Above, the notation $\eqp$ means that the convergence holds in probability. Next, we establish the convergence results given in \eqref{eq:detailed_consistency}.

\textbf{Step 1.} Let $t = 1$. We begin by proving the first result in \eqref{eq:detailed_consistency}. To this end, we occasionally add the notation $(N)$ to the quantities associated with the system containing $N$ experimental units to avoid ambiguity. We have
\begin{equation}
    \label{eq:consistency_proof_1}
    \begin{aligned}
        \EAVO{N}{}{1}
        &=
        \frac{1}{N} \sum_{i=1}^N \outcomeg{0}{}\big(\outcomeD{}{i}{0}, \OVtreatment{i}{u}, \Vcovar{i}; \Eparam{\outcomeg{0}{}}(N) \big)
    \end{aligned}
\end{equation}
Note that, by Assumption~\ref{asmp:continuous_parameterization}, the right-hand side of \eqref{eq:consistency_proof_1} can be seen as a continuous function of $\Eparam{\outcomeg{0}{}}(N)$. Therefore, applying the continuous mapping theorem, e.g., Theorem 2.3 in \cite{van2000asymptotic}, implies that
\begin{equation}
    \label{eq:consistency_proof_2}
    \begin{aligned}
        \lim_{N \rightarrow \infty} \EAVO{N}{}{1}
        =
        \lim_{N \rightarrow \infty} \frac{1}{N} \sum_{i=1}^N \outcomeg{0}{}\big(\outcomeD{}{i}{0}, \OVtreatment{i}{u}, \Vcovar{i}; \Eparam{\outcomeg{0}{}}(N) \big)
        \eqp
        \lim_{N \rightarrow \infty} \frac{1}{N} \sum_{i=1}^N \outcomeg{0}{}\big(\outcomeD{}{i}{0}, \OVtreatment{i}{u}, \Vcovar{i}; \param{\outcomeg{0}{}} \big)
        \eqas
        \MAVO{}{}{1}.
    \end{aligned}
\end{equation}
In the last equality, we used the result of Theorem~\ref{thm:Batch_SE}. The second result also follows immediately.

Fixing an arbitrary function $\psi \in \poly{\frac{k}{2}}$ which is also affine in its first argument, we also need the following intermediary result:
\begin{align}
    \label{eq:consistency_proof_intermediary}
    \lim_{N \rightarrow \infty} \frac{1}{N} \sum_{i=1}^N \psi\big( \EHoutcome{i}{N}{0}, \OVtreatment{i}{u}, \Vcovar{i}\big)
    \eqp
    \E \left[
    \psi\big(\Houtcome{}{}{0}, \Vtreatment{}{u}, \Vcovar{}\big)
    \right],
\end{align}
where $\Vtreatment{}{u}$ represents the weak limit of $\OVtreatment{i}{u}$'s. To obtain this result, let $\widetilde{\psi}$ be the function such that
\begin{align*}
    \lim_{N \rightarrow \infty} \frac{1}{N} \sum_{i=1}^N \psi\big(\EHoutcome{i}{N}{0}, \OVtreatment{i}{u}, \Vcovar{i}\big)
    &=
    \lim_{N \rightarrow \infty} \frac{1}{N} \sum_{i=1}^N \psi\big( \outcomeh{0}{}\big(\outcomeD{}{i}{0}, \OVtreatment{i}{u}, \Vcovar{i}; \Eparam{\outcomeh{0}{}}(N)\big), \OVtreatment{i}{u}, \Vcovar{i}\big)
    \\
    &=
    \lim_{N \rightarrow \infty} \frac{1}{N} \sum_{i=1}^N \widetilde\psi\big(\outcomeD{}{i}{0}, \OVtreatment{i}{u}, \Vcovar{i}, \Eparam{\outcomeh{0}{}}(N)\big)
    \\
    &\eqp
    \lim_{N \rightarrow \infty} \frac{1}{N} \sum_{i=1}^N \widetilde\psi\big(\outcomeD{}{i}{0}, \OVtreatment{i}{u}, \Vcovar{i}, \param{\outcomeh{0}{}}\big)
    \\
    &\eqas
    \E \left[ \widetilde\psi\big(\outcomeD{}{}{0}, \Vtreatment{}{u}, \Vcovar{}, \param{\outcomeh{0}{}}\big)
    \right].
\end{align*}
Above, we used the continuous mapping theorem and Theorem~\ref{thm:SLLN-2} in view of Assumption~\ref{asmp:weak_limits}. Note that $\widetilde{\psi} \in \poly{\frac{k}{2}}$ because of Assumption~\ref{asmp:BL}-\ref{asmp:BL-pl h-functions}.

\textbf{Induction Hypothesis (IH).} Suppose that the limits in \eqref{eq:detailed_consistency} hold true for $t$ and also
\begin{align}
    \label{eq:consistency_proof_intermediaryP_IH}
    \lim_{N \rightarrow \infty} \frac{1}{N} \sum_{i=1}^N \psi\big(\EHoutcome{i}{N}{t-1}, \OVtreatment{i}{u}, \Vcovar{i}\big)
    \eqp
    \E \left[
    \psi\big(\Houtcome{}{}{t-1}, \Vtreatment{}{u}, \Vcovar{}\big)
    \right].
\end{align}

\textbf{Step 2.} We show that $\EAVO{N}{}{t+1} \xrightarrow{P} \MAVO{}{}{t+1}$. By reusing the continuous mapping theorem, the induction hypothesis, and Assumption~\ref{asmp:affine_functions}, we have
\begin{equation}
    \label{eq:consistency_proof_4}
    \begin{aligned}
        \lim_{N \rightarrow \infty} \EAVO{N}{}{t+1} 
        &=
        \lim_{N \rightarrow \infty} \frac{1}{N} \sum_{i=1}^N \outcomeg{t}{}\big(\EAVO{N}{}{t} + 
        \EHoutcome{i}{N}{t-1}, \OVtreatment{i}{u}, \Vcovar{i}; \Eparam{\outcomeg{t}{}}(N)\big)
        \\
        &\eqp
        \lim_{N \rightarrow \infty} \frac{1}{N} \sum_{i=1}^N \outcomeg{t}{}\big(\MAVO{}{}{t} + 
        \EHoutcome{i}{N}{t-1}, \OVtreatment{i}{u}, \Vcovar{i}; \param{\outcomeg{t}{}}\big)
        \\
        &\eqp
        \E \left[
        (\MAVO{}{}{t} + \Houtcome{}{}{t-1}) \outcomeg{t}{1}\big(\Vtreatment{}{u}, \Vcovar{}\big)
        +
        \outcomeg{t}{2}\big(\Vtreatment{}{u}, \Vcovar{}\big)
        \right]
        \\
        &\eqp
        \E \left[
        (\MAVO{}{}{t} + \MVVO{}{}{t}Z_t + \Houtcome{}{}{t-1}) \outcomeg{t}{1}\big(\Vtreatment{}{u}, \Vcovar{}\big)
        +
        \outcomeg{t}{2}\big(\Vtreatment{}{u}, \Vcovar{}\big)
        \right]
        \\
        &\eqp
        \E \left[
        \outcomeg{t}{}\big(\MAVO{}{}{t} + \MVVO{}{}{t}Z_t + \Houtcome{}{}{t-1}, \Vtreatment{}{u}, \Vcovar{}\big)
        \right]
        =
        \MAVO{}{}{t+1},
    \end{aligned}
\end{equation}
where $Z_t$ is as given in Theorem~\ref{thm:Batch_SE} and in the last line we used the fact that $Z_t$ is independent of $\Vtreatment{}{u}$ and $\Vcovar{}$.
Similarly, one can establish $\frac{1}{N} \sum_{i=1}^N \EHoutcome{i}{}{t}\xrightarrow{P} \E\left[ \Houtcome{}{}{t} \right]$. We therefore conclude the proof by demonstrating the following result:
\begin{align}
    \label{eq:consistency_proof_intermediaryP_step2}
    \lim_{N \rightarrow \infty} \frac{1}{N} \sum_{i=1}^N \psi\big( \EHoutcome{i}{N}{t}, \OVtreatment{i}{u}, \Vcovar{i}\big)
    \eqp
    \E \left[
    \psi\big(\Houtcome{}{}{t}, \Vtreatment{}{u}, \Vcovar{}\big)
    \right].
\end{align}
For this purpose, considering $\EHoutcome{i}{N}{t} = \outcomeh{t}{}\big(\EAVO{N}{}{t} + \EHoutcome{i}{N}{t-1}, \OVtreatment{i}{u}, \Vcovar{i}; \Eparam{\outcomeh{t}{}}(N)\big)$ and the fact that $\psi$ and $\outcomeh{t}{}$ are affine, for any $i$, we have: 
\begin{equation}
\label{eq:consistency_proof_intermediaryP_step3}
\begin{aligned}
    &\quad\;
    \psi\big( \EHoutcome{i}{N}{t}, \OVtreatment{i}{u}, \Vcovar{i}\big)
    \\
    &=
    \EHoutcome{i}{N}{t} \psi^1\big(\OVtreatment{i}{u}, \Vcovar{i}\big)
    +
    \psi^2\big(\OVtreatment{i}{u}, \Vcovar{i}\big)
    \\
    &=
    \outcomeh{t}{}\Big(\EAVO{N}{}{t} + \EHoutcome{i}{N}{t-1}, \OVtreatment{i}{u}, \Vcovar{i}; \Eparam{\outcomeh{t}{}}(N)\Big) \psi^1\big(\OVtreatment{i}{u}, \Vcovar{i}\big)
    +
    \psi^2\big(\OVtreatment{i}{u}, \Vcovar{i}\big)
    \\
    &=
    \left(
    \Big( \EAVO{N}{}{t} + \EHoutcome{i}{N}{t-1} \Big)
    \outcomeh{t}{1}\Big(\OVtreatment{i}{u}, \Vcovar{i}; \Eparam{\outcomeh{t}{}}(N)\Big)
    +
    \outcomeh{t}{2}\Big(\OVtreatment{i}{u}, \Vcovar{i}; \Eparam{\outcomeh{t}{}}(N)\Big)
    \right)
    \psi^1\big(\OVtreatment{i}{u}, \Vcovar{i}\big)
    +
    \psi^2\big(\OVtreatment{i}{u}, \Vcovar{i}\big),
\end{aligned}
\end{equation}
where $\psi^1$ and $\psi^2$ are the affine decomposition of the $\psi$.
Then, applying the continuous mapping theorem and the induction hypothesis yields the following result:
\begin{align*}
    &\;\lim_{N \rightarrow \infty} \frac{1}{N} \sum_{i=1}^N
    \left(
    \Big( \EAVO{N}{}{t} + \EHoutcome{i}{N}{t-1} \Big)
    \outcomeh{t}{1}\Big(\OVtreatment{i}{u}, \Vcovar{i}; \Eparam{\outcomeh{t}{}}(N)\Big)
    \psi^1\big(\OVtreatment{i}{u}, \Vcovar{i}\big)
    \right)
    \\
    \eqp
    &\;\lim_{N \rightarrow \infty} \frac{1}{N} \sum_{i=1}^N
    \left(
    \Big( \AVO{}{}{t} + \EHoutcome{i}{N}{t-1} \Big)
    \outcomeh{t}{1}\Big(\OVtreatment{i}{u}, \Vcovar{i}; \param{\outcomeh{t}{}}\Big)
    \psi^1\big(\OVtreatment{i}{u}, \Vcovar{i}\big)
    \right)
    \\
    \eqp
    &\;\E
    \left[
    \Big( \AVO{}{}{t} + \Houtcome{}{}{t-1} \Big)
    \outcomeh{t}{1}\Big(\Vtreatment{}{u}, \Vcovar{}; \param{\outcomeh{t}{}}\Big)
    \psi^1\big(\Vtreatment{}{u}, \Vcovar{}\big)
    \right].
\end{align*}
Using Theorem~\ref{thm:Batch_SE} once again, we get:
\begin{align*}
    &\;\lim_{N \rightarrow \infty} \frac{1}{N} \sum_{i=1}^N
    \left(
    \Big( \EAVO{N}{}{t} + \EHoutcome{i}{N}{t-1} \Big)
    \outcomeh{t}{1}\Big(\OVtreatment{i}{u}, \Vcovar{i}; \Eparam{\outcomeh{t}{}}(N)\Big)
    \psi^1\big(\OVtreatment{i}{u}, \Vcovar{i}\big)
    \right)
    \\
    \eqp
    &\;\E
    \left[
    \Big( \AVO{}{}{t} + \VVO{}{}{t} Z_t + \Houtcome{}{}{t-1} \Big)
    \outcomeh{t}{1}\Big(\Vtreatment{}{u}, \Vcovar{}; \param{\outcomeh{t}{}}\Big)
    \psi^1\big(\Vtreatment{}{u}, \Vcovar{}\big)
    \right].
\end{align*}
Repeating a similar approach for other terms on the right-hand side of \eqref{eq:consistency_proof_intermediaryP_step3} and then reversing the steps in \eqref{eq:consistency_proof_intermediaryP_step3}, we get the desired result, and the proof is complete. \ep

\subsection{Application to Bernoulli Randomized Design}
\label{sec:application_to_BRD}
We showcase the applicability of our framework by considering a Bernoulli randomized design, where each unit $i$ at time $t$ receives treatment with a probability denoted by $\expr_t$. Therefore, $\treatment{i}{t} \sim Bernoulli(\expr_t)$, and the $\treatment{i}{t}$'s are independent across experimental units.

We consider a first-order yet non-linear approximation of functions $\outcomeg{t}{}$ and $\outcomeh{t}{}$ in the outcome specification in \eqref{eq:outcome_function_matrix}. Specifically, we let $\Vcovar{i} = (\BE_{g}^i, \CE_{gl}^i, \ldots, \CE_{g1}^i, \DE_g^i, \PE_g^i, \BE_{h}^i, \CE_{hl}^i, \ldots, \CE_{h1}^i, \DE_h^i, \PE_h^i)^\top$ and define
\begin{equation}
    \label{eq:function_structure}
    \begin{aligned}
        \outcomeg{t}{} \left(
        \outcomeD{}{i}{t-l+1}, \ldots, \outcomeD{}{i}{t},
        \Vtreatment{i}{}, \Vcovar{i}
        \right)
        &= \BE_{g}^i + \CE_{gl}^i \outcomeD{}{i}{t-l+1} + \ldots + \CE_{g1}^i \outcomeD{}{i}{t} + 
        \DE_g^i \treatment{i}{t+1} + \PE_g^i \outcomeD{}{i}{t} \treatment{i}{t+1}
        \\
        \outcomeh{t}{} \left(
        \outcomeD{}{i}{t-l+1}, \ldots, \outcomeD{}{i}{t},
        \Vtreatment{i}{}, \Vcovar{i}
        \right)
        &= \BE_{h}^i + \CE_{hl}^i \outcomeD{}{i}{t-l+1} + \ldots + \CE_{h1}^i \outcomeD{}{i}{t} + 
        \DE_h^i \treatment{i}{t+1} + \PE_h^i \outcomeD{}{i}{t} \treatment{i}{t+1}.
    \end{aligned}
\end{equation}
\begin{remark}
    In \eqref{eq:function_structure}, we allow the parameters of functions $\outcomeg{t}{}$ and $\outcomeh{t}{}$ to vary across units. These parameters can be viewed as unobserved unit-specific covariates that characterize how each unit responds to interventions. Additional unit-specific characteristics can be incorporated as observed covariates in the vectors $\Vcovar{i},\; i \in [N]$. We omit these details for brevity.
\end{remark}

We continue by considering a subpopulation of experimental units denoted by $\batch$. We assume that the sampling rule determining $\batch$ depends \emph{only} on the treatment allocations. Because the treatment allocation is independent of other variables, we can assume in the state evolution equations outlined in Eq.~\eqref{eq:state evolution} that $\MIM{\cdot}{\batch}$ and $\Vcovar{\batch}$ are equal to their global counterparts $\MIM{\cdot}{}$ and $\Vcovar{}$, respectively. This reflects the idea that sampling based on treatment allocation is equivalent to random sampling from the experimental population. Thus, by the state evolution equations, for $t = 0, 1, \ldots, l-1$, we can write,
\begin{equation}
    \label{eq:SE_BRD_1}
    \begin{aligned}
        \AVO{}{}{t}
        &=
        \lim_{N \rightarrow \infty}
        \frac{1}{N}
        \sum_{i=1}^N
        \outcomeD{}{i}{t},
        \quad
        \AVO{}{\batch}{t}
        =
        \lim_{N \rightarrow \infty}
        \frac{1}{\cardinality{\batch}}
        \sum_{i \in \batch}
        \outcomeD{}{i}{t},
    \end{aligned}
\end{equation}
and for $t \geq l-1$,
\begin{equation}
    \label{eq:SE_BRD_2}
    \begin{aligned}
        \MAVO{}{}{t+1}
        &=
        \ABE_{g} +
        \ACE_{gl} \AVO{}{}{t-l+1} + \ldots + \ACE_{g1} \AVO{}{}{t} +  
        \ADE_g \expr_{t+1} +
        \APE_g \expr_{t+1} \AVO{}{}{t}
        \\
        \AVO{}{\batch}{t+1}
        &=
        \MAVO{}{}{t+1} +
        \ABE_{h} +
        \ACE_{hl} \AVO{}{\batch}{t-l+1} + \ldots + \ACE_{h1} \AVO{}{\batch}{t} +  
        \ADE_h \expr_{t+1}^\batch +
        \APE_h \expr_{t+1}^\batch \AVO{}{\batch}{t},
    \end{aligned}
\end{equation}
where
\begin{equation}
\label{eq:apndx_BRD_parameters_mean}
\begin{aligned}
    &(\ABE_{g}, \ACE_{gl}, \ldots, \ACE_{g1}, \ADE_g, \APE_g, \ABE_{h}, \ACE_{hl}, \ldots, \ACE_{h1}, \ADE_h, \APE_h)^\top
    \\
    :=
    &\E \left[ 
    \Vcovar{} =
    (\BE_{g}, \CE_{gl}, \ldots, \CE_{g1}, \DE_g, \PE_g, \BE_{h}, \CE_{hl}, \ldots, \CE_{h1}, \DE_h, \PE_h)^\top
    \right].
\end{aligned}
\end{equation}
Here, $\Vcovar{}$ represents the weak limit of $\Vcovar{1}, \ldots, \Vcovar{N}$ when $N \rightarrow \infty$, as specified by Assumption~\ref{asmp:weak_limits}. Then, Equation~\eqref{eq:SE_BRD_2} follows from Assumption~\ref{asmp:independent_interference_mean} and the additional assumption about the elements of $\Vcovar{}$. For example, \eqref{eq:SE_BRD_2} holds when the elements of $\Vcovar{}$ are random variables independent of all other sources of randomness in the model.

To proceed with counterfactual estimation, we consider $b$ distinct subpopulations, denoted by $\batch_1, \ldots, \batch_b$, each determined solely by treatment allocations. With convention $\ABE := \ABE_{g} + \ABE_{h}$ in \eqref{eq:SE_BRD_2}, we can write the following linear regression model:
\begin{equation}
    \label{eq:apndx_BRD_model}
    \begin{aligned}
        \begin{bmatrix}
            \AVO{}{\batch_1}{l}
            \\
            \AVO{}{\batch_1}{l+1}
            \\
            \vdots
            \\
            \AVO{}{\batch_1}{T}
            \\
            \\
            \vdots
            \\
            \\
            \AVO{}{\batch_b}{l}
            \\
            \AVO{}{\batch_b}{l+1}
            \\
            \vdots
            \\
            \AVO{}{\batch_b}{T}
        \end{bmatrix}
        =
        \ABE
        \begin{bmatrix}
            1
            \\
            1
            \\
            \vdots
            \\
            1
            \\
            \\
            \vdots
            \\
            \\
            1
            \\
            1
            \\
            \vdots
            \\
            1
        \end{bmatrix}
        &+ \ACE_{gl}
        \begin{bmatrix}
            \AVO{}{}{0}
            \\
            \AVO{}{}{1}
            \\
            \vdots
            \\
            \AVO{}{}{T-l}
            \\
            \\
            \vdots
            \\
            \\
            \AVO{}{}{0}
            \\
            \AVO{}{}{1}
            \\
            \vdots
            \\
            \AVO{}{}{T-l}
        \end{bmatrix}
        +
        \ldots
        + \ACE_{g1}
        \begin{bmatrix}
            \AVO{}{}{l-1}
            \\
            \AVO{}{}{l}
            \\
            \vdots
            \\
            \AVO{}{}{T-1}
            \\
            \\
            \vdots
            \\
            \\
            \AVO{}{}{l-1}
            \\
            \AVO{}{}{l}
            \\
            \vdots
            \\
            \AVO{}{}{T-1}
        \end{bmatrix}
        + \ADE_g 
        \begin{bmatrix}
            \expr_{l}
            \\
            \expr_{l+1}
            \\
            \vdots
            \\
            \expr_{T}
            \\
            \\
            \vdots
            \\
            \\
            \expr_{l}
            \\
            \expr_{l+1}
            \\
            \vdots
            \\
            \expr_{T}
        \end{bmatrix}
        + \APE_g  
        \begin{bmatrix}
            \expr_{l}\AVO{}{}{l-1}
            \\
            \expr_{l+1}\AVO{}{}{l}
            \\
            \vdots
            \\
            \expr_{T}\AVO{}{}{T-1}
            \\
            \\
            \vdots
            \\
            \\
            \expr_{l}\AVO{}{}{l-1}
            \\
            \expr_{l+1}\AVO{}{}{l}
            \\
            \vdots
            \\
            \expr_{T}\AVO{}{}{T-1}
        \end{bmatrix}
        \\
        &+ \ACE_{hl}
        \begin{bmatrix}
            \AVO{}{\batch_1}{0}
            \\
            \AVO{}{\batch_1}{1}
            \\
            \vdots
            \\
            \AVO{}{\batch_1}{T-l}
            \\
            \\
            \vdots
            \\
            \\
            \AVO{}{\batch_b}{0}
            \\
            \AVO{}{\batch_b}{1}
            \\
            \vdots
            \\
            \AVO{}{\batch_b}{T-l}
        \end{bmatrix}
        +
        \ldots
        + \ACE_{h1}
        \begin{bmatrix}
            \AVO{}{\batch_1}{l-1}
            \\
            \AVO{}{\batch_1}{l}
            \\
            \vdots
            \\
            \AVO{}{\batch_1}{T-1}
            \\
            \\
            \vdots
            \\
            \\
            \AVO{}{\batch_b}{l-1}
            \\
            \AVO{}{\batch_b}{l}
            \\
            \vdots
            \\
            \AVO{}{\batch_b}{T-1}
        \end{bmatrix}
        + \ADE_h 
        \begin{bmatrix}
            \expr_{l}^{\batch_1}
            \\
            \expr_{l+1}^{\batch_1}
            \\
            \vdots
            \\
            \expr_{T}^{\batch_1}
            \\
            \\
            \vdots
            \\
            \\
            \expr_{l}^{\batch_b}
            \\
            \expr_{l+1}^{\batch_b}
            \\
            \vdots
            \\
            \expr_{T}^{\batch_b}
        \end{bmatrix}
        + \APE_h
        \begin{bmatrix}
            \expr_{l}^{\batch_1} \AVO{}{\batch_1}{l-1}
            \\
            \expr_{l+1}^{\batch_1} \AVO{}{\batch_1}{l}
            \\
            \vdots
            \\
            \expr_{T}^{\batch_1} \AVO{}{\batch_1}{T-1}
            \\
            \\
            \vdots
            \\
            \\
            \expr_{l}^{\batch_b} \AVO{}{\batch_b}{l-1}
            \\
            \expr_{l+1}^{\batch_b} \AVO{}{\batch_b}{l}
            \\
            \vdots
            \\
            \expr_{T}^{\batch_b} \AVO{}{\batch_b}{T-1}
        \end{bmatrix},
    \end{aligned}
\end{equation}
or equivalently in a matrix form
\begin{align*}
    \Vec{\Yc} = \Xc (\ABE, \ACE_{gl}, \ldots, \ACE_{g1}, \ADE_g, \APE_g, \ACE_{hl}, \ldots, \ACE_{h1}, \ADE_h, \APE_h)^\top,
\end{align*}
where $\Vec{\Yc}$ represents the vector on the left-hand side of \eqref{eq:apndx_BRD_model}, while $\Xc$ denotes the matrix formed by the columns corresponding to the vectors on the right-hand side of \eqref{eq:apndx_BRD_model}.

For the regression model outlined in \eqref{eq:apndx_BRD_model}, we now show that the least squares estimator provides a strongly consistent estimate for the unknown coefficients. To this end, upon observing $\OMtreatment{}{}$ and $\Moutcome{}{}{}(\OMtreatment{}{})$ within a system of $N$ experimental units, we define the following:
\begin{align}
    \label{eq:apndx_BRD_reg_model}
    \Vec{\Yc}(N) \sim \Xc(N) \Vec{\Bc}(N),
\end{align}
where $\Vec{\Yc}(N)$ denotes the sample mean of observed outcomes over time and across various subpopulations, corresponding to the vector on the left-hand side of \eqref{eq:apndx_BRD_model}. The vector $\Vec{\Bc}(N)$ represents the unknown coefficients on the right-hand side of \eqref{eq:apndx_BRD_model}, while $\Xc(N)$ denotes a $b(T-l+1)$ by $1+l+2+l+2$ matrix. This matrix is aligned with the vectors on the right-hand side of \eqref{eq:apndx_BRD_model}, but with the corresponding sample means of observed outcomes and treatments replacing the asymptotic terms, see \eqref{eq:reg_cover_matrix}.
\begin{proposition}
    \label{prp:BRD_consistency}
    Suppose that $\Xc(N)^\top \Xc(N)$ is invertible. Let $\Vec{\widehat{\Bc}}(N) := \big( \Xc(N)^\top \Xc(N) \big)^{-1} \Xc(N)^\top \Vec{\Yc}(N)$ be the least squares estimator. Then, $\Vec{\widehat{\Bc}}(N)$ provides a strongly consistent estimator for the coefficient vector $(\ABE, \ACE_{gl}, \ldots, \ACE_{g1}, \ADE_g, \APE_g, \ACE_{hl}, \ldots, \ACE_{h1}, \ADE_h, \APE_h)^\top$.
\end{proposition}
\textbf{Proof.}
Because the matrix $\Xc(N)^\top \Xc(N)$ is invertible, the estimator $\Vec{\widehat{\Bc}}(N)$ is a continuous function of the input data. Therefore, we can pass the limit through the estimator function as:
\begin{align*}
    \lim_{N \rightarrow \infty} \Vec{\widehat{\Bc}}(N)
    =
    \left( \Big(\lim_{N \rightarrow \infty} \Xc(N)\Big)^\top \Big(\lim_{N \rightarrow \infty} \Xc(N)\Big) \right)^{-1} \Big(\lim_{N \rightarrow \infty} \Xc(N)\Big)^\top \Big(\lim_{N \rightarrow \infty} \Vec{\Yc}(N)\Big)
    \eqas
    ( \Xc^\top \Xc )^{-1} \Xc^\top \Vec{\Yc},
\end{align*}
where we used the result of Theorem~\ref{thm:Batch_SE}. Now, note that \eqref{eq:apndx_BRD_model} defines a deterministic regression model, and $( \Xc^\top \Xc )^{-1} \Xc^\top \Vec{\Yc} = (\ABE, \ACE_{gl}, \ldots, \ACE_{g1}, \ADE_g, \APE_g, \ACE_{hl}, \ldots, \ACE_{h1}, \ADE_h, \APE_h)^\top$. This concludes the proof. \ep

To ensure the invertibility condition of the matrix $\Xc(N)^\top \Xc(N)$, we need to set some simple conditions on the experimental design. Basically, conducting the experiment in more than one stage (equivalent to having two distinct values for elements of the vector $\big( \hat\expr_l, \ldots, \hat\expr_T \big)^\top$ should suffice. This is needed to ensure that the first column of $\Xc(N)$ (i.e., $\Vec{1}_{b(T-l+1)}$) is linearly independent of the column $\big(\hat\expr_l, \ldots, \hat\expr_T, \ldots, \hat\expr_l, \ldots, \hat\expr_T\big)^\top$. 

Assuming a non-zero treatment effect (i.e., $\ADE_h \neq 0$), we can choose the batches to ensure enough variation across different batches. Therefore, upon a careful batching, columns of $\Xc(N)$ are linearly independent as each has its own specific variation patterns over time and/or subpopulations. 

Although the exact value of $\ADE$ is unknown, we suppose that contextual information suggests the presence of a non-zero direct treatment effect. In the case where $\ADE = 0$, according to contextual information, both the subpopulation sample mean $\AVO{}{\batch}{t}$ and the population sample mean $\AVO{}{}{t}$ are equal in \eqref{eq:SE_BRD_1} and  \eqref{eq:SE_BRD_2}, allowing for further simplification of the underlying model.
\setcounter{MaxMatrixCols}{11}
\begin{equation}
\label{eq:reg_cover_matrix}
    \begin{aligned}
        \Xc(N)
        =
        \begin{bmatrix}
        1
        &\HAVO{}{}{0}
        &\ldots
        &\HAVO{}{}{l-1}
        &\Oexpr{}{}{l}
        &\Oexpr{}{}{l} \HAVO{}{}{l-1} 
        &\HAVO{}{\batch_1}{0}
        &\ldots
        &\HAVO{}{\batch_1}{l-1}
        &\Oexpr{}{\batch_1}{l}
        &\Oexpr{}{\batch_1}{l} \HAVO{}{\batch_1}{l-1} 
        \\
        1
        &\HAVO{}{}{1}
        &\ldots
        &\HAVO{}{}{l}
        &\Oexpr{}{}{l+1}
        &\Oexpr{}{}{l+1} \HAVO{}{}{l} 
        &\HAVO{}{\batch_1}{1}
        &\ldots
        &\HAVO{}{\batch_1}{l}
        &\Oexpr{}{\batch_1}{l+1}
        &\Oexpr{}{\batch_1}{l+1} \HAVO{}{\batch_1}{l}
        \\
        \vdots
        &\vdots
        &\vdots
        &\vdots
        &\vdots
        &\vdots
        &\vdots
        &\vdots
        &\vdots
        &\vdots
        &\vdots
        \\
        1
        &\HAVO{}{}{T-l}
        &\ldots
        &\HAVO{}{}{T-1}
        &\Oexpr{}{}{T}
        &\Oexpr{}{}{T} \HAVO{}{}{T-1} 
        &\HAVO{}{\batch_1}{T-l}
        &\ldots
        &\HAVO{}{\batch_1}{T-1}
        &\Oexpr{}{\batch_1}{T}
        &\Oexpr{}{\batch_1}{T} \HAVO{}{\batch_1}{T-1}
        \\
        \\
        \vdots
        &\vdots
        &\vdots
        &\vdots
        &\vdots
        &\vdots
        &\vdots
        &\vdots
        &\vdots
        &\vdots
        &\vdots
        \\
        \\
        1
        &\HAVO{}{}{0}
        &\ldots
        &\HAVO{}{}{l-1}
        &\Oexpr{}{}{l}
        &\Oexpr{}{}{l} \HAVO{}{}{l-1} 
        &\HAVO{}{\batch_b}{0}
        &\ldots
        &\HAVO{}{\batch_b}{l-1}
        &\Oexpr{}{\batch_b}{l}
        &\Oexpr{}{\batch_b}{l} \HAVO{}{\batch_b}{l-1} 
        \\
        1
        &\HAVO{}{}{1}
        &\ldots
        &\HAVO{}{}{l}
        &\Oexpr{}{}{l+1}
        &\Oexpr{}{}{l+1} \HAVO{}{}{l} 
        &\HAVO{}{\batch_b}{1}
        &\ldots
        &\HAVO{}{\batch_b}{l}
        &\Oexpr{}{\batch_b}{l+1}
        &\Oexpr{}{\batch_b}{l+1} \HAVO{}{\batch_b}{l}
        \\
        \vdots
        &\vdots
        &\vdots
        &\vdots
        &\vdots
        &\vdots
        &\vdots
        &\vdots
        &\vdots
        &\vdots
        &\vdots
        \\
        1
        &\HAVO{}{}{T-l}
        &\ldots
        &\HAVO{}{}{T-1}
        &\Oexpr{}{}{T}
        &\Oexpr{}{}{T} \HAVO{}{}{T-1} 
        &\HAVO{}{\batch_b}{T-l}
        &\ldots
        &\HAVO{}{\batch_b}{T-1}
        &\Oexpr{}{\batch_b}{T}
        &\Oexpr{}{\batch_b}{T} \HAVO{}{\batch_b}{T-1}
        \end{bmatrix}.
    \end{aligned}
\end{equation}

\subsection{First-order Estimators}
\label{apndx:estimators}
Given the observed outcomes $\Moutcome{}{}{}(\OMtreatment{o}{})$, we can leverage Theorem~\ref{thm:consistency} and Proposition~\ref{prp:BRD_consistency} to consistently estimate counterfactuals under a desired treatment allocation $\OMtreatment{u}{}$. To this end, we propose two closely related families of estimators, which we detail below. Both family of estimators require that the delivered treatment allocation $\OMtreatment{o}{}$ and the desired treatment allocation $\OMtreatment{u}{}$ match during the first $l$ periods. These initial $l$ periods serve as the common foundation from which counterfactual trajectories are constructed.

Given subpopulation $\batch$, Algorithms~\ref{alg:FO-semi-recursive} and \ref{alg:FO-recursive} aim to estimate counterfactuals under the desired treatment allocation over $\batch$ using $b$ distinct subpopulations $\batch_1, \ldots, \batch_b$ as the estimation samples. Both algorithms share their first two steps. In the first step, they compute sample means of observed outcomes for both the entire population and each subpopulation, along with sample means of delivered and desired treatment allocations. The second step estimates unknown parameters in the state evolution equation \eqref{eq:SE_BRD_2} using least squares estimation as detailed in Proposition~\ref{prp:BRD_consistency}. The algorithms then diverge in their third step, applying these results through two distinct approaches detailed below. The consistency proofs for both algorithms follow directly from earlier results and are omitted for brevity.

\subsubsection{Semi-recursive Estimation Method}
\label{apndx:semi-recursive_estimators}
This estimator, outlined in Algorithm~\ref{alg:FO-semi-recursive}, builds on the observed sample means in its third step. It uses the parameter estimates from the second step and the state evolution equation \eqref{eq:SE_BRD_2} to modify the observed sample means by adjusting the treatment level to the desired one. This approach transfers the original complexities of the observed outcomes to the estimated counterfactual, making it particularly suitable for scenarios with strong time trends or complex baselines. This estimator generalizes the algorithm proposed in \cite{shirani2024causal} by accommodating broader model classes and providing more general estimands.

\begin{algorithm}
\caption{First-order semi-recursive counterfactual estimator}
\label{alg:FO-semi-recursive}
\begin{algorithmic}
\Require $\Moutcome{}{}{}(\OMtreatment{o}{}), \OMtreatment{o}{}$, $\OMtreatment{u}{}$, estimation batch $\batch$, sample batches $\batch_1, \ldots, \batch_b$, and $l$

\State \hspace{-1.3em} \textbf{Step 1: Data processing}
\For{$t = 0, \ldots, T$}
    \State $\HAVO{}{}{t}
        \gets
        \frac{1}{N} \sum_{i=1}^N \outcomeD{}{i}{t} (\OMtreatment{o}{})$
    \State $\HAVO{}{\batch}{t}
        \gets
        \frac{1}{\cardinality{\batch}} \sum_{i \in  \batch} \outcomeD{}{i}{t} (\OMtreatment{o}{})$
    \State $\Oexpr{}{}{t}
        \gets
        \frac{1}{N} \sum_{i=1}^N \Otreatment{i}{o,t}{}$
    \State $\Oexpr{}{\batch}{t}
        \gets
        \frac{1}{\cardinality{\batch}} \sum_{i \in  \batch} \Otreatment{i}{o,t}{}$
    \State $\Dexpr{}{}{t}
        \gets
        \frac{1}{N} \sum_{i=1}^N \Otreatment{i}{u,t}{}$
    \State $\Dexpr{}{\batch}{t}
        \gets
        \frac{1}{\cardinality{\batch}} \sum_{i \in  \batch} \Otreatment{i}{u,t}{}$
    \For{$j = 1, \ldots, b$}
        \State $\HAVO{}{\batch_j}{t}
        \gets
        \frac{1}{\cardinality{\batch_j}} \sum_{i \in  \batch_j} \outcomeD{}{i}{t} (\OMtreatment{o}{})$
        \State $\Oexpr{}{\batch_j}{t}
        \gets
        \frac{1}{\cardinality{\batch_j}} \sum_{i \in  \batch_j} \Otreatment{i}{o,t}{}$
    \EndFor    
\EndFor

\State \hspace{-1.3em} \textbf{Step 2: Parameters estimation}
\State $(\EBE, \ECE_{gl}, \ldots, \ECE_{g1}, \EDE_g, \EPE_g, \ECE_{hl}, \ldots, \ECE_{h1}, \EDE_h, \EPE_h)^\top \gets \big( \hat{\Xc}^\top \hat{\Xc} \big)^{-1} \hat{\Xc}^\top \Vec{\hat{\Yc}}$

\State \hspace{-1.3em} \textbf{Step 3: Counterfactual estimation}

\State $\Big(\ECF{}{0}{\OMtreatment{u}{}}, \ldots, \ECF{}{l-1}{\OMtreatment{u}{}} \Big) \gets \left(\HAVO{}{}{0}, \ldots, \HAVO{}{}{l-1} \right)$

\State $\Big(\ECF{\batch}{0}{\OMtreatment{u}{}}, \ldots, \ECF{\batch}{l-1}{\OMtreatment{u}{}} \Big) \gets \left(\HAVO{}{\batch}{0}, \ldots, \HAVO{}{\batch}{l-1} \right)$

\For{$t = l, \ldots, T$}
    \State $\Rc_g \gets \sum_{j=1}^l \ECE_{gj} (\ECF{}{t-j}{\OMtreatment{u}{}} - \HAVO{}{}{t-j}) + \EDE_g (\Dexpr{}{}{t}-\Oexpr{}{}{t}) + \EPE_g (\Dexpr{}{}{t} \ECF{}{t-1}{\OMtreatment{u}{}} - \Oexpr{}{}{t} \HAVO{}{}{t-1})$
    \State $\Rc_h \gets \sum_{j=1}^l \ECE_{hj} (\ECF{}{t-j}{\OMtreatment{u}{}} - \HAVO{}{}{t-j}) + \EDE_h (\Dexpr{}{}{t}-\Oexpr{}{}{t}) + \EPE_h (\Dexpr{}{}{t} \ECF{}{t-1}{\OMtreatment{u}{}} - \Oexpr{}{}{t} \HAVO{}{}{t-1})$
    \State $\Rc_h^\batch \gets \sum_{j=1}^l \ECE_{hj} (\ECF{\batch}{t-j}{\OMtreatment{u}{}} - \HAVO{}{\batch}{t-j}) + \EDE_h (\Dexpr{}{\batch}{t}-\Oexpr{}{\batch}{t}) + \EPE_h (\Dexpr{}{\batch}{t} \ECF{\batch}{t-1}{\OMtreatment{u}{}} - \Oexpr{}{\batch}{t} \HAVO{}{\batch}{t-1})$
    \State $\ECF{}{t}{\OMtreatment{u}{}} \gets \HAVO{}{}{t} + \Rc_g + \Rc_h$
    \State $\ECF{\batch}{t}{\OMtreatment{u}{}} \gets \HAVO{}{\batch}{t} + \Rc_g + \Rc_h^\batch$
\EndFor

\Ensure $\ECF{}{t}{\OMtreatment{u}{}}$ and $\ECF{\batch}{t}{\OMtreatment{u}{}}$, for $t = 0, \ldots, T$.
\end{algorithmic}
\end{algorithm}

\subsubsection{Recursive Estimation Method}
\label{apndx:recursive_estimators}
This estimator, outlined in Algorithm~\ref{alg:FO-recursive}, directly leverages the state evolution equation and parameter estimates from the second step. Specifically, it estimates counterfactuals recursively, using only the past $l$ terms and desired treatment levels. Unlike Algorithm~\ref{alg:FO-semi-recursive}, this algorithm can estimate counterfactuals even for time blocks where no outcome data were collected. For example, having observed data until December, it can predict counterfactual outcomes for January without requiring any observations during this month.

\begin{algorithm}
\caption{First-order recursive counterfactual estimator}
\label{alg:FO-recursive}
\begin{algorithmic}
\Require $\Moutcome{}{}{}(\OMtreatment{o}{}), \OMtreatment{o}{}$, $\OMtreatment{u}{}$, estimation batch $\batch$, sample batches $\batch_1, \ldots, \batch_b$, and $l$

\State \hspace{-1.3em} \textbf{Step 1: Data processing}
\For{$t = 0, \ldots, T$}
    \State $\HAVO{}{}{t}
        \gets
        \frac{1}{N} \sum_{i=1}^N \outcomeD{}{i}{t} (\OMtreatment{o}{})$
    \State $\HAVO{}{\batch}{t}
        \gets
        \frac{1}{\cardinality{\batch}} \sum_{i \in  \batch} \outcomeD{}{i}{t} (\OMtreatment{o}{})$
    \State $\Oexpr{}{}{t}
        \gets
        \frac{1}{N} \sum_{i=1}^N \Otreatment{i}{o,t}{}$
    \State $\Oexpr{}{\batch}{t}
        \gets
        \frac{1}{\cardinality{\batch}} \sum_{i \in  \batch} \Otreatment{i}{o,t}{}$
    \State $\Dexpr{}{}{t}
        \gets
        \frac{1}{N} \sum_{i=1}^N \Otreatment{i}{u,t}{}$
    \State $\Dexpr{}{\batch}{t}
        \gets
        \frac{1}{\cardinality{\batch}} \sum_{i \in  \batch} \Otreatment{i}{u,t}{}$
    \For{$j = 1, \ldots, b$}
        \State $\HAVO{}{\batch_j}{t}
        \gets
        \frac{1}{\cardinality{\batch_j}} \sum_{i \in  \batch_j} \outcomeD{}{i}{t} (\OMtreatment{o}{})$
        \State $\Oexpr{}{\batch_j}{t}
        \gets
        \frac{1}{\cardinality{\batch_j}} \sum_{i \in  \batch_j} \Otreatment{i}{o,t}{}$
    \EndFor    
\EndFor

\State \hspace{-1.3em} \textbf{Step 2: Parameters estimation}
\State $(\EBE, \ECE_{gl}, \ldots, \ECE_{g1}, \EDE_g, \EPE_g, \ECE_{hl}, \ldots, \ECE_{h1}, \EDE_h, \EPE_h)^\top \gets \big( \hat{\Xc}^\top \hat{\Xc} \big)^{-1} \hat{\Xc}^\top \Vec{\hat{\Yc}}$

\State \hspace{-1.3em} \textbf{Step 3: Counterfactual estimation}

\State $\Big(\ECF{}{0}{\OMtreatment{u}{}}, \ldots, \ECF{}{l-1}{\OMtreatment{u}{}} \Big) \gets \left(\HAVO{}{}{0}, \ldots, \HAVO{}{}{l-1} \right)$

\State $\Big(\ECF{\batch}{0}{\OMtreatment{u}{}}, \ldots, \ECF{\batch}{l-1}{\OMtreatment{u}{}} \Big) \gets \left(\HAVO{}{\batch}{0}, \ldots, \HAVO{}{\batch}{l-1} \right)$

\For{$t = l, \ldots, T$}
    \State $\Rc_g \gets \sum_{j=1}^l \ECE_{gj} \ECF{}{t-j}{\OMtreatment{u}{}} + \EDE_g \Dexpr{}{}{t} + \EPE_g \Dexpr{}{}{t} \ECF{}{t-1}{\OMtreatment{u}{}}$
    \State $\Rc_h \gets \sum_{j=1}^l \ECE_{hj} \ECF{}{t-j}{\OMtreatment{u}{}} + \EDE_h \Dexpr{}{}{t} + \EPE_h \Dexpr{}{}{t} \ECF{}{t-1}{\OMtreatment{u}{}}$
    \State $\Rc_h^\batch \gets \sum_{j=1}^l \ECE_{hj} \ECF{\batch}{t-j}{\OMtreatment{u}{}} + \EDE_h \Dexpr{}{\batch}{t} + \EPE_h \Dexpr{}{\batch}{t} \ECF{\batch}{t-1}{\OMtreatment{u}{}}$
    \State $\ECF{}{t}{\OMtreatment{u}{}} \gets \Rc_g + \Rc_h$
    \State $\ECF{\batch}{t}{\OMtreatment{u}{}} \gets \Rc_g + \Rc_h^\batch$
\EndFor

\Ensure $\ECF{}{t}{\OMtreatment{u}{}}$ and $\ECF{\batch}{t}{\OMtreatment{u}{}}$, for $t = 0, \ldots, T$.
\end{algorithmic}
\end{algorithm}

\subsection{Higher-order Recursive Estimators}
\label{apndx:HO_recursive_estimators}
In light of Theorem~\ref{thm:consistency}, we can extend our approach to utilize higher-order approximations of $\outcomeg{t}{}$ and $\outcomeh{t}{}$. The estimator, outlined in Algorithm~\ref{alg:HO-recursive} for $l=1$ lag terms, incorporates up to order $m \geq 2$ moments of the unit outcomes. Precisely, we introduce two families of feature functions: $\Vec\phi = (\phi_1, \ldots, \phi_{n_1})^\top$ for population-level moments and $\Vec\psi = (\psi_1, \ldots, \psi_{n_2})^\top$ for subpopulation-level moments. The approach employs linear regression to estimate weights for a linear combination of these features. This can be realized as a generalization of \cite{bayati2024higher}'s method and enables capturing more complex patterns in counterfactuals by leveraging richer information about the outcomes' distributions over time. When the feature functions $\Vec\phi$ and $\Vec\psi$ are continuous, consistency follows from Theorem~\ref{thm:consistency}, though we defer rigorous treatment to future work.

\begin{algorithm}
\caption{Higher-order recursive counterfactual estimator}
\label{alg:HO-recursive}
\small
\begin{algorithmic}
\Require $\Moutcome{}{}{}(\OMtreatment{o}{}), \OMtreatment{o}{}$, $\OMtreatment{u}{}$, $\batch_1, \ldots, \batch_b$, $m \geq 2$, $\Vec\phi = (\phi_1, \ldots, \phi_{n_1})^\top$, and $\Vec\psi = (\psi_1, \ldots, \psi_{n_2})^\top$

\State \hspace{-1.3em} \textbf{Step 1: Data processing}
    \For{$t = 0, \ldots, T$}
    \State $\HAVO{}{}{t}
        \gets
        \frac{1}{N} \sum_{i=1}^N \outcomeD{}{i}{t} (\OMtreatment{o}{})$
    \State $\HAVO{}{\batch}{t}
        \gets
        \frac{1}{\cardinality{\batch}} \sum_{i \in \batch} \outcomeD{}{i}{t} (\OMtreatment{o}{})$
    \For{$k = 2, \ldots, m$}
        \State $\HVVO{}{(k)}{t}
        \gets
        \frac{1}{N} \sum_{i=1}^N 
        \left(
        \outcomeD{}{i}{t} (\OMtreatment{o}{})
        -
        \HAVO{}{}{t}
        \right)^k$
        \State $\HVVO{}{\batch,(k)}{t}
        \gets
        \frac{1}{\cardinality{\batch}} \sum_{i \in \batch}
        \left(
        \outcomeD{}{i}{t} (\OMtreatment{o}{})
        -
        \HAVO{}{\batch}{t}
        \right)^k$
    \EndFor
    \State $\Oexpr{}{}{t}
        \gets
        \frac{1}{N} \sum_{i=1}^N \Otreatment{i}{o,t}{}$
    \State $\Oexpr{}{\batch}{t}
        \gets
        \frac{1}{\cardinality{\batch}} \sum_{i \in \batch} \Otreatment{i}{o,t}{}$
    \State $\Dexpr{}{}{t}
        \gets
        \frac{1}{N} \sum_{i=1}^N \Otreatment{i}{u,t}{}$
    \State $\Dexpr{}{\batch}{t}
        \gets
        \frac{1}{\cardinality{\batch}} \sum_{i \in \batch} \Otreatment{i}{u,t}{}$
    \For{$j = 1, \ldots, b$}
        \State $\HAVO{}{\batch_j}{t}
        \gets
        \frac{1}{\cardinality{\batch_j}} \sum_{i \in  \batch_j} \outcomeD{}{i}{t} (\OMtreatment{o}{})$
        \For{$k = 2, \ldots, m$}
            \State $\HVVO{}{\batch_j,(k)}{t}
            \gets
            \frac{1}{\cardinality{\batch_j}} \sum_{i \in  \batch_j} 
            \left(
            \outcomeD{}{i}{t} (\OMtreatment{o}{})
            -
            \HAVO{}{\batch_j}{t}
            \right)^k$
        \EndFor
        \State $\Oexpr{}{\batch_j}{t}
        \gets
        \frac{1}{\cardinality{\batch_j}} \sum_{i \in  \batch_j} \Otreatment{i}{o,t}{}$
    \EndFor    
\EndFor

\State \hspace{-1.3em} \textbf{Step 2: Parameters estimation}
\State Estimate $\bm{\Theta}_g \in \R^{m \times n_1}$ and $\bm{\Theta}_h  \in \R^{m \times n_2}$, respectively, as $\widehat{\bm{\Theta}}_g$ and $\widehat{\bm{\Theta}}_h$: 
\begin{equation*}
    \begin{aligned}
        (\HAVO{}{\batch_j}{t+1}, \HVVO{}{\batch_j,(2)}{t+1}, \ldots, \HVVO{}{\batch_j,(m)}{t+1})^\top
        =
        \bm{\Theta}_g \Vec{\phi}(\HAVO{}{}{t}, \HVVO{}{(2)}{t}, \ldots, \HVVO{}{(m)}{t}, \Oexpr{}{}{t+1})
        +
        \bm{\Theta}_h \Vec{\psi}(\HAVO{}{\batch_j}{t}, \HVVO{}{\batch_j,(2)}{t}, \ldots, \HVVO{}{\batch_j,(m)}{t}, \Oexpr{}{\batch_j}{t+1}),
    \end{aligned}
\end{equation*}
where $j = 1, \ldots, b$ and $t = 0, \ldots, T-1$.

\State \hspace{-1.3em} \textbf{Step 3: Counterfactual estimation}

\State $\ECF{}{0}{\OMtreatment{u}{}} \gets \HAVO{}{}{0}$ and $(\DHVVO{}{(2)}{0}, \ldots, \DHVVO{}{(m)}{0}) \gets (\HVVO{}{(2)}{0}, \ldots, \HVVO{}{(m)}{0})$
\State $\ECF{\batch}{0}{\OMtreatment{u}{}} \gets \HAVO{}{\batch}{0}$ and $(\DHVVO{}{\batch,(2)}{0}, \ldots, \DHVVO{}{\batch,(m)}{0}) \gets (\HVVO{}{\batch,(2)}{0}, \ldots, \HVVO{}{\batch,(m)}{0})$
\For{$t = 1, \ldots, T$}
    \State $\Vec\Rc_g \gets
    \widehat{\bm{\Theta}}_g \Vec{\phi}(\ECF{}{t-1}{\OMtreatment{u}{}}, \DHVVO{}{(2)}{t-1}, \ldots, \DHVVO{}{(m)}{t-1}, , \Dexpr{}{}{t})$
    \State $\Vec\Rc_h \gets
    \widehat{\bm{\Theta}}_h \Vec{\psi}(\ECF{}{t-1}{\OMtreatment{u}{}}, \DHVVO{}{(2)}{t-1}, \ldots, \DHVVO{}{(m)}{t-1}, \Dexpr{}{}{t})$
    \State $\Vec\Rc_h^\batch \gets
    \widehat{\bm{\Theta}}_h \Vec{\psi}(\ECF{\batch}{t-1}{\OMtreatment{u}{}}, \DHVVO{}{\batch,(2)}{t-1}, \ldots, \DHVVO{}{\batch,(m)}{t-1}, \Dexpr{}{\batch}{t})$
    \State $(\ECF{}{t}{\OMtreatment{u}{}}, \DHVVO{}{(2)}{t}, \ldots, \DHVVO{}{(m)}{t})^\top \gets
    \Vec\Rc_g + \Vec\Rc_h$
    \State $(\ECF{\batch}{t}{\OMtreatment{u}{}}, \DHVVO{}{\batch,(2)}{t}, \ldots, \DHVVO{}{\batch,(m)}{t})^\top \gets
    \Vec\Rc_g + \Vec\Rc_h^\batch$
\EndFor

\Ensure $\ECF{}{t}{\OMtreatment{u}{}}$ and $\ECF{\batch}{t}{\OMtreatment{u}{}}$, for $t = 0, \ldots, T$.
\end{algorithmic}
\end{algorithm}

\section{Detrending for Temporal Patterns}
\label{sec:preprocessing}
While semi-recursive estimators (see \S\ref{apndx:semi-recursive_estimators}) can capture complex temporal patterns in unit outcomes, they face two major limitations. First, they cannot estimate out-of-sample counterfactuals because their architecture relies directly on observed sample means. Second, unlike recursive estimators (see Algorithm~\ref{alg:HO-recursive}), they cannot accommodate higher-order approximations of the outcome functions ($\outcomeg{t}{}$ and $\outcomeh{t}{}$). This limitation stems from their dependence on the explicit form of the state evolution equation (as outlined in \eqref{eq:SE_BRD_2}), which may not exist for more complex characterizations of the outcome functions.

This section develops a two-stage estimation method that combines the advantages of both semi-recursive and recursive estimators. Although it requires an additional structural assumption on the outcome specification, this approach can handle complex temporal patterns while enabling both out-of-sample counterfactual estimation and higher-order approximations of the outcome functions. The method proceeds as follows: first, we employ a semi-recursive estimator with sufficient lag terms to accurately estimate temporal patterns. We use this to estimate the baseline outcome means (the counterfactual for all control units: $\CFE{\mathbf{0}}{}{t},\; t =0, 1, \ldots, T$). Next, we detrend the observed outcomes by subtracting the estimated baseline from them. We then apply a recursive estimator focused specifically on estimating treatment effects in the absence of temporal patterns. Finally, we add back the subtracted baseline to obtain the desired estimand (see Algorithm~\ref{alg:FO_with_preprocessing}). The following sections provide more detailed expositions of the algorithm.

\subsection{Baseline Outcome Estimation}
\label{sec:Y0_estimation}
Letting $\VOoutcomeD{}{}{t} := \VoutcomeD{}{}{t}(\Mtreatment{}{}=\mathbf{0})$ denote the vector of baseline outcomes at time $t= 0, 1, \ldots, T$ under no treatment, we can write from \eqref{eq:outcome_function_matrix}:
\begin{equation}
\label{eq:outcome_function_NoTreatment}
\begin{aligned}
    \VOoutcomeD{}{}{t+1}
    =
    \VoutcomeD{}{}{t+1}(\Mtreatment{}{}=\mathbf{0})
    &=
    \big(\IM+\IMatT{t}\big)\outcomeg{t}{}\left(\VOoutcomeD{}{}{t}, \mathbf{0}, \covar\right)
    +
    \outcomeh{t}{}\left(\VOoutcomeD{}{}{t}, \mathbf{0}, \covar\right).
\end{aligned}
\end{equation}
Thus, the matrix $\MOoutcomeD{}{}{} = [\VOoutcomeD{}{}{0}| \ldots | \VOoutcomeD{}{}{T}]$ represents the panel data of baseline outcomes that would be observed in the absence of any intervention.

In the first step of Algorithm~\ref{alg:FO_with_preprocessing}, we employ a semi-recursive algorithm to estimate the sample means of the columns of $\MOoutcomeD{}{}{}$, denoted by $\ECF{}{t}{\mathbf{0}}$. The consistency of this estimation follows directly from the consistency of Algorithm~\ref{alg:FO-semi-recursive}.

\subsection{Augmented Causal Message-Passing Model}
\label{sec:augmented_CMP}
The next two steps of Algorithm~\ref{alg:FO_with_preprocessing} require the following assumption.
\begin{assumption}
    \label{asmp:additive_baseline}
    For $t=0,1,\ldots,T-1$, we assume there exist families of functions $\Toutcomeg{t}$ and $\Toutcomeh{t}$ such that the potential outcomes $\VoutcomeD{}{}{t}(\Mtreatment{}{})$ satisfy:
    \begin{equation}
        \label{eq:aug_outcome_function}
    \begin{aligned}    
        \VoutcomeD{}{}{t+1}(\Mtreatment{}{})
        =
        \VOoutcomeD{}{}{t+1} +
        \big(\IM+\IMatT{t}\big)\Toutcomeg{t}\left(\VoutcomeD{}{}{t}(\Mtreatment{}{}) - \VOoutcomeD{}{}{t} ,\Mtreatment{}{}, \covar\right)
        +
        \Toutcomeh{t} \left(\VoutcomeD{}{}{t}(\Mtreatment{}{}) - \VOoutcomeD{}{}{t} ,\Mtreatment{}{}, \covar\right).
    \end{aligned}
    \end{equation}
    Additionally, $\Toutcomeg{t}\left(\Vec{0} , \mathbf{0}, \covar\right) = \Toutcomeh{t}\left(\Vec{0} , \mathbf{0}, \covar\right) = \Vec{0}$, and functions $\Toutcomeg{t}$ and $\Toutcomeh{t}$ satisfy the conditions of Assumption~\ref{asmp:BL}.
\end{assumption}
Note that enforcing conditions $\Toutcomeg{t}\left(\Vec{0} , \mathbf{0}, \covar\right) = \Toutcomeh{t}\left(\Vec{0} , \mathbf{0}, \covar\right) = \Vec{0}$ ensures that $\VoutcomeD{}{}{t+1}(\mathbf{0}) = \VOoutcomeD{}{}{t+1}$, aligning with \eqref{eq:outcome_function_NoTreatment}. 
Letting $\PPVoutcomeD{}{}{t}(\Mtreatment{}{}) := \VoutcomeD{}{}{t}(\Mtreatment{}{}) - \VOoutcomeD{}{}{t}$, we can then rewrite the augmented model \eqref{eq:aug_outcome_function} to match the dynamics of the original outcome specification:
\begin{equation}
\label{eq:preprocessed_outcome_function}
\begin{aligned}    
    \PPVoutcomeD{}{}{t+1}(\Mtreatment{}{})
    &=
    \big(\IM+\IMatT{t}\big)\Toutcomeg{t}\left(\PPVoutcomeD{}{}{t}(\Mtreatment{}{}) ,\Mtreatment{}{}, \covar\right)
    +
    \Toutcomeh{t}\left(\PPVoutcomeD{}{}{t}(\Mtreatment{}{}) ,\Mtreatment{}{}, \covar\right).
\end{aligned}
\end{equation}
We emphasize that Equations \eqref{eq:outcome_function_NoTreatment} and \eqref{eq:preprocessed_outcome_function} provide distinct characterizations of the outcomes, and Algorithm~\ref{alg:FO_with_preprocessing} requires both to hold simultaneously. Specifically, assuming the conditions of \S\ref{apndx:batch_state_evolution} hold in both settings, the baseline outcomes $\VOoutcomeD{}{}{t}$ satisfy the state evolution equation corresponding to \eqref{eq:outcome_function_NoTreatment}. However, in the context of \eqref{eq:preprocessed_outcome_function}, state evolution becomes relevant only when treatment is delivered (i.e., $\Mtreatment{}{} \neq \mathbf{0}$). Indeed, under Assumption~\ref{asmp:additive_baseline}, the third condition in Assumption~\ref{asmp:BL} indicates that state evolution can be derived only when there exists a non-zero treatment effect.

Then, the consistency of the second step estimation in Algorithm~\ref{alg:FO_with_preprocessing} holds in the context of the outcome specification \eqref{eq:aug_outcome_function}. Finally, the consistency of the ultimate estimator follows from both the consistency of individual steps and the fact that Algorithm~\ref{alg:FO_with_preprocessing} can be viewed as a combination of continuous functions.

\begin{algorithm}
\caption{First-order counterfactual estimator with preprocessing}
\label{alg:FO_with_preprocessing}
\begin{algorithmic}
\Require $\Moutcome{}{}{}(\OMtreatment{o}{}), \OMtreatment{o}{}$, $\OMtreatment{u}{}$, estimation batch $\batch$, sample batches $\batch_1, \ldots, \batch_b$, and $l$

\State \hspace{-1.3em} \textbf{Step 1: Detrending}
\State Use Algorithm~\ref{alg:FO-semi-recursive} with $\OMtreatment{u}{} = \mathbf{0}$ to obtain $\ECF{}{t}{\mathbf{0}}$, $t = 0, \ldots, T$. 
\For{$t = 0, \ldots, T$}
    \State $\VoutcomeD{}{'}{t} \gets \VoutcomeD{}{}{t}(\Mtreatment{}{}=\OMtreatment{o}{}) - \ECF{}{t}{\mathbf{0}}$
\EndFor

\State \hspace{-1.3em} \textbf{Step 2: Counterfactual estimation with proprocessed data}
\State Use Algorithm~\ref{alg:FO-recursive} with $\Moutcome{}{'}{}$ to obtain $\ECF{'\batch}{t}{\OMtreatment{u}{}}$, $t = 0, \ldots, T$.

\State \hspace{-1.3em} \textbf{Step 3: Post-processing}
\For{$t = 0, \ldots, T$}
    \State $\ECF{\batch}{t}{\OMtreatment{u}{}} \gets \ECF{'\batch}{t}{\OMtreatment{u}{}} + \ECF{}{t}{\mathbf{0}}$
\EndFor

\Ensure $\ECF{\batch}{t}{\OMtreatment{u}{}}$, $t = 0, \ldots, T$.
\end{algorithmic}
\end{algorithm}

\section{Auxiliary Results}
\label{apndx:auxiliary_results}
We need the following strong law of large numbers (SLLN) for triangular arrays of independent but not identically distributed random variables. The form stated below is Theorem~3 in \cite{bayati2011dynamics} that is adapted from Theorem~2.1 in \cite{hu1997strong}.
\begin{theorem}[SLLN]
    \label{thm:SLLN}
    Let $\left\{X_{n,i}:1\leq i \leq n,\; n \geq 1\right\}$ be a triangular array of random variables such that $(X_{n,1},\ldots,X_{n,n})$ are mutually independent with a mean equal to zero for each $n$ and $\frac{1}{n} \sum_{i=1}^n E\left[|X_{n,i}|^{2+\kappa}\right] \leq c n^{\kappa/2}$ for some $0 < \kappa < 1$ and $c < \infty$. Then, we have
    \begin{align}
        \label{eq:SLLN}
        \lim_{n \rightarrow \infty}
        \frac{1}{n} \sum_{i=1}^n X_{n,i} \eqas 0.
    \end{align}
\end{theorem}
We also need the following form of the law of large numbers, which is an extension of Lemma~4 in \cite{bayati2011dynamics}, and the proof can be found in the work of \cite{shirani2024causal}.
\begin{theorem}
    \label{thm:SLLN-2}
    Fix $k\geq 2$ and an integer $l$ and let $\left\{\bm v(N)\right\}_{N \geq 1}$ be a sequence of vectors that $\bm v(N) \in \R^{N\x l}$. That means, $\bm v(N)$ is a matrix with $N$ rows and $l$ columns. Assume that the empirical distribution of $\bm v(N)$, denoted by $\hat{p}_{N}$, converges weakly to a probability measure $p_v$ on $\R^l$ such that $\E_{p_v}\left[\norm{\Vec{V}}^k\right] < \infty$ and $\E_{\hat{p}_{N}}\left[\norm{\Vec{V}}^k\right] \rightarrow \E_{p_v}\left[\norm{\Vec{V}}^k\right]$ as $N \rightarrow \infty$. Then, for any continuous function $f:\R^l \mapsto \R$ with at most polynomial growth of order $k$, we have
    \begin{align}
        \label{eq:SLLN-2}
        \lim_{N \rightarrow \infty}
        \frac{1}{N} \sum_{n=1}^N f\big(\bm v_n(N)\big)
        \eqas \E_{p_v} [f(\Vec{V})].
    \end{align}
\end{theorem}
Next, we present Lemma 9 from \cite{li2022non}, and then employ it to prove Lemma~\ref{lm:Gap_with_Gaussian}.
\begin{lemma}
    \label{lm:Gap_with_Gaussian_vector}
    Fixing $N$ and $t < N$, consider a set of i.i.d. random vectors $\Vec{\phi_i} \sim \Nc(0,\frac{1}{N}\I_N),\; i=1, \ldots, t$, and any unit vector $\VNPC{}{} = (\NPC{1}{}, \ldots, \NPC{t}{})^\top$ that might be statistically dependent on $\{\Vec{\phi_i}\}_{i=1}^t$. Then, the 1-Wasserstein distance between the distribution of $\sum_{i=1}^t \NPC{i}{} \Vec{\phi_i}$, denoted by $\law(\sum_{i=1}^t \NPC{i}{} \Vec{\phi_i})$, and $\Nc(0,\frac{1}{N}\I_N)$ obeys
    \begin{equation*}
        W_1\left(\law\left(\sum_{i=1}^t \NPC{i}{} \Vec{\phi_i}\right),\Nc\left(0,\frac{1}{N}\I_N\right)\right) \leq c \sqrt{\frac{t \log N}{N}},
    \end{equation*}
    for some constant $c$ that does not depend on $N$.
\end{lemma}

\begin{lemma}
    \label{lm:Gap_with_Gaussian}
    Fixing $N$ and $t < N$, consider a set of i.i.d. random vectors $\Vec{\phi_i} \sim \Nc(0,\frac{1}{N}\I_N),\; i=1, \ldots, t$, and any unit vector $\VNPC{}{} = (\NPC{1}{}, \ldots, \NPC{t}{})^\top$ that might be statistically dependent on $\{\Vec{\phi_i}\}_{i=1}^t$. Let $\Vec{\Phi} := \sum_{i=1}^t \NPC{i}{} \Vec{\phi_i}$ such that $\Vec{\Phi} = (\Phi^1, \ldots, \Phi^N)^\top$ and $\batch \subset [N]$ be subset of the indices. Then, the 1-Wasserstein distance between the distribution of $\frac{1}{\sqrt{\cardinality{\batch}}}\sum_{n \in \batch} \Phi^n$, denoted by $\law(\frac{1}{\sqrt{\cardinality{\batch}}}\sum_{n \in \batch} \Phi^n)$, and $\Nc(0,{1}/ {N})$ satisfies
    \begin{equation*}
        W_1\left(\law\left(\frac{1}{\sqrt{\cardinality{\batch}}}\sum_{n \in \batch} \Phi^n\right),\Nc\left(0,\frac{1}{N}\right)\right) \leq c \sqrt{\frac{t \log N}{N}},
    \end{equation*}
    for some constant $c$ that does not depend on $N$ and $\batch$.
\end{lemma}
\textbf{Proof}. Considering Kantorovich-Rubinstein duality, we use the dual representation of the 1-Wasserstein distance:
\begin{equation}
    \label{eq:apndx_proof_Gap_with_Gaussian_1}
    W_1\left(\law\left(\frac{1}{\sqrt{\cardinality{\batch}}}\sum_{n \in \batch} \Phi^n\right),\Nc\left(0,\frac{1}{N}\right)\right)
    = \sup\left\{\E\left[f\left(\frac{1}{\sqrt{\cardinality{\batch}}}\sum_{n \in \batch} \Phi^n\right)\right] - \E\left[f\left(\frac{Z}{\sqrt{N}}\right)\right]\;\Bigg|\; f \text{ is 1-Lipschitz}\right\},
\end{equation}
where $Z \sim \Nc(0,1)$. To proceed, fix the 1-Lipschitz function $f$ arbitrarily. Further, define the function $\psi: \R^N \mapsto \R$ such that for any vector $\Avec{}{} = (\avec{1}{}, \ldots, \avec{N}{})^\top$, we have $\psi(\Avec{}{}) = \frac{1}{\sqrt{\cardinality{\batch}}} \sum_{n \in \batch} \avec{n}{}$. Then, we define $\tilde{f} = f\circ \psi : \R^N \mapsto \R$. Note that both $f$ and $\psi$ are continuous functions; as a result,
the function $\tilde{f}$ is measurable. We show it is also $1$-Lipschitz. To this end, for vectors $\Avec{}{1}, \Avec{}{2} \in \R^N$, we write:
\begin{equation*}
    \begin{aligned}
        \left|\tilde{f}\left(\Avec{}{2}\right) - \tilde{f}\left(\Avec{}{1}\right)\right|
        = \left|f\left(\psi(\Avec{}{2})\right) - f\left(\psi(\Avec{}{1})\right)\right|
        \leq \left|\psi(\Avec{}{2}) - \psi(\Avec{}{1})\right|
        \leq
        \frac{\sum_{n \in \batch} \left|\avec{n}{2} - \avec{n}{1} \right|}{\sqrt{\cardinality{\batch}}} 
        \leq
        \sqrt{\sum_{n \in \batch} \left|\avec{n}{2} - \avec{n}{1} \right|^2}
        \leq
        \norm{\Avec{}{2} - \Avec{}{1}},
    \end{aligned}
\end{equation*}
where we used the fact that $f$ is 1-Lipschitz and the Cauchy–Schwarz inequality. Therefore, the function $\tilde{f}$ is 1-Lipschitz and by the result of Lemma~\ref{lm:Gap_with_Gaussian_vector}, we get
\begin{equation*}
    \E\left[f\left(\frac{1}{\sqrt{N}}\sum_{n=1}^N \Phi^n\right)\right] - \E\left[f(\frac{Z}{\sqrt{N}})\right]
    = \E\left[\tilde{f}\left(\Vec{\Phi}\right)\right] - \E\left[\tilde{f}\left(\frac{1}{\sqrt{N}}\Vec{Z}\right)\right]
    \leq 
    c\sqrt{\frac{t \log N}{N}},
\end{equation*}
where $\Vec{Z} \sim \Nc(0,\I)$. Because $f$ is chosen arbitrarily, by Eq.~\eqref{eq:apndx_proof_Gap_with_Gaussian_1}, we obtain the desired result. \ep